\documentclass[a4paper,10pt,twoside,openright]{book}

\usepackage{structure-digital}

\hypersetup{
pdftitle={Data augmentation and image understanding},
pdfsubject={The role of data augmentation as implicit regularisation of artificial neural networks and inductive bias from visual perception and biological vision},
pdfauthor={Alex Hernandez-Garcia},
pdfkeywords={data augmentation; regularization; deep learning; neuroscience; visual perception; salience; attention; vision; self-supervised learning}
}

\begin{document}

\pagenumbering{arabic}

{
\begin{titlepage}
\newgeometry{top=3cm,bottom=3.5cm,left=2.0cm,right=2.0cm}

    \begin{center}

	\Huge{\textbf{Data augmentation\\and image understanding}} \\
	
	\vspace{4.0cm}
	
	\Large{%
    PhD Thesis\\
    Institute of Cognitive Science\\
    University of Osnabrück}\\
	\vspace{2.0cm}
	
	\Large{\textbf{Alex Hern\'andez-Garc\'ia}} \\
	
	\vspace*{0.2cm}
	\rule{80mm}{0.1mm} \\ 
	\vspace*{0.1cm}
      
	\large{Advisor: Prof. Peter K\"onig} \\
      
	\vspace{0.5cm}

	\large{July 2020}

	\vspace{1cm}

	\large{\textbf{Doctoral committee:}}\\
	\vspace{0.1cm}
	\large{Jeffrey Bowers}\\
	\large{Konrad P. Kording}\\
	\large{Graham W. Taylor}\\
  \large{Peter K\"onig (advisor)}\\
  \large{Gordon Pipa (chair)}

  \vspace{3.0cm}

  \includegraphics[width= 0.2 \linewidth]{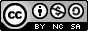}\\\vspace{0.5cm}
    \footnotesize{This work is licensed under a Creative Commons Attribution-NonCommercial-ShareAlike 4.0 International License: \texttt{http:}//creativecommons.org/licenses/by-nc-sa/4.0}
  \end{center}
  
\end{titlepage}
\cleardoublepage
}

\phantomsection
\addcontentsline{toc}{chapter}{Foreword}
\begingroup
\vspace*{\fill}
	\begin{center} 
        \Large{\textbf{Foreword}}
	\vspace{10pt}
	\end{center}
\small
There are certainly many people I am grateful to for their support and love. I hope I show them and they feel my gratitude and love often enough. Instead of naming them here, I will make sure to thank them personally. I am also grateful to my PhD adviser and collaborators. I acknowledge their specific contributions at the beginning of each chapter. I wish to use this space for putting down in words some thoughts that have been recurrently present throughout my PhD and are especially strong now. 

The main reason why I have been able to write a PhD thesis is \textit{luck}. Specifically, I have been very lucky to be a privileged person. In my opinion, we often overestimate our own effort, skills and hard work as the main factors for professional success. I would not have got here without my privileged position, in the first place. 

There are some obvious comparisons. To name a few: During the first months of my PhD, Syrians had undergone several years of war\footnote{\href{https://en.wikipedia.org/wiki/Syrian_civil_war\#Timeline}{https://en.wikipedia.org/wiki/Syrian\_civil\_war\#Timeline}} and citizens of Aleppo were suffering one of the most cruel sieges ever\footnote{\href{https://en.wikipedia.org/wiki/Battle_of_Aleppo_(2012-2016)}{https://en.wikipedia.org/wiki/Battle\_of\_Aleppo\_(2012-2016)}}. Hundreds of thousands were killed and millions had to flee their land and seek refuge. While I was visiting Palestine after a PhD-related workshop in Jerusalem, 17 Palestinian people were killed by the Israeli army in Gaza during the Land Day\footnote{\href{https://en.wikipedia.org/wiki/Land_Day}{https://en.wikipedia.org/wiki/Land\_Day}} protests. Almost 200 hundred more human beings were killed and thousands were injured by snipers in the weeks thereafter\footnote{\href{https://en.wikipedia.org/wiki/2018-19_Gaza_border_protests}{https://en.wikipedia.org/wiki/2018-19\_Gaza\_border\_protests}}. Also during my PhD, hundreds of thousands of Rohingya\footnote{\href{https://en.wikipedia.org/wiki/Rohingya_refugees_in_Bangladesh}{https://en.wikipedia.org/wiki/Rohingya\_refugees\_in\_Bangladesh}} nationals of Myanmar (formerly Burma) were forced to flee their country due to ethnic and religious persecution. Many people were killed, tortured and thousands of women and girls were raped. An enormous fraction of the population in Africa live below the poverty line, threatened by starvation, conflict, diseases and climate change. Many are forced to cross the continent to seek a safer life in Europe, going through one of the most dangerous migration routes in the world. More than 12,000 human beings have lost their lives in the Mediterranean Sea\footnote{\href{https://data2.unhcr.org/en/situations/mediterranean}{https://data2.unhcr.org/en/situations/mediterranean}} and many more are thought to have died in the Sahara Desert, only since I started my PhD. Around the world, thousands of women are killed by men every year and many live under extremely unsafe conditions\footnote{\href{https://en.wikipedia.org/wiki/Violence_against_women}{https://en.wikipedia.org/wiki/Violence\_against\_women}}. People in these circumstances, regardless of how intelligent and hard-working, can hardly consider the idea of pursuing a PhD. These are only a few well-known, extreme situations around the globe. The list would go much longer, highlighting the privilege of those who are born at the lucky side of the world. I have also been lucky in subtler ways.

First of all, I am a man, and my skin is white. I grew up in Madrid, the capital of a member state of the European Union. This is already privileged with respect to most of the rest of the world, and to most of the rest of my country, Spain. Even though I was not born to a rich family, my parents went to university and became a nurse and a teacher. They always encouraged me and my siblings to study, while giving us the freedom to choose what we wanted. They showed us the pleasure of reading and learning. Not all children, even in the privileged areas of the privileged countries, grow in such positive environments. I have also been lucky that I have not had to go through a separation of my parents. I found supportive friends and had most of the rest of my family close by during my childhood. I worked in the summers to earn some pocket money and certain independence, but I was lucky enough to be able to focus on my studies during the year. Some of my friends could not afford to go to uni. My parents could also afford to cover a year abroad as an Erasmus student. Nothing of this was a merit of mine, but pure \textit{luck}. Without all these positive initial conditions I may not have gained sufficient intellectual and emotional development. Of course I did study and work hard, but many work a lot harder just to survive.

Knowing of my privileged position encourages me to keep the focus and try to make the best out of my time and work. I strongly believe that our everyday actions make a change in the world and that the progress of science will lead us to a fairer planet. This dissertation will make little change, if anything, but the collective effort of many has the potential to make a big impact. May this keep on steering my motivation.
\begin{flushright}
\vspace{5pt}
July 2020
\end{flushright}
\vspace*{\fill}
\endgroup

\phantomsection
\addcontentsline{toc}{chapter}{Abstract}
\begingroup
\vspace*{\fill}
	\begin{center} 
        \Large \textbf{Abstract}
	\vspace{15pt}
	\end{center}
Interdisciplinary research is often at the core of scientific progress. As an example, artificial neural networks, currently an essential tool in many applications that require learning from data, were originally inspired by insights from biological neural networks. Since its inception as a field, the progress of artificial intelligence has at times converged and at times diverged from the field of neuroscience. While occasional divergence can be fruitful, in this dissertation we will explore some advantageous synergies between machine learning, cognitive science and neuroscience.

In particular, this thesis focuses on vision and images. The human visual system has been widely studied from both behavioural and neuroscientific points of view, as vision is the dominant sense of most people. In turn, machine vision has also been an active area of research, currently dominated by the use of artificial neural networks. Despite their origin and some similarities with biological networks, the recent progress in neural networks for image understanding has shown signs of divergence from neuroscience. One likely cause is the focus on benchmark performance, regardless of \textit{what} the models learn. This work focuses instead on \textit{learning representations} that are more aligned with visual perception and the biological vision. For that purpose, I have studied tools and aspects from cognitive science and computational neuroscience, and attempted to incorporate them into machine learning models of vision.

A central subject of this dissertation is data augmentation, a commonly used technique for training artificial neural networks to \textit{augment} the size of data sets through transformations of the images. Although often overlooked, data augmentation implements transformations that are perceptually plausible, since they correspond to the transformations we see in our visual world---changes in viewpoint or illumination, for instance. Furthermore, neuroscientists have found that the brain invariantly represents objects under these transformations. Throughout this dissertation, I use these insights to analyse data augmentation as a particularly useful inductive bias, a more effective regularisation method for artificial neural networks, and as the framework to analyse and improve the invariance of vision models to perceptually plausible transformations. Overall, this work aims to shed more light on the properties of data augmentation and demonstrate the potential of interdisciplinary research.
\par\vspace{30pt}
\noindent\rule{\linewidth}{0.1mm}\vspace{5pt} 
\small{The code produced in this thesis is open source and available at \href{https://github.com/alexhernandezgarcia}{\texttt{www.github.com/alexhernandezgarcia}}}
\vspace*{\fill}
\endgroup

\tableofcontents
\addcontentsline{toc}{chapter}{Contents}
\clearpage

{
\chapter{Introduction}
\label{ch:intro}
\renewcommand{\chapterpath}{includes/intro}
Visual information plays a remarkably prominent role in how humans and many other animals perceive the world. By way of illustration, about 27 \% of the cerebral cortex of humans and 52 \% of the macaque monkey's is devoted to vision \citep{vanessen2003visualcortex}. This involves billions of neurons and connections which give rise to very complex and sophisticated mechanisms, such as visual attention and object recognition. Understanding the way visual information is processed in the brain and how it affects our behaviour is a very active area of research in cognitive science and neuroscience. In turn, developing artificial algorithms that mimic some aspects of biological vision by learning patterns from collections of image data is another active and currently fruitful area of research in machine learning and computer vision. While all these disciplines are rooted in significantly different origins and make use of distinct tools, there are grounds to defend that the interdisciplinary study of these areas of research can highly benefit each other \citep{bengio2015dlandneuroscience, marblestone2016dlandneuroscience, hassabis2017aiandneuroscience, bowers2017pdp, richards2019dlandneuroscience, kietzmann2019dnncompneuro, lindsay2020dlandneuroscience, saxe2020dlandneuroscience}.

In this thesis, we study several aspects of vision and image understanding, such as visual object recognition and visual attention, using the methods and techniques of various disciplines. In particular, we approach machine visual object recognition through deep artificial neural networks, compare some of their properties with the visual cortex, and draw inspiration from visual perception and biological vision to improve the artificial models; we employ eye tracking to analyse the global salience of images when humans look at competing stimuli; and we study the effectiveness of visual salience to identify images from fMRI data. Overall, this dissertation aims at integrating knowledge and methodologies of various disciplines to further advance our understanding of biological and artificial vision.

Summarised, the specific contributions of this thesis are the following: we shed light on the heavily understudied role of data augmentation---transformations of the input data---on training artificial neural networks for image object recognition, and show that it is more effective than the most popular regularisation techniques. Additionally, we found that models trained with heavier image transformations may learn internal representations more aligned with the activations measured in the visual cortex of the brain. Further, we demonstrate how data augmentation can be used to incorporate inductive biases---useful priors---from visual perception and biological vision, in particular the invariance to identity-preserving transformations. Separately, we propose the concept of image global salience, a property of natural images that reflects the likelihood of a stimulus to attract the gaze of a human observer when presented in competition with other stimuli. Finally, we studied the correlation of image salience with the measured activations in the visual cortex.

The interdisciplinary nature of this thesis is rooted in the breadth of my\footnote{Except in the cases where I refer explicitly to me---the author of this thesis---as an individual or intend to express my personal opinion, I will use, in general, the plural form of the first person in order to acknowledge the contribution of some collaborators to certain parts of the work and keep a consistency throughout the whole thesis.} personal interests as well as in the opportunities that the characteristics of my PhD programme offered to me. Having a background on computer science and electrical engineering, already my Bachelor's thesis \citep{hernandez2014bscthesis} addressed the problem of predicting subjective perception from audiovisual content, work continued in my Master's thesis, where I used psychophysical measurements \citep{hernandez2017mscthesis}. My journey towards cognitive science went on when I was admitted as a PhD candidate in the Institute of Cognitive Science of the University of Osnabrück, with Professor Peter König---although in practice I lived and worked in Berlin. The PhD programme also gave me the opportunity to get in touch, for the first time, with neuroscience, as I became a fellow of the Marie Skłodowska-Curie Innovative Training Network ``Training the next generation of European visual neuroscientists" (\href{http://nextgenvis.eu}{NextGenVis\footnote{\href{http://nextgenvis.eu}{www.nextgenvis.eu}}}). In spite of the challenges of being an outlier in the cohort of PhD students as a machine learning scientist, the breadth of my interests certainly expanded. As a mandatory aspect of the Marie Skłodowska-Curie grant, I had to carry out two 2-months internships at laboratories of the partnership or related. This gave me the opportunity to work at the Spinoza Centre for Neuroimaging in Amsterdam with Dr. Serge Dumoulin, where I worked on the project that is described in Chapter~\ref{ch:imageid}; and at the Cognition and Brain Sciences Unit of the University of Cambridge with Dr. Tim Kietzmann, where I started the work that yielded Chapter~\ref{ch:invariance}. I am very greateful to Serge, Tim and Peter for these opportunities.

The breadth of one's interests and work is certainly at odds with depth. However, there is probably not a single sweet spot in the trade-off between breadth and depth valid for all scientists. I understand science as a collaborative ecosystem that is most effective if scientists are distributed across the whole spectrum of the breadth-depth dichotomy. This thesis---and my work so far---is an attempt to explore the interdisciplinary approach to science, in particular to the study of learning systems---artificial and biological---specialised in visual information.

\section{Learning to see}
\label{sec:intro-learning_to_see}
Vision has evolved as one of the most advanced perceptual mechanisms in humans and many other animals, and is the main source of information for most of us to navigate and understand the world around us. Among the multiple high-level perceptual abilities achieved by our visual system, one of the most sophisticated and best studied is visual object recognition: under an enormous variety of conditions, humans can easily distinguish objects of different categories as well as identify individual objects within the same class \citep{logothetis1996objectrecognition}. Being such an important aspect of human perception, automatically finding and identifying objects from digital images has been as well an active area of research in computer vision and a technological challenge for several decades.

The task of image object recognition in computer vision can be defined in its simplest case as the categorisation of an image into one class from a set of pre-defined object classes, according to the main object present in the image by extracting and processing the visual information, that is the pixel values. Object recognition is closely related to other tasks such image object detection \citep{zhao2019objectdetection} and content-based image retrieval \citep{latif2019imageretrieval, zhou2017imageretrieval}. The generalisation of this broader set of of tasks is known as \textit{image understanding}. Ultimately, the goal of artificial image understanding is to extract rich and complex information from a digital image, as close as possible to biological vision and visual perception.

During several decades, object recognition and the related tasks of image understanding were predominantly approached by extracting handcrafted features from the images to then train machine learning classifiers. As a result, much of the computer vision literature was dominated by the proposal and refinement of image processing techniques, such as edge detectors, line and curve detectors like the Hough transform \citep{duda1972hough}, scale-invariant feature descriptors such like SIFT \citep{lowe2004sift} and HOG \citep{dalal2005hog} or Haar-like features \citep{lienhart2002haar}, among many others. These features are usually combined using bag-of-visual-words techniques \citep{sivic2003bog} to train classifiers such as support vector machines (SVM) \citep{cortes1995svm}; or more sophisticated approaches such as Fisher kernels \citep{perronnin2007fisher}. Today, these techniques are informally known as \textit{traditional} machine learning or \textit{traditional} computer vision.

What currently is considered \textit{modern} machine learning are \textit{deep artificial neural networks} (ANN). A major breakthrough in computer image object recognition occurred in 2012 when \citet{krizhevsky2012alexnet} presented results of a neural network algorithm that nearly halved the previous error measure on the large-scale benchmark data set ImageNet \citep{russakovsky2015imagenet}. This attracted an increasing amount of attention towards this kind of algorithms, leading to a rapid development of the subfield of machine learning known as \textit{deep learning}.

ANNs are, however, not exactly \textit{modern}. The history of neural networks in the scientific literature can be traced back at least to the 1940s, when the first theories of biological learning were proposed \citep{mcculloch1943biologicallearning, hebb1949biologicallearning}. Biological learning and biological neural networks inspired the development of artificial models such as the perceptron \citep{rosenblatt1958perceptron} and later the Neocognitron \citep{fukushima1982neocognitron}---a precursor of current convolutional neural networks---directly inspired by the findings by \citet{hubelwiesel1959} about the receptive fields of neurons in the cat's visual cortex. ANNs gained some popularity in the 1980s and 1990s with the development and application of the \textit{backpropagation} principle \citep{rumelhart1986backprop} and other successful proposals, such as the long-short term memory (LSTM) recurrent neural networks \citep{hochreiter1997lstm}. Nonetheless, the progress and adoption of neural networks until 2012 was slow and minor, overshadowed by other algorithms such as the SVM.

One of the main differences---and arguably, advantages---of deep learning with respect to traditional machine learning algorithms is that they are able to extract relevant features for a certain task, such as image object recognition, more directly from the data. This allows to apply similar learning principles, techniques and architectures to a more general set of data modalities and tasks. This can be regarded as a step towards finding a general learning principle, a hypothesis suggested in biological learning, upon the observation of the highly regular structure found in the neocortex across different brain areas and species \citep{douglas1989neocortex, harris2015neocortex}. This \textit{philosophy} of letting an algorithm learn the relevant features---\textit{representation learning}---from the available data can be regarded as the other end of \textit{symbolic artificial intelligence} (AI) \citep{haugeland1989ai}, whose approach was to programme an agent to perform certain tasks by directly manipulating symbols and manually implementing rules to simulate some behaviour. The traditional machine learning and computer vision algorithms used for many years would be somewhere in between symbolic AI and the philosophy of deep learning.

Despite the significant practical and conceptual step forward originated by the recent progress and adoption of deep learning, the claims or beliefs that ANNs are or will be able to automatically learn anything from data without human intervention are overstated. Naturally, artificial neural networks are also subject to the \textit{no free lunch theorem}: no learning algorithm is better than any other at classifying unobserved data points, when averaged over all possible data distributions \citep{wolpert1996nofreelunch}. In other words, in order to learn anything about the world it is necessary to introduce some \textit{priors} or \textit{inductive biases} about the world.

In this light, although it is often claimed that deep learning is a shift away from the traditional approach of hand-engineering features to ``automatically'' learning them with ``a general-purpose learning procedure'' \citep{lecun2015deeplearningnature}, it can also be seen as a shift of inductive biases. The priors used by deep learning are definitely \textit{more} general-purpose than those used in traditional machine learning, but neural networks do not remove the need for inductive biases---and they will never do. While a few years ago we handcrafted visual descriptors that could discriminate object classes in a set of images, we now handcraft types of layers, architectures, parameter initialisation strategies, regularisers and optimisation methods, to name a few elements of deep learning\footnote{Even approaches such as neural architecture search \citep{zoph2016nas}, which aim at further automating machine learning, not only use a tremendous amount of computational resources---with a proportional environmental impact---but their efficacy has been questioned, as reviewed by \citet{gencoglu2019hark}, among others.}. The current success of deep learning is the result of a collective effort in exploring the combinations of these elements that work best for different tasks, presented in a large body of scientific literature.

We argue that the new landscape opened by deep learning is undoubtedly a significant step towards a more natural and general way of processing data, especially because it allows to train learning algorithms almost end-to-end from nearly naturalistic sensory signals, such as digital images or sound \citep{saxe2020dlandneuroscience}. However, we should not neglect the need for searching better inductive biases, that is incorporating prior information about the tasks we want to solve, since without it learning would be much less efficient or not possible at all. We know this from statistical learning theory, but also from the innate genetic inductive biases provided by evolution in nature, which seem to predispose organisms to quickly learn and adapt to their specific environment \citep{zador2019purelearning}. Hence, studying biological learning systems seems like a natural approach to draw inspiration for improving artificial learning algorithms \citep{hassabis2017aiandneuroscience, nayebi2017bioconstraints, lindsay2018bioconstraints, malhotra2020bioconstraints}.

A key pair of ingredients in the development and success of deep learning was the increase in available computational power, alongside the publication of large data sets. The comparably much smaller data sets and reduced computational resources several decades ago has been argued to be one of the reasons why researchers could not prove the capabilities of ANNs until recently. Compared to other machine learning models, neural networks excel at learning from large data sets, but in some cases require much and specialised computer power, like graphical processing units (GPU). In computer vision specifically, the efforts in hand-labelling large data sets in the last decades set the grounds for deep learning to unleash its potential. Some of these data sets, which we have used in the experiments presented in this thesis, are MNIST, 70,000 small greyscale images of digits \citep{lecun1998mnist}; CIFAR-10 and CIFAR-100, 60,000 small colour images of objects from 10 and 100 classes, respectively \citep{krizhevsky2009cifar}; and ImageNet (ILSVRC 2012), 1.3 million high-resolution natural images of 1,000 object classes \citep{russakovsky2015imagenet}.

With the availability of these benchmark data sets to train and compare different algorithms, most of the research efforts in the last years have been used to improve different aspects of the neural network architectures and the training process: parameter initialisation strategies \citep{glorot2010glorot, he2015he}, activation functions \citep{glorot2011relu}, normalisation layers \citep{ioffe2015batchnorm}, stochastic optimisation methods \citep{duchi2011adagrad, kingma2014adam}, network architectures \citep{springenberg2014allcnn, simonyan2014vgg, he2016resnet, huang2017densenet}, and so on. The collaborative effort of many researchers has been essential to develop these and many other methods that have advanced the performance and our understanding of deep neural networks. Meanwhile, beyond the creation and publication of data sets, little attention has been paid to the data, which was considered fixed and given, since the goal is generally to improve the state of the art results on the common benchmark data sets. However, we have noted that the availability of data was key for the success of deep learning in image object recognition.

The need for larger amounts of labelled examples to train neural networks led some researchers to create data sets, and also led many to think of ways of easily create new examples. A straightforward way to extend a data set, especially in the image domain, is through \textit{data augmentation}. Data augmentation, broadly defined, consists of synthetically expanding a data set by applying transformations on the available examples that preserve the ground truth labels. Data augmentation has been used in image object recognition at least since the 1990s \citep{abumostafa1990hints, simard1992daug}; has been identified as critical component of many successful models, such as AlexNet \citep{krizhevsky2012alexnet}; and is ubiquitous in both deep learning research and application. Nonetheless, despite its popularity, data augmentation has surprisingly remained a largely understudied component of machine learning: many in the field use it, know that it helps generalisation and intuitively understand why, but little is known about its interaction with other techniques or whether its actual potential goes beyond simply increasing the number of training examples.

A significant part of this thesis aims at filling this gap and shedding light on the role of data augmentation for visual object recognition with deep neural networks. We argue that data augmentation has been heavily understudied because it has been looked down on by the machine learning community, regarded almost as a \textit{cheating} technique, rather than as a method that deserves analysis and attention. In this dissertation, we contend that data augmentation is interesting beyond simply providing new examples. We analyse data augmentation as regularisation, but draw significant differences with respect to classical regularisation. We also analyse the potential of data augmentation for training models that learn robust visual representations, taking inspiration from properties of the visual cortex. Overall, we here aim to explore the connections between data augmentation, visual perception and biological vision.

\section{Why data augmentation?}
Having a background in image processing and having carried out research projects with traditional, handcrafted visual descriptors \citep{hernandez2017mscthesis}, when I first started training neural networks with image data at the beginning of my PhD, it was the most natural approach for me to apply transformations on the images in my data sets to get some more data points \textit{for free}. As I was learning about deep learning, I became increasingly interested in the generalisation properties of neural networks and in regularisation methods. In my toy experiments, I noticed that while the most popular regularisation methods, that is weight decay and dropout, did improve the test performance of the models, the true boost in generalisation seemed to be provided by the image transformations that I had coded, that is data augmentation. This made sense to me: we know from statistical learning theory that generalisation can improve by finding the right complexity of the model, that is by accurately tuning the regularisation; but generalisation should always improve with more training examples (see Section~\ref{sec:background-generalisation}).

I became even more intrigued about these observations and got additional insights after reading ``Understanding deep learning requires rethinking regularization'', by \citet{zhang2016understandingdl}. This article, among other results, included the performance of a few models trained with and without weight decay, dropout and data augmentation. The superiority of data augmentation was also apparent by carefully analysing the results in the tables, but this fact is nearly ignored in the paper, since all three methods---weight decay, dropout and data augmentation---are considered the same type of regularisation (see Chapter~\ref{ch:daugreg}). I had the chance to chat with Dr. Chiyuan Zhang at his poster presentation at the International Conference on Learning Representations in 2017. While he was arguing how the generalisation bounds based on the Rademacher complexity (see Section~\ref{sec:background-rademacher}) may not explain the generalisation performance of deep neural networks, I asked what would the $n$---the number of training examples---be in the formula if they trained with data augmentation. Of course there is no straightforward answer to that question, so that got me thinking.

Since data augmentation was so remarkably effective in improving the test performance of neural networks, I assumed there should be literature on the topic that explained how exactly data augmentation affects generalisation and empirical comparisons with other regularisation methods. Unfortunately, I could not find much, except than corroborating that everybody seemed to use data augmentation in their papers to push the performance of their models towards the state of the art \citep{krizhevsky2012alexnet, springenberg2014allcnn}. I could think of three types of reasons for this lack of literature on data augmentation: One, ``it is \textit{obvious} that data augmentation is beneficial''. Two, ``it is \textit{too complicated} to study''. Three, ``it is \textit{not interesting}''. The first reason was not very satisfying, scientifically. The second one was actually a reason for me to try to shed some light. While it is probably a mix of all, I think the third point was the actual reason why data augmentation was disregarded: As mentioned in Section~\ref{sec:intro-learning_to_see}, the philosophy of the re-emerging deep learning field was to let a model learn ``good features \dots automatically using a general-purpose learning procedure'' \citep{lecun2015deeplearningnature}. Handcrafting some image transformations to augment a data set seemed closer to the \textit{old-fashioned} traditional machine learning methods and against the deep learning philosophy. As a matter of fact, data augmentation seemed to be considered \textit{cheating} and many papers would include results of their new method or architecture \textit{without} data augmentation to show that it actually worked \citep{goodfellow2013maxout, graham2014fracmaxpool, larsson2016fractalnet}---this ablation studies were however not carried out with other techniques, such as weight decay or dropout. 

The fact that data augmentation increases the number of training examples, even though breaking the assumption of independent and identically distributed sampling, is only the most obvious advantage: the tip of the iceberg. In what follows, we will outline the interpretation of data augmentation as a remarkably useful inductive bias directly connected to visual perception, and make explicit some connections with properties of the visual cortex.

In the application of deep learning we have considered so far, image understanding and, in particular, object recognition, it is easy to get stuck in the specific goal of the task at hand that is often performed: to obtain a high classification performance in the test set of the benchmark data set. However, it is always useful to take a step back to see the forest for the trees. The objective for, at least, the research community is not to incrementally improve the state of the art performance on the benchmark data sets, but to truly develop good models of image understanding. Furthermore, when we talk about object recognition, in general we do not mean object recognition for any arbitrary visual system or modality of the many in nature, but we mean \textit{human} object recognition, that is recognising the object classes that are relevant for humans, from photos or videos that actually resemble how we perceive the world, that is natural images\footnote{Some particular subfields of computer vision are devoted to non-human vision, for example multispectral imagery \citep{audebert2019multispectral}.}.

Since we are interested in human object recognition, we argue that we should always keep an eye on what we know from visual perception and, ideally, biological vision too. The transformations that are most commonly applied in data augmentation schemes are \textit{perceptually plausible}: the resulting image preserves the properties of the perceived visual world as well as, in the case of object recognition, the object class\footnote{Other types of transformations in which perceptual plausibility is not preserved have been successfully used in computer vision. One popular example is \textit{mixup} \citep{zhang2017mixup}, which performs the weighted average of the pixels of two images---and their labels. Another example is data augmentation in feature space \citep{devries2017daugfeatspace}. However, in this thesis we are interested in exploring connections between machine learning and visual perception and thus we will consider only perceptually plausible image transformations.}. Especially in the image domain, it is straightforward to identify a large number of perceptually plausible transformations---equivariant transformations, some colour adjustments, blurring, etc. Some examples are shown in Figure~\ref{fig:intro-daug_imagenet}---and it is well-known since decades ago how to apply them to digital images \citep{gonzalez2018imageprocessing}. Hence, the combination of having tools for performing the transformations and expert domain knowledge---visual perception---provides us with a large-capacity generator of new examples and an approximate \textit{oracle} of the target function, for a relevant subset of the input space. 

The access to an oracle of the target function serves as a remarkably effective inductive bias, which is at the very essence of machine learning. Data augmentation exploits this oracle, constructed from our knowledge about perception to densely populate the relevant regions of the high-dimensional input space. That is, the different views of the objects that are perceptually plausible. \citet{richards2019dlandneuroscience} argue that in order to train neural networks ``the three components specified by design are the objective functions, the learning rules and the architectures''. Throughout this dissertation we will argue that incorporating inductive biases through the data, possibly in combination with the objective function (Chapter~\ref{ch:invariance}), is another effective way to learn better, more robust representations.

\begin{figure}[htb]
  \begin{center}
    \includegraphics[width = \linewidth]{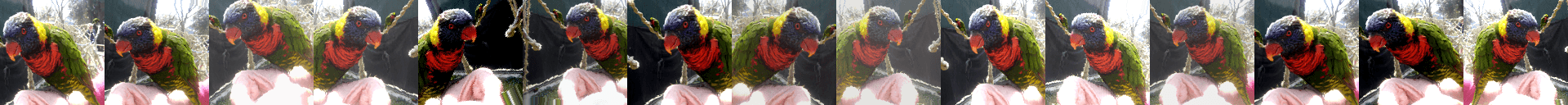}
  \end{center}
  \caption{Some examples of perceptually plausible image transformations applied on the same image, from ImageNet.}
\label{fig:intro-daug_imagenet}
\end{figure}

\section{Rethinking supervised learning}
\label{sec:intro-rethinking_supervised}
The re-emergence of deep learning during the last decade due to the noteworthy achievements of artificial neural networks built up a sort of philosophy that anything could be automatically learnt from data without human intervention, in contrast to the previous approach that required, indeed, a higher degree of manual design. However, this promise is misleading, since it misses the fact that the success of deep networks has been partly due to the human effort of manually labelling thousands of images and other data modalities \citep{russakovsky2015imagenet, cao2018vggface2}. As a matter of fact, the need for much data has been at the core of some strong criticisms of deep learning \citep{marcus2018critiquedl}. We argue instead that neither is deep learning some sort of exceptional solution to learn without inductive biases, nor is it a hopeless model class because it requires large data sets. On the contrary, we contend that the competitive advantage of artificial neural networks is that they are deep universal function approximators \citep{hornik1991functionapproximation} that have been proven to \textit{be able to} learn from large collections of data. While other models are known to scale poorly as the amount of training data increases, ANNs excel at fitting the training data and applying interpolation on unseen examples \citep{belkin2019biasvariance, hasson2020directfit}. This is a feature, not a bug. We will make faster progress if we exploit the advantages of deep learning, while being aware of its limitations.

Data augmentation, as we have seen, is an effective way of using this property: it generates examples in the regions of the input space where the model should learn how to interpolate. A commonly seen argument to defend that this should not be necessary and neural networks should be able to generalise from few examples is that humans and other animals learn---to categorise objects, for example---with very little supervision \citep{vinyals2016oneshot, marcus2018critiquedl, morgenstern2019oneshot, zador2019purelearning}. In this section, we will elaborate on our views on why we think this may be a misconception that can lead us astray and how we can benefit from rethinking the notions of supervised and unsupervised learning. We will do so by taking a look into learning theory, visual perception and biological vision and how these ideas relate to data augmentation and the contributions of this dissertation.

\subsection{Supervised biological learning}
The comparison between artificial intelligence---specifically artificial neural networks---and biological learning systems is intrinsic to the field, since one long-term goal of artificial intelligence is to mirror the capabilities of human intelligence. However, these capabilities are sometimes overestimated. One example is the argument that intelligence in nature evolves without supervision. In what follows, we will discuss three aspects of biological learning to argue against this view, in order to gain insights that better inform our progress in machine learning: first, we will discuss how generalisation requires exposure to relevant \textit{training data}; second, the role of evolution and brain development; third, the variety of supervised signals that the brain may use. 

First, in the argument that machine learning models should generalise from a few examples, there seems to be a promise or aspiration that with future better methods it will be possible for a neural network---for instance---to perform robust visual object categorisation among many object classes after being trained on one or a few examples per class. While we agree that a challenge for the near future is to \textit{more efficiently} learn from fewer examples, we should also remind ourselves that no machine learning algorithm can robustly learn anything that cannot be inferred from the data it has been trained on. While this may seem to be against the ultimate goal of deep learning and a reason to look for radically different approaches, we should also bear in mind that learning in nature is not too different. 

Even though the human visual system is remarkably robust, its capabilities are optimised for the tasks it needs to perform and largely shaped by experience, that is the \textit{training data}---and years of evolution \citep{hasson2020directfit}, as we will discuss below. For instance, from the literature on human visual perception, we know that object recognition is viewpoint dependent \citep{tarr1998viewpointdependence}. A well-studied property of human vision is that our face recognition ability is severely impaired if faces are presented upside down \citep{yin1969invertedfaces, valentine1988invertedfaces}. Setting aside the specific complexity of face processing in the brain, a compelling explanation for this impairment is that we are simply not used to seeing and recognising inverted faces. More generally, human perception of objects and our recognition ability is greatly affected when we see objects from unfamiliar viewpoints \citep{edelman1992viewpointdependence, tarr1998viewpointdependence, bulthoff2006viewpointdependence, milivojevic2012viewpointdependence}. Furthermore, although better than the \textit{one-shot} or \textit{few-shot} generalisation of current ANNs, humans also have limited ability to recognise novel classes \citep{morgenstern2019oneshot}. Interestingly, experiments with some novel classes of objects, known as Greebles, showed that with sufficient experience and training, humans can acquire expertise in recognising new objects from different viewpoints, even making use of an area of the brain---the fusiform face area---that typically responds strongly with face stimuli \citep{gauthier1999greebles}. This provides evidence that \textit{generalisation} to multiple viewpoints is only developed after exposure to similar conditions. In this regard, data augmentation seems like a straightforward way to provide certain degree of input variability.

Second, the commonplace comparison of artificial neural networks with the brain often misses a fundamental component of biology, recently brought to the fore by \citet{zador2019purelearning} and \citet{hasson2020directfit}, although considered since the early days of artificial intelligence \citep{turing1948intelligentmachinery}: the role that millions of years of evolution have played in developing the nervous systems of organisms in nature, including the human brain. For example, some similarities have been observed between the representational geometry of the internal features learnt by ANNs trained for visual object recognition and that of the neural activations measured in the visual cortex of the brain \citep{yamins2014annsbrains, khaligh2014annbrains, gucclu2015annbrains}. These articles were followed up by numerous studies that postulate ANNs as models of the visual cortex. However, what exactly drives better similarity with the brain remains an open question and this line of research presents several challenges. 

While artificial neural networks are typically trained from scratch, from random initialisation, the neural activations are often measured in the adult brain. A reasonable approach would be to at least consider insights from developmental neuroscience and the infant brain \citep{harwerth1986criticalperiods, atkinson2002developmental, gelman2011childcategorization}. Moreover, as mentioned above, an interesting avenue is to also take into account the role of evolution. A large part of the brain connectivity is encoded genetically, and some properties of the visual cortex are known to be innate, that is without prior exposure to visual stimuli \citep{zador2019purelearning}. Since neural networks are expected to learn some of these properties through extensive training on image data sets from scratch\footnote{Some interesting and promising areas in machine learning research deviate from this standard approach. For example, transfer learning studies and domain adaptation study the potential of features learnt on one task to be reused in different, related tasks \citep{zhuang2019transferlearning}, and continual learning studies the ways in which machine learning models can indefinitely sustain the acquisition of new knowledge without detriment of the previously learnt tasks \citep{aljundi2019continuallearning, mundt2019continuallearning}. These approaches are inspired by biological learning or share interesting properties with it.}, part of the artificial learning process may have more to do with evolution than with the visual learning capabilities of an adult brain. This may be another reason to temper the expectations that neural networks be able to learn from a few examples, without \textit{hard-wiring} some of the innate properties of the brain \citep{lindsey2019bioconstraints, malhotra2020bioconstraints} or, alternatively, simulating part of the evolutionary process.
%
%

Third, another commonly found argument has it that children, and other animals in general, learn robust object recognition without supervision. First of all, we should recall again the role of evolution, which can be interpreted as a pre-trained model, optimised through millions of years of data with natural selection as a supervised signal \citep{zador2019purelearning}. Second, we will argue against the very claim that children learn in fully unsupervised fashion. Obviously, the kind of supervision that humans make use of is not identical to that of machine learning algorithms: we do not see a class label on top of every object we look at. However, we receive supervision from multiple sources. Even though not for every visual stimulus, children do frequently receive information about the object classes they see---parents would point at objects and name them, for instance. Furthermore, humans usually follow a guided hierarchical learning: children do not directly learn to tell apart breeds of dogs, but rather start with umbrella terms and then progressively learn down the class hierarchy \citep{bornstein2010hierarchychildren}. \citet{hasson2020directfit} mention other examples of supervision from \textit{social cues}, that is from other humans, such as learning to recognise individual faces, produce grammatical sentences, read and write; as well as from embodiment and action, such as learning to balance the body while walking or grasping objects. In all these actions, we can identify a supervised signal that surely influences learning in the brain \citep{shapiro2019embodied}.

Moreover, besides this kind of explicit supervision, the brain surely makes use of more subtle, implicit supervised signals, such as temporal stability \citep{becker1999temporalstability, wyss2003temporalstability}: The light that enters the retina is not a random signal from a sequence of rapidly changing arbitrary photos, but a highly coherent and regular flow of slowly changing stimuli, especially at the higher, semantical level \citep{kording2004complexcells}. At the very least, this is how we perceive it and if such a smooth perception is a consequence instead of a cause, then it should be, again, a by-product of a long process of evolution. In the light of these insights from evolutionary theory and the explicit and implicit supervision that drives biological learning, we argue that we should temper the claims and aspirations that artificial neural networks should learn without supervision and from very few examples. Instead, we may benefit from rethinking the concept of supervision, embrace it and try to incorporate the forms of supervision present in nature into machine learning algorithms.

\subsection{Supervised machine learning}
If we open a machine learning textbook \citep{murphy2012machinelearning, abu2012learningfromdata, goodfellow2016dlbook}, we will most surely find a taxonomy of learning algorithms with a clear distinction between \textit{supervised} and \textit{unsupervised} learning. If we take a look at the deep learning literature of the past years, we will also find abundant work on some variants \textit{in between}: semi-supervised learning, self-supervised learning, etc. However, while this taxonomy can be useful, the boundaries are certainly not clear. As a matter of fact, strictly speaking, unsupervised learning is an illusion. If we recall the \textit{no free lunch} theorem \citep{wolpert1996nofreelunch}, averaged over all possible distributions, all classification algorithms are equivalent. Therefore, we need to constrain the distributions or, in other words, introduce prior knowledge---that is \textit{supervision}. Recently, \citet{locatello2018disentanglement} obtained a similar result for the case of unsupervised learning of disentangled representations: without inductive biases for both the models and the data sets, unsupervised disentanglement learning is impossible. These results are purely theoretical and do not hold in real world applications, precisely because in practice we use multiple inductive biases, even when we do so-called unsupervised learning.  

In a strict sense, even the classical, \textit{purely} unsupervised methods, such as independent component analysis or nearest neighbours classifiers, make use of priors, such as independence or distance, respectively. Without inductive bias, learning is not possible. Consider the data points in Figure~\ref{fig:unsupervised} (middle). With no prior information, all possible point categorisations are possible and equally valid. Depending on the inductive bias used, one model may find the configuration on the left, on the right, or any other. Which one is better depends on the task.

\begin{figure}[htb]
  \centering
  \begin{subfigure}{0.32 \linewidth}
      \includegraphics[keepaspectratio=true, width=\linewidth]{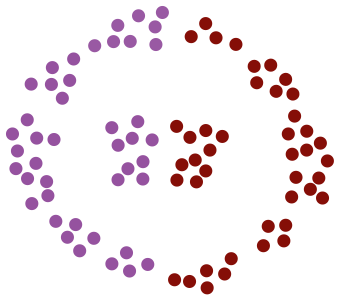}
      \caption{One possible categorisation}
      \label{fig:lateral}
  \end{subfigure}
  \begin{subfigure}{0.32 \linewidth}
      \includegraphics[keepaspectratio=true, width=\linewidth]{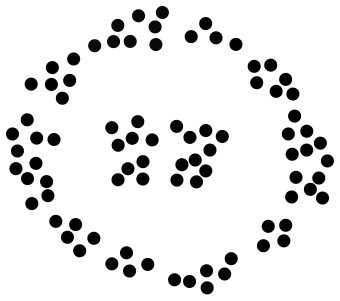}
      \caption{A set of data points}
      \label{fig:allblack}
  \end{subfigure}
  \begin{subfigure}{0.32 \linewidth}
      \includegraphics[keepaspectratio=true, width=\linewidth]{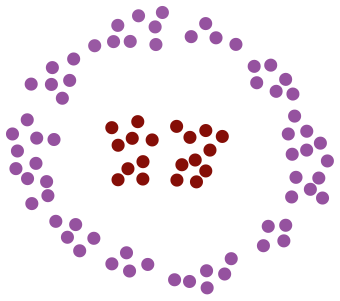}
      \caption{One possible categorisation}
      \label{fig:concentric}
  \end{subfigure}
  \caption{An illustration of the need for inductive biases. Without any prior knowledge about the objective, any clustering of the data points is valid and can be potentially realised by a learning algorithm}
  \label{fig:unsupervised}
\end{figure}

Although this is not news, the terminology used in the machine learning literature seems to neglect these nuances and evidences that the field suffers from \textit{catastrophic forgetting}\footnote{I am borrowing the expression from Prof. Irina Rish, who used it at a panel discussion of the UNIQUE Student Symposium 2020.}. Particularly in deep learning and computer vision, the term \textit{supervised} learning is often used to actually refer to \textit{classification}, that is models trained on examples labelled according to the object classes, for instance. In turn, \textit{unsupervised} learning is used for any model that does not use the labels, regardless of what other kind of supervision may be used. Further, the term \textit{semi-supervised} learning refers to models that are trained with a fraction of the labels. While this terminology may be useful in some cases, it is not well defined and misses the fact that supervision can come in multiple flavours, not only as classification labels.

We have seen examples of different forms of supervision used by humans and other animals. What forms of supervision are common in machine learning? The relatively recent explosion of deep learning has brought about the development of several libraries for automatic differentiation\footnote{Automatic differentiation is also known as algorithmic differentiation and differentiable programming, among other names. It is the set of techniques that allows to calculate the derivatives of numeric functions algorithmically. The development of automatic differentiation libraries has played a key role in the progress of deep learning in the past 10--15 years. Examples are Theano \citep{theano2016}, TensorFlow \citep{tensorflow2015} and PyTorch \citep{pytorch2019}.} \citep{baydin2017automaticdifferentiation}, which in turn have enabled the proposal of multiple loss functions with other types of supervision that can effectively be optimised by artificial neural networks through backpropagation and stochastic gradient descent. However, these approaches are often termed in the literature \textit{semi-supervised}, \textit{self-supervised} or even \textit{unsupervised} learning. The term \textit{semi-supervised} learning has gained popularity recently \citep{jing2020selfsupervised}. On the one hand, this term acknowledges the fact that supervision is used---as opposed to unsupervised learning. On the other hand, it draws a hard line between classification and the other forms of supervised learning. The most likely reason for this separation is that labels are more costly to obtain than defining and implementing the tasks of self-supervised learning, rather than a formal, conceptual reason. From a theoretical point of view, both the conventional classification models and the recent wave of self-supervised objectives can be formalised within the category of supervised learning.

Importantly, the bulk of statistical learning theory (see a brief review in Section~\ref{sec:background-generalisation}) has been developed for binary classification loss functions, then extended for multiclass classification, and in part for regression loss functions such as the mean squared error. Critically, the mapping of many results from statistical learning theory onto the various objective functions used in semi-supervised learning is far from trivial. Given the success of this kind of supervised objectives and their connection with perception, the study of these methods from a theoretical point of view might be a fruitful direction for future work.

\section{Data augmentation, regularisation and inductive biases}
After discussing the conflict in the terminology and acknowledging that purely unsupervised learning is an illusion, we can return to the role of inductive biases from visual perception and biological vision in defining useful forms of supervision for training artificial neural networks and, in particular, the role of data augmentation. 

If we consider the particular, well-studied case of classification, to which the problem of visual object categorisation belongs, we can recall again the well-known \textit{no free lunch} theorem, which establishes that learning is not possible without any prior. The field of statistical learning theory, whose fundamental results we review in Section~\ref{sec:background-generalisation}, studies the conditions that make learning from data possible. One of its key results is that the space of possible solutions where an algorithm can search for a solution, the hypothesis set, has to be finite \citep{vapnik1971vc}. Otherwise, the problem of inferring a function from a finite set of data is ill-posed. The classical way to ensure the problem is well-posed is through regularisation \citep{phillips1962regularisation, tikhonov1963regularisation, ivanov1976regularisation}. Essentially, regularisation imposes a constraint on the hypothesis set---in the classical sense, a smoothness constraint---so that the inferred function cannot vary too rapidly around the training data points (see Section~\ref{sec:background-regularisation} for more details). Therefore, we can think of regularisation as an inductive bias. 

Being such an essential ingredient of the learning problem, regularisation has been widely studied and multiple forms of regularisation have been proposed. Two ubiquitously used regularisation techniques in deep learning are weight decay and dropout (reviewed in Section~\ref{sec:background-regularisation}). The former is a classical constraint on the norm of the learnable weights. The latter is a procedure to randomly turn off a subset of neurons during training. Both techniques have been shown to improve the generalisation of ANNs on the test data and hence their wide adoption. Thinking of regularisation techniques as inductive biases allows us to analyse the type of prior knowledge they incorporate, as well as widen the notion of regularisation to include other methods, such as data augmentation.

One of the contributions of this thesis is the comparison of weight decay, dropout and data augmentation. In Chapter~\ref{ch:daugreg}, we show that neural networks trained on image object recognition tasks \textit{without} weight decay and dropout achieve better performance than their counterparts, provided data augmentation is used during training. However, the common practice is to include all three: weight decay, dropout and data augmentation, among other methods. If we think of the inductive biases each technique introduces, data augmentation seems intuitively more advantageous: Weight decay assumes that models with smaller parameters should generalise better. Dropout assumes that neural networks that are forced to perform well with a subset of the neurons should perform better when all the neurons are used. Data augmentation makes use of an approximation of the oracle function, derived from prior knowledge about visual perception, that generates examples in regions of the input space where the model should learn a mapping. Although the inductive biases introduced by weight decay and dropout are clearly beneficial, the contribution of data augmentation seems more powerful. Nonetheless, these results were received with scepticism by the machine learning community.

In order to shed more light on the debate, we also draw theoretical insights from statistical learning theory that support the empirical findings and provide a grounded distinction between explicit regularisation methods, to which weight decay and dropout belong, and the implicit regularisation effect provided by data augmentation and other methods (Chapters~\ref{ch:reg}~and~\ref{ch:daugreg}). Explicit regularisation methods operate by directly constraining the hypothesis set of the model, known as representational capacity. However, many other methods that cannot be considered explicit regularisation provide an implicit regularisation effect, since they improve generalisation. Drawing a connection with the previously discussed ideas about supervision and inductive biases, we conclude that while explicit regularisation may indeed improve generalisation, it generally involves the optimisation of sensitive hyperparameters and at least the same or even better returns can be obtained by exploring ways of incorporating more meaningful inductive biases from visual perception and biological vision.

\section{Invariance}
In search of better inductive biases, we subscribe to the increasing trend in the machine community of exploring ways of training artificial neural networks beyond classification\footnote{Note we use the term \textit{classification}, not \textit{supervised learning}, after the discussion in Section~\ref{sec:intro-rethinking_supervised}.}. A pillar of this thesis is the attempt to integrate different disciplines. Therefore, in order to look for ways of improving visual object recognition models, we searched for inspiration in the mechanisms the brain has evolved in the visual cortex to solve object recognition.

The visual cortex is the part of the brain that processes visual information. One of its fundamental properties is the hierarchical organisation: while the primary areas of the visual cortex, which first process the information from the retina, respond to low-level properties such as the location in the visual field and the orientation of small parts of the stimuli \citep{hubelwiesel1962}, the inferior temporal (IT) cortex, later in the processing pipeline, responds to higher-level properties such as the object category of the stimulus \citep{gross1972it}. This organisation of the biological networks in the brain greatly inspired the development of artificial neural networks.

Related to the hierarchical organisation of the visual cortex, \citet{desimone1984invariantitmacaque} found that some neurons in the inferior temporal cortex of the macaque monkey responded consistently to the presentation of the same faces, regardless of their size and position. In contrast, neurons in earlier areas of the visual cortex are very sensitive to small changes in low-level properties of the visual stimuli, such as the orientation of edges. This invariance property was found to generalise to other objects besides faces \citep{booth1998invariantitmacaque} in the late 1990s and in the 2000s was observed in the human brain too \citep{quiroga2005invariantithuman}. Importantly, the invariance to identity-preserving transformations has been proposed to be an essential ingredient of robust visual object recognition in the brain \citep{dicarlo2007untangling, tacchetti2018invariance}. Hence, a reasonable question is whether artificial neural networks trained for visual object recognition are also invariant to such transformations.

The question of invariance in artificial neural networks has been addressed almost since their inception and studied from multiple perspectives. A large body of work has aimed at encoding different types of transformations into the networks (see a short review by \citet{cohen2016groupequivcnns}), such as translation or rotation invariance. One contribution of this thesis is the study of the invariance of ANNs towards identity-preserving transformations using data augmentation. The use of data augmentation is convenient: Not coincidentally, the kind of stimulus transformations tested by computational neuroscientists to study the invariance to identity-preserving transformations in the inferior temporal cortex and the image transformations typically included in data augmentation schemes are the same, or very similar. These are the transformations that we encounter naturally in the visual world, as we perceive it along the temporal dimension \citep{kording2004complexcells, einhauser2005viewpointinvariance, taylor2011temporalstability}, which perceptually preserve the object identity. For this reason we have named them \textit{perceptually plausible} transformations.

First, we measured the invariance to identity-preserving transformations of models trained on image object recognition data sets (Chapter~\ref{ch:invariance}). Intuitively and also taking the insights from the properties of the IT cortex, two images that represent different views of the same object should produce similar activations at the higher layers of a neural network. However, we found that the similarity is not better than at the pixel space, even though the models do classify the images correctly and were exposed to such transformations during training. This finding contradicts in part the general intuition that neural networks learn hierarchical representations, ranging from specific to more abstract, object-related features. We hypothesised that this is a sign of a lack of perceptual inductive bias. Given the large representational capacity of neural networks, training them with the sole objective of classifying a data set of images into the right classes does not seem enough to learn perceptually invariant representations, despite the theoretical results showing that invariance to \textit{nuisances} should emerge naturally \citep{achille2018emergence}. In general, there exist multiple possible solutions for the classification problem within the hypothesis space spanned by modern deep neural networks and the models do not seem to naturally converge to solutions well aligned with some crucial aspects of visual perception and biological vision \citep{sinz2019dlvsbrain, geirhos2020shortcutlearning, dujmovic2020adversarial}. This led us to propose \textit{data augmentation invariance}, an objective function that encourages robust representations, inspired by the invariance observed in the visual cortex.

In Chapter~\ref{ch:invariance}, we discuss the details of data augmentation invariance. We trained artificial neural networks on image object recognition data sets by jointly optimising the categorisation objective and a new data augmentation invariance objective. The latter is a layer-wise objective that encourages that the representations of transformations of the same image---generated through perceptually plausible data augmentation---cluster together. We attempted to simulate the increasing invariance along the visual cortex hierarchy by distributing the invariance loss exponentially along the neural network layers. Models trained with data augmentation invariance effectively learnt increasingly invariant representations without detriment to the classification accuracy, which even improved in some cases. 

In view of these results we argue that replacing or complementing the standard classification objectives with perceptually and biologically inspired objectives, such as data augmentation invariance, is a promising avenue to both improve computer vision algorithms and obtain better models of natural vision. Such objectives are likely not biologically plausible, in the sense that the brain does not optimise an equivalent objective \citep{pozzi2018biologicallyplausible}. However, since properties like invariance to identity-preserving transformations are at least a by-product of either evolution or early brain development (or both), it is reasonable and potentially fruitful to optimise artificial neural networks with objectives that simulate key properties of the brain.
\section{Visual salience}
So far our discussion around visual perception, biological vision and machine image understanding algorithms has revolved mainly around object recognition. However, animal vision encompasses a broader range of capabilities that allow us to navigate and understand the world. As part of the interdisciplinary pursuit of this work, we here studied some aspects of another central component of vision: visual attention.

Visual attention is a complex brain mechanism that enables us to coherently process the sheer amount of light that enters our eyes. At any given time, even though the retina receives stimulation from the whole visual field, only a small fraction of the information is processed in detail \citep{desimone1995visualattention}. Specifically, the visual system preferentially processes the information located in the centre of the visual field---the \textit{fovea}---as the central part of the retina has a larger density of photoreceptors than in the surroundings---the \textit{visual periphery} \citep{wassle1990fovea, azzopardi1993fovea}. For instance, a recent study has shown that many people fail to notice when up to 95 \% of the visual field is presented without colour \citep{cohen2020colour}. What particular area of the available information in front of us is processed in most detail at a time is mediated by eye movements, and what exactly drives eyes movement is a complex, widely studied question, which remains largely open.

For example, it is known that eye movements can be driven by both low-level properties of the stimuli---\textit{bottom-up}---and by cognitive processes derived from, for instance, a task or desire---\textit{top-down} \citep{vonstein2000topdown, munoz2004bottomup, connor2004buttomuptopdown, betz2010topdown, kollmorgen2010topdownbottomup, schutt2019bottomuptopdown}. An interesting subject of study is the relationship between object recognition and visual attention. There is strong evidence for the role of object recognition in the direction of eye movements \citep{zhaoping2007topdown}, but visual attention has been also suggested to predict object perceptual awareness \citep{holm2008attentionprecedes, kietzmann2011attentionprecedes}. 

From a behavioural perspective, the majority of the research work that studies visual attention makes use of eye tracking devices, which are able to map the gaze of an observer at any given time with the location on a stimulus. Another active area of research, at the intersection between vision science and computer vision, is the modelling of visual salience, first proposed by \citet{itti1998salience}. Salience models shift the focus to the stimulus side and aim to answer the question ``what parts of a stimulus are most salient to a human observer?''. Adhering to the definitions by \citet{kuemmerer2018salience}, a salience model predicts the probability that a pixel on a given image is fixated, which can be expressed through salience maps that represent the distribution of salience for specific tasks. For this thesis we made use of both eye tracking and salience maps to study some aspects of human vision.

In one project, presented in detail in Chapter~\ref{ch:globsal}, we were interested in studying the global salience of competing stimuli, that is stimuli presented side by side. The bulk of the research on computational models of visual salience addresses the question of what parts of a stimulus are more likely to attract the gaze of observers. In this case, we aimed at quantifying the salience of images as a whole, to seek answers for the questions: Are some types of images more likely to attract the gaze of observers? If so, is this global salience related to the local salience properties of the images? Do other factors, such as familiarity with one of the images, play a role in the gaze direction of observers? In order to answer these questions, we conducted eye tracking experiments in which we recorded the gaze direction of participants who were shown pairs of images side by side. We then modelled the behavioural data with a machine learning algorithm and computed the local salience properties of the images with representative salience models from the literature. As a main finding, we concluded that natural images intrinsically have a global salience that varies widely across different types of images and is independent of the local salience properties. 

In another study, we combined computational models of visual salience with brain measurements of functional magnetic resonance imaging (fMRI) to analyse properties of the human visual cortex. We followed up the work by \citet{zuiderbaan2017imageidentification}, where the authors showed that it is possible to identify which natural image was shown to a participant in the scanner from fMRI recordings. The predictions were made by comparing the brain activations elicited by each image on areas V1, V2 and V3, and a combination of a contrast map of the images with the receptive field properties of the cortical areas, obtained through the population receptive field (pRF) model \citep{dumoulin2008prf}. Here, we studied whether brain activity was better predicted by salience maps than by contrast maps, and extended the analysis to a broader range of visual cortical areas: V1, V2, V3, hV4, LO12 and V3AB \citep{wandell2007visualfield}. We studied two distinct models of visual salience, ICF and DeepGaze \citep{kuemmerer2017icfdeepgaze}, and concluded that salience is more predictive of brain activations than contrast, especially the salience model based on intensity and contrast information only (ICF), rather than on high-level features, suggesting that salience information is still present in the neural activations of the visual cortex.

\section{Overview of contributions and outline}
\label{sec:intro-contributions}
The overarching objective of this thesis is to explore and exploit the connections between deep artificial neural networks and the visual cognitive and neural sciences. We believe that all three fields can benefit from mutual collaboration and synergies.

In order to facilitate the understanding of the thesis to a broader audience and set the grounds for the discussion throughout the dissertation, Chapter~\ref{ch:background} provides an introduction to the fundamentals of both machine learning and visual object recognition in the brain, as well as other relevant concepts. This chapter serves also as a review of related scientific literature.

Then, a first block from Chapter~\ref{ch:reg} to \ref{ch:invariance}, has data augmentation as the central theme, starting from the rather machine learning-centred Chapter~\ref{ch:reg}, towards gradually incorporating aspects from visual perception and biological vision in the subsequent chapters.

Specifically, in Chapter~\ref{ch:reg}, we discuss the concepts of explicit and implicit regularisation and provide definitions of these terms that have been widely but ambiguously used in the literature. Part of this chapter is based on the publication \citep{hergar2018daugreg} and we here provide a longer discussion about the taxonomy of regularisation and examples of explicit and implicit regularisation, arguing in particular that data augmentation is not explicit regularisation, as considered before in the literature.

Chapter~\ref{ch:daugreg} is focused on the comparison of explicit regularisation techniques---weight decay and dropout---and data augmentation and much of the content has been published in several articles \citep{hergar2018daugadvantages, hergar2018daugreg, hergar2018wddropout}. We present the results of a systematic empirical evaluation, alongside some insights from statistical learning theory, to conclude that data augmentation alone can provide the same generalisation gain than combined with explicit regularisation, and is remarkably more flexible. In view of these results, we discuss the need for weight decay and dropout in deep learning and propose to rethink the status of data augmentation.

In Chapter~\ref{ch:daugit}, we compare the representations learnt by neural networks trained with data augmentation and the activations in the inferior temporal cortex of the human brain \citep{hergar2018daugit}. We found that models trained with heavier transformations learn features more aligned with the representations in the higher visual cortex.

Following up the connection between data augmentation and biological vision, in Chapter~\ref{ch:invariance}, we study one of the fundamental properties of the visual cortex: the increasing invariance along the ventral stream to identity-preserving transformations of visual objects. Using data augmentation as a framework to generate such transformations, we first show that standard artificial neural networks trained optimised for object categorisation are hardly robust in terms of representational similarity. Then, we propose \textit{data augmentation invariance} as a simple, yet effective and efficient way of learning robust features, while preserving the categorisation performance \citep{hergar2019dauginv}.

The last two chapters of the dissertation can be seen as a separate block, in which data augmentation and artificial neural networks are not the main subject, although machine learning is still used as a tool. Chapter~\ref{ch:globsal} is closer to the field of cognitive science, as we study some aspects of visual behaviour through an eye-tracking experiment \citep{hergar2019globsal}. In particular, we propose \textit{global visual salience} as a metric of the likelihood of competing natural images to attract the gaze of an observer. Chapter~\ref{ch:imageid} is closer to neuroimaging, as we compare models of image identification from brain data to study properties of the early visual cortex. Part of the results of this study were presented as a poster contribution at the Annual Meeting of the Visual Sciences Society in 2019 \citep{hergar2019imageid}, and we here extended the analysis.

We conclude the dissertation with a general discussion in Chapter~\ref{ch:discussion}, where we provide an overview of the main results, outline the connections between the different parts, discuss future lines of work and the broader potential impact of this work.

\chapterbibliography
}

{
\chapter{Background}
\label{ch:background}
\renewcommand{\chapterpath}{includes/background}
The purpose of this chapter is two-fold: First, it aims at providing the fundamental background about the most relevant aspects of machine learning for this thesis, such as the theory of generalisation and regularisation. Second, it is also intended to serve as a review of the relevant and related scientific literature.

The introduction to the core aspects of machine learning in this chapter is deliberately non-exhaustive, as it is intended to provide only a sufficient background to enable the understanding of the rest of the thesis to a wider audience. The interested reader may follow the references to scientific literature provided throughout the chapter.

\section{Machine learning fundamentals}
Humans perceive the world and make sense of it in a great variety of ways: we see light as visual information, hear sounds and create music, use language to produce written text and speech and organise much of what we know into collections of numbers, to name a few. As diverse as light, sound, language and numbers are, they can all be conceptualised as \textit{data}. Thinking of what we perceive and know about the world as information that can be organised into data is a powerful formal conceptualisation that allows us to better understand and analyse the input and output we perceive and generate.

Machine learning is the discipline that studies how to automatically discover patterns from data \citep{murphy2012machinelearning}. Much as humans and other animals are able to make sense of the world out of what they perceive, machine learning aims at providing the methods to make sense out of formally defined data points, that is learning from data \citep{abu2012learningfromdata}. Machine learning has its roots in mathematics and statistical inference, but has developed as a distinct field after the development and spread of computers and computer science, which have provided the means to store and process data efficiently and automatically. 

\subsection{Elements of machine learning}
\label{sec:background-elements_ml}
The fundamental component of machine learning is the data and, as motivated above, the field itself arises from the ability to map any observation of the world into data points. Formally, one can define one data point as an $M$-dimensional vector $\mathbf{x} = x_{j}, \ldots, x_{M}$. Then, a set of $N$ observations $\{\mathbf{x}_{i}\}_{i}^{N} \in \mathcal{X}$ is said to be the \textit{input} data set. Most of the machine learning literature \citep{alpaydin2009machinelearning, abu2012learningfromdata, murphy2012machinelearning} makes a broad distinction amongst machine learning methods depending on the \textit{output} or \textit{target} data. A non-exhaustive taxonomy of the main types of machine learning methods as commonly found in the literature is the following:

\begin{itemize}
  \item \textbf{Supervised learning}: in a supervised learning setting, every observation $\mathbf{x}_{i}$ is paired with an output variable $y_{i}$, also referred to as \textit{ground truth}, and thus the data set is considered $\mathcal{D} = \{(\mathbf{x}_{i}, y_{i})\}_{i}^{N}$. The goal of supervised learning is discovering the relationship between the input data $\mathbf{x}_{i}$ and the target variables $y_{i}$. Depending on the nature of $y_{i}$, the learning problem can be \textit{classification} or \textit{regression}:
  \begin{itemize}
    \item Classification: $y_{i} \in \{1, \ldots, C\}$ is a discrete or categorical value. The possible values of $y$ are also referred to as \textit{classes} or \textit{labels}.
    \item Regression: $y_{i} \in \mathbb{R}$ is a continuous variable.
  \end{itemize}
  \item \textbf{Unsupervised learning}: in an unsupervised learning setting, there is not explicit access to target variables. Therefore the data set is simply $\mathcal{D} = \{\mathbf{x}_{i}\}_{i}^{N}$, and the goal is to discover patterns in the input data. A prototypical example of unsupervised learning is clustering.
  \item \textbf{Reinforcement learning} is a broad class of machine learning methods, initially inspired by behavioural psychology and the concept of trial-and-error learning. Instead of a mapping between input and output variables, in reinforcement learning typically there is access to a \textit{reward} signal that might not be available for every input data point. The goal is to learn policy that maximises the expected rewards by seeking a balance between exploration---the acquisition of new knowledge---and exploitation---the use of that knowledge to improve performance.
\end{itemize}

While this distinction is useful, the boundaries are sometimes blurred, in practice. For example, some problems often labelled as unsupervised learning could be considered particular cases of supervised learning, as we have discuss in the Introduction (Section~\ref{sec:intro-rethinking_supervised}). Most of the problems that we will address in this thesis can be defined within a supervised framework and, more specifically, classification. For these reasons, in the remaining of this section we will focus on classification, unless specified otherwise.

As introduced above, the goal of a (supervised) machine learning method is to discover the relationship between the input data and the target variables. This assumes that such relationship is determined by an underlying, unknown function $f \colon \mathcal{X} \mapsto \mathcal{Y}$. Since $f$ is a latent function, the task is to find a function $g \in \mathcal{H} \colon \mathcal{X} \mapsto \mathcal{Y}$ from a set of candidate functions $h \in \mathcal{H}$---the hypothesis set---that approximates $f$ according to certain error or loss measure $L(h, f)$. In order to find $g$, the \textit{learning algorithm} $\mathcal{A}$ uses the available \textit{training} data $\mathcal{D} = \{(\mathbf{x}_{i}, y_{i})\}_{i}^{N} = (\mathbf{x}_{1}, y_{1}), \ldots, (\mathbf{x}_{N}, y_{N})$ to solve an optimisation problem by adjusting a set of learnable parameters $\boldsymbol{\theta}$. Hence, the task is to determine from the set of functions $h(\mathbf{x}; \boldsymbol{\theta})$, the one which best approximates the data $\mathcal{D}$.

Nonetheless, if the relationships found apply only to the training data $\mathcal{D}$, then the process could not be considered \textit{learning}, but at best memorisation. Crucially, the ultimate objective of machine learning is to learn relationships and make correct predictions beyond the observed data. This is called \textit{generalisation}. This notion of learning is only feasible in a probabilistic way. The probabilistic view introduces the important assumption that the relationship between the targets $y$ and the input $\mathbf{x}$ is not deterministic, but probabilistic and there exists an unknown, underlying joint probability distribution $P_{X,Y}(\mathbf{x}, y)$ on $\mathcal{X} \times \mathcal{Y}$---thus also a marginal input distribution $P_{X}(\mathbf{x})$ and a conditional output distribution $P_{Y|X}(y|\mathbf{x})$, in Bayesian terms. Furthermore, it is also generally assumed that the observed available data points were sampled independently from $P_{X,Y}$. A summary schematic of the main elements of supervised learning is shown in Figure~\ref{fig:background-elements_learning}.

\begin{figure}[htb]
  \begin{center}
    \includegraphics[width = \linewidth]{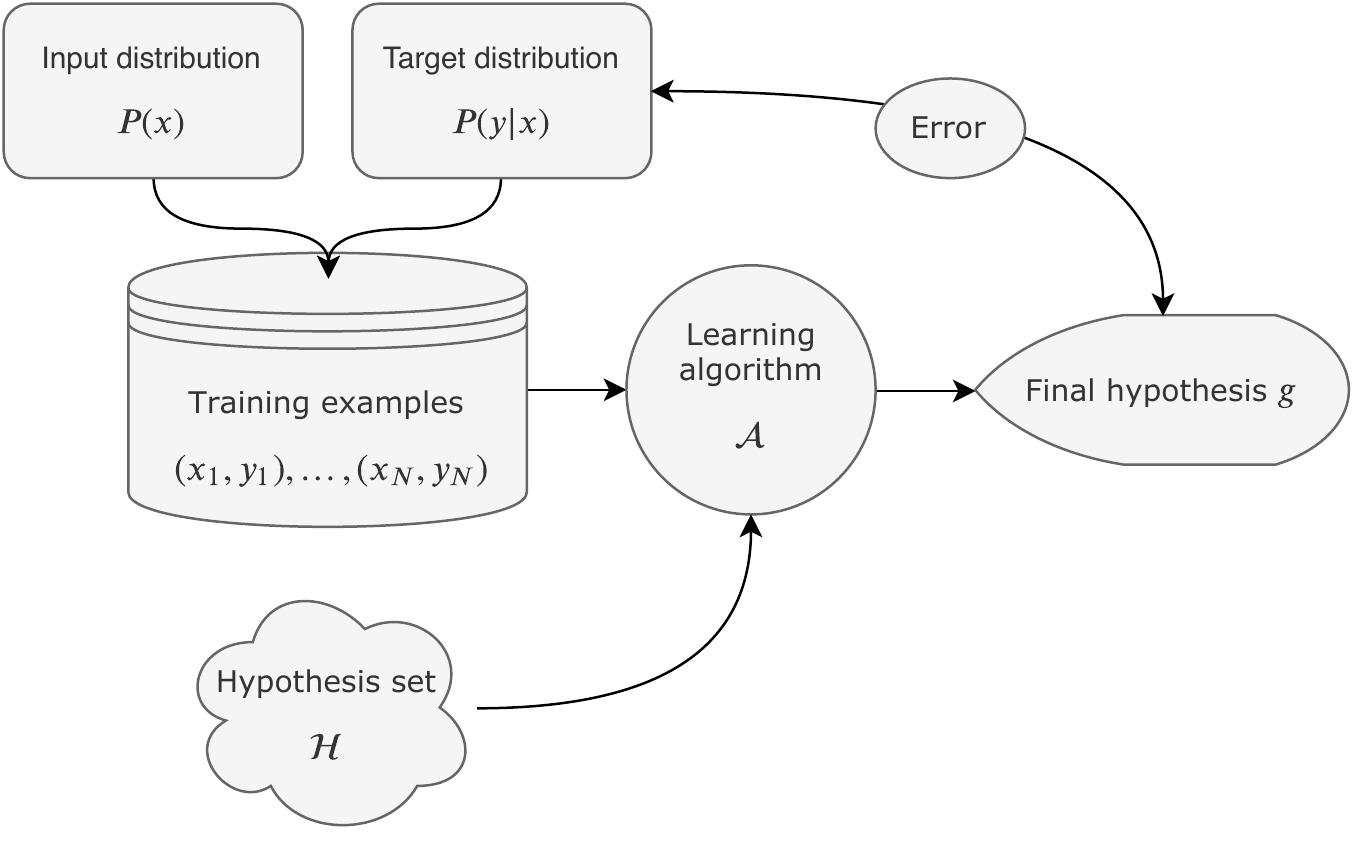}
  \end{center}
  \caption{Schematic of the main elements of (supervised) learning. Adapted from \citet{abu2012learningfromdata}.}
\label{fig:background-elements_learning}
\end{figure}

The feasibility of learning from a mathematical and probabilistic perspective is studied by the field of statistical learning theory \citep{vapnik1995learningtheory, bousquet2003learningtheory, vonluxburg2011learningtheory}. In the next section, we introduce the important concept of generalisation in machine learning and some notions from learning theory that are relevant to this thesis. 

\subsection{Theory of generalisation}
\label{sec:background-generalisation}
Generalisation is one of the most important concepts in machine learning and statistical learning theory. It refers to the idea that the ultimate goal in the learning process is not to minimise the error computed on the available training data in and of itself, but to perform well on unseen data, that is to generalise. This raises the reasonable first question of whether generalisation is possible at all. The probabilistic perspective not only provides a positive answer to the question but also the tools to analyse the generalisation guarantees of learning algorithms.

\subsubsection{Empirical risk minimisation}
In order to elaborate this idea, let us first introduce some important concepts that we will use in this section and throughout the thesis. To formally describe the problem at hand, we recall that we consider data that consist of the inputs $\mathbf{x}_{i} \in \mathcal{X}$ and the outputs or targets $y_{i} \in \mathcal{Y} = \{-1, 1\}$, that is binary classification, for simplicity of the exposition. Furthermore, we assume that the pairs $(\mathbf{x}_{i}, y_{i}) \in \mathcal{X} \times \mathcal{Y}$ are independently and identically sampled according to an unknown probability distribution $P_{X,Y}$. So as to measure the discrepancy between the target variables $y$ and the outcome of the hypotheses $h(x)$,\footnote{The hypotheses depend on both $\mathbf{x}$ and $\boldsymbol{\theta}$, that is $h(\mathbf{x}; \boldsymbol{\theta})$, but we will in general abuse notation and write simply $h(x)$, or even just $h$, for better readability.} we assume that we are given a real-valued \textit{loss function} $L(y, h(x))$. Since we are considering binary classification, the loss function will be the classification error: $L(y, h(x)) = \mathbbm{1}_{h(x) \neq y}$. Then, the \textit{risk} associated with a hypothesis $h$ is given by the expectation of the loss function, defined by the following \textit{risk functional}:
\begin{equation}
    R(h) = \mathbb{E} \left[ L(y, h(x)) \right] = \int L(y, h(x))dP_{X,Y}(x, y)
\end{equation}

Ultimately, the goal of a learning algorithm is to find the optimal hypothesis $h^{*} \in \mathcal{H}$ that minimises the risk $R(h)$:
\begin{equation}
    h^{*} = \argmin_{h \in \mathcal{H}} R(h)
\end{equation}

However, because the joint probability distribution $P_{X,Y}$ is unknown, it is not possible to exactly calculate $R(h)$. In practice, the risk functional is replaced by the computation of the \textit{empirical risk} on the set of $N$ available data points:
\begin{equation}
    R_{N}(h) = \frac{1}{N} \sum_{i=1}^{N} L(y_{i}, h(x_{i}, \theta))
\end{equation}
and the learning algorithm chooses the hypothesis $g$ by minimising the empirical risk:
\begin{equation}
    g = \hat{h} = \argmin_{h \in \mathcal{H}} R_{N}(h)
\end{equation}

This method is known as the \textit{Empirical Risk Minimisation} inductive principle (ERM) \cite{vapnik1982erm, vapnik1992erm}. The ERM principle was thoroughly studied during the 1960--1990s by Vladimir Vapnik and Alexey Chervonenkis, as well as other scientists, and it is summarised in the book \textit{The nature of statistical learning theory} \cite{vapnik1995learningtheory}. Empirical risk minimisation is a general principle and many classical estimation methods, such as least squares regression and maximum likelihood estimation, can be formulated as realisations of the ERM principle.

Some important aspects of the theory explained in the book and studied in a number of publications are the necessary and sufficient conditions for the consistency of algorithms based on the ERM, that is under what conditions minimising the empirical risk converges in the minimisation of the risk, when the number of examples tends to infinity \cite{vapnik1991necessary}; the rate of convergence of learning processes based on the ERM. Here, we will summarise the most important aspects and concepts of the learning theory based on the ERM.

\subsubsection{Uniform bounds}
Having outlined the notion of empirical risk minimisation and its components, we can define generalisation as the ability of a learning algorithm to achieve small risk $R(h)$. We can now return to the question of the feasibility of generalisation or the consistency of a learning process. An important result from probability theory that serves as starting point to study the feasibility of learning is Hoeffding's inequality \cite{hoeffding1963hoeffding}. It states that for any $\varepsilon > 0$ and a set of $N$ i.i.d. random variables $Z_{1} \ldots Z_{N}$, such that $\mathbb{P} \left[a \leq Z_{i} \leq b \right] \forall i$, then:
\begin{equation}
\label{eq:background-hoeffdings}
    \mathbb{P} \left[ \left| \frac{1}{N} \sum_{i=1}^{N}Z_{i} - \mathbb{E}\left[Z\right] \right| > \varepsilon \right] \leq 2 \exp \left( -\frac{2\varepsilon^{2}N}{(b-a)^{2}} \right)
\end{equation}

Hoeffding's inequality can be regarded as quantitative version of the law of large numbers\footnote{The law of large numbers is a fundamental theorem from probability theory. It states that the sample average converges in probability towards the expected value as the sample size increases.} for the case when the variables are bounded. Equation~\ref{eq:background-hoeffdings} can be developed into a more useful form for our purposes, relating the risk of a hypothesis, $R(h)$, and the empirical risk, $R_{N}(h)$. For any $\delta > 0$, at least with probability $1 - \delta$:
\begin{equation}
\label{eq:background-hoeffdings_risk_single}
    R(h) \leq R_{N}(h) + \sqrt{\frac{1}{2N}\log\frac{2}{\delta}}
\end{equation}

The interpretation of Equation~\ref{eq:background-hoeffdings_risk_single} is that the risk of hypothesis $h$ is bounded by a quantity that depends linearly and on the empirical risk plus a constant that depends on the number of samples used to obtain the empirical risk and the confidence $\delta$. On the one hand, this is good news as it establishes that the empirical risk is indicative of the true risk when the number of samples is large. This confirms the feasibility of learning. 

Nonetheless, on the other hand, the bound in Equation~\ref{eq:background-hoeffdings_risk_single} is highly limited and useless in practice. Essentially, it says that for each hypothesis $h$, there exists a set of samples for which the bound holds. However, the function which will be chosen by the learning algorithm is unknown before training it on the data set. For instance, there exists, as well, a function for which the empirical risk is not indicative of the true risk at all. In order to derive tighter, more useful bounds we need to take into account all the possible hypotheses $\mathcal{H}$ that the learning algorithm may choose. One of the best studied approaches is to consider \textit{uniform bounds}.

The idea is to find an upper bound of the \textit{supremum} of $R(h) - R_{N}(h)$, as that will clearly provide an upper bound on $R(g)$:
\begin{equation}
    R(g) - R_{N}(g) \leq \sup_{h \in \mathcal{H}} \left[ R(h) - R_{N}(h) \right]
\end{equation}

The simplest way is to consider the disjunction of all events $\abs{R(h_{m}) - R_{N}(h_{m})} > \varepsilon$ for a finite set of hypothesis $\mathcal{H}$ of size $M$ and $m = 1, \ldots, M$. Then, by applying the union bound\footnote{$\mathbb{P}(\bigcup_{I}A_{i}) \leq \mathbb{P}(\sum_{i}A_{i})$} to the application of Hoeffding's inequality (see Equation~\ref{eq:background-hoeffdings}) to the union of the difference of the risk and the empirical risk for all hypotheses, we get that
\begin{equation}
\label{eq:background-uniform_bounds}
\begin{split}
    \mathbb{P} \left[ \left| R(h) - R_{N}(h) \right| > \varepsilon \right]& \leq \sum_{m=1}^{M} \mathbb{P} \left[ \abs{R(h_{m}) - R_{N}(h_{m})} > \varepsilon \right]\\ 
                                                                            & \leq 2M \exp(-2\varepsilon^{2}N)
\end{split}
\end{equation}
and hence, equivalent to Equation~\ref{eq:background-hoeffdings_risk_single}, for $\delta > 0$,  with probability at least $1 - \delta$:
\begin{equation}
\label{eq:background-hoeffdings_risk_all}
    R(g) \leq R_{N}(g) + \sqrt{\frac{1}{2N}\log\frac{2M}{\delta}}
\end{equation}

This is finally a practical error bound that guarantees that the risk of the final hypothesis $g$ found via empirical risk minimisation will be bounded by the empirical risk measured on the training set, as long as the hypothesis set is finite, that is $\abs{\mathcal{H}} = M$. A subtler implication of Equation~\ref{eq:background-uniform_bounds}, besides the fact that it leads to Equation~\ref{eq:background-hoeffdings_risk_all}, is that $R(g) \geq R_{N}(g) - \varepsilon$ also holds. Hence, the ERM principle ensures that there are not much better hypotheses than $g$ in the set $\mathcal{H}$.

While the generalisation bound based on uniform deviations is a key result in statistical learning theory and it is theoretically relevant, it is only valid for learning algorithms that operate with finite hypothesis sets. For many machine learning algorithms, the hypothesis sets are infinitely large and additional theoretical results are necessary to describe their generalisation guarantees. Below we present some of the most important results.

\subsubsection{Vapnik-Chervonenkis theory}
When the hypothesis set $\mathcal{H}$ is uncountable and thus the right hand side of Equation~\ref{eq:background-hoeffdings_risk_all} is unbounded, a classical approach is to use the notion of \textit{growth function}, also known as \textit{shatter coefficient} or \textit{shattering number}. The growth function $m_{\mathcal{H}}(N)$ was introduced by \citet{vapnik1971vc} and it denotes the maximal \textit{effective} size of $\mathcal{H}$ on a set of $N$ examples, that is the maximum number of ways into which $N$ data points can be classified by the function class. Formally: 
\begin{equation}
\label{eq:background-growth_function}
    m_{\mathcal{H}}(N) = \sup_{x_1, \ldots, x_N \in \mathcal{X}}\left| (h(x_1), \ldots, h(x_N)) : h \in \mathcal{H} \right|
\end{equation}

For the case of binary classification that we are considering, the growth function $m_{\mathcal{H}}(N) \leq 2^N$. If the hypothesis set is capable of generating all possible dichotomies (binary labellings) of $x_1, \ldots, x_N$, then $\mathcal{H}$ is said to \textit{shatter} the data set. If no data set of size $k$ can be shattered by $\mathcal{H}$, then $k$ is said to be a \textit{break point} for $\mathcal{H}$. 

One important concept in statistical learning theory is the Vapnik-Chervonenkis dimension, known as \textit{VC dimension} for short. The VC dimension of a hypothesis set $\mathcal{H}$, denoted by $d_{VC}(\mathcal{H}$) or simply $d_{VC}$ is the largest $N$ such that $m_{\mathcal{H}}(N) = 2^N$, that is the largest data set size that the hypothesis can shatter. Hence, if $d_{VC}$ is the VC dimension of $\mathcal{H}$, then $k = d_{VC} + 1$ is a break point for the growth function.

It can be shown that if a hypothesis class $\mathcal{H}$ has finite VC dimension $d_{VC}$, then the growth function can be upper bounded by a polynomial:
\begin{equation}
\label{eq:background-sauer_lemma}
m_{\mathcal{H}}(N) \leq \sum_{i=0}^{d_{VC}}{\binom{N}{i}}
\end{equation}

This allows us to bound the risk of a hypothesis in terms of the empirical risk and the growth function, what is known as the \textit{VC generalisation bound}. For $\delta > 0$,  with probability at least $1 - \delta$:
\begin{equation}
\label{eq:background-vc_bound}
    R(g) \leq R_{N}(g) + \sqrt{\frac{8}{N}\log\left(\frac{4m_{\mathcal{H}}(2N)}{\delta}\right)}
\end{equation}

The VC generalisation bound is a key result in statistical learning theory as it establishes the feasibility of learning with infinite hypothesis sets: with enough data, all hypotheses in an infinite $\mathcal{H}$ with finite VC dimension will generalise from the empirical risk. The bound holds for all hypothesis sets, learning algorithms, input spaces, probability distributions and binary targets \citep{abu2012learningfromdata}. Such generality comes at the expense of being quite a loose bound to be used in practice.

One interpretation of the generalisation bound in Equation~\ref{eq:background-vc_bound} that we will use in this thesis is that the right hand side consists of two terms: the empirical risk and a term that is usually interpreted as a penalty for model complexity:
\begin{equation}
\label{eq:background-vc_model_complexity}
\Omega(N, \delta, \mathcal{H}) = \sqrt{\frac{8}{N}\log\left(\frac{4m_{\mathcal{H}}(2N)}{\delta}\right)} \leq \sqrt{\frac{8}{N}\log\left(\frac{4(2N)^{d_{VC}}}{\delta}\right)}
\end{equation}

$\Omega$ depends on the number of examples $N$, the confidence parameter $\delta$ and the hypothesis class $\mathcal{H}$. The bound gets tighter (better) as the number of examples increases, as the confidence constant $\delta$ increases and as the complexity of the hypothesis set decreases (lower $d_{VC}$). This form of the bound on the risk, $R(g) \leq R_{N}(g) + \Omega(N, \delta, \mathcal{H})$, is found in most methods to estimate the theoretical generalisation guarantees of learning algorithms.

\subsubsection{Rademacher complexity}
\label{sec:background-rademacher}
Given these limitations of the Vapnik-Chervonenkis theory and, in particular, of the VC dimension, other measures of complexity have been developed \citep{bartlett2002complexity}. One relatively recent, popular example is the \textit{Rademacher complexity}, which allows to define generalisation bounds that are not restricted to binary classification and hold for any class of real-valued functions. 

Let $\sigma_1, \ldots, \sigma_N$ be a set of independent random variables such that $P(\sigma_i = 1) = P(\sigma_i = -1) = \frac{1}{2}$. These are known as \textit{Rademacher variables}, hence the name of the complexity measure. As before, we consider a sample of $N$ independent data points $x_1, \ldots, x_N$ defined on $\mathcal{X}$. Now, instead of being restricted to binary classification, we let $\mathcal{F}$ be the class of real-valued functions $f \colon \mathcal{X} \mapsto \mathbb{R}$. Then, the \textit{empirical Rademacher complexity} of $\mathcal{F}$ with respect to the sample of size $N$ is defined as:
\begin{equation}
\label{eq:background-empirical_rademacher_complexity}
  \hat{\mathcal{R}}_{N}(\mathcal{F}) = \mathbb{E}_{\sigma} \left[ \underset{f \in \mathcal{F}}{\mathrm{sup}} \left| \frac{1}{N} \sum_{i=1}^{N} \sigma_{i}f(x_{i}) \right| \right]
\end{equation}
where $\mathbb{E}_{\sigma}$ denotes the expectation with respect to the Rademacher variables. The \textit{Rademacher complexity}, also found in the literature as \textit{Rademacher average}, is defined as the expectation of the empirical Rademacher complexity over all data sets of size $N$ on $\mathcal{X}$:
\begin{equation}
\label{eq:background-rademacher_complexity}
  \mathcal{R}_{N}(\mathcal{F}) = \mathbb{E} \left[ \hat{\mathcal{R}}_{N}(\mathcal{F}) \right]
\end{equation}

The interpretation of the Rademacher average as a complexity measure is intuitive: It is a measure of the ability of the function class $\mathcal{F}$ to fit random noise, introduced by the Rademacher variables $\sigma_i$. For a very large and complex $\mathcal{F}$, there will be a function $f$ that can fit the noise, making $\hat{\mathcal{R}}_{N}(\mathcal{F})$ larger. 

For the case of binary classification that we have considered so far, in which $\mathcal{H} \subseteq \{h \colon \mathcal{X} \mapsto \mathcal{Y} = \{-1, 1\}\}$, it can be easily shown that $\hat{\mathcal{R}}_{N}(\mathcal{H}) = \frac{1}{2}\hat{\mathcal{R}}_{N}(\mathcal{F})$, using the fact that $\sigma_i$ and $\sigma_i Y_i$ have the same distribution. Finally, we can use the Rademacher complexity to bound the risk of the final hypothesis. For $\delta > 0$,  with probability at least $1 - \delta$:
\begin{equation}
\label{eq:background-rademacher_risk}
    R(g) \leq R_{N}(g) + \hat{\mathcal{R}}_{N} + \sqrt{\frac{2\log\frac{2}{\delta}}{N}}
\end{equation}



\subsection{Regularisation}
\label{sec:background-regularisation}
In Section \ref{sec:background-generalisation}, we have seen that the goal of a machine learning algorithm is to find a hypothesis $h(\mathbf{x})$ that, given some data $\mathcal{D} = \{(\mathbf{x}_{i}, y_{i})\}_{i}^{N}$, minimises the risk functional $R(h)$, which in turn depends on a loss function $L(y, h(\mathbf{x}))$ chosen as a criterion for the optimisation problem. We have also seen that, since it is not possible to exactly calculate the risk, in practice we optimise the empirical risk $R_{N}(h)$, which is calculated on the available training data:
\begin{equation}
\label{eq:background-argmin_empirical_risk}
    g = \argmin_{h \in \mathcal{H}} R_{N}(h)
\end{equation}

This method is known as empirical risk minimisation (ERM), and in Section~\ref{sec:background-generalisation} we have summarised the theory that describes the convergence of the empirical risk to the actual risk of the model. Nonetheless, despite the importance of this theory to confirm the feasibility of learning, the bounds on the generalisation error are not always applicable in practice: they are quite loose, depend on hypothesis sets with finite VC dimension and, in general, the plain ERM principle is intended to deal with large sample sizes. In practice, learning algorithms rely on extensions of ERM, such as the principle of structural risk minimisation (SRM) \citep{vapnik1974srm}. A particular case of SRM is regularisation, a widely used technique in machine learning and one of its cornerstones \citep{poggio1990regularisation, girosi1995regularization}. In this section we review the fundamentals of regularisation and present some of its most common forms, which are relevant for this thesis.

The concept of regularisation of learning algorithms is closely related to the mathematical problem of approximating a function from sparse data, that is finding $f \in \mathcal{F}$ such that $Af = F$. \citet{hadamard1902illposed} demonstrated that under some general circumstances this is an ill-posed problem. That is, an arbitrarily small deviation $\varepsilon$ of $F$ ($F_{\varepsilon}$ instead of $F$, where $\norm{F - F_{\varepsilon}} < \varepsilon$) can cause large deviations in the solution of the equation. Formally, minimising the functional
\begin{equation}
\label{eq:background-ill_posed_functional}
    \rho(f) = \norm{Af - F_{\varepsilon}}^2
\end{equation}
is not guaranteed to provide a good approximation even if $\varepsilon$ tends to zero. This closely resembles the learning problem that we have described above, where the task is to find the function $g$ that best approximates the data $\mathcal{D}$, using the empirical risk, as summarised in Equation~\ref{eq:background-argmin_empirical_risk} and detailed in Section~\ref{sec:background-generalisation}. As a matter of fact, finding $g$ in the presence of noise is also ill-posed, as there is an infinite number of solutions. In order to find a suitable solution with access to only limited data, it is necessary to constrain the hypothesis space $\mathcal{H}$ with some a priori information, for instance assuming that the function is smooth. This is the idea of the regularisation principles discovered in the 1960s \citep{phillips1962regularisation, tikhonov1963regularisation, ivanov1976regularisation}. In particular, they found that if instead of minimising the functional $\rho(f)$ of Equation~\ref{eq:background-ill_posed_functional}, one minimises the so-called regularised functional
%
\begin{equation}
\label{eq:background-reg_functional}
    \rho^*(f) = \norm{Af - F_{\varepsilon}}^2 + \lambda(\varepsilon)\Omega(f)
\end{equation}
where $\Omega(f)$ is the regularisation functional, and $\lambda(\varepsilon)$ is a constant that determines the level of noise, then the sequence of solutions converges as $\varepsilon \rightarrow 0$. In our particular case of learning from data, the principles of regularisation translate into adding a similar regularisation term to the objective function. The similarity between the two problems is most obvious if we consider the mean squared error loss, instead of binary classification. In this case, the optimisation problem becomes the following:
\begin{equation}
\label{eq:background-reg_objective}
\begin{split}
	g &= \argmin_{h \in \mathcal{H}}R_{reg}(h) = \argmin_{h \in \mathcal{H}}\left[ R_{N}(h) + \lambda\Omega(h) \right]\\
      &= \argmin_{h \in \mathcal{H}}\left[ \sum_{i=1}^{N}(h(\mathbf{x}_{i}) - y_{i})^2 + \lambda\Omega(h) \right]
\end{split}
\end{equation}

$\Omega(h)$ is the regularisation functional or regulariser, which incorporates prior information or desired properties of the model. In general, the regulariser is chosen to encourage smooth functions. The constant $\lambda$ is the regularisation parameter, which controls the strength of the regularisation. 

The concepts of generalisation and particularly regularisation are closely related to the widely used concept of \textit{overfitting}, that is the tendency of a learning algorithm to excessively fit the training data points, to the detriment of its generalisation. Broadly speaking, a complex hypothesis function is more likely to \textit{overfit} the training data than a simpler function. The idea of function smoothness introduced by the regularisation term can be seen as a way to counteract overfitting, in favour of better generalisation. This establishes a trade-off where ideally the learning algorithm should strike the right balance between fitting the data, that is minimising the empirical risk, and finding a smooth enough function that generalises well. This trade-off can be controlled by the value of the regularisation parameter, which is often determined through \textit{cross-validation} \citep{stone1974crossval, allen1974crossval}. 

The choice of $\Omega(h)$ leads to different forms of regularisation and there is a very large body of literature on this topic. Popular choices are constraints on the norm of the parameters, which we will discuss in Section~\ref{sec:background-weight_decay} or constraints on the curvature of $h$. In modern machine learning, the concept of regularisation is very broad and regularisation is considered to be any mechanism that prevents overfitting, hence improving generalisation. In Chapter~\ref{ch:reg} we compare different forms of regularisation and discuss the distinction between implicit and explicit regularisation---key in Chapter~\ref{ch:daugreg}---and other regularisation taxonomies.

In the remaining of this section we introduce two specific regularisation techniques, weight decay and dropout, which are arguably the two most common forms of regularisation in modern neural networks.

\subsubsection{Weight decay}
\label{sec:background-weight_decay}
Weight decay is the common name used to refer to \textit{$L^2$--norm regularisation}\footnote{Note, however, that in part of the machine literature, the term weight decay refers to a form of regularisation in which the $L^2$--norm penalty is added directly to the update rule of gradient descent. This results in a conceptually equivalent form of regularisation, but with a slight numerical difference. See \citet{babenko2018wdvsl2} for more details.}, which is in turn a particular case of $L^p$--norm regularisation. In this section we will first review $L^p$--norm regularisation as a direct realisation of the type of regularisation described above, and then present the specific aspects of weight decay.

In the previous section we have seen that the concept of regularisation derived from the mathematical tool for solving ill-posed problems, results in a modification of the objective function (see Equation~\ref{eq:background-reg_objective}). We will denote the regularised objective by $\hat{J}$:
\begin{equation}
\label{eq:background-reg_objective_explicit}
    \hat{J}(\boldsymbol{\theta}; \mathbf{x}, y) = J(\boldsymbol{\theta}; \mathbf{x}, y) + \lambda\Omega(\boldsymbol{\theta})
\end{equation}

$L^p$--norm regularisation refers to the family of techniques which apply a penalty on the norm of the parameters:
\begin{equation}
\label{eq:background-l_p_norm}
\Omega(\boldsymbol{\theta}) = \phi(\norm{\boldsymbol{\theta}}_p) = \phi\left(\left(\sum_{i=1}^{d}\abs{\theta_i}^p\right)^{\frac{1}{p}}\right)
\end{equation}
where $\phi(\cdot)$ is an optional function applied on the norm, for example the squared function. The most commonly used $L^p$--norm penalties are $L^1$ and $L^2$ regularisation. $L^2$ regularisation is probably one of the most widely used regularisation techniques in machine learning. In deep learning, it is commonly referred to as \textit{weight decay}, but it is also known as Tikhonov regularisation \citep{tikhonov1963regularisation} or ridge regression when applied to linear regression. In this thesis, we will analyse weight decay in neural networks and compare it to other regularisation techniques in Chapter~\ref{ch:daugreg}, and we use it to regularise a logistic regression algorithm in Chapter~\ref{ch:globsal}.

The specific regularisation term typically used for weight decay is $\Omega(\boldsymbol{\theta}) = \frac{1}{2}\norm{\boldsymbol{\theta}}_2^2$ because it allows to implement and express the objective function in a convenient and efficient way using the dot product between the vector of parameters and its transpose:
\begin{equation}
\label{eq:background-weight_decay_objective}
    \hat{J}(\boldsymbol{\theta}; \mathbf{x}, y) = J(\boldsymbol{\theta}; \mathbf{x}, y) + \frac{\lambda}{2}\boldsymbol{\theta}^T\boldsymbol{\theta}
\end{equation}

Weight decay has been long used, at least since the 1980s \citep{hinton1987wd}, and widely studied, both empirically \citep{zhang2018wd} and theoretically \citep{krogh1992wd, neyshabur2015regularization}, especially in the context of neural networks. Intuitively, the mechanism provided by weight decay is to restrict the norm of the trainable parameters, by decreasing the weight vector at every iteration of a model trained with gradient descent, in the directions that do not contribute much to reducing the objective function. Relevant to this dissertation is the result by \citet{bishop1995tikhonov}, which showed that optimising a squared error loss with weight decay is equivalent to training with random noise in the inputs. In Chapter~\ref{ch:daugreg}, we will use this result to derive some theoretical insights into the comparison of weight decay and data augmentation.

\subsubsection{Dropout}
\label{sec:background-dropout}
Dropout is a regularisation technique first described by this name in \citep{hinton2012dropout, srivastava2014dropout}, although closely related to \textit{dilution} \citep{hertz1991dilution}. It is very widely used in modern neural networks due to its simplicity and effectiveness. In practice, dropout is implemented and hence can be described as a method that omits every unit---parameter, feature detector, etc.---of a model with probability $p$, at every iteration of the optimisation process (training). At inference (test) time, the whole set of units is considered. While dropout can be applied to a broad class of models, it is most often used to train deep neural networks and for simplicity we will also consider neural networks in this section.

Dropout is often described as a practical approximation of training an ensemble of models through bootstrap aggregation, commonly known as \textit{bagging} \citep{breiman1994bagging}, in which $M$ models are trained on $M$ subsets of a data set of size $N$ uniformly sampled with replacement (bootstrap sample). At inference time, the outputs of the $M$ models on each data point are averaged (for regression) or combined through majority voting (for classification). Bagging is a widely used technique, known to reduce the variance and overfitting of learning algorithms. However, it is computationally expensive as it requires training multiple models. Dropout efficiently approximates a form of bagging with an exponentially large number of sub-networks (models). Since neural networks are typically trained with mini-batch iterative methods (such as stochastic gradient descent), the parameters of the model are updated by computing the loss of a sub-network on a sub-sample of the data set. An important difference between standard bagging and dropout is that while in bagging the models are independent, with dropout the models share a subset of the parameters from the parent neural network.

In \citep{srivastava2014dropout}, dropout training is connected with a theory by \citet{livnat2010sex} about the superiority of sexual over asexual reproduction in nature. According to this theory, a criterion for natural selection would be enhancing the robust combination of different genes for better adaptation to changes, as opposed to the optimisation of the individual fitness through a slight mutation of one parent's genes. Sexual reproduction would favour this criterion by preventing co-adaptations of the available genes in one individual. With dropout, the units of a network are forced to learn useful combinations with other subsets of random units, hence preventing co-adaptation and increasing robustness.

Dropout has greatly impacted the deep learning community\footnote{The two original papers have been increasingly cited almost 25,000 times at the time of writing, according to Google Scholar.}. It is widely used for training neural networks in both research and application and it has been deeply studied both empirically and theoretically \citep{gal2016dropout}, sometimes uncovering contradictory and surprising properties. While it is out of the scope to review the vast literature on dropout training, we can mention some relevant findings. Since shortly after it was proposed, dropout has been analysed as adaptive form of regularisation \citep{wager2013dropout}. \citet{baldi2013dropout} found that the dynamics of gradient descent with dropout training approximate that of a regularised error function, while \citet{helmbold2017dropout} showed that in deeper networks, the behaviour of dropout differs significantly from standard regularisation. More recently, \citet{mou2018dropout} derived generalisation bounds based on the Rademacher complexity for deep neural networks trained with dropout. Finally, an interesting and relevant finding for this thesis is that dropout applied to the intermediate units of a neural network has been shown to be equivalent to training with noise in the input \citep{bouthillier2015dropoutasdaug}.

\chapterbibliography
}

{
\chapter{Explicit and implicit regularisation}
\label{ch:reg}
\renewcommand{\chapterpath}{includes/regularisation}
\begin{outreach}
    \item \textit{Data augmentation instead of explicit regularization.} Alex Hern{\'a}ndez-Garc{\'i}a, Peter K{\"o}nig. arXiv preprint arXiv:1806.03852, 2018.
\end{outreach}
One of the central issues in machine learning research and application is finding ways of improving generalisation. Regularisation, broadly defined as any modification applied to a learning algorithm that helps the model generalise better, plays therefore a key role in machine learning\footnote{In Chapter~\ref{ch:background} we review the fundamentals of machine learning and in particular, Section~\ref{sec:background-regularisation} reviews the essential aspects of regularisation to understand this and the upcoming chapters.}. In the case of deep learning, where neural networks tend to have several orders of magnitude more parameters than training examples, statistical learning theory (Section~\ref{sec:background-generalisation}) indicates that regularisation becomes even more crucial. Accordingly, a myriad of techniques have been proposed as regularisers (Section~\ref{sec:background-regularisation}): weight decay \citep{hanson1989wd} and other $L^p$ penalties on the learnable parameters; dropout---random dropping of units during training---\citep{srivastava2014dropout} and stochastic depth---random dropping of whole layers---\citep{huang2016stochasticdepth}, to name a few. 

Moreover, whereas in simpler machine learning algorithms the regularisers can be easily identified as explicit terms in the objective function, in modern deep neural networks the sources of regularisation are not only explicit but implicit \citep{neyshabur2014implicitreg}.  In this regard, many techniques have been studied for their regularisation effect, despite not being explicitly intended as such. Examples are convolutional layers \citep{lecun1990conv}, batch normalisation \citep{ioffe2015batchnorm} and data augmentation. In sum, there are multiple elements in deep learning that contribute to reducing overfitting and thus improve generalisation.

It is common practice in both the scientific literature and application to incorporate several of these regularisation techniques in the training procedure of neural networks. For instance, weight decay, dropout and data augmentation have been used jointly in multiple well-known architectures \citep{tan2019efficientnet, huang2017densenet, zagoruyko2016wrn, springenberg2014allcnn}. It is therefore implicitly assumed that each technique  is necessary and contributes additively to improving generalisation. However, the interplay between regularisation techniques is yet to be well understood and might be an important piece for the puzzle of why and when deep networks generalise. 

In Chapter~\ref{ch:daugreg}, we will focus on contrasting some specific forms of regularisation, namely weight decay, dropout and data augmentation. This chapter serves as a preamble of the following one. Here, we will provide definitions of two terms that have been widely but ambiguously used in the machine learning literature: explicit and implicit regularisation. We contend that these terms are useful to understand and explain the role of regularisation in artificial neural networks, as reflected by their use in the literature. Therefore, it is important to settle the precise meaning of the terms and provide examples. Besides being helpful to interpret the results in Chapter~\ref{ch:daugreg}, we also hope that this is a useful contribution to the machine learning community.

\section{Why do we need definitions?}
\label{sec:reg-why}
While several regularisation taxonomies have been proposed (see Section~\ref{sec:reg-taxonomies}), to the best of our knowledge there is no formal definitions of explicit and implicit regularisation in the machine literature. Nonetheless, the terms have been widely used \cite{neyshabur2014implicitreg, zhang2016understandingdl, wilson2017neurips, mesnil2011transferlearning, poggio2017theory3, martin2018selfregularisation, achille2018emergence}. This could suggest that the concepts are ingrained in the field and well understood by the community. However, by analysing the use of the terms explicit and implicit regularisation in the literature and in discussions with practitioners one can see that there is a high degree of ambiguity. In this section we will review some examples and motivate the need for formal definitions.

The PhD thesis by \citet{neyshabur2017thesis} is devoted to the study of implicit regularisation in deep learning. For instance, Neyshabur shows that common optimisation methods for deep learning, such as stochastic gradient descent (SGD), introduce an inductive bias that lead to better generalisation. That is, SGD \textit{implicitly} regularises the learning process. However, the notion and definition of implicit regularisation is only implied in Neyshabur's PhD thesis and related works. 

Some may argue that the definitions are not necessary. By looking at one single piece of work, even without a formal definition, the meaning may be inferred from the use. However, when one considers a larger body of work by multiple authors, differences and even contradictions emerge. In the work by Neyshabur and colleagues \citep{neyshabur2017thesis, neyshabur2014implicitreg}, it can be interpreted that implicit regularisation refers to the generalisation improvement provided by techniques such as stochastic gradient descent (SGD) that are not \textit{typically} considered as regularisation. By extension, explicit regularisation would refer to those other techniques: ``we are not including any explicit regularisation, neither as an explicit penalty term nor by modifying optimisation through, e.g., drop-outs, weight decay, or with one-pass stochastic methods'' \citep{neyshabur2017thesis}. In \citet{poggio2017theory3}, it can be interpreted that implicit regularisation refers to techniques that lead to minimisation of the parameter norm without explicitly optimising for it. By extension, explicit regularisation would refer to classical penalties on the parameter norm, such as weight decay. It is therefore unclear whether other methods such as dropout should be considered explicit or implicit regularisation according to \citet{poggio2017theory3}.

\citet{zhang2016understandingdl} raised the thought-provoking idea that ``explicit regularisation may improve generalisation performance, but is neither necessary nor by itself sufficient for controlling generalisation error.'' The authors came to this conclusion from the observation that turning off the ``explicit regularisers'' of a model does not prevent the model from generalising reasonably well. In their experiments, the explicit regularisation techniques they turned off were, specifically, weight decay, dropout and data augmentation. In this case, it seems that \citet{zhang2016understandingdl} made a distinction based on the mere intention of the practitioner. Under that logic, because data augmentation has to be designed and applied \textit{explicitly}, it would be explicit regularisation. 

These examples illustrate that the terms explicit and implicit regularisation have been used subjectively and inconsistently in the literature. In order to help avoid ambiguity, settle the concepts and facilitate the discussion, in the next section we propose definitions and provide examples to illustrate each category. Further, we will argue that data augmentation is not explicit regularisation and introduce some key differences with respect to explicit regularisation, which will set the grounds for Chapter~\ref{ch:daugreg}. 

\section{Definitions and examples}
\label{sec:reg-definitions}
We propose the following definitions of explicit and implicit regularisation:

\begin{itemize}
\item \textbf{Explicit regularisation techniques} are those techniques which reduce the \textit{representational} capacity of the model class they are applied on. That is, given a model class $\mathcal{H}_0$, for instance a neural network architecture, the introduction of explicit regularisation will span a new hypothesis set $\mathcal{H}_1$,  which is a \textit{proper subset} of the original set, that is $\mathcal{H}_1 \subsetneq \mathcal{H}_0$.
\item \textbf{Implicit regularisation} is the reduction of the generalisation error or overfitting provided by means other than explicit regularisation techniques. Elements that provide implicit regularisation do not reduce the \textit{representational} capacity, but may affect the \textit{effective} capacity of the model: the \textit{achievable} set of hypotheses given the model, the optimisation algorithm, hyperparameters, etc.
\end{itemize}

Note that we define explicit and implicit regularisation by using the concepts of \textit{representational} and \textit{effective} capacity. Although these terms are also used ambiguously by some practitioners, definitions of these concepts can be found in the literature. For instance, the textbook Deep Learning \citep{goodfellow2016dlbook} clearly defines the representational capacity as the ``the family of functions the learning algorithm can choose from'' and explains that the effective capacity ``may be less than the representational capacity'' because the learning algorithm does not always find the ``best function'' due to ``limitations, such as the imperfection of the optimisation algorithm''. Thinking of these \textit{limitations} as implicit regularisation denotes that this can be beneficial. In any case, we here adopt these definitions of representational and effective capacity.

One of the most common explicit regularisation techniques in machine learning is $L^p$-norm regularisation \citep{tikhonov1963regularisation}, of which weight decay is a particular case, widely used in deep learning. Weight decay sets a penalty on the $L^2$ norm of the model's learnable parameters, thus constraining the representational capacity of the model. Dropout \citep{srivastava2014dropout} is another common example of explicit regularisation, where the hypothesis set is reduced by stochastically deactivating a number of neurons during training. Similar to dropout, stochastic depth \citep{huang2016stochasticdepth}, which drops whole layers instead of neurons, is also an explicit regularisation technique.

Regarding implicit regularisation, note first that the above definition does not refer to \textit{techniques}---as in the definition of explicit regularisation---but to a regularisation \textit{effect}, as it can be provided by multiple elements of different nature. For instance, stochastic gradient descent (SGD) is known to have an implicit regularisation effect---reduction of the generalisation error---without constraining the representational capacity \citep{zhang2017sgd}. Batch normalisation neither reduces the capacity, but it improves generalisation by smoothing the optimisation landscape \citep{santurkar2018bn}. Of quite a different nature, but still implicit, is the regularisation effect provided by early stopping \citep{yao2007earlystopping}, which does not reduce the representational but the effective capacity.

In these examples and all other cases of implicit regularisation, we can think of the effect on the capacity in the following way: we start by defining our model class, for instance a neural network, which spans a set of functions $\mathcal{H}_0$ (see Section~\ref{sec:background-elements_ml}). If we decide to train with explicit regularisation, for instance weight decay or dropout, then the model will have access to a smaller set of functions $\mathcal{H}_1 \subsetneq \mathcal{H}_0$, that is the representational capacity. On the contrary, if we decide to train with SGD, batch normalisation or early stopping, the set of functions spanned by the model stays identical. Due to the dynamics and limitations imposed by these techniques, some functions may never be found, but theoretically they could be. In other words, the effective capacity may be smaller but not the representational capacity.

Central to this thesis is data augmentation, a technique that provides an implicit regularisation effect. As we have discussed, \citet{zhang2016understandingdl} considered data augmentation an explicit regularisation technique and was analysed as equivalent in terms of category to weight decay and dropout. However, data augmentation does not reduce the representational capacity of the models and hence, according to our definitions, cannot be considered explicit regularisation. This is relevant to understand the differences between weight decay, dropout and data augmentation that we will present in Chapter~\ref{ch:daugreg}, especially in the context of artificial neural networks.

\section{On the taxonomy of regularisation}
\label{sec:reg-taxonomies}
As in most disciplines, many taxonomies of regularisation techniques for machine learning have been proposed. Being a key ingredient of machine learning theory and practice, machine learning textbooks include a review of regularisation methods. In the case of deep learning, besides the classical regularisation methods used in \textit{traditional} machine learning, multiple new regularisation techniques have been proposed in recent years, and many techniques have been analysed because of their implicit regularisation effect. In this section, we briefly review some taxonomies of regularisation proposed in the literature and discuss their similarity with our definitions. 

In their textbook, \citet{goodfellow2016dlbook} review some of the most common regularisation techniques used to train deep neural networks, but do not discuss the concepts of explicit and implicit regularisation. More recently, \citet{kukavcka2017regularization} provided an extensive review of regularisation methods for deep learning. Although they mention the implicit regularisation effect of techniques such as SGD, no further discussion of the concepts is provided. Nonetheless, they define the category \textit{regularisation via optimisation}, which is somewhat related to implicit regularisation. However, regularisation via optimisation is more specific than our definition; hence, methods such as data augmentation would not fall into that category.

Recently, \citet{guo2018mixup} provided a distinction between \textit{data-independent} and \textit{-dependent} regularisation. They define data-independent regularisation as those techniques that impose certain constraint on the hypothesis set, thus constraining the optimisation problem. Examples are weight decay and dropout. We believe this is closely related to our definition of explicit regularisation. On the other hand, they define data-dependent regularisation as those techniques that make assumptions on the hypothesis set with respect to the training data, as is the case of data augmentation. While we acknowledge the usefulness of such taxonomy, we argue that the division between data-independent and -dependent regularisation leaves some ambiguity about other techniques, such as batch-normalisation, which neither imposes an explicit constraint on the representational capacity nor on the training data. 

On the contrary, our distinction between explicit and implicit regularisation aims at being complete, since implicit regularisation refers to any regularisation effect that does not come from explicit---or data-independent---techniques.
%

\section{Discussion}
\label{sec:reg-discussion}
The main contribution of this chapter has been the proposal of definitions of explicit and implicit regularisation. These terms that have been widely used in the machine learning literature without being formally defined, hence giving rise to subjective and ambiguous use. With the definition of these important concepts we have set the grounds for our discussion on the rest of this thesis, especially in Chapter~\ref{ch:daugreg}, but we also hope to help settle the concepts and reduce the ambiguity in the literature.

Besides this contribution, it is interesting to draw some connections between the concept of implicit regularisation, the discussion about inductive biases in the Introduction (Chapter~\ref{ch:intro}) and data augmentation. According to our definition above, implicit regularisation is the improvement in generalisation provided by elements that are not explicit regularisation techniques. This is a broad definition that includes many possible sources of implicit regularisation. A concept that underlies many of them is that of inductive bias. The inductive bias encoded in explicit regularisation techniques is simply that smaller models generalise better, which is reminiscent of Occam's razor. While this is a powerful inductive bias, we have discussed that many other sources of inductive bias are possible and are worth exploring.

In this regard, a clear distinction between explicit and implicit regularisation may help in the analysis, especially in the case of deep learning. A striking difference between neural networks and other machine learning algorithms is that deep networks easily scale to an (almost) arbitrarily large number of parameters and still generalise well on a held out test data. Only recently are we starting to understand this phenomenon, which seemed to be at odds with the results of statistical learning theory \citep{belkin2019biasvariance}. 

First, the concept of implicit regularisation may help explain the generalisation of deep neural networks. Second, the fact that very large neural networks can generalise well directly casts doubts on the need for explicit regularisation \citep{zhang2016understandingdl}, that is to constrain the representational capacity. However, most artificial neural networks are still trained with explicit regularisation methods such as weight decay and dropout. In the next chapter, we follow up this idea and directly address the question of whether explicit regularisation is necessary in deep learning, provided enough implicit regularisation is included, specifically data augmentation.

\chapterbibliography
}

{
\chapter[Data augmentation instead of explicit regularisation]{Data augmentation\\instead of explicit regularisation}
\label{ch:daugreg}
\renewcommand{\chapterpath}{includes/daug-reg}
\begin{outreach}
    \item \textit{Data augmentation instead of explicit regularization.} \textbf{Alex Hern{\'a}ndez-Garc{\'i}a}, Peter K{\"o}nig. arXiv preprint arXiv:1806.03852, 2018.
    \item \textit{Do deep nets really need weight decay and dropout?.} \textbf{Alex Hern{\'a}ndez-Garc{\'i}a}, Peter K{\"o}nig. arXiv preprint arXiv:1802.07042, 2018.
    \item \textit{Further advantages of data augmentation on convolutional neural networks.} \textbf{Alex Hern{\'a}ndez-Garc{\'i}a}, Peter K{\"o}nig. International Conference on Artificial Neural Networks (ICANN, Best Paper Award), 2018.
\end{outreach}
Data augmentation in machine learning refers to the techniques that synthetically create new examples from a data set by applying possibly stochastic transformations on the existing examples. In the image domain, these transformations can be, for instance, slight translations or rotations, which preserve the perceptual appearance of the original images, but significantly alter the actual pixel values. Despite being an old technique \citep{abumostafa1990hints, simard1992daug} and ubiquitous in the deep learning literature and practice, data augmentation has often been regarded as a sort of \textit{cheating}, \textit{lower class} technique\footnote{As a result, the machine learning scientific community has heavily ignored data augmentation as a subject of study until recently. By way of illustration, the textbook Deep Learning \citep{goodfellow2016dlbook} dedicates one and a half pages to data augmentation, of which one third is devoted to the caveats of data augmentation. Only in the last few years has data augmentation started to receive increasing attention, likely due to the success of some data augmentation techniques, such as \textit{cutout} \citep{devries2017cutout} and \textit{mixup} \citep{zhang2017mixup}, and especially by the popularisation by Google of \textit{automatic} data augmentation \citep{cubuk2018autoaugment}, previously proposed by various university groups \citep{hauberg2016learningdaug, antoniou2017dagan, ratner2017learningdaug, lemley2017smartdaug}. We first submitted the results presented in this chapter in 2017 \citep{hergar2018daugregopenreview} and other authors have also presented surveys on data augmentation techniques \citep{perez2017dauganalysis, shorten2019daugsurvey}. Promisingly, very recently has data augmentation started to be analysed as well from a theoretical point of view \citep{rajput2019daug, chen2019invariance, lyle2020daug}}, which should not be used in order to assess the actual strength of a new proposal \citep{goodfellow2013maxout, graham2014fracmaxpool, larsson2016fractalnet, goodfellow2016dlbook}. A common criticism is that data augmentation usually requires domain or expert knowledge and it cannot be easily generalised across data domains \citep{devries2017daugfeatspace}.

Explicit regularisation methods such as weight decay \citep{hanson1989wd} and dropout \citep{srivastava2014dropout} are also nearly ubiquitous. In contrast, they are considered intrinsic parts of the learning algorithm and thus have remained unquestioned. However, in Chapter~\ref{ch:reg}, we have introduced the differences between explicit and implicit regularisation and raised the question of whether explicit regularisation is necessary in deep learning. On the other hand, in the Introduction (Chapter~\ref{ch:intro}) we have discussed the view of data augmentation as a powerful inductive bias from visual perception. Building upong these insights, in this chapter, we analyse the role of data augmentation in neural networks trained for image object categorisation and the need for weight decay and dropout when data augmentation is used. We first derive some theoretical insights from statistical learning theory and then present the results of a large empirical study in which we contrast the contributions of each technique.

\section{Theoretical insights}
\label{sec:daugreg-theoretical_insights}
As we have reviewed in Section~\ref{sec:background-generalisation}, the generalisation of a model class $\mathcal{H}$ can be analysed through complexity measures such as the VC-dimension or, more generally, the Rademacher complexity $\mathcal{R}_{N}(\mathcal{H}) = \mathbb{E} \left[ \hat{\mathcal{R}}_{N}(\mathcal{H}) \right]$, where:
\begin{equation}
\label{eq:daugreg-rademacher}
  \hat{\mathcal{R}}_{N}(\mathcal{H}) = \mathbb{E}_{\sigma} \left[ \underset{h \in \mathcal{H}}{\mathrm{sup}} \left| \frac{1}{N} \sum_{i=1}^{N} \sigma_{i}h(x_{i}) \right| \right]
\end{equation}
is the empirical Rademacher complexity, defined with respect to a specific set of $N$ data samples. Then, in the case of binary classification and the class of linear separators, the generalisation error of a hypothesis, $\hat{\epsilon}_{N}(h)$, can be bounded using the Rademacher complexity:

\begin{equation}
\label{eq:daugreg-genbound}
 \hat{\epsilon}_{N}(h) \leq \mathcal{R}_{N}(\mathcal{H}) + \mathcal{O} \left( \sqrt{\frac{\ln \sfrac{1}{\delta}}{N}} \right)
\end{equation}
with probability $1 - \delta$. Tighter bounds for some model classes, such as fully connected neural networks, can be obtained \citep{bartlett2002rademacher}, but it is not trivial to formally analyse the influence on generalisation of specific architectures or techniques. Nonetheless, we can use these theoretical insights to discuss the differences between explicit regularisation---specifically weight decay and dropout---and data augmentation. 

A straightforward yet very relevant conclusion from the analysis of any generalisation bound is the strong dependence on the number of training examples $N$. Increasing $N$ drastically improves the generalisation guarantees, as reflected by the second term in the right hand side of Equation~\ref{eq:daugreg-genbound} and by the dependence of the Rademacher complexity (Equation~\ref{eq:daugreg-rademacher}) on the sample size too. Data augmentation exploits prior knowledge about the data domain and aspects of visual perception---in the case of image object recognition---to create new examples and its impact on generalisation is related to an increment in $N$, as stochastic data augmentation can generate virtually infinite different samples. Admittedly, the augmented samples are not independent and identically distributed and thus, the effective increment of samples does not strictly correspond to the increment in $N$. This is why formally analysing the impact of data augmentation on generalisation is complex. Recent studies have made progress in this direction by analysing the effect of simple data transformations on generalisation from a theoretical point of view \citep{chen2019invariance, rajput2019daug}.

Explicit regularisation methods aim, in contrast, at improving the generalisation error by constraining the hypothesis class $\mathcal{H}$ to reduce its complexity $\mathcal{R}_{N}(\mathcal{H})$ and, in turn, the generalisation error $\hat{\epsilon}_{N}(h)$. Crucially, while data augmentation exploits domain knowledge, most explicit regularisation methods only \textit{naively} constrain the hypothesis class, by simply reducing the representational capacity, as we have discussed in the previous chapter. For instance, weight decay constrains the learnable models $\mathcal{H}$ by setting a penalty on the weights norm. Interestingly, \citet{bartlett2017boundsnn} showed that weight decay has little impact on the generalisation bounds and confidence margins. Dropout has been extensively used and studied as a regularisation method for neural networks \citep{wager2013dropout}, but the exact way in which it impacts generalisation is still an open question. In fact, it has been stated that the effect of dropout on neural networks is ``somewhat mysterious'', complicated and its penalty highly non-convex \citep{helmbold2017dropout}. Recently, \citet{mou2018dropout} have established new generalisation bounds on the variance induced by a specific type of dropout on feedforward networks. 

An interesting observation is that dropout can be analysed as a random form of data augmentation without domain knowledge \citep{bouthillier2015dropoutasdaug}. This implies that any generalisation bound derived for dropout can be regarded as a pessimistic bound for domain-specific, standard data augmentation. A similar argument applies for weight decay, which, as first shown by \citet{bishop1995tikhonov}, is equivalent to training with noisy examples if the noise amplitude is small and the objective is the sum-of-squares error function. Therefore, some forms of explicit regularisation are at least approximately equivalent to adding random noise to the training examples, which is the simplest form of data augmentation\footnote{Note that the opposite view---domain-specific data augmentation as explicit regularisation---does not apply. In Section~\ref{sec:reg-taxonomies} we discuss about the taxonomies of regularisation, including the difference between data augmentation and data-dependent regularisation}. Thus, it is reasonable to argue that more sophisticated data augmentation can overshadow the benefits provided by explicit regularisation.

In general, we argue that the reason why explicit regularisation may not be necessary is that neural networks are already implicitly regularised by many elements---stochastic gradient descent (SGD), convolutional layers, normalisation and data augmentation, to name a few---that provide a more successful inductive bias \citep{neyshabur2014implicitreg}. For instance, it has been shown that linear models optimised with SGD converge to solutions with small norm, without any explicit regularisation \citep{zhang2016understandingdl}. Furthermore, as discussed in Section~\ref{sec:reg-discussion}, if overparameterised neural networks are able to generalise well, the need for constraining their capacity is questionable. In the rest of the chapter we present an empirical study to contrast data augmentation and explicit regularisation---weight decay and dropout.

\section{Methods}
\label{sec:daugreg-methods}
This section describes the main aspects of the experimental setup for systematically analysing the role of data augmentation in deep neural networks compared to weight decay and dropout.

\subsection{Data}
\label{sec:daugreg-methods_data}
We performed the experiments on the highly benchmarked data sets ImageNet \citep{russakovsky2015imagenet} ILSVRC 2012, CIFAR-10 and CIFAR-100 \citep{krizhevsky2009cifar}. We resized the 1.3 M images from ImageNet into $150\times200$ pixels, as a compromise between keeping a high resolution and speeding up the training. Both on ImageNet and on CIFAR, the pixel values were mapped into the range $[0, 1]$.  

So as to analyse the role of data augmentation, we trained every model with two different augmentation schemes as well as with no data augmentation at all. The two augmentation schemes are the following:

\subsubsection{\textit{Light} augmentation}
This scheme is common in the literature, for example \citep{goodfellow2013maxout, springenberg2014allcnn}, and performs only horizontal flips and horizontal and vertical translations of 10\% of the image size. 

\subsubsection{\textit{Heavier} augmentation}
This scheme performs a larger range of affine transformations such as scaling, rotations and shear mappings, as well as contrast and brightness adjustment. On ImageNet we additionally performed random crops of $128\times128$ pixels. The choice of the allowed transformations is arbitrary and the only criterion was that the objects be still recognisable in general. We deliberately avoided designing a particularly successful scheme. The details of the transformations are presented below and the range of the parameters are specified in Table~\ref{tab:daugreg-heavier_params} and some visual examples are shown in Figure~\ref{fig:daugreg-cifar10_daug}.

\begin{itemize}
  \item Affine transformations: 
  \vspace{5pt} \\
    $
      \begin{bmatrix}
      x^\prime \\
      y^\prime \\
      1
      \end{bmatrix}
      = 
      \begin{bmatrix}
      f_h z_x \cos(\theta) & -z_y \sin(\theta + \phi) & t_x \\
      z_x \sin(\theta) & z_y \cos(\theta + \phi) & t_y \\
      0 & 0 & 1
    \end{bmatrix}
    \begin{bmatrix}
      x \\
      y \\
      1
      \end{bmatrix}
    $
  \item Contrast adjustment: $x^\prime = \gamma (x - \overline{x}) + \overline{x}$

  \item Brightness adjustment: $x^\prime = x + \delta$
\end{itemize}

\begin{table}[ht]
\begin{center}
\begin{tabular}{cll}
\textbf{Parameter} & \textbf{Description}   & \textbf{Range}                                            \\
\hline \\
$f_h$              & Horiz. flip        & $1 - 2 B(0.5)$                                            \\
$t_x$              & Horiz. translation & $\mathcal{U}(-0.1, 0.1)$                                  \\
$t_y$              & Vert. translation   & $\mathcal{U}(-0.1, 0.1)$                                  \\
$z_x$              & Horiz. scale       & $\mathcal{U}(0.85, 1.15)$                                 \\
$z_y$              & Vert. scale         & $\mathcal{U}(0.85, 1.15)$                                 \\
$\theta$           & Rotation angle         & $\mathcal{U}(-22.5^\circ, 22.5^\circ)$  \\
$\phi$             & Shear angle            & $\mathcal{U}(-0.15, 0.15)$                                \\
$\gamma$           & Contrast               & $\mathcal{U}(0.5, 1.5)$                                   \\
$\delta$           & Brightness             & $\mathcal{U}(-0.25, 0.25)$                                
\end{tabular}
\end{center}
\caption{Description and range of possible values of the parameters used for the heavier augmentation scheme. $B(p)$ denotes a Bernoulli distribution and $\mathcal{U}(a, b)$ a uniform distribution.}
\label{tab:daugreg-heavier_params}
\end{table}

\begin{figure}[htb]
  \begin{center}
    \includegraphics[width = \textwidth]{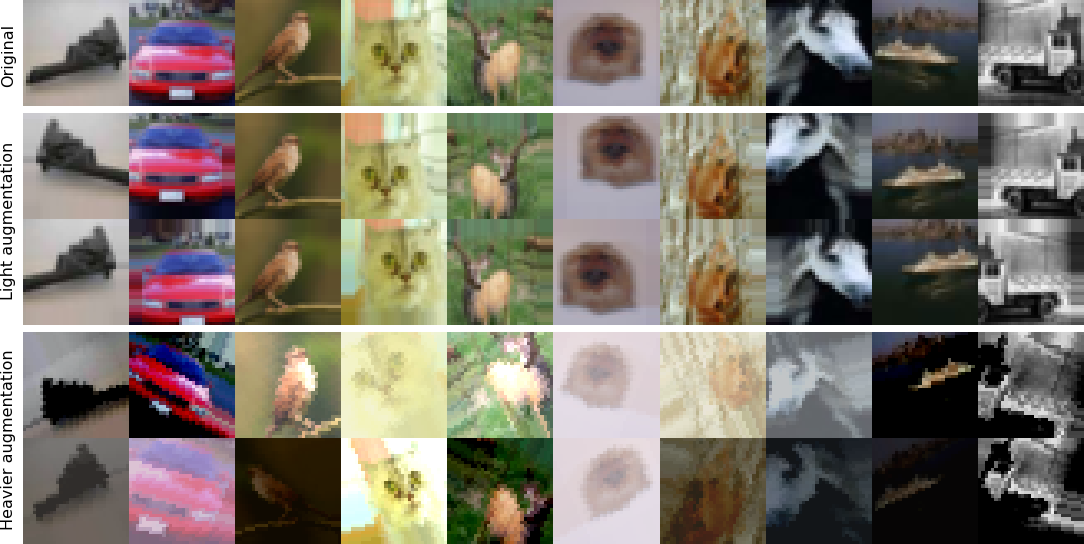}
  \end{center}
  \caption{Illustration of the most extreme transformations performed by the data augmentation schemes on ten images---one per class---from CIFAR-10.}
  \label{fig:daugreg-cifar10_daug}
\end{figure}

\subsection{Network Architectures}
\label{sec:daugreg-methods_archs}
We trained three distinct, popular architectures that have achieved successful results in visual object recognition: the all convolutional network, All-CNN \citep{springenberg2014allcnn}; the wide residual network, WRN \citep{zagoruyko2016wrn}; and the densely connected network, DenseNet \citep{huang2017densenet}. Importantly, we kept the training hyperparameters---learning rate, training epochs, batch size, optimiser, etc.---as in the original papers. Table~\ref{tab:architectures} summarises the main features of each network and below we specify further details.

\begin{table}[ht]
\begin{center}
\begin{tabular}{rccc}
    & \textbf{All-CNN} & \textbf{WRN} & \textbf{DenseNet}\\
    Ref. in original paper & \textit{All-CNN-C} & \textit{WRN-28-10} & \textit{DenseNet-BC}\\
    Main feature & Only conv. layers & Residual connections & Dense connectivity\\
    Number of layers & 16 / 12 & 28 & 101\\
    Number of parameters & 9.4 / 1.3 M & 36.5 M & 0.8 M\\
    Training hours & 35--45 / 2.5 & 100--145 / 14--15  & 24--27\\
    CO2e emissions\footnotemark (kg) & 4.17--5.36 / 0.29 & 11.91--17.27 / 1.66--1.78  & 2.86--3.21\\
\end{tabular}
\end{center}
\caption{Key aspects of the network architectures. Cells with two values correspond to ImageNet  / CIFAR.}
\label{tab:architectures}
\end{table}

\footnotetext{The carbon emissions were computed using the online calculator at \href{http://www.green-algorithms.org/}{green-algorithms.org} \cite{lannelongue2020carbonemissions}. The whole set of experiments in this chapter emmited an estimated total of 390.45 CO2e. The details about how the carbon emissions were calculated and about the impact of this study on global warming were made available as supplementary material of the main publication of this chapter .}

\subsubsection{All Convolutional Network}
All-CNN consists exclusively of convolutional layers with ReLU activation \citep{glorot2011relu}, it is relatively shallow and has few parameters. For ImageNet, the network has 16 layers and 9.4 million parameters; for CIFAR, it has 12 layers and about 1.3 million parameters. In our experiments to compare the adaptability of data augmentation and explicit regularisation to changes in the architecture (Section~\ref{sec:daugreg-depth}), we also tested a \textit{shallower} version, with 9 layers and 374,000 parameters, and a \textit{deeper} version, with 15 layers and 2.4 million parameters. The four architectures can be described as in Table~\ref{tab:allcnn}, where $K$\textbf{C}$D$($S$) is a $D \times D$ convolutional layer with $K$ channels and stride $S$, followed by batch normalisation and a ReLU non-linearity. \textit{N.Cl.} is the number of classes and Gl.Avg. refers to global average pooling. 

\begin{table}[ht]
\begin{center}
\begin{tabular}{l|l}
\multirow{1}{*}{ImageNet}  & 96\textbf{C}11(2)--96\textbf{C}1(1)--96\textbf{C}3(2)--256\textbf{C}5(1) \\
                           & --256\textbf{C}1(1)--256\textbf{C}3(2)--384\textbf{C}3(1) \\
                           & --384\textbf{C}1(1)--384\textbf{C}3(2)--1024\textbf{C}3(1) \\
                           & --1024\textbf{C}1(1)--\textit{N.Cl}.C1(1) \\
                           & --Gl.Avg.--Softmax \\ [3pt]
\multirow{1}{*}{CIFAR}     & ~2$\times$96\textbf{C}3(1)--96\textbf{C}3(2)--2$\times$192\textbf{C}3(1) \\
                           & --192\textbf{C}3(2)--192\textbf{C}3(1)--192\textbf{C}1(1) \\
                           & --\textit{N.Cl}.C1(1)--Gl.Avg.--Softmax \\ [3pt]
\multirow{1}{*}{Shallower} & ~2$\times$96\textbf{C}3(1)--96\textbf{C}3(2)--192\textbf{C}3(1) \\
                           & --192\textbf{C}1(1)--\textit{N.Cl}.C1(1)--Gl.Avg.--Softmax \\ [3pt]
\multirow{1}{*}{Deeper}    & ~2$\times$96\textbf{C}3(1)--96\textbf{C}3(2)--2$\times$192\textbf{C}3(1) \\
                           & --192\textbf{C}3(2)--2$\times$192\textbf{C}3(1)--192\textbf{C}3(2) \\
                           & --192\textbf{C}3(1)--192\textbf{C}1(1) \\
                           & --\textit{N.Cl}.C1(1)--Gl.Avg.--Softmax \\
\end{tabular}
\end{center}
\caption{Specification of the All-CNN architectures.}
\label{tab:allcnn}
\end{table}

The CIFAR network is identical to the All-CNN-C architecture in the original paper, except for the introduction of the batch normalisation layers \citep{ioffe2015batchnorm}, which we included because they generally improve performance, but had not been proposed at the time of publication of All-CNN \citep{springenberg2014allcnn}. The ImageNet version also includes batch normalisation layers and a stride of 2 instead of 4 in the first layer to compensate for the reduced input size. 

Importantly, we kept the same training parameters as in the original paper in the cases they were reported. Specifically, the All-CNN networks were trained using stochastic gradient descent, with fixed Nesterov momentum 0.9, learning rate of 0.01 and decay factor of 0.1. The batch size for the experiments on ImageNet was 64 and we trained during 25 epochs decaying the learning rate at epochs 10 and 20. On CIFAR, the batch size was 128, we trained for 350 epochs and decayed the learning rate at epochs 200, 250 and 300. The kernel parameters were initialised according to the Xavier uniform initialisation \citep{glorot2010glorot}.

\subsubsection{Wide Residual Network}
WRN is a modification of ResNet \citep{he2016resnet} that achieves better performance with fewer layers, but more units per layer. Here, we chose for our experiments the WRN-28-10 version (28 layers and about 36.5 M parameters), which was reported to achieve the best results on CIFAR. It has the following architecture:

\begin{center}
\centering
16\textbf{C}3(1)--4$\times$160\textbf{R}--4$\times$320\textbf{R}--4$\times$640\textbf{R}--BN--ReLU--Avg.(8)--FC--Softmax
\end{center}
where $K$\textbf{R} is a residual block with residual function  BN--ReLU--$K$\textbf{C}3(1)--BN--ReLU--$K$\textbf{C} 3(1). BN is batch normalisation, Avg.(8) is spatial average pooling of size 8 and FC is a fully connected layer. On ImageNet, the stride of the first convolution is 2. The stride of the first convolution within the residual blocks is 1 except in the first block of the series of 4, where it was set to 2 in order to subsample the feature maps. 

Similarly, we kept the training parameters of the original paper: we trained with SGD, with fixed Nesterov momentum 0.9 and learning rate of 0.1. On ImageNet, the learning rate was decayed by 0.2 at epochs 8 and 15 and we trained for a total of 20 epochs with batch size 32. On CIFAR, we trained with a batch size of 128 during 200 epochs and decayed the learning rate at epochs 60, 120 and 160. The kernel parameters were initialised according to the He normal initialisation \citep{he2015he}.

\subsubsection{DenseNet}

The main characteristic of DenseNet \citep{huang2017densenet} is that the architecture is arranged into blocks whose layers are connected to all the layers below, forming a dense graph of connections, which permits training very deep architectures with fewer parameters than, for instance, ResNet. Here, we used a network with bottleneck compression rate $\theta = 0.5$ (DenseNet-BC), growth rate $k = 12$ and 16 layers in each of the three blocks. The model has nearly 0.8 million parameters. The specific architecture can be described as follows:

\begin{center}
\centering
2$\times k$\textbf{C}3(1)--DB(16)--TB--DB(16)--TB--DB(16)--BN--Gl.Avg.--FC--Softmax
\end{center}
where DB($c$) is a dense block, that is a concatenation of $c$ convolutional blocks. Each convolutional block is a set of layers whose output is concatenated with the input to form the input of the next convolutional block. A convolutional block with bottleneck structure has the following layers:

\begin{center}
\centering
BN--ReLU--4$\times k$\textbf{C}1(1)--BN--ReLU--$k$\textbf{C}3(1)--Concat.
\end{center}

TB is a transition block, which downsamples the size of the feature maps, formed by the following layers:

\begin{center}
\centering
BN--ReLU--$k$\textbf{C}1(1)--Avg.(2).
\end{center}

Like with All-CNN and WRN, we kept the training hyperparameters of the original paper. On the CIFAR data sets, we trained with SGD, with fixed Nesterov momentum 0.9 and learning rate of 0.1, decayed by 0.1 on epochs 150 and 200 and training for a total of 300 epochs. The batch size was 64 and the parameters were initialised with He initialisation.

\subsection{Train and Test}
Every architecture was trained on each data set both with explicit regularisation---weight decay and dropout as specified in the original papers---and without. Furthermore, we trained each model with the three data augmentation schemes: no augmentation, light and heavier. Figure~\ref{fig:experimental_setup} shows a summary of this experimental setup. The performance of the models was computed on the held out test tests. As in previous works \citep{krizhevsky2012alexnet, simonyan2014}, we averaged the softmax posteriors over 10 random \textit{light} augmentations, since slightly better results are obtained. Then we computed the classification accuracy for the models trained on CIFAR and the top-5 accuracy for the ImageNet models.

All the experiments were performed on Keras \citep{chollet2015keras} on top of TensorFlow \citep{tensorflow2015}, with a single GPU NVIDIA GeForce GTX 1080 Ti.

\begin{figure}[htb]
  \begin{center}
    \includegraphics[width = \textwidth]{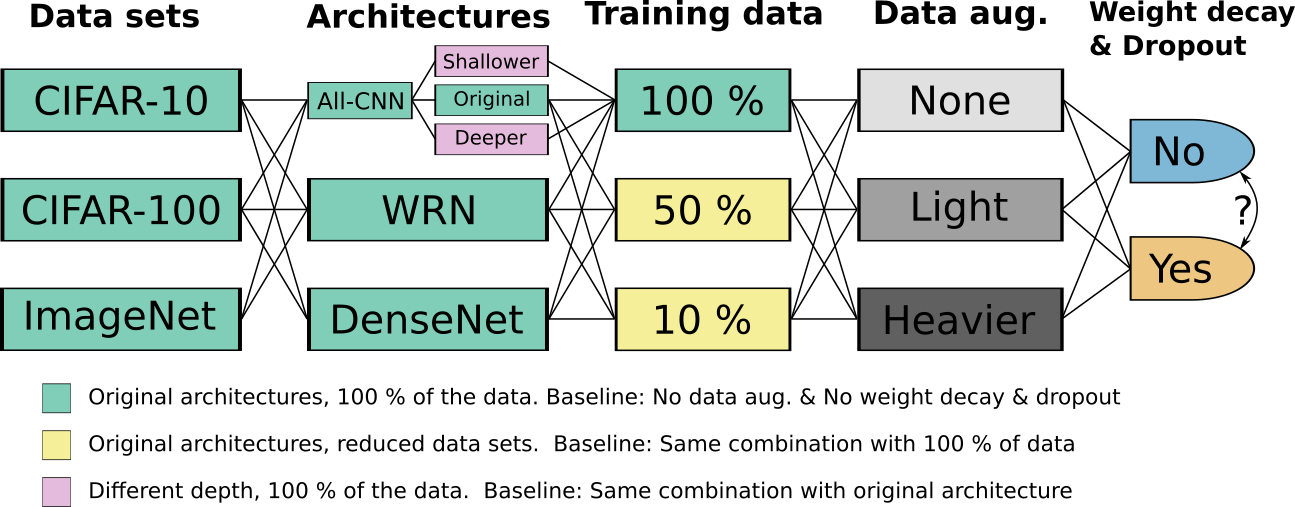}
  \end{center}
  \caption{Visual summary of the experimental setup. The figure represents the factors of variation in our experiments: data sets, architectures, amount of training data, data augmentation scheme and inclusion of explicit regularization. Comparisons within a factor of variation are most relevant on the factors on the right, like the performance of the models train with and without explicit regularization.}
  \label{fig:experimental_setup}
\end{figure}

\subsection{Carbon footprint of the computational experiments}
\label{sec:daugreg-carbon_footprint}

Training artificial neural networks effectively on large, non-trivial data sets consumes a considerable amount of energy \citep{strubell2019energydl}. Crucially, the amount of compute of the largest models has been increasing exponentially during the last decade \citep{amodei2018energyai}. Therefore, the contribution of deep learning research to global warming and climate change should not be neglected \citep{schwartz2019greenai, lacoste2019carbonemissions, lannelongue2020carbonemissions}. As the experimental design of this chapter required training multiple neural network models, we wanted to be both aware and transparent about the environmental impact of our experimental study and in this section we will report calculations of the estimated carbon emissions associated with training our models for this chapter and the details of how these are computed.

In Table~\ref{tab:architectures} of Section~\ref{sec:daugreg-methods_archs} we report the estimated carbon emissions associated with training each architecture on each data set, given the specific characteristics of our computing hardware. These estimations rely on the online calculator available at \href{http://www.green-algorithms.org/}{green-algorithms.org}, developed by \citet{lannelongue2020carbonemissions}. In order to estimate the carbon emissions of each model, we took into consideration the following information: all the models were trained in a local desktop computer, located in Germany, with a single graphic processing unit (GPU), model GTX 1080 Ti, with 11 GB of memory. We assumed full usage of the 11 GB of memory and of the processing core for all models---a conservative estimation. 

\begin{table}[htbp]
\caption{Summary of estimated carbon emissions associated to training the models for our experimental setup}
\begin{center}
\begin{tabular}{ccccccc}
Network                                       & Data set                                                            & Depth                                & \% Data & N. models    & Total h.        & Total CO2e      \\ \hline
\multirow{8}{*}{All-CNN}                      & \multirow{5}{*}{\makecell{CIFAR\\2.5 h\\0.29 CO2e}}                 & \multirow{3}{*}{\makecell{original}} & 100~\%  & 36           & 90              & 10.73           \\
                                              &                                                                     &                                      & 50~\%   & 24           & 30              & 3.58            \\
                                              &                                                                     &                                      & 10~\%   & 24           & 6               & 0.72            \\ \cline{3-7}
                                              &                                                                     & shallower                            & 100~\%  & 12           & 25.2            & 3.0             \\ \cline{3-7}
                                              &                                                                     & deeper                               & 100~\%  & 12           & 36              & 4.29            \\ \cline{2-7}
                                              & \multirow{3}{*}{\makecell{ImageNet\\35--45 h\\4.17--5.36 CO2e}}     & \multirow{3}{*}{\makecell{original}} & 100~\%  & 6            & 270             & 32.18           \\
                                              &                                                                     &                                      & 50~\%   & 6            & 135             & 16.09           \\
                                              &                                                                     &                                      & 10~\%   & 6            & 27              & 3.22            \\ \hline
\multirow{6}{*}{WRN}                          & \multirow{3}{*}{\makecell{CIFAR\\14--15 h\\1.66--1.78 CO2e}}        & \multirow{3}{*}{\makecell{original}} & 100~\%  & 36           & 540             & 64.35           \\
                                              &                                                                     &                                      & 50~\%   & 12           & 90              & 10.73           \\
                                              &                                                                     &                                      & 10~\%   & 12           & 18              & 2.15            \\ \cline{2-7}
                                              & \multirow{3}{*}{\makecell{ImageNet\\100--145 h\\11.91--17.27 CO2e}} & \multirow{3}{*}{\makecell{original}} & 100~\%  & 6            & 870             & 103.68          \\
                                              &                                                                     &                                      & 50~\%   & 6            & 435             & 51.84           \\
                                              &                                                                     &                                      & 10~\%   & 6            & 87              & 10.37           \\ \hline
\multirow{3}{*}{DenseNet}                     & \multirow{3}{*}{\makecell{CIFAR\\24--27 h\\2.86--3.21 CO2e}}        & \multirow{3}{*}{\makecell{original}} & 100~\%  & 12           & 324             & 38.61           \\
                                              &                                                                     &                                      & 50~\%   & 6            & 81              & 9.65            \\
                                              &                                                                     &                                      & 10~\%   & 6            & 16.2            & 1.93            \\ \hline
\vspace{-5pt}\\
\multicolumn{4}{c}{\textbf{Total}}                                                                                                                                   & \textbf{228} & \textbf{3080.4} & \textbf{367.12} 
\end{tabular}
\end{center}
\label{tab:carbon_emissions}
\end{table}

As reported in Table~\ref{tab:carbon_emissions}, the complete set of experiments reported in this chapter needed a total of 3,276 GPU hours (136.5 days), which correspond to actual real time, since we had access to a single GPU. With our hardware, this corresponds, according to \cite{lannelongue2020carbonemissions}, to an estimate of 832.53 kWh or 390.45 carbon dioxide equivalent (CO2e). Carbon dioxide equivalent represents the equivalent CO2 that would have the same global warming impact than a mixture of gases. By way of comparison, 390.45 CO2e corresponds to 34.25 tree-years---the time taken by a mature tree to absorb the CO2---, 69~\% of a flight New York City--San Francisco or 2,231 km in a passenger car.

\section{Results}
\label{sec:daugreg-results}
Here we present the results of the empirical study. In the first set of experiments (Section~\ref{sec:daugreg-orig}) we trained the architectures as in the original papers with the full data sets. A relevant characteristic of explicit regularisation methods is that they typically require the specification of hyperparameters. These are usually fine-tuned by the authors of research papers to achieve higher performance, as demanded by the dynamics of the scientific publication environment in the machine learning community. However, the sensitivity of the results towards these hyperparameters is often not made available. In order to gain insight on the role of explicit regularisation and data augmentation in more real world cases, where the hyperparameters have not been highly optimised, we varied the amount of training data (Section~\ref{sec:daugreg-less_data}) and the depth of the architectures (Section~\ref{sec:daugreg-depth}), while keeping all other hyperparameters untouched.

The objective of the study is to contrast the performance gained by training the models with both explicit regularisation and data augmentation, which is the common practice in the literature \citep{tan2019efficientnet, huang2017densenet, zagoruyko2016wrn, springenberg2014allcnn}, against training with only data augmentation. Hence, the presentation of the results in the figures aims at facilitating this comparison. In the performance plots, we represent the relative performance gain of each model with respect to the relevant baseline, which we specify at each section. We plot the results in pairs: the squared blue dots on the top, blue-shaded area correspond to the models trained with only data augmentation and the round orange dots on the bottom, orange-shaded area to the models trained with both data augmentation and explicit regularization. Additionally, the results of training with different levels of data augmentation are represented with dots in three lightness and saturation shades and we connected with dotted lines the models trained with the same level of augmentation.

In order to assess the statistical significance of the differences between models trained with and without explicit regularisation, we carried out percentile bootstrap analyses \citep{efron1992bootstrap}, that is simulations based on sampling with replacement. We followed the guidelines by \citet{rousselet2019bootstrap}. In all cases, the values of the distribution correspond to the difference between the performance---with respect to the baseline---of the models trained without explicit regularisation minus the performance of the models trained with explicit regularisation---the pairs of dots connected by a dotted line. We then compared the distribution of this difference in the bootstrap samples and with respect to the null hypothesis, that is no difference ($H_0 = 0$). For each experiment we sampled all possible bootstrap samples with replacement or a maximum of one million.

\subsection{Original architectures}
\label{sec:daugreg-orig}
\begin{figure}[htb]
  \begin{center}
    \includegraphics[width = \textwidth]{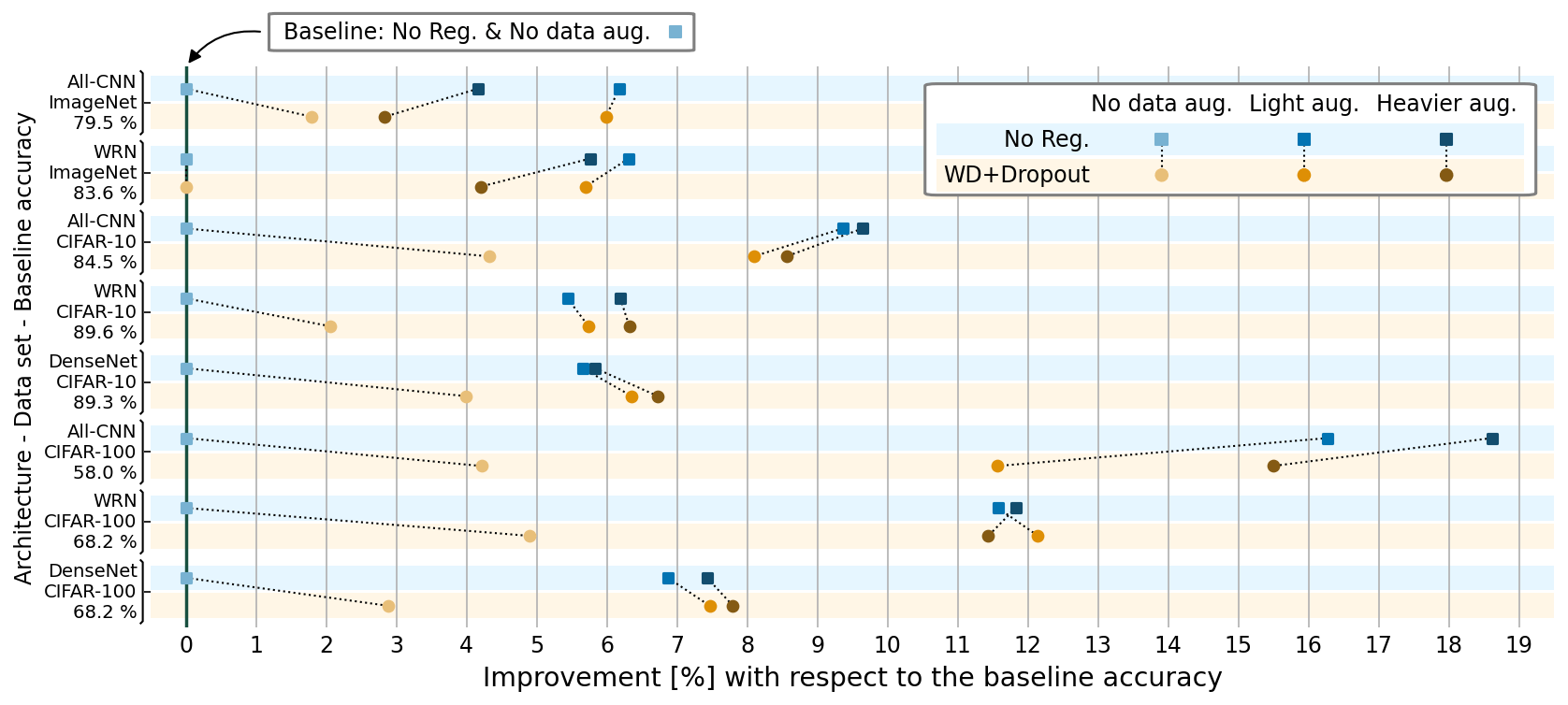}
  \end{center}
  \caption{Relative improvement of adding data augmentation and explicit regularization to the baseline models, $(accuracy - baseline)/accuracy * 100$. The baseline accuracy is shown on the left. The results suggest that data augmentation alone (in blue) can achieve even better performance than the models trained with both weight decay and dropout (in orange).}
  \label{fig:daugreg-orig}
\end{figure}

First, we contrast the regularisation effect of data augmentation and weight decay and dropout on the original networks trained with the complete data sets, and show the results in Figure~\ref{fig:daugreg-orig}. As a baseline, we consider the ``bare bone'' models, that is the model trained with neither explicit regularisation nor data augmentation. We report the accuracy of the baseline on the left axis of the plot in Figure~\ref{fig:daugreg-orig}. To assess the relevant comparisons, we show the relative improvement in test performance achieved by adding each technique or combination of techniques to the baseline model. Table~\ref{tab:daugreg-orig_nets} shows the mean and standard deviation of each combination on the architecture and data set and Figure~\ref{fig:daugreg-bootstrap_orig} the results of the bootstrap analysis, which considers the differences of all pairs---squared blue dots minus round orange dots, connected with dotted lines\footnote{The relative performance of WRN on ImageNet trained with weight decay and dropout with respect to the baseline is negative (-6.22~\%) and is neither depicted in Figure~\ref{fig:daugreg-orig} nor taken into consideration to compute the average improvements in Table~\ref{tab:daugreg-orig_nets} and the bootstrap analysis in Figure~\ref{fig:daugreg-bootstrap_orig}.}.

\begin{figure}[ht]
  \centering
  \begin{center}
    \includegraphics[width = \textwidth]{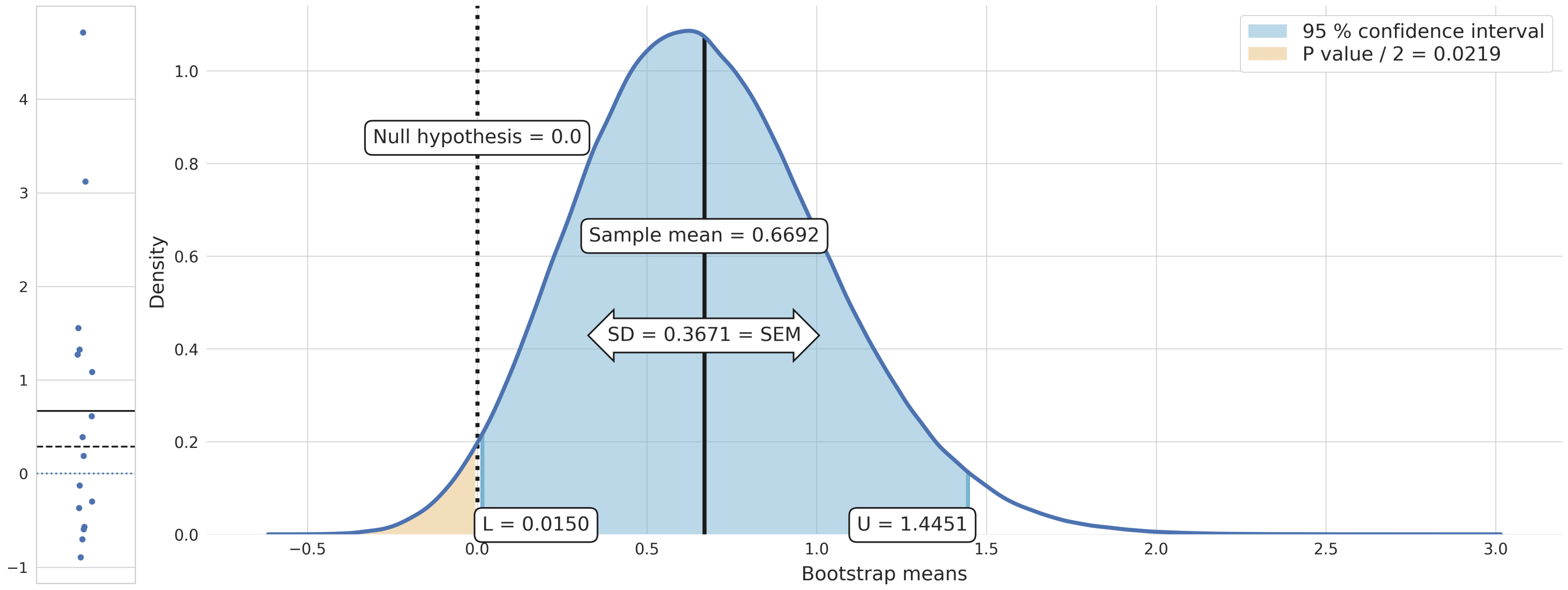}
  \end{center}
  \caption{Bootstrap analysis to assess the difference in performance gain provided by training without and with weight decay and dropout, on the original architectures and using the full data sets. On the left of the figure we plot the bootstrap values---differences---with the mean and median as a solid and dashed line, respectively. The main figure shows the distribution of the mean of the bootstrap samples, the standard error of the sample mean, the 95~\% confidence intervals and the $P$ value with respect to the null hypothesis ($H_0=0$).}
  \label{fig:daugreg-bootstrap_orig}
\end{figure}

\begin{table}[hb]
\begin{center}
\begin{tabular}{rcc}
            & No explicit reg.  & Weight decay + dropout \\
    None    & \textit{baseline} & 3.02 (1.65)            \\
    Light   & 8.46 (3.80)       & 7.88 (2.60)            \\
    Heavier & 8.68 (4.69)       & 7.92 (4.03) 
\end{tabular}
\end{center}
\caption{Average accuracy improvement over the baseline model of each combination of data augmentation level and presence of weight decay and dropout.}
\label{tab:daugreg-orig_nets}
\end{table}

The first conclusion from Figures~\ref{fig:daugreg-orig} and \ref{fig:daugreg-bootstrap_orig} as well as Table~\ref{tab:daugreg-orig_nets} is that training with data augmentation alone (blue dots on the top, blue-shaded areas) is better than training with both augmentation and explicit regularisation (in orange). This is the case in more than half of the cases (9/16) and the bootstrap analysis reveals that the difference is positive with 95~\% confidence and $P$ value $=0.022$. On average, adding data augmentation to the baseline model improved the accuracy on 8.57~\%, and adding both augmentation and explicit regularisation on 7.90~\% respectively.

At first glance, one may think that this is not remarkable, since the differences are small and data augmentation alone is not better in 100~\% of the cases. However, this result is surprising and remarkable for the following reason: note that the studied architectures achieved state-of-the-art results at the moment of their publication and the models included all light augmentation, weight decay and dropout, whose parameters were presumably finely tuned to optimise the accuracy. The replication of these results corresponds to the mid-orange dots in Figure~\ref{fig:daugreg-orig}. Here, we have show that simply removing weight decay and dropout---while keeping all other hyperparameters intact, see Section~\ref{sec:daugreg-methods_archs}---improves the \textit{then state-of-the-art} accuracy in 4 of the 8 studied cases. Why did not the authors trained without explicit regularisation and obtain better results?

Second, it can also be observed that the regularisation effect of weight decay and dropout, an average improvement of 3.02~\% with respect to the baseline, is much smaller than that of data augmentation: simply applying light augmentation increased the accuracy in 8.46~\% on average. Although the heavier augmentation scheme was deliberately not designed to optimise the performance, in both CIFAR-10 and CIFAR-100 it improved the test performance with respect to the light augmentation scheme. This was not the case on ImageNet, probably due to the larger complexity of the data set. 

\begin{figure}[htb]
  \begin{center}
    \includegraphics[width = \textwidth]{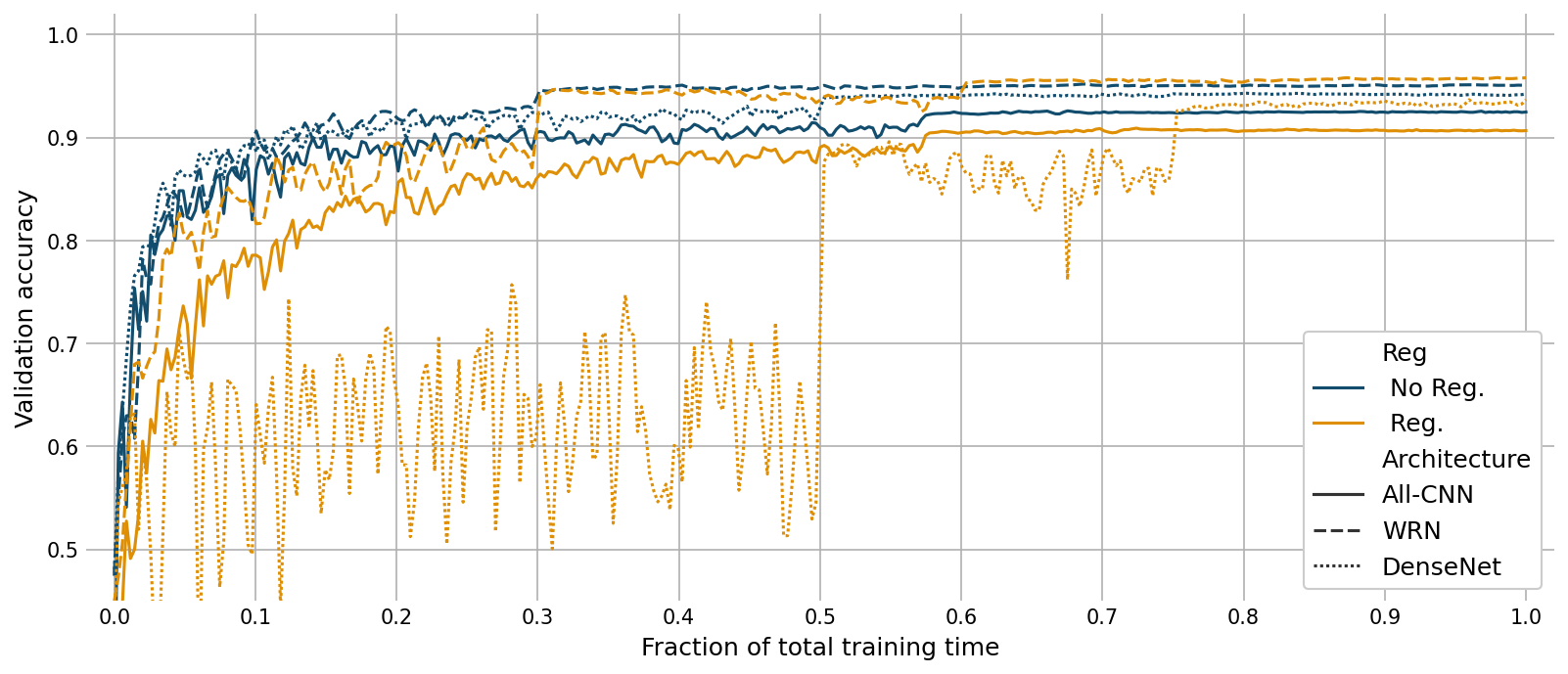}
  \end{center}
  \caption{Dynamics of the validation accuracy during training of All-CNN, WRN and DenseNet, trained on CIFAR-10 with heavier data augmentation, contrasting the models trained with explicit regularization (orange lines) and the models trained with only data augmentation (in blue). The regularized models heavily rely on the learning rate decay to obtain the boost of performance, while the models trained without explicit regularization quickly approach the final performance.}
  \label{fig:daugreg-dynamics}
\end{figure}

Further, it can be observed that the results are in general more consistent in the models trained without explicit regularisation. Finally, an additional advantage of training without explicit regularisation is that the learning dynamics (Figure~\ref{fig:daugreg-dynamics}) is much faster and predictive of the final performance. Typically, regularisers such as weight decay and dropout effectively prevent the model from fitting the training data during the first epochs and heavily rely on the learning rate decay to obtain the boost that yields the final performance. On the contrary, models trained with only data augmentation reach very high validation performance after a few epochs. This effect is particularly acute on DenseNet, which performs heavier weight decay.

In sum, it seems the performance gain provided by weight decay and dropout can be achieved and often improved by data augmentation alone. Besides, the models trained without explicit regularisation presented additional advantages, which we will further discuss in Section~\ref{sec:daugreg-discussion}.

\subsection{When the available training data changes}
\label{sec:daugreg-less_data}
\begin{figure}[ht]
  \centering
  \begin{subfigure}{\linewidth}
      \includegraphics[keepaspectratio=true, width=\columnwidth]{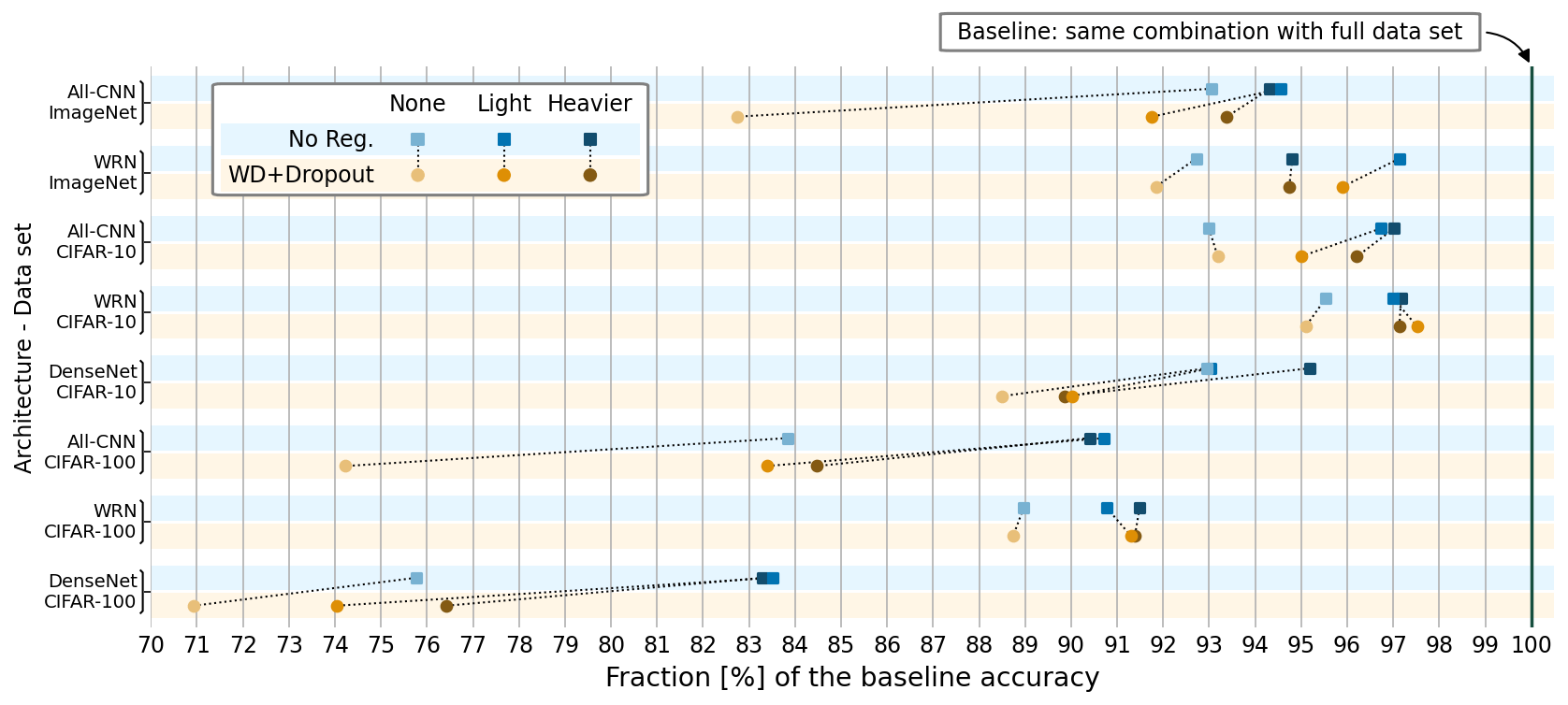}
      \caption{50~\% of the available training data}
	  \label{fig:daugreg-less_data_50}
  \end{subfigure}
  \\ 
  \begin{subfigure}{\linewidth}
      \includegraphics[keepaspectratio=true, width=\columnwidth]{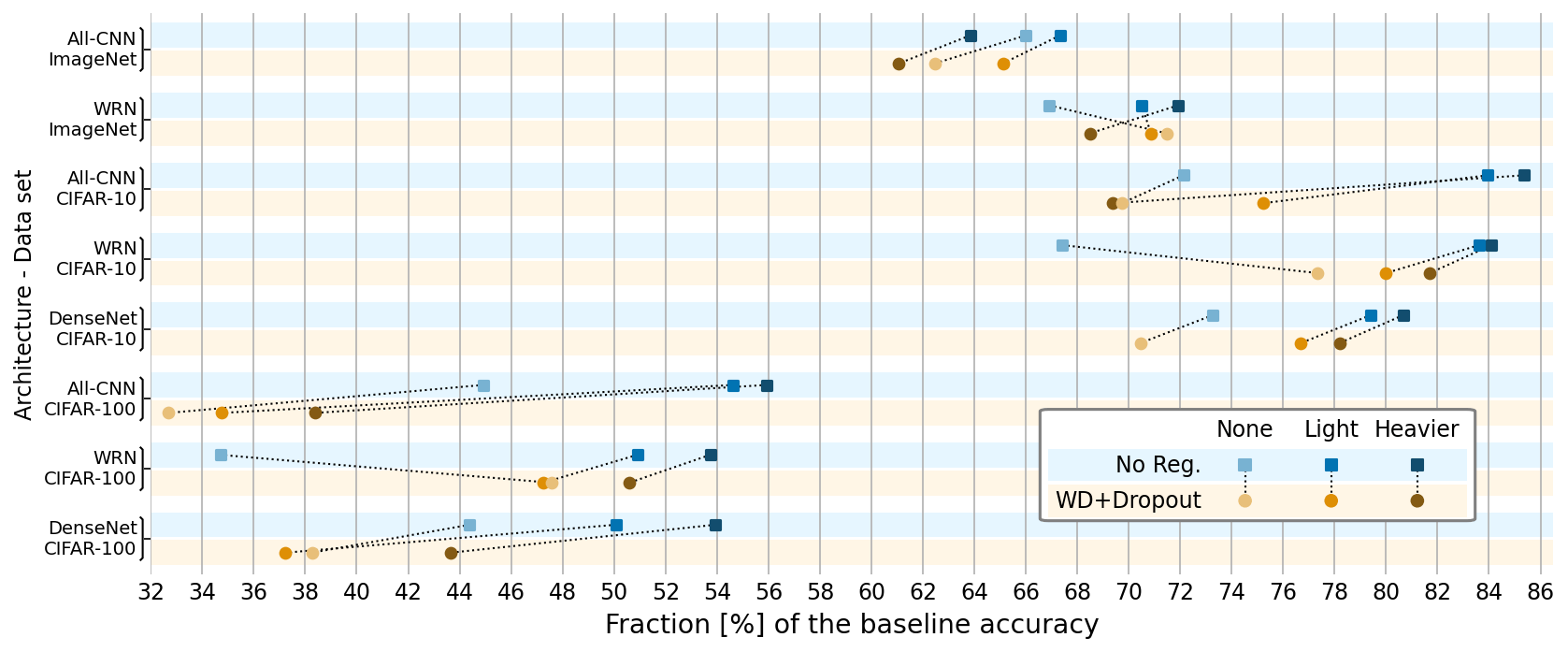}
      \caption{10~\% of the available training data}
	  \label{fig:daugreg-less_data_10}
  \end{subfigure}
  \caption{Fraction of the baseline performance when the amount of available training data is reduced, $accuracy/baseline * 100$. The models trained wit explicit regularisation present a significant drop in performance as compared to the models trained with only data augmentation. The differences become larger as the amount of training data decreases.}
  \label{fig:daugreg-less_data}
\end{figure}

We argue that one of the main drawbacks of explicit regularisation techniques is their poor adaptability to changes in the conditions with which the hyperparameters were tuned. To test this hypothesis and contrast it with the adaptability of data augmentation, we extended the analysis by training the same networks with fewer examples. All models were trained with the same random subset of data and evaluated in the same test set as the previous experiments. In order to better visualise how well each technique resists the reduction of training data, in Figure~\ref{fig:daugreg-less_data} we show the fraction of baseline accuracy achieved by each model when trained with 50~\% and 10~\% of the available data. In this case, the baseline is thus each corresponding model trained with the complete data set. Table~\ref{tab:daugreg-less_data} summarises the mean and standard deviation of each combination and Figure~\ref{fig:daugreg-bootstrap_less_data} shows the result of the bootstrap analysis.

\begin{table}[htb]
  \begin{center}
    \begin{tabular}{rcc}
      & \multicolumn{2}{c}{50~\% of the training data}    \\
      \cline{2-3} 
              & No explicit reg. & Weight decay + dropout \\
      None    & 88.11 (6.27)     & 83.20 (9.83)           \\
      Light   & 91.47 (4.31)     & 88.27 (7.39)           \\
      Heavier & 91.82 (4.63)     & 89.28 (6.63)           \\
      \vspace{-7pt}\\
      & \multicolumn{2}{c}{10~\% of the training data}    \\
      \cline{2-3} 
              & No explicit reg. & Weight decay + dropout \\
      None    & 58.72 (14.93)    & 58.75 (16.92)          \\
      Light   & 67.55 (14.27)    & 60.89 (18.39)          \\
      Heavier & 68.69 (13.61)    & 61.43 (15.90) 
    \end{tabular}
  \end{center}
  \caption{Average fraction of the original accuracy of each corresponding combination of data augmentation level and presence of weight decay and dropout.}
  \label{tab:daugreg-less_data}
\end{table}

\begin{figure}[ht]
  \centering
  \begin{subfigure}{\linewidth}
      \includegraphics[width = \textwidth]{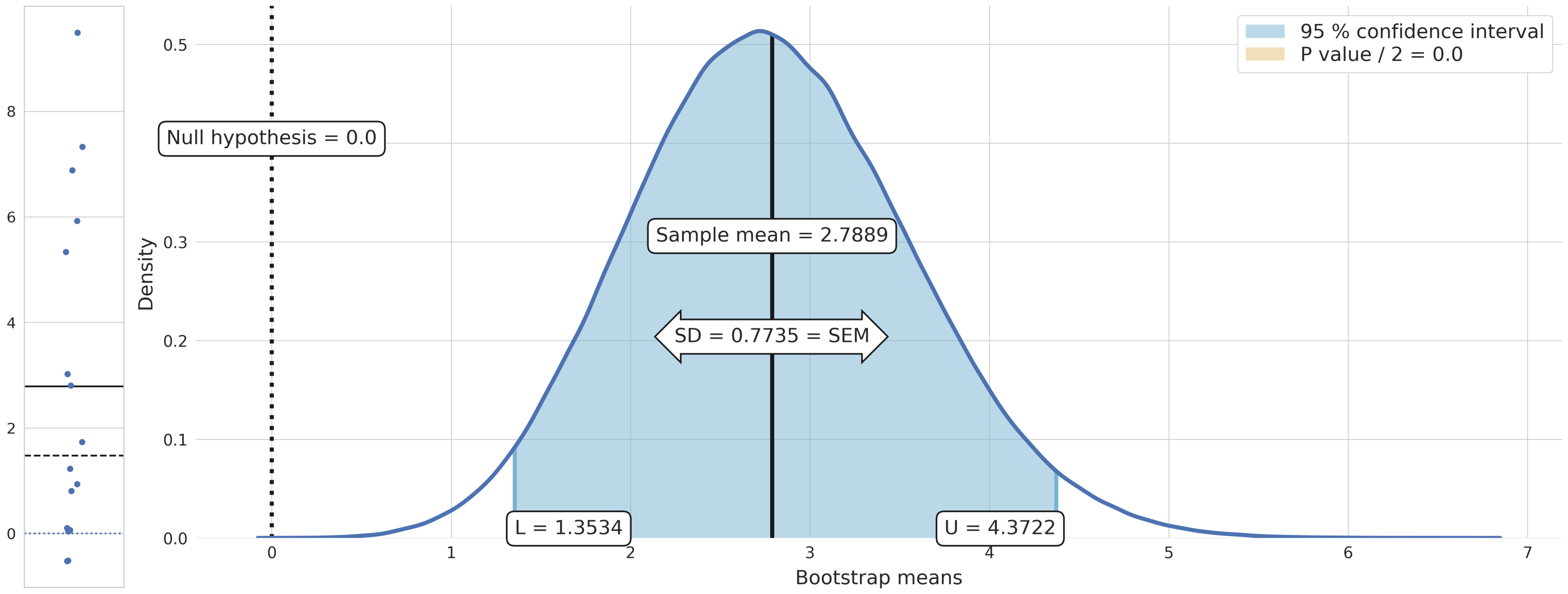}
      \caption{50~\% of the available training data}
      \label{fig:daugreg-bootstrap_means_50}
  \end{subfigure}
  \\
  \begin{subfigure}{\linewidth}
      \includegraphics[width = \textwidth]{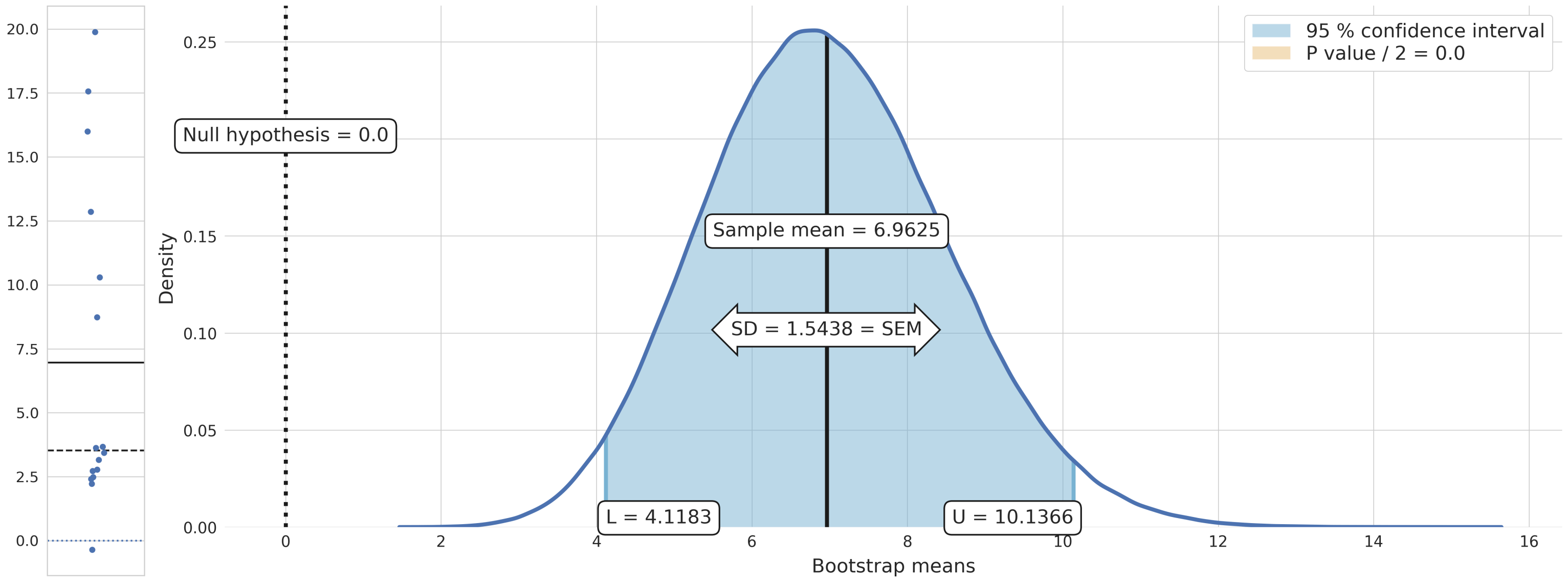}
      \caption{10~\% of the available training data}
      \label{fig:daugreg-bootstrap_means_10}
  \end{subfigure}
  \caption{Bootstrap analysis analogous to the one detailed in Section~\ref{sec:daugreg-orig} and Figure~\ref{fig:daugreg-bootstrap_orig}, to analyse the statistical significance of the performance difference of models trained with only 10 and 50~\% of the data.}
  \label{fig:daugreg-bootstrap_less_data}
\end{figure}

One of the main conclusions of this set of experiments is that if no data augmentation is applied, explicit regularisation hardly resists the reduction of training data by itself. On average, with 50~\% of the available data, these models only achieve 83.20~\% of the original accuracy (Table~\ref{tab:daugreg-less_data}), which, remarkably, is even worse than the models trained without any explicit regularisation (88.11~\%). On 10~\% of the data, the average fraction is the same (58.75 and 58.72~\%, respectively). This implies that training with explicit regularisation is even detrimental for the performance.

When combined with data augmentation, the models trained with explicit regularisation (orange dots) also perform worse (88.78 and 61.16~\% with 50 and 10~\% of the data, respectively), than the models without explicit regularisation (blue dots, 91.64 and 68.12~\% on average). Note that the difference becomes larger as the amount of available data decreases. Even more decisive are the results of the bootstrap analysis (Figure~\ref{fig:daugreg-bootstrap_less_data}): the mean difference of the fraction of the performance achieved by the models trained without and with explicit regularisation is 2.78 and 6.96, with 50 and 10~\% of the training data, respectively; the confidence intervals are well above the null hypothesis and the $P$ values are exactly 0.

Importantly, it seems that the combination of explicit regularisation and data augmentation is only slightly better than training without data augmentation. Two reasons may explain this: first, the original regularisation hyperparameters seem to adapt poorly to the new conditions. The hyperparameters were specifically tuned for the original setup and they would require re-tuning to obtain comparable results. Second, since explicit regularisation reduces the representational capacity, this might prevent the models from taking advantage of the augmented data.

In contrast, the models trained without explicit regularisation but with data augmentation more naturally adapt to the reduced availability of data. With 50~\% of the data, these models achieve about 91.5~\% of the performance with respect to training with the complete data sets. With only 10~\% of the data, they achieve nearly 70~\% of the baseline performance, on average. This highlights the suitability of data augmentation to serve, to a great extent, as true, useful data \citep{vinyals2016oneshot}.

\subsection{When the architecture changes}
\label{sec:daugreg-depth}
\begin{figure}[tb]
  \begin{center}
    \includegraphics[width = \linewidth]{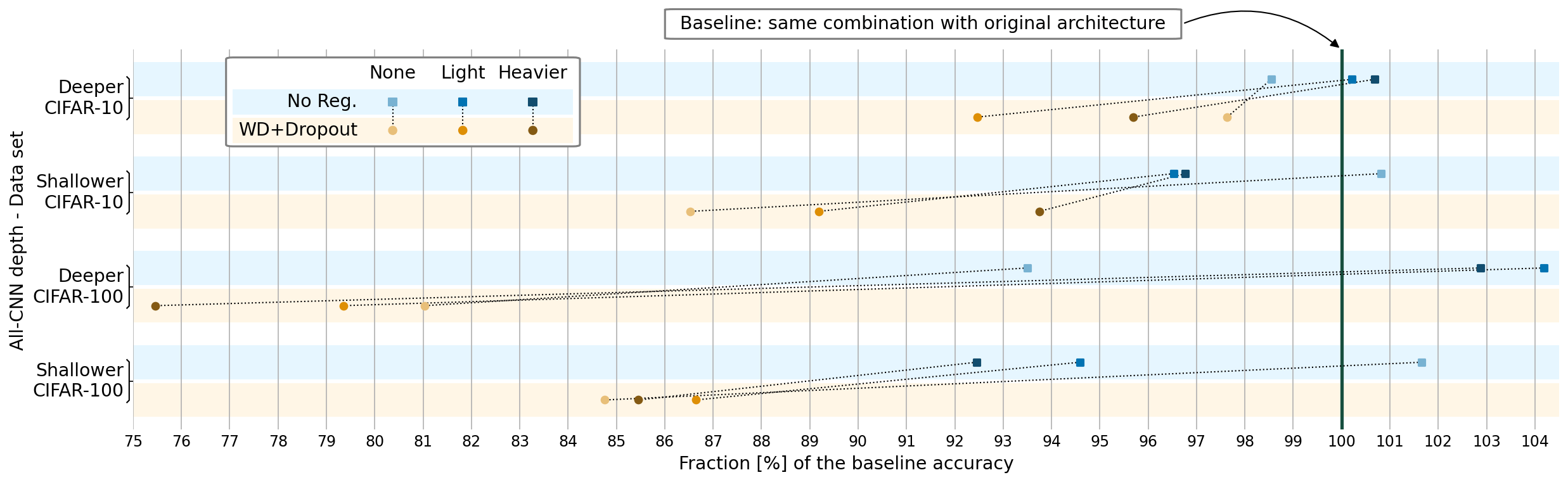}
  \end{center}
  \caption{Fraction of the original performance when the depth of the All-CNN architecture is increased or reduced in 3 layers. In the explicitly regularised models, the change of architecture implies a dramatic drop in the performance, while the models trained without explicit regularisation present only slight variations with respect to the original architecture.}
  \label{fig:daugreg-depth}
\end{figure}

Finally, in the same spirit, we tested the adaptability of data augmentation and explicit regularisation to changes in the depth of the All-CNN architecture, by training shallower (9 layers) and deeper (15 layers) versions of the architecture. We show the fraction of the performance with respect to the original architecture in Figure~\ref{fig:daugreg-depth} and the bootstrap analysis in Figure~\ref{fig:daugreg-bootstrap_depth}.

\begin{figure}[ht]
  \begin{center}
    \includegraphics[width = \textwidth]{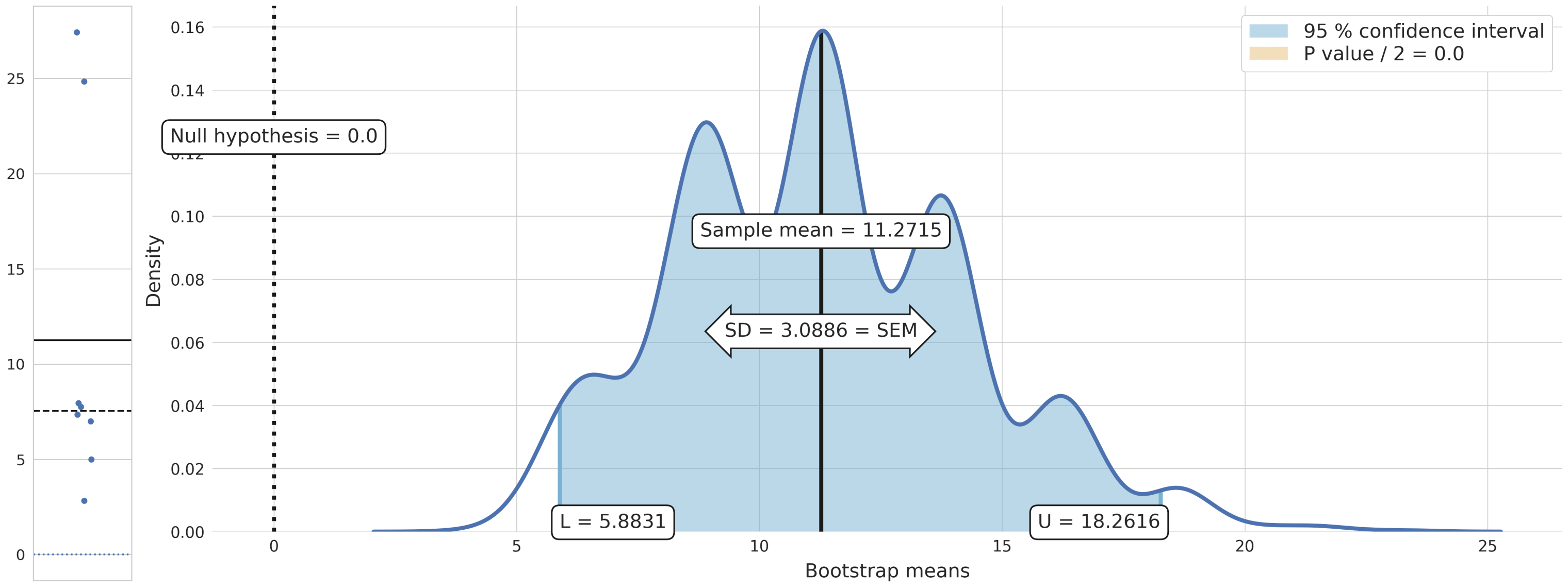}
  \end{center}
  \caption{Bootstrap analysis analogous to the one detailed in Section~\ref{sec:daugreg-orig} and Figure~\ref{fig:daugreg-bootstrap_orig}, to analyse the statistical significance of the performance difference of All-CNN trained with 3 more and 3 fewer layers.}
  \label{fig:daugreg-bootstrap_depth}
\end{figure}

A noticeable result from these experiments is that all the models trained with weight decay and dropout (round orange dots) suffered a dramatic drop in performance when the architecture changed, regardless of whether deeper or shallower and of the amount of data augmentation. As a matter of fact, the models trained without explicit regularisation performed on average 11.23~\% ($SD = 3.06$) better. As in the case of reduced training data, this may be explained by the poor adaptability of the regularisation hyperparameters, which strongly depend on the architecture.

This highly contrasts with the performance of the models trained without explicit regularisation (top, squared blue dots). With a deeper architecture, these models achieve slightly better performance, effectively exploiting the increased capacity. With a shallower architecture, they achieve only slightly worse performance\footnote{Note that the shallower models trained with neither explicit regularisation nor data augmentation achieve even better accuracy than their counterpart with the original architecture, probably due to the reduction of overfitting provided by the reduced capacity.}. Thus, these models seem to more naturally adapt to the new architecture and data augmentation becomes beneficial.

It is worth commenting on the particular case of the CIFAR-100 benchmark, where the difference between the models with and without explicit regularisation is even more pronounced, in general. It is common practice in object recognition papers to tune the parameters for CIFAR-10 and then test the performance on CIFAR-100 with the same hyperparameters. Therefore, these are typically less suitable for CIFAR-100. We believe this is the reason why the benefits of data augmentation seem even more pronounced on CIFAR-100 in our experiments.

In sum, these results highlight another crucial advantage of data augmentation: the effectiveness of its hyperparameters, that is the type of image transformations, depend mostly on the type of data, rather than on the particular architecture or amount of available training data, unlike explicit regularisation hyperparameters. Therefore, removing explicit regularisation and training with data augmentation increases the flexibility of the models.

\section{Discussion}
\label{sec:daugreg-discussion}
In this section we summarise our findings and discuss their relevance. In particular, we challenge the need for weight decay and dropout to train artificial neural networks, and propose to rethink data augmentation as a \textit{first class} technique instead of a \textit{cheating} method.

As an empirical analysis, one caveat of our work is the limited number of experiments (over 200 models trained). In order to increase the generality of our conclusions, we chose three significantly distinct network architectures and three data sets. Importantly, we also took a conservative approach in our experimentation: all the hyperparameters were kept as in the original models, which included both weight decay and dropout, as well as light augmentation. This setup is clearly suboptimal for models trained without explicit regularisation. Besides, the heavier data augmentation scheme was deliberately not optimised to improve the performance as it was not the scope of this work to propose a specific data augmentation technique. We leave for future work to explore data augmentation schemes that can more successfully be exploited by any deep model. Finally, in order to strengthen the conclusions from the empirical analysis, we have also discussed some theoretical insights in Section~\ref{sec:daugreg-theoretical_insights}, concluding that the generalisation gain provided by weight decay can be seen as a lower bound of what can be achieved by domain-specific data augmentation. We also hope that this work inspires researchers in other application domains, such as natural language processing, to further contrast data augmentation and explicit regularisation.

\subsection{Do deep nets really need weight decay and dropout?}
\label{sec:daug_vs_reg-wd_drop}
In Section~\ref{sec:daugreg-results} we have presented the results of a systematic empirical analysis of the role of weight decay, dropout and data augmentation in deep convolutional neural networks for object recognition. Our results have shown that explicit regularisation is not only unnecessary \citep{zhang2016understandingdl}, but also that its performance gain can be achieved by data augmentation alone: in most cases, training with data augmentation only was better than training with both data augmentation and explicit regularisation. In the few cases where that was not the case, the difference was very small. Moreover, unlike data augmentation, models trained with weight decay and dropout exhibited poor adaptability to changes in the architecture and the amount of training data. Why do researchers and practitioners keep training their neural networks with weight decay and dropout? Do deep nets really need weight decay and dropout?

The relevance of these findings lies in the fact that weight decay and dropout are almost ubiquitously present in convolutional neural networks \citep{huang2017densenet, zagoruyko2016wrn, springenberg2014allcnn}, including recent, state of the art models \citep{tan2019efficientnet}. Certainly, it has been shown in multiple research papers that weight decay and dropout can boost the performance of neural networks, and we here do not challenge the usefulness of weight decay and dropout, but the convenience to use it, given the associated cost and risk, and available alternatives. 

First, not only add weight decay and dropout extra computations during training, but also they typically require training the models several times with different hyperparameters: the coefficient of the penalty for weight decay; for dropout, the location of the dropout mask and the amount of units to drop. These hyperparameters are arguably very sensitive to changes in elements of the learning process. Here we have studied changes in the amount of training data (Section~\ref{sec:daugreg-less_data}) and the depth of the architecture (Section~\ref{sec:daugreg-depth}). Consider, for instance, our results in Section~\ref{sec:daugreg-less_data}: All-CNN trained on CIFAR-10 with weight decay, dropout and light augmentation reaches about 92~\% accuracy. If we were in the development process and were unsure about what architecture to use, we could simply try our network with three more layers and we would obtain about 85~\% accuracy. We could also try an architecture with three fewer layers and obtain about 82~\% accuracy. This may lead us to conclude that the first architecture has the right number of layers, because adding or removing layers drastically reduces the performance; or perhaps that by adding or removing layers there is some negative interaction between the layers sizes, or any other of the many hypothesis we could think of.

Consider now what happens if we train without weight decay and dropout: All-CNN trained on CIFAR-10 with light augmentation---but without weight decay and dropout---obtains about 93.3~\% accuracy. This is slightly better than the explicitly regularised model, but we will ignore this now. If we train this model with three more layers, we obtain 93.4~\% accuracy, that is the same or slightly better---as opposed to the drop of 7 points we have seen before. If we train with three fewer layers, we obtain about 90~\% accuracy, a drop of 3 points---as opposed to a drop of 10 points. In this case, we would not conclude that adding or removing layers creates negative interactions. Note that the only difference between these two cases is that the first models are trained with weight decay and dropout. Therefore, it may be reasonable to only include explicit regularisation in the final version of a model, in order to potentially obtain a slight boost in performance prior to publication or production---provided the hyperparameters are adequately fine-tuned---but keeping weight decay and dropout as intrinsic part of our models can certainly lead us astray.

Finally, we can draw some connections between the results from this chapter and the insights from the previous chapter. In Chapter~\ref{ch:reg} we discussed that the role of explicit regularisation techniques, such as weight decay and dropout, is to reduce the representational capacity of the models. This, according to statistical learning theory, can reduce overfitting and in turn improve generalisation. However, artificial neural networks have usually orders of magnitude more parameters than training examples, and they still generalise well. While this phenomenon is still not well understood, one working hypothesis is that overparameterisation does not cause negative overfitting, but rather smooth fitting that can be suitable for accurate interpolation \citep{belkin2019biasvariance, hasson2020directfit}. If overparameterisation is not a problem for artificial neural networks, is it then necessary to constrain the representational capacity through explicit regularisation? Is it reasonable to train very large models, that require a lot of memory and computation---and negatively impact the environment---and at the same time constrain their capacity?

We also hypothesise that a reason why artificial neural networks generalise well in many tasks is due to the fact that the models include many sources of implicit regularisation or, in other words, inductive biases. For example, it is known that stochastic gradient descent naturally converges to solutions with small norm \citep{zhang2016understandingdl, neyshabur2014implicitreg}, batch normalisation also contributes to better generalisation, convolutional layers are particularly efficient to process image data---and not only---to name a few examples. In our case, we argue that data augmentation has the potential to encode very powerful inductive biases that improve generalisation. We conclude that in the presence of many other sources of implicit regularisation and more effective inductive biases, weight decay and dropout may not be necessary to train large deep artificial neural networks\footnote{Previous work has suggested interesting connections between weight decay and other types of regularisation and improved adversarial robustness \citep{galloway2018wdadversarial, jakubovitz2018regadversarial}. An interesting avenue for future work is studying whether this effects are also provided by data augmentation}.

\subsection{Rethinking Data Augmentation}
\label{sec:daugreg-rethink_daug}
Data augmentation is often regarded by authors of machine learning papers as \textit{cheating}, suggesting it should not be used in order to test the potential of newly proposed methods \citep{goodfellow2013maxout, graham2014fracmaxpool, larsson2016fractalnet}. In contrast, weight decay and dropout are considered intrinsic elements of the algorithms \citep{tan2019efficientnet}. In view of our results, we propose to rethink data augmentation and switch roles with explicit regularisation: good models should effectively exploit data augmentation and explicit regularisation should only be applied, if at all, once all other elements are fixed. This approach improves the performance and saves computational resources.

In this regard, it is worth highlighting some advantages of data augmentation: Not only does it not reduce the representational capacity, unlike explicit regularisation, but also, since the transformations reflect plausible variations of the real objects, it increases the robustness of the model \cite{novak2018sensitivity, rusak2020robustness}. Interestingly, in Chapter~\ref{ch:daugit} we will also show that models trained with heavier data augmentation learn representations more aligned with the inferior temporal (IT) cortex, highlighting its connection with visual perception and biological vision. Deep nets are especially well suited for data augmentation because they do not rely on pre-computed features. Moreover, unlike explicit regularisation, it can be performed on the CPU, in parallel to the gradient updates. Finally, from Sections~\ref{sec:daugreg-less_data} and~\ref{sec:daugreg-depth} we concluded that data augmentation naturally adapts to architectures of different depth and amounts of available training data, without the need for specific fine-tuning of hyperparameters.

A commonly cited disadvantage of data augmentation is that it depends on expert knowledge and it cannot be applied to all domains \citep{devries2017daugfeatspace}. However, we argue instead that expert and domain knowledge should not be disregarded but exploited. Expert and domain knowledge are, in fact, useful inductive biases. A remarkable advantage of data augmentation is that a single augmentation scheme can be designed for a broad family of data---for example, natural images, using our knowledge about visual perception---and effectively applied to a broad set of tasks---object recognition, segmentation, localisation, etc. We hope that these insights encourage more research on data augmentation and, in general, highlight the importance of using the available data more effectively. In the following chapters, we explore additional properties of models trained with data augmentation (Chapter~\ref{ch:daugit}) and how it can be used as part of the objective function to learn representations more aligned with the properties of the primate visual cortex (Chapter~\ref{ch:invariance}).

\chapterbibliography
}

{
\chapter[Data augmentation and object representation in the brain]{Data augmentation\\and object representation in the brain}
\label{ch:daugit}
\renewcommand{\chapterpath}{includes/daug-it}
\begin{contributors}
    Johannes Mehrer performed the representational similarity analysis. Nikolaus Kriegeskorte, Peter K{\"onig} and Tim C. Kietzmann reviewed and edited the manuscript submitted to CCN.
\end{contributors}
\begin{outreach}
    \item \textit{Deep neural networks trained with heavier data augmentation learn features closer to representations in hIT.} \textbf{Alex Hern{\'a}ndez-Garc{\'i}a}, Johannes Mehrer, Nikolaus Kriegeskorte, Peter K{\"o}nig, Tim C. Kietzmann. Cognitive Computational Neuroscience (CCN), 2018. 
\end{outreach}
%
%
One of the central goals of computational neuroscience is to develop better models of the human brain. The re-emergence of deep artificial neural networks, which now excel at many artificial intelligence tasks by automatically learning hierarchical representations \citep{girshick2014dlhierarchy}, has also had a positive impact on computational neuroscience. For instance, the features learnt by models trained for image object classification have been found to correlate better with the representations in the human inferior temporal cortex (hIT) than  traditional hand-crafted features or shallow models \citep{khaligh2014annbrains, yamins2014annsbrains, gucclu2015annbrains}. Further, convolutional neural networks are currently the most accurate models for multiple regions across the primate visual cortex \citep{kietzmann2019dnncompneuro, yamins2016computneuro}. However, while the similarity between artificial and biological neural networks is promising, a crucial question remains: what makes neural networks learn representations that more closely mirror activations in the brain?

Delving into this question is one of the goals of this thesis because of its potential implications on our understanding of the brain and learning systems in general. Previous work has revealed that networks performing better in classification tasks correlate more strongly with neural representations in high level areas \citep{yamins2014annsbrains}. However, the network architecture seems to play a crucial role \citep{storrs2017ccn} and \citet{mehrer2017ccn} showed that training with more ecologically relevant image categories yields more similar representations. Inspired by the apparent importance of the training data, and the properties of data augmentation discussed in Chapter~\ref{ch:daugreg}, we here explore the influence of data augmentation on the representational similarity between artificial neural networks and the human inferior temporal cortex.

As we have discussed in the Introduction (Chapter~\ref{ch:intro}), the transformations included in (perceptually plausible) data augmentation schemes are inspired by the properties of visual perception. We perform translations, rotations, scaling and changes in the illumination of images (see Section~\ref{sec:daugreg-methods_data}) because these transformations are part of the variance we observe in the visual real-world. Transformations of this kind within certain ranges do not change the perceived object class and even identity. In Chapter~\ref{ch:daugreg} we have seen that applying these transformations to the training images of a neural network model is highly beneficial for generalisation. In this chapter we test the hypothesis that training with heavier data augmentation may encourage learning representations more aligned representations with the inferior temporal cortex.
%

\section{Methods}
\label{sec:daugit-methods}
This section presents the experimental setup to analyse the role of data augmentation on the similarity between artificial neural networks and neural representations in hIT. We describe the network architectures, the augmentation schemes and the methodology employed to compare both systems.

\subsection{Network architectures}
To increase the generality of our results, we analysed two distinct, well-known convolutional neural networks, which reach high-performance on image object-classification: the all convolutional network, All-CNN \citep{springenberg2014allcnn} and the wide residual network, WRN \citep{zagoruyko2016wrn}. We used the same architectures for the experiments in Chapter~\ref{ch:daugreg} and they are described in detail in Section~\ref{sec:daugreg-methods_archs}, so we here only provide a brief overview of the most important properties:

\begin{itemize}
 \item \textbf{All-CNN} consists only of 12 convolutional layers, each followed by batch normalisation and a ReLU activation. It has a total of 9.4 million parameters.
 \item \textbf{WRN} is a modification of ResNet \citep{he2016resnet} that achieves better performance with fewer layers, but more units per layer. We chose the WRN-28-10 version of the original paper, which has 28 layers and about 36.5 million parameters.
\end{itemize}

Following the conclusions from Chapter~\ref{ch:daugreg}, we did not train the models with either weight decay or dropout, but we kept the rest of the hyperparameters as in the original papers.

\subsection{Data augmentation}
The goal of this work was to study the impact of data augmentation on the similarity of the representations with the activations in the inferior temporal cortex. For that purpose, we considered two data augmentation schemes: \textit{light} and \textit{heavier} augmentations, as described in Section~\ref{sec:daugreg-methods_data}. Below we summarise the transformations included in each scheme:

\begin{itemize}
    \item The \textbf{light} augmentation scheme has been widely used in the literature, for instance \citep{springenberg2014allcnn}. It performs only random horizontal flips and horizontal and vertical translations of maximum 10\% of the image size. Additionally, we performed random crops of $128\times128$ pixels.
    \item The \textbf{heavier} scheme performs a larger range of random affine transformations such as scaling, rotations and shear mapping, as well as contrast and brightness adjustment and random crops.
\end{itemize}
 
We used these schemes to augment the highly benchmarked ImageNet ILSVRC 2012 data set \citep{russakovsky2015imagenet}. We used ImageNet instead of CIFAR-10---for instance---because its higher resolution images more closely match the stimulus statistics of the human visual system. We resized the images into $150\times200$ pixels. Examples of the light and heavier augmentations on ImageNet photos are shown in Figure~\ref{fig:daugit-daugimagenet}

\begin{figure}[ht]
  \begin{center}
    \includegraphics[width = 0.8 \linewidth]{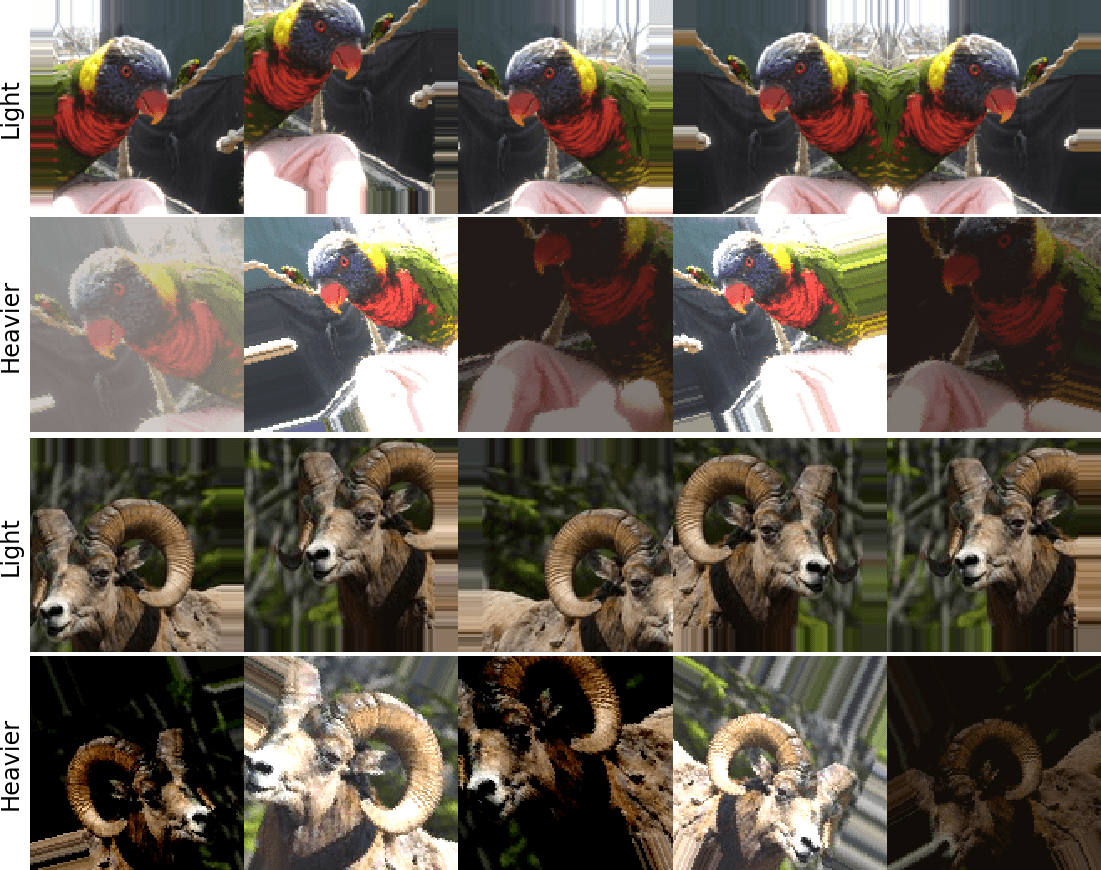}
  \end{center}
  \caption{Illustration of the transformations performed by the light and heavier augmentation schemes on two example images. Note that the five transformations of the images in this figure have been produced by setting extreme values of the parameters, so as to highlight the characteristics of the schemes and the differences between them.}
  \label{fig:daugit-daugimagenet}
\end{figure}

The performance of All-CNN and WRN trained with light and heavier augmentation is shown in Figure~\ref{fig:daugit-performance}. Note that training with light augmentation provides better results, specially on All-CNN. As pointed out in Chapter~\ref{ch:daugreg}, this is likely explained by first, the fact that the heavier augmentation scheme was not designed to optimise classification, but rather as an arbitrary larger set of plausible transformations; and second, because the limited capacity of the models---especially All-CNN---may prevent them from exploiting the aggressive transformations of the already large ImageNet data set. Nonetheless, the objective of this study was to analyse the learnt representations given a reasonably accurate performance. Ideally, we would also analyse the representations of a model trained with no augmentation. However, the performance without data augmentation is significantly worse and this would likely impact the representations \citep{yamins2014annsbrains}.

\begin{figure}[ht]
  \begin{center}
    \includegraphics[width = 0.7 \linewidth]{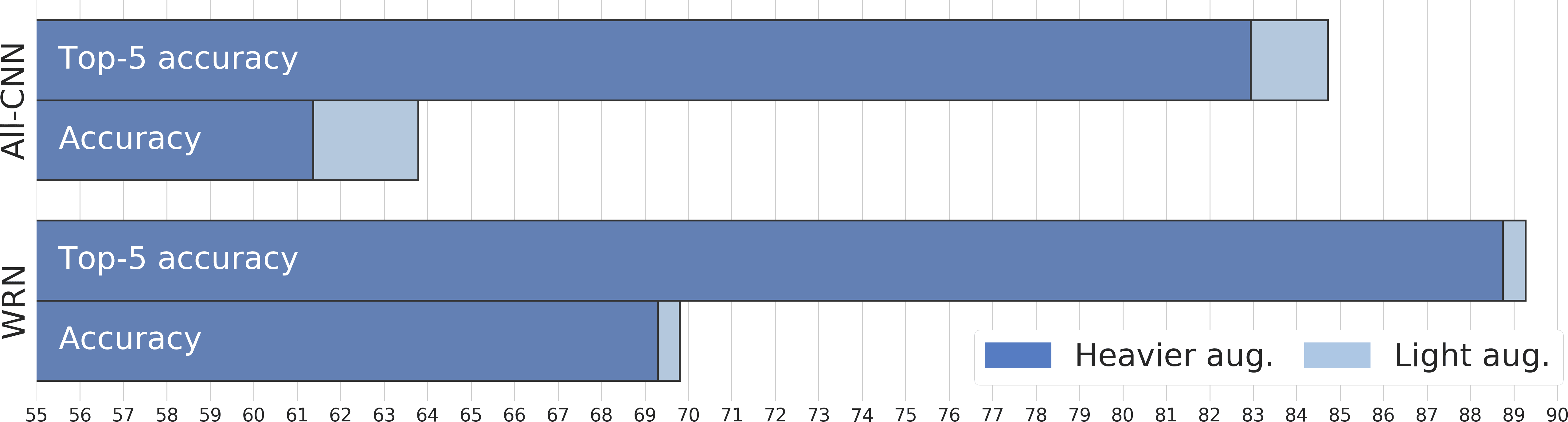}
  \end{center}
  \caption{Test performance of All-CNN and WRN trained with light and heavier data augmentation.}
  \label{fig:daugit-performance}
\end{figure}

\subsection{Representational similarity analysis}
In order to compare the representations learnt by the neural networks and the activations measured in the inferior temporal cortex, we made use of representational similarity analysis (RSA) \citep{kriegeskorte2008rsa, nili2014rsatoolbox}. The main advantage of RSA is that it allows direct comparisons across different model systems without having to explicitly align the different measurement types. This is accomplished by constructing representational dissimilarity matrices (RDMs) to express the pairwise similarity between stimuli, instead of directly comparing the representations of single stimuli. Across a set of input images, RDMs characterise the internal representations of a given system by storing all pairwise distances. The resulting matrix therefore expresses the representational geometry in the learnt activation space. By relying on distances, RDMs remain unchanged, if the space over which they are computed is rotated.

To characterise the representations in hIT, functional magnetic resonance imaging (fMRI) was used to measure BOLD responses while 15 participants were presented with 92 images of isolated objects. The images originate from a wide variety of categories and levels of abstraction. On the broadest level, they can be separated into animate and inanimate. Inanimate objects can either be natural or artificial, whereas animate objects are divided into human stimuli---heads and body parts---and animals---full body and heads only. This fMRI data set has been used in multiple studies and the details of the data acquisition can be found in \citep{kriegeskorte2008manandmonkey}. In Figure~\ref{fig:daugit-rdms} we show the RDM of the brain data, and the RDMs of the WRN model for illustration. As in \citep{kriegeskorte2008manandmonkey}, in order to better visualise the differences across the RDM, the colour code represents the percentiles of the actual RDMs.

To compare artificial neural networks and hIT representations, the network activation profiles for the 92 images were extracted. In particular, we computed the activations at the outputs of the 12 ReLU layers of All-CNN and at the outputs of the residual blocks of WRN. We then computed the RDM of these activations using the Pearson correlation, as well as the RDM of the fMRI responses in hIT. To obtain a more compact representation of the CNN models, we combined the RDMs of all layers into a single RDM as a linear combination of the individual layer RDMs with respect to the hIT RDM using non-negative least squares and a cross-validation procedure, which avoids overfitting the image set.

\begin{figure}[ht]
  \begin{center}
    \includegraphics[width = \linewidth]{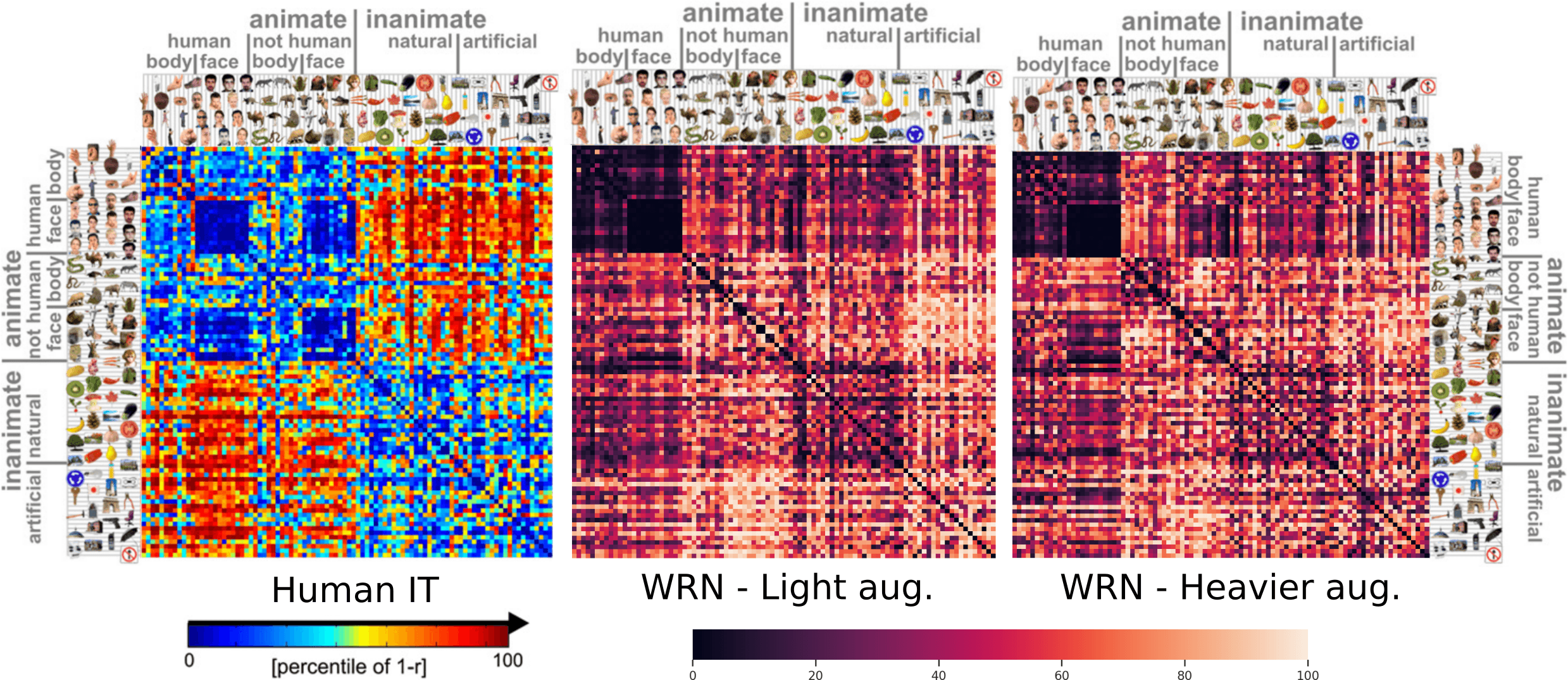}
  \end{center}
  \caption{RDMs of the systems compared. Left: adapted from \citep{kriegeskorte2008manandmonkey}, the RDM of the human inferior temporal cortex. Centre and right, the RDMs of the WRN model, trained with light and heavier augmentation, respectively. As in \citep{kriegeskorte2008manandmonkey}, the colour code in the matrices shown in the figure represents the percentile of the dissimilarity.}
  \label{fig:daugit-rdms}
\end{figure}

Finally, we characterise the similarity between the artificial neural networks and hIT by computing the Kendall's rank correlation coefficient $\tau_{A}$ between the RDM of the hIT representations and the RDM of the convolutional models. Standard errors were obtained from the similarity estimates of the 15 human subjects.

\section{Results and discussion}
\label{sec:daugit-results}
We show the results of the representational similarity analysis to compare the representations learnt by the neural networks and the fMRI data in Figure~\ref{fig:kendall}. As a main conclusion, we found that the correlation with the hIT representations is significantly higher for the models trained with heavier data augmentation. Not only is this indicated by the Kendall correlation, but also by visual inspection, the RDM of the model trained with heavier augmentations seems more similar to the RDM of the human IT. For example, face images---both human and non-human---form clearer similarity clusters in the model trained with heavier data augmentation, which is a well-studied property of the primate visual cortex. 

In the case of the wide residual network (WRN) the difference between the two levels of augmentation is considerably larger, while in the All-CNN models, although statistically significant ($p<0.05$), the difference is smaller. However, recall that the classification performance of the models trained with heavier augmentations is worse, especially in the case of All-CNN (Figure-~\ref{fig:daugit-performance}. Therefore, it seems that even though the more aggressive transformations do not improve the classification performance, they do increase the similarity with the inferior temporal cortex. 

\begin{figure}[ht]
  \begin{center}
    \includegraphics[width = 0.8 \linewidth]{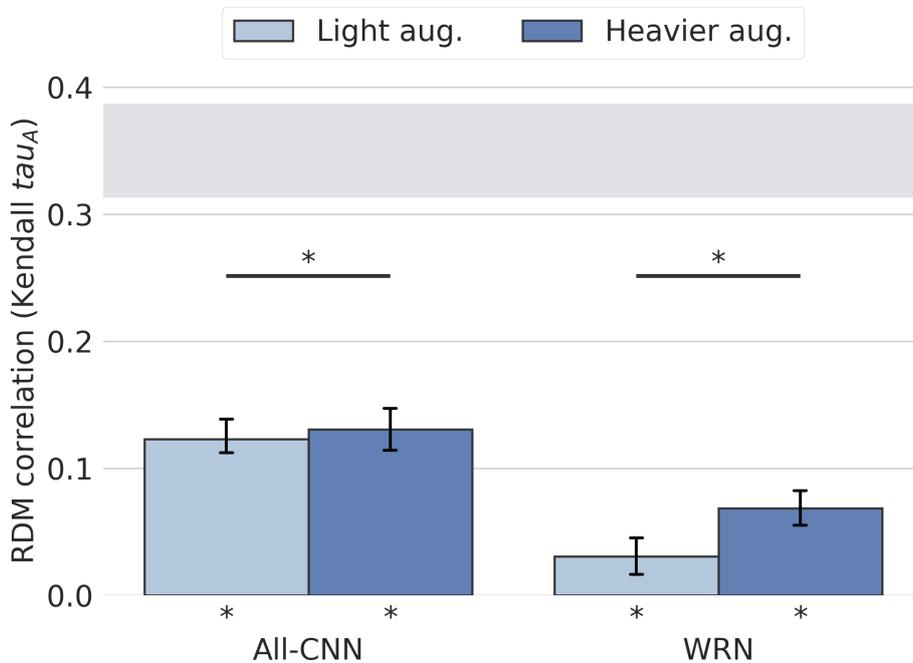}
  \end{center}
  \caption{Comparison of the Kendall's $\tau_{A}$ coefficient of the hIT RDM and the RDM of the networks trained with light and heavier data augmentation. Both on All-CNN and WRN, the correlation of the model trained with heavier transformations is significantly higher than the light counterpart. The grey shaded area indicates the maximum possible correlation of a model given the noise in the measured data.}
  \label{fig:kendall}
\end{figure}

It is also interesting that the similarity with hIT is higher for All-CNN, again despite the lower performance, as opposed to previous work that indicated a correlation between performance and similarity with the visual cortex \citep{yamins2014annsbrains}. The overall lower correlation of WRN adds more evidence to the conclusion of \citet{storrs2017ccn}, who showed that residual networks exhibit a particularly low correlation with hIT compared to other architectures.

Given the exploratory nature of this study, it is not yet clear what exact mechanisms lead to the better match between representational geometries in higher level visual cortex and networks trained with heavier data augmentation. One hypothesis is that the larger variety during training may be more biologically plausible than training with constant images or very light transformations. Humans develop robust object representations based on highly variable input, while freely exploring the world. Sources of variation include different orientations, lighting conditions, backgrounds and occlusion. Eye-movements, including drifts and \textit{microsaccades}, may further contribute to the variability in the sensory input to which the recognition has to be robust. As we will further discuss in Chapter~\ref{ch:invariance}, this robustness is reflected in invariant activations towards identity-preserving transformations in the higher visual cortex.

Our experiments addressed the question as to which factors drive computational models to learn features closer to the brain representations. Given the superiority in visual robustness of the human brain, these insights may have implications for artificial vision systems based on deep neural networks, and for ANNs as a model system for visual processing in the brain. Finding that heavier data transformations leads to more IT-like representations further supports the notion that the input distribution plays a crucial role during the learning of representations in both the brain artificial networks. 

\section{Conclusions}
In this chapter, we have explored how far light and heavier augmentation of the training set can affect the internal representations of deep neural networks and their alignment with human IT. To compare the neural and model system, we used representational similarity analysis, which allows for straightforward comparisons across different modalities---in this case, fMRI BOLD signal and neural network activations. RSA revealed that the neural networks trained with heavier transformations learn representations more similar to those observed in higher visual cortex.

Future work should analyse a larger range of network architectures and data sets to gain better insights into the mechanisms driving the internal representations. It will be also interesting to study the different components of data augmentation in order to understand which particular transformations play a bigger role in better explaining hIT.

\chapterbibliography
}

{
\chapter{Data augmentation invariance}
\label{ch:invariance}
\renewcommand{\chapterpath}{includes/invariance}
\begin{contributors}
    Tim C. Kietzmann supervised the work during my internship at his lab. Tim and Peter K{\"o}nig reviewed and edited the manuscripts submitted to conferences.
\end{contributors}
\begin{outreach}
    \item \textit{Learning robust visual representations using data augmentation invariance.} \textbf{Alex Hern{\'a}ndez-Garc{\'i}a}, Peter K{\"o}nig, Tim C. Kietzmann. arXiv preprint arXiv:1906.04547 \& Cognitive Computational Neuroscience (CCN), 2019 \& Workshop on Bridging AI and Cognitive Science, International Conference on Learning Representations (ICLR), 2019.
    \item \textit{Learning representational invariance instead of categorization.} \textbf{Alex Hern{\'a}ndez-Garc{\'i}a}, Peter K{\"o}nig. Workshop on pre-registration in computer vision, International Conference on Computer Vision (ICCV), 2019.
\end{outreach}
Deep artificial neural networks (ANNs) have borrowed much inspiration from neuroscience and are, at the same time, the current best model class for predicting neural responses across the visual system in the brain \citep{kietzmann2019dnncompneuro, kubilius2018cornet}. Yet, despite consensus about the benefits of a closer integration of deep learning and neuroscience \citep{bengio2015dlandneuroscience, marblestone2016dlandneuroscience, richards2019dlandneuroscience}, important differences remain \citep{sinz2019dlvsbrain, geirhos2020shortcutlearning, dujmovic2020adversarial}.

Here, we investigate a representational property that is well established in the neuroscience literature on the primate visual system: the increasing robustness of neural responses to identity-preserving image transformations. While early areas of the ventral stream (V1-V2) are strongly affected by variation in object size, position, viewpoint and other factors, later levels of processing are increasingly robust to such changes, as measured first in single neurons of the inferior temporal (IT) cortex of macaques \cite{booth1998invariantitmacaque} and then in humans' \citep{quiroga2005invariantithuman, isik2013dynamics}. The cascaded achievement of invariance to such identity-preserving transformations has been proposed as a key mechanism for robust object recognition \citep{dicarlo2007untangling, tacchetti2018invariance}.

Learning such invariant representations has been a desired objective since the early days of artificial neural networks \citep{simard1992daug}. Accordingly, a myriad of techniques have been proposed to attempt to achieve tolerance to different types of transformations (\citet{cohen2016groupequivcnns} briefly reviewed some of these efforts). Interestingly, recent theoretical work \citep{achille2018emergence} has shown that invariance to ``nuisance factors'' should naturally emerge from the optimisation process of deep models that minimise the information of the representations about the inputs, while retaining the minimum information about the target, as proposed by \citet{tishby2015infobottleneck} in the information bottleneck principle.

Nevertheless, artificial neural networks are still not robust to identity-preserving transformations, including simple image translations \citep{zhang2019convolutions}. A remarkable extreme example are adversarial attacks \citep{szegedy2013adversarial}, in which small changes, imperceptible to the human brain, can alter the classification output of the network. Extending this line of research, we used data augmentation as a framework to generate the transformations to which vision models should be invariant to, according to visual perception and biological vision. As a first contribution, we propose a simple metric, \textit{data augmentation invariance score}, to measure the invariance of neural networks to identity-preserving transformations.
%

Second, inspired by the increasing invariance observed along the primate ventral stream of the visual cortex, we here propose \textit{data augmentation invariance}, a simple, yet effective and efficient mechanism to improve the robustness of the representations: we include an additional contrastive term in the objective function that encourages the similarity between augmented examples within each batch. We will argue that this objective encodes a useful inductive bias that exploits prior knowledge from visual perception and biological vision.

Finally, we explore the possibility of fully replacing the categorisation objective that is commonly used to train neural networks for classification, the categorical cross-entropy, by objective functions purely based on invariance learning.

\section{Invariance score}
\label{sec:invariance-eval}
To measure the invariance of the learnt features under the influence of identity-preserving image transformations we compare the activations of a given image with the activations of an augmented version of the same image. 

Consider the activations of an input image $x$ at layer $l$ of a neural network, which can be described by a function $f^{(l)}(x) \in \mathbb{R}^{D^{(l)}}$. We can define the distance between the activations of two input images $x_{i}$ and $x_{j}$ by their mean squared difference:

\begin{equation}
\label{eq:invariance-mse}
 d^{(l)}(x_{i}, x_{j}) = \frac{1}{D^{(l)}}\sum_{k=1}^{D^{(l)}}(f_{k}^{(l)}(x_{i}) - f_{k}^{(l)}(x_{j}))^2
\end{equation}

Following this, we are interested in the mean squared distance between $f^{(l)}(x_i)$ and a randomly sampled transformation of $x_i$, that is $d^{(l)}(x_{i}, G(x_{i}))$. $G(x)$ refers to the stochastic function that transforms the input images according to a pre-defined, parameterised data augmentation scheme.

In order to assess the similarity between the activations of an image $x_i$ and of its augmented versions $G(x_{i})$ we normalise it by the average similarity in the (test) set. We define the \textit{data augmentation invariance score} $S_{i}^{(l)}$ of image $x_i$ towards the transformation $G(x)$ at layer $l$ of a model, with respect to a data set of size $N$, as follows:

\begin{equation}
\label{eq:invariance-invariance}
 S_{i}^{(l)} = 1 - \frac{d^{(l)}(x_{i}, G(x_{i}))}{\frac{1}{N}\sum_{j=1}^{N}d^{(l)}(x_{i}, x_{j})}
\end{equation}

Note that the invariance $S_{i}^{(l)}$ takes the maximum value of 1 if the activations of $x_{i}$ and its transformed version $G(x_{i})$ are identical, and the value of 0 if the distance between transformed examples, $d^{(l)}(x_{i}, G(x_{i}))$ (numerator), is equal to the average distance in the set (denominator).
%

\subsection{Learning objective}
\label{sec:invariance-daug_invariance}
Most ANNs trained for object categorisation are optimised through mini-batch stochastic gradient descent (SGD), that is the weights are updated iteratively by computing the loss of a batch $\mathcal{B}$ of examples, instead of the whole data set at once. The models are typically trained for a number of \textit{epochs} $E$ which is a whole pass through the entire training data set of size $N$. That is, the weights are updated $K=\frac{N}{|\mathcal{B}|}$ times each epoch.

Data augmentation introduces variability into the process by performing a different, stochastic transformation of the data every time an example is fed into the network. However, with standard data augmentation, the model has no information about the \textit{identity} of the images. In other words, different augmented examples, seen at different epochs, separated by $\frac{N}{|\mathcal{B}|}$ iterations on average, correspond to the same seed data point. This information is potentially valuable and useful to learn better representations. For example, in biological vision, the high temporal correlation of the stimuli that reach the visual cortex may play a crucial role in the creation of robust connections \citep{becker1999temporalstability, kording2004complexcells, wyss2006temporalstability}. However, this is generally not exploited in supervised settings. In semi-supervised learning, where the focus is on learning from fewer labelled examples, data augmentation has been used as a source of variability together with dropout or random pooling, among others \citep{laine2016ssl}.

In order to make use of this information and improve the robustness, we first propose \textit{in-batch} data augmentation by constructing the batches with $M$ transformations of each example---\citet{hoffer2019batchaugmentation} recently discussed a similar idea. Additionally, we propose a new objective function that accounts for the invariance of the feature maps across multiple image samples. Considering the difference between the activations at layer $l$ of two images, $d^{(l)}(x_{i}, x_{j})$, defined in Equation~\ref{eq:invariance-mse}, we define the \textit{data augmentation invariance} loss at layer $l$ for a given batch $\mathcal{B}$ as follows:

\begin{equation}
\label{eq:invariance-data_aug_inv}
 \mathcal{L}_{inv}^{(l)} = \frac{\sum_{k}\frac{1}{|\mathcal{S}_{k}|^2}\sum_{x_i, x_j \in \mathcal{S}_{k}}d^{(l)}(x_{i}, x_{j})}{\frac{1}{|\mathcal{B}|^2}\sum_{x_i, x_j \in \mathcal{B}}d^{(l)}(x_{i}, x_{j})}
\end{equation}
where $\mathcal{S}_{k}$ is the set of samples in the batch $\mathcal{B}$ that are augmented versions of the same seed sample $x_k$. This loss term intuitively represents the average difference of the activations between the sample pairs that correspond to the same source image, relative to the average difference of all pairs. A convenient property of this definition is that $\mathcal{L}_{inv}$ does not depend on either the batch size or the number of in-batch augmentations $M=|\mathcal{S}_{k}|$. Furthermore, it can be efficiently implemented using matrix operations. Our data augmentation invariance can be seen as a contrastive loss \citep{hadsell2006contrastive}, since the aim is to bring closer the representations of related examples---transformations of the same source image---and push apart the representations from other examples.

In order to simultaneously achieve certain representational invariance at $L$ layers of the network and high object recognition performance at the network output, we define the total loss as follows:

\begin{equation}
 \mathcal{L} = (1 - \alpha)\mathcal{L}_{obj} + \sum_{l=1}^{L}\alpha^{(l)}\mathcal{L}_{inv}^{(l)}
\end{equation}
where $\alpha = \sum_{l=1}^{L}\alpha^{(l)}$ and $\mathcal{L}_{obj}$ is the loss associated with the object recognition objective, typically the cross-entropy between the object labels and the output of a softmax layer. For all the experiments presented in this chapter, we set $\alpha=0.1$ and distributed the coefficients across the layers according to an exponential law, such that $\alpha^{(l=L)}= 10\alpha^{(l=1)}$. This aims at simulating a probable response along the ventral visual stream, where higher regions are more invariant than the early visual cortex\footnote{It is beyond the scope of this study to analyse the sensitivity of the hyperparameters $\alpha^{(l)}$, but we did not observe a significant impact in the classification performance by using other distributions.}.

\subsection{Architectures and data sets}
\label{sec:invariance-arch-and-data}
As test beds for our hypotheses and proposal we trained three neural network architectures: all convolutional network, All-CNN-C \citep{springenberg2014allcnn}; wide residual network, WRN-28-10 \citep{zagoruyko2016wrn}; and DenseNet-BC \citep{huang2017densenet}. All three architectures have been widely used in the deep learning literature, including our experiments in Chapter~\ref{ch:daugreg}, where the details can be consulted. They provide generality of the results, as they have distinctive architectural characteristics: only convolutional layers, residual blocks and dense connectivity, respectively.

We trained the three architectures on the highly benchmarked data set for object recognition CIFAR-10 \citep{krizhevsky2009cifar}. Additionally, in order to test our proposal on higher resolution images and a larger number of classes, we also trained All-CNN and WRN on the \textit{tiny} ImageNet data set, a subset of ImageNet \citep{russakovsky2015imagenet} with 100,000 64x64 training images that belong to 200 classes. All models were trained using the \textit{heavier} data augmentation scheme as defined in Section~\ref{sec:daugreg-methods_data}, which consists of affine transformations, contrast adjustment and brightness adjustment.

For the models trained on CIFAR-10, the training hyperparameters---learning rate, decay, number of epochs, etc.---were set as in the original papers, except that, following the conclusions from Chapter~\ref{ch:daugreg}, we did not use explicit regularisation---weight decay and dropout---since comparable performance is obtained without them if data augmentation is used. For all three architectures, we performed $M=|\mathcal{S}_{k}|=8$ augmentations per image in the batch, while keeping the real batch size as in the original papers.

On tiny ImageNet, All-CNN included three additional layers and was trained for 150 epochs, with batch size 128 and initial learning rate 0.01 decayed by 0.1 at epochs 100 and 125. We included $M=8$ augmentations per image in the batches. WRN was trained for 50 epochs, with batch size 32 and initial learning rate 0.01 decayed by 0.2 at epochs 30 and 40. To train WRN with data augmentation invariance, we performed $M=4$ augmentations per image. In all cases, the models were trained with stochastic gradient descent with Nesterov momentum 0.9. 

Note that the hyperparameters were fine-tuned for the original papers by training only with the standard categorical cross-entropy and with standard epoch-wise data augmentation. Therefore, they were likely suboptimal for our proposed data augmentation invariance. However, our aim was not achieving the best possible classification performance, but rather demonstrate the suitability of data augmentation invariance and analyse the learnt representations.

The invariance loss defined in Equation~\ref{eq:invariance-data_aug_inv} was computed after the ReLU activation of each convolutional layer for All-CNN, at the output of each residual block for WRN, and after the first convolution and the output of each dense block for DenseNet.
%

\section{Results}
\label{sec:invariance-results}
The first contribution of this chapter is to empirically test in how far convolutional neural networks trained for object categorisation produce invariant representations under the influence of identity-preserving transformations of the input images. Figures~\ref{fi:invariance-invariance_allcnn}--\ref{fi:invariance-invariance_densenet} show the invariance scores, as defined in Equation~\ref{eq:invariance-invariance}, across the network layers. Since we do not compute the invariance score at every single layer of the architectures, the numbering of the layers simply indicate increasing depth in the hierarchy (see Section~\ref{sec:invariance-arch-and-data} for the details). The red boxes correspond to the baseline models and the blue boxes to the models trained with our data augmentation invariance objective.

\begin{figure}[ht]
  \begin{center}
    \includegraphics[width = \linewidth]{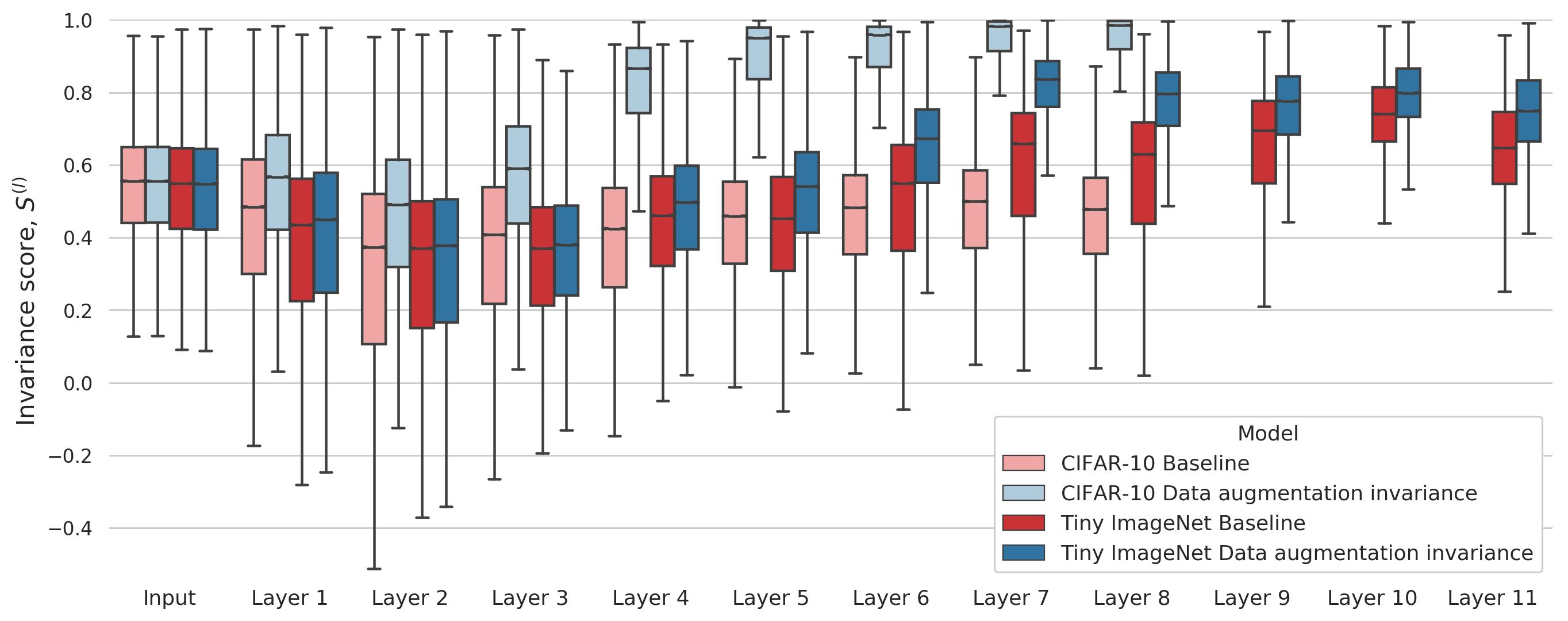}
  \end{center}
  \caption{All-CNN: distribution of the invariance score at each layer of the baseline model and the model trained data augmentation invariance (higher is better).}
  \label{fi:invariance-invariance_allcnn}
\end{figure}

\begin{figure}[ht]
  \begin{center}
    \includegraphics[width = \linewidth]{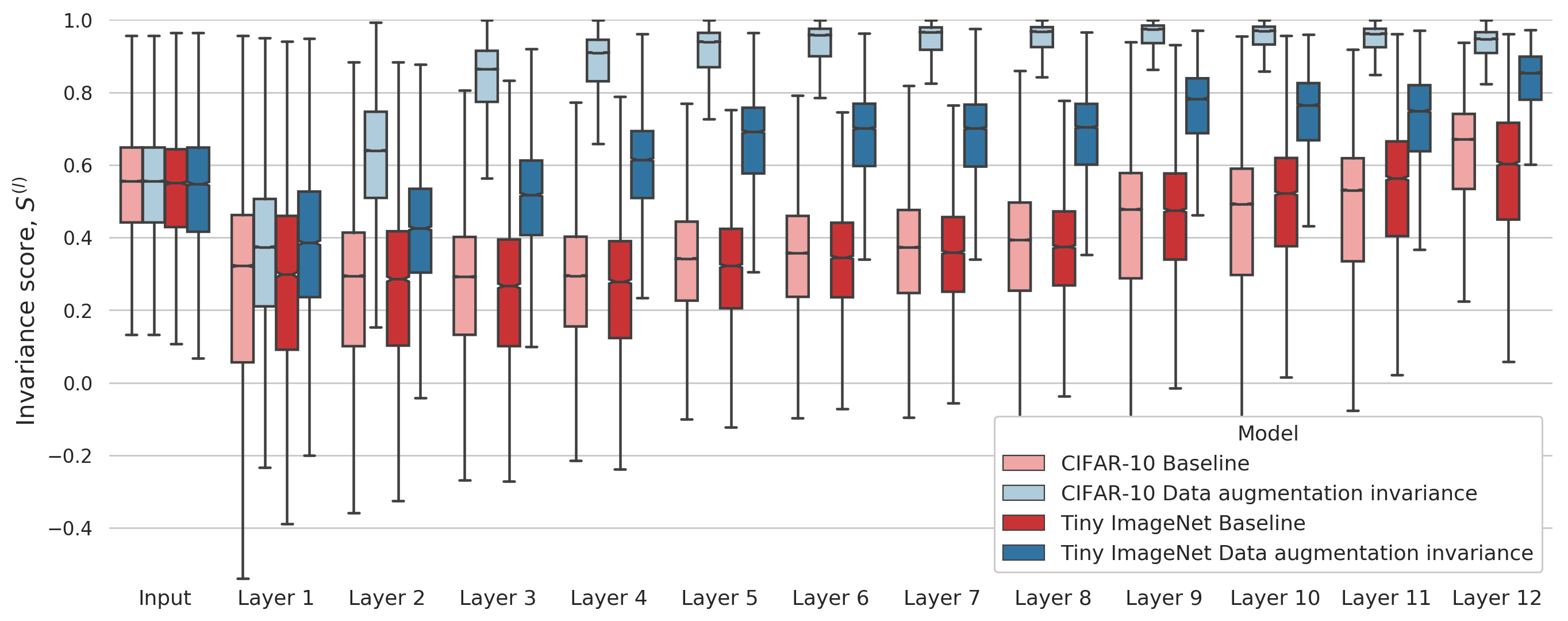}
  \end{center}
  \caption{WRN: distribution of invariance score at each layer of the baseline model and the model trained data augmentation invariance (higher is better).}
  \label{fi:invariance-invariance_wrn}
\end{figure}

\begin{figure}[ht]
  \begin{center}
    \includegraphics[width = \linewidth]{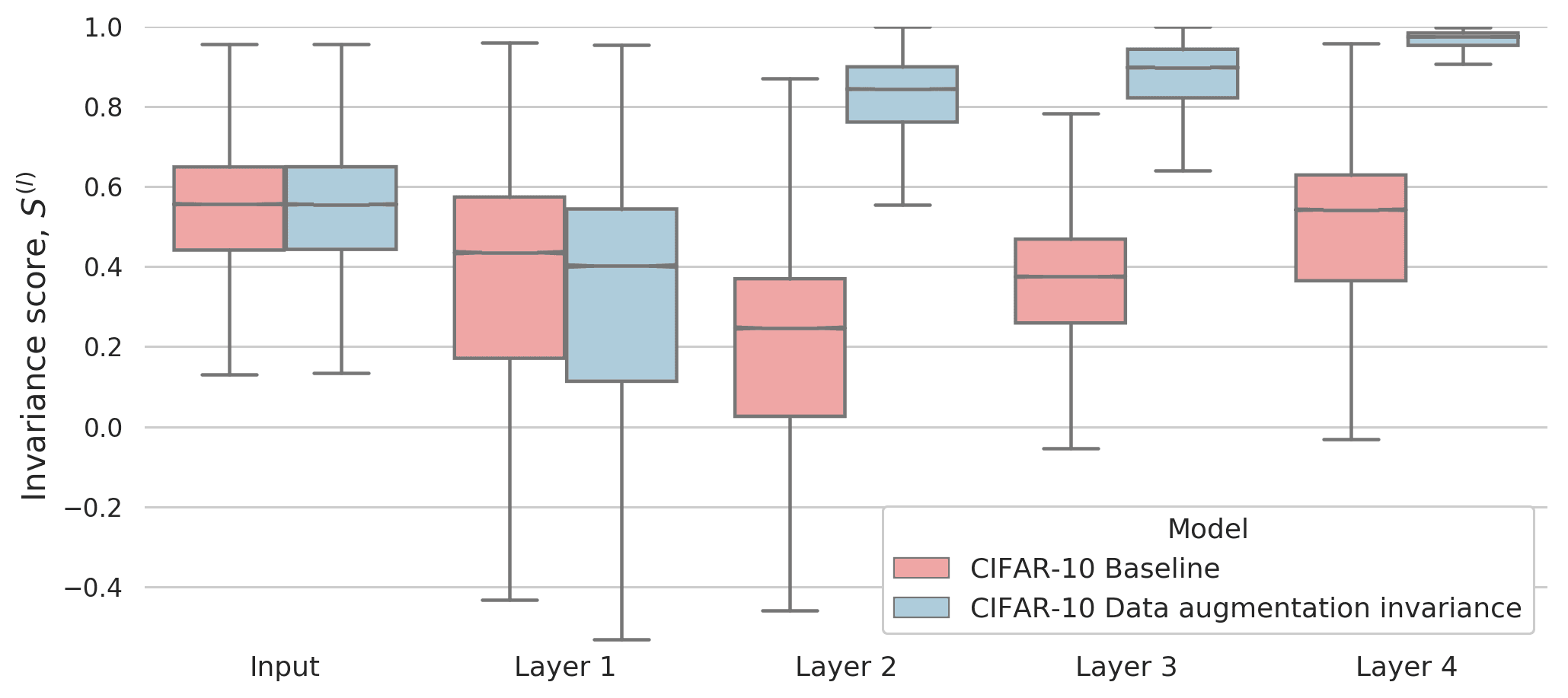}
  \end{center}
  \caption{DenseNet: distribution of invariance score at each layer of the baseline model and the model trained data augmentation invariance (higher is better).}
  \label{fi:invariance-invariance_densenet}
\end{figure}

The distributions of the invariance score shown in the figures were computed using the test partitions of the data. For each image, we performed five random transformations using as parameters the values at exactly half of the range used for training (see Section~\ref{sec:daugreg-methods_data}). Despite the presence of data augmentation during training, which implies that the networks \textit{sees} and may learn augmentation-invariant transformations, the representations of the baseline models (red boxes) do not increase substantially beyond the invariance observed in the pixel space (left-most boxes). To illustrate this, see the images in Figure~\ref{fig:invariance-sample_images}, whose representations are all equally distant to the reference image, despite the perceptual similarity of the transformed images. 

\begin{figure}[htb]
  \begin{center}
    \includegraphics[width = 0.8 \linewidth]{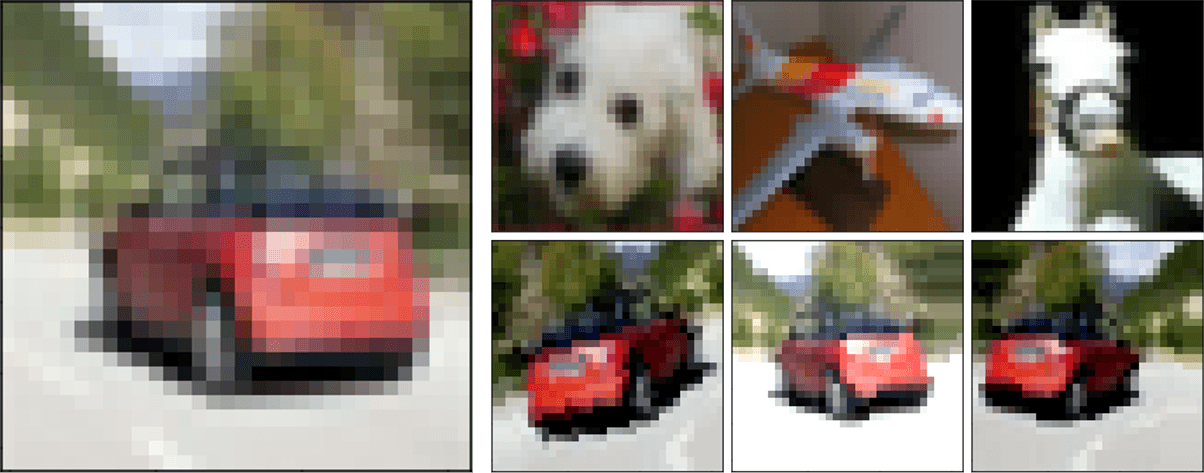}
  \end{center}
  \caption{The top layer representations of the six images on the right learnt by All-CNN are equally (dis)similar to the reference image (left), even though the images at the bottom row are transformations of it.}
  \label{fig:invariance-sample_images}
\end{figure}

As a solution, we have proposed a simple, label-free modification of the loss function to encourage the learning of data augmentation-invariant features. The blue boxes in Figures~\ref{fi:invariance-invariance_allcnn}--\ref{fi:invariance-invariance_densenet} show that our invariance mechanism pushed network representations to become increasingly more robust with network depth\footnote{Both All-CNN and WRN seem to more easily achieve the representational invariance on CIFAR-10 than on Tiny ImageNet. This may indicate that the complexity of the data set not only makes the object categorisation more challenging, but also the learning of invariant features.}. As discussed above, this is a well-studied property of the visual ventral stream in the primate brain. 

In order to better understand the effect of the data augmentation invariance objective, we analysed the learning dynamics of the invariance loss at each layer of All-CNN trained on CIFAR-10. In Figure~\ref{fig:dynamics}, we see that in the baseline model, the invariance loss keeps increasing over the course of training. In contrast, in the models trained with data augmentation invariance, the loss drops for all but the last layer. Perhaps unexpectedly, the invariance loss increases during the first epochs and only then starts to decrease. While further investigation is required, these two phases may correspond to the fitting and compression-diffusion phases proposed in the framework of the information bottleneck principle by \citet{shwartz2017bottleneck}. According to the authors, during the first epochs of optimisation with SGD, the model increases the information about the labels (fitting) and during the rest of training, the model reduces the information on the input (compression). However, note that \citet{saxe2019informationbottleneck} have argued that this occurs only in some cases that depend on the non-linearities.

\begin{figure}[ht]
  \begin{center}
    \includegraphics[width = \linewidth]{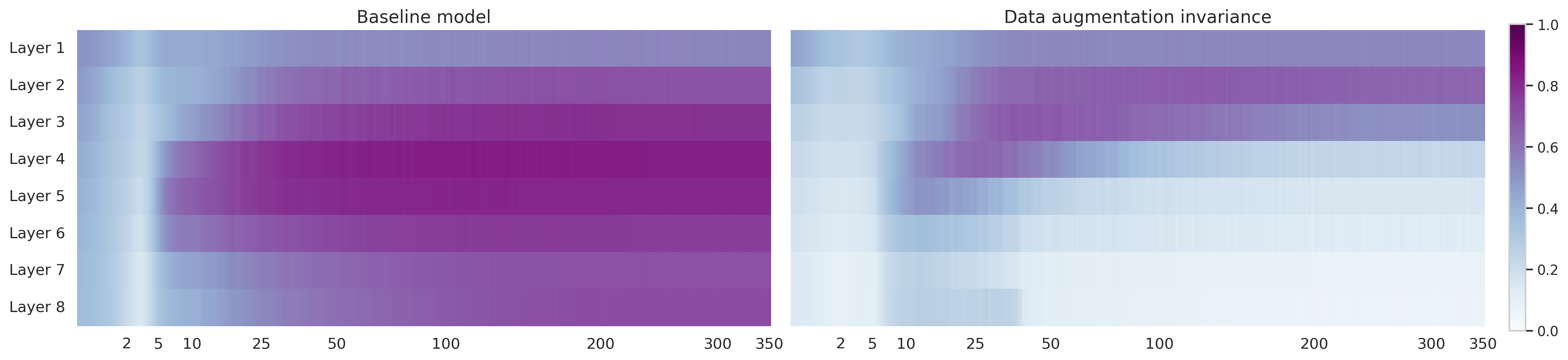}
  \end{center}
  \caption{Dynamics of the data augmentation invariance loss $\mathcal{L}_{inv}^{(l)}$ during training (lower is better). The axis of abscissas (epochs) is scaled quadratically to better appreciate the dynamics at the first epochs. The same random initialisation was used for both models.}
  \label{fig:dynamics}
\end{figure}

In terms of efficiency, adding terms to the objective function implies an overhead of the computations. However, since the pairwise distances can be efficiently computed at each batch through matrix operations, the training time was only increased by about 10 \% on average. 

Finally, as reported in Table~\ref{tab:accuracy}, the improved invariance comes at little or no cost in the categorisation performance, as the networks trained with data augmentation invariance achieved similar classification performance to the baseline model and in some cases it clearly improved it. This is remarkable as the hyperparameters used in all cases were optimised to maximise the performance in the original models, trained without data augmentation invariance. Therefore, it is reasonable to expect an improvement in the classification performance if, for instance, the batch size or the learning rate schedule are better tuned for this new learning objective. Learning increasingly invariant features could lead to more robust categorisation, as exemplified by the increased test performance observed for the All-CNN models---despite no hyperparameter tuning.

\begin{table}[tb]
  \begin{center}
    \begin{tabular}{rccccc}
	  & \multicolumn{3}{c}{CIFAR-10} & \multicolumn{2}{c}{Tiny ImageNet (acc. | top5)}          \\
      \cline{2-6} 
	                               & All-CNN & WRN   & DenseNet & All-CNN       & WRN   \\
	  Baseline                     & 91.48   & 94.58 & 94.88    & 51.09 | 73.48 & 61.49 | 82.99 \\
	  Data aug. invariance         & 92.47   & 94.86 & 93.98    & 52.57 | 76.53 & 61.23 | 83.23
    \end{tabular}
  \end{center}
  \caption{Classification accuracy on the test set of the baseline models and the models trained with data augmentation invariance.}
  \label{tab:accuracy}
\end{table}

\section[Learning representational invariance \textit{instead} of categorisation]{Learning representational invariance \textit{instead}\\of categorisation}
\label{sec:invariance-instead_categorisation}
In view of the effectivity of the data augmentation invariance learning objective, it is reasonable to wonder whether such an objective could fully replace the standard categorisation objective commonly used in so-called\footnote{See the discussion about supervised learning in the Introduction (Section~\ref{sec:intro-rethinking_supervised})} supervised learning. There are multiple reasons why exploring alternatives to classification objectives is attractive. In the Introduction of this thesis we have discussed the benefits of incorporating inductive biases from visual perception and biological vision in the form of objective functions and the results presented above in this chapter have demonstrated the usefulness of data augmentation to promote invariant representations. 

A remarkable example of the mismatch between ANNs and primate visual perception is the well-known vulnerability of the former to adversarial perturbations \citep{szegedy2013adversarial, dujmovic2020adversarial}. Recent work by \citet{ilyas2019advfeatures} has suggested that adversarial vulnerability might be caused by highly discriminative features present in the data yet incomprehensible to humans. In the same line, it has been recently shown \cite{wang2019highfreq} that ANNs learn high-frequency components of images, imperceptible to humans, but useful for categorisation. A related idea was suggested earlier by \citet{jo2017surfaceregularities}. Notably, this is only one example of the differences between current artificial and biological visual object perception \citep{sinz2019dlvsbrain, geirhos2020shortcutlearning, malhotra2020bioconstraints}. We hypothesise that some of these undesired properties might be caused by the optimisation of the specific task of classification. 

In this section we present the results of a exploratory, preliminary study where we aim to replace the standard categorical-cross entropy objective by a combination of data augmentation invariance and a similarly defined \textit{class-wise invariance}, which uses the labels of the images.

\subsection{Class-wise invariance}
\label{sec:invariance-classwise}
Class-wise invariant representation learning \citet{belharbi2017classinvariance} was introduced as a regularisation term that encourages similarity in the representations of objects from the same class. The authors showed that class-wise invariance helps improve generalisation, especially when few examples are available. Related ideas have been proposed in the metric learning and clustering literature.

Class-wise invariance is interesting because, in spite of simply optimising the prediction of the object labels, it sets the learning objective on how the intermediate features should be like. However, used on its own it would possibly be subject to some of the same undesirable properties of purely supervised methods. We hypothesise that combined, data augmentation and class-wise invariance alone---without a categorisation objective---may learn robust, discriminative features. 

We define the class-wise invariance loss at layer $l$ of a neural network $\mathcal{L}_{C}^{(l)}$ as a parallel to the data augmentation invariance loss (Equation~\ref{eq:invariance-data_aug_inv}):

\begin{equation}
\label{eq:invariance-class_wise_inv}
 \mathcal{L}_{C}^{(l)} = \frac{\sum_{r}\frac{1}{|\mathcal{T}_{r}|^2}\sum_{x_i, x_j \in \mathcal{T}_{r}}d^{(l)}(x_{i}, x_{j})}{\frac{1}{|\mathcal{B}|^2}\sum_{x_i, x_j \in \mathcal{B}}d^{(l)}(x_{i}, x_{j})}
\end{equation}
where $\mathcal{T}_{r}$ are the subsets from $\mathcal{B}$ formed by images of the same object class $r$. Let us denote by $\mathcal{L}_{DA}^{(l)}$ the data augmentation invariance loss (Equation~\ref{eq:invariance-data_aug_inv}). We propose to optimise, through stochastic gradient descent, the following overall objective:

\begin{equation}
\label{eq:invariance-daug_class_objective}
 \mathcal{L} = \sum_{l=1}^{L}\alpha^{(l)}\mathcal{L}_{DA}^{(l)} + \sum_{l=1}^{L}\beta^{(l)}\mathcal{L}_{C}^{(l)}
\end{equation}
where $\alpha^{(l)}$ and $\beta^{(l)}$ are scalars that control the degree of similarity between the features of augmented samples and of objects of the same category, respectively, at each layer $l$ of the architecture. Summarised, by jointly optimising $\mathcal{L}_{DA}^{(l)}$ and $\mathcal{L}_{C}^{(l)}$, we expect the model to learn robust features---as encouraged by the data augmentation invariance---while still allowing for categorisation---driven by the class-wise invariance.

\subsection{Results}
\label{sec:invariance-classwise_results}
In order to test this idea, we trained All-CNN on CIFAR-10 with the objective defined in Equation~\ref{eq:invariance-daug_class_objective}. We found that optimising this objective function as is, with the hyperparameters as in the models trained with standard categorical cross-entropy was \textit{not} able to perform multi-class classification. One hypothesis is that the learnt representations collapse along one single dimension, since the explained variance by the first principal component gets close to 100~\% and, therefore, the data points get separated in two clusters only. We have not found yet the exact cause of the undesired behaviour, but it may be due to the inability to escape local minima.

Despite the limitation of the fact that a 10-classes problem could not be optimised, in order to gain insights on the effect of the method, we analysed the representations learnt through the optimisation of invariance, instead of categorisation. In Figure~\ref{fig:invariance-mse_matrices}, we plot the representational dissimilarity matrices of the invariance model, alongside the baseline model trained with categorical cross-entropy and an additional model that jointly optimised both objectives. The latter model did achieved test performance on CIFAR-10 comparable to the baseline model. 

Interestingly, we found that the models optimised with invariance objectives learnt representations that naturally create meaningful clusters that separate animals and vehicle classes, that is animate and inanimate objects. This categorisation has been reported multiple times to be distinctively represented in the inferior temporal cortex of the primate brain \citep{kriegeskorte2008manandmonkey, bao2020itmaps}. This connection is still speculative and in the future we will further explore the representations learnt through this invariance objectives.

\begin{figure}[ht]
  \begin{center}
    \includegraphics[width = \linewidth]{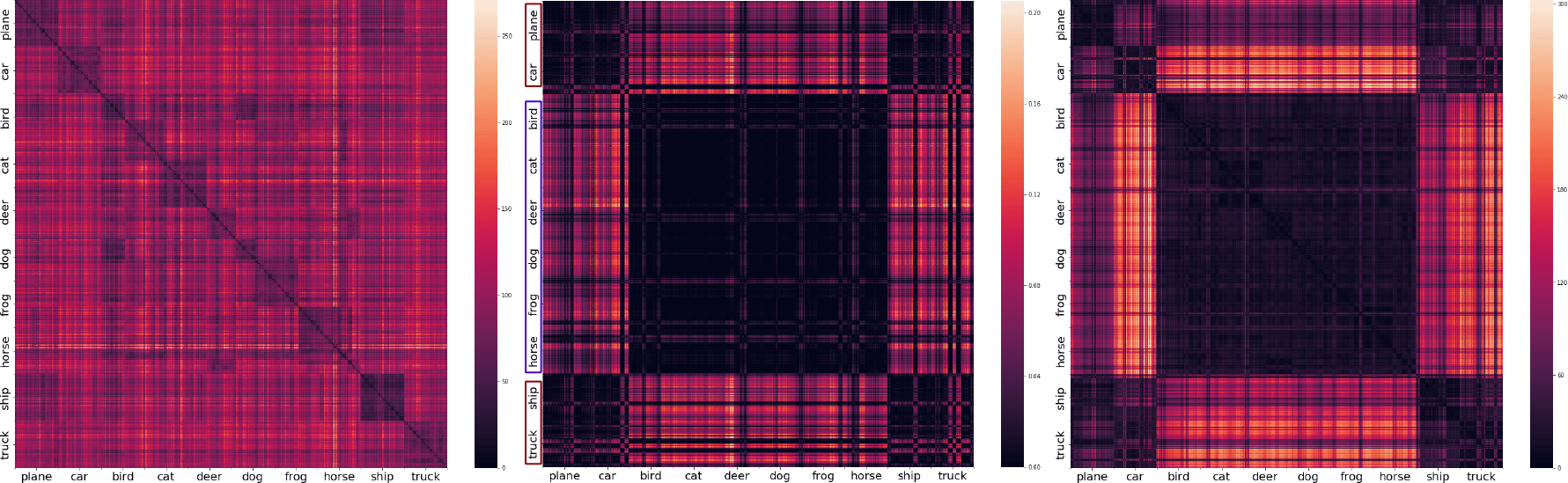}
  \end{center}
  \caption{Representational similarity matrices of All-CNN trained on CIFAR-10 with (left) standard categorical cross-entropy, (middle) the invariance loss defined in Equation~\ref{eq:invariance-daug_class_objective} and (right) the invariance loss plus a categorical-cross entropy term. In the models trained with invariance losses, meaningful hierarchical clusters (animals vs. vehicles) emerge.}
  \label{fig:invariance-mse_matrices}
\end{figure}

In an additional effort to further understand the learnt representations without the limitation of the low accuracy on 10 classes, we created a subset of CIFAR-10 with the images of cars and dogs only, that is a binary classification problem. This data set could be classified correctly with high accuracy. In Figure~\ref{fig:invariance-2d_representations}, we see that in the model trained with the invariance objective most examples of the two classes are separated by a larger margin than in the baseline model, where the clusters of the two classes are closer to each other.

\begin{figure}[ht]
  \centering
  \begin{subfigure}{0.4 \linewidth}
      \includegraphics[width = \linewidth]{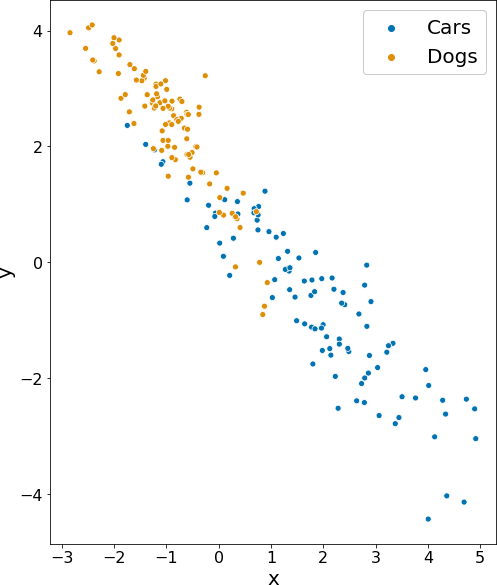}
      \caption{Baseline model}
  \end{subfigure}
  \begin{subfigure}{0.4 \linewidth}
      \includegraphics[width = \linewidth]{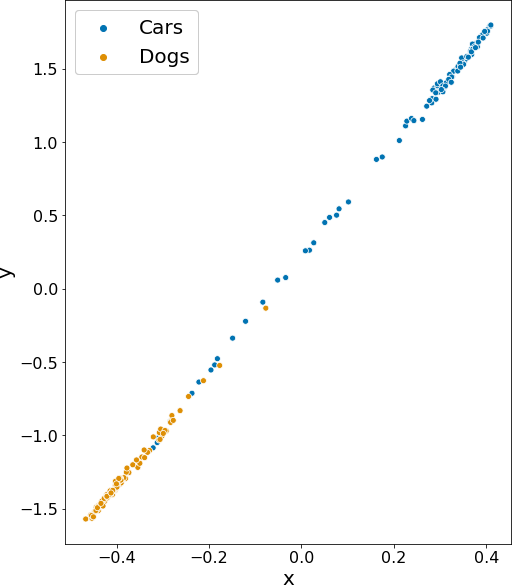}
      \caption{Invariance model}
  \end{subfigure}
  \caption{Representations of the test \textit{CIFAR-2} (cars and dogs) examples along the two principal components of the top layer representations of All-CNN.}
  \label{fig:invariance-2d_representations}
\end{figure}

Finally, we evaluated the adversarial robustness of the models. We had hypothesised that one of the reasons for adversarial vulnerability is that models optimised for categorisation only are highly unconstrained, with no specification of how the features should be. Therefore, the model focuses on learning any features that allows for linearly separable classes at the output of the network \citep{malhotra2020bioconstraints}. This has been shown to be prone to adversarial vulnerability, that is small perturbations largely affect the classification. In contrast, our invariance objective can be seen as the opposite strategy. It specifies how the features should be---increasingly invariant to identity-preserving transformations and relatively invariant within object classes---and expects good classification as a by-product. This could improve the sensitivity to adversarial perturbations and is what our results in Table~\ref{tab:invariance-adversarial_robustness} suggest: the model trained with the invariance objectives is remarkably more robust that the baseline models, without sacrificing performance.

\begin{table}[htb]
    \begin{center}
    \begin{tabular}{rcc}
                & Baseline (only cat.)  & Only invariance \\
        \hline
        Clean examples                 & 98.9~\% & 98.7~\%\\
        Attack: PGD $\varepsilon=0.03$  & 7.6~\%  & 62.0~\%\\
        Attack: FGSM $\varepsilon=0.03$ & 31.5~\% & 88.5~\%\\
    \end{tabular}
    \end{center}
    \caption{Adversarial robustness of the baseline (only categorisation) and invariance models trained on \textit{CIFAR-2}. Without any adversarial training, the model trained with the invariance objective only is highly robust to adversarial perturbations, in contrast to the standard, categorisation model.}
\label{tab:invariance-adversarial_robustness}
\end{table}

\section{Discussion}
\label{sec:invariance-discussion}
In this chapter, we have first proposed an invariance score that assesses the robustness of the features learnt by a neural network towards the identity-preserving transformations typical of common data augmentation schemes (see Equation~\ref{eq:invariance-invariance}). Intuitively, the more similar the representations of transformations of the same image, with respect to other images in a data set, the higher the data augmentation invariance score. The data augmentation score is meaningful in that the transformations performed by perceptually plausible data augmentation schemes---which we consider here exclusively---are motivated by human visual perception and coincide largely with the transformations to which the higher visual cortex is invariant.

Using this score, we have analysed the representations learnt by three popular neural network architectures---All-CNN, WRN and DenseNet---trained on image object recognition tasks---CIFAR-10 and Tiny ImageNet. The analysis revealed that their features are less invariant than commonly assumed, despite sufficient exposure to matching image transformations during training. In most cases, the representational invariance did not even increase with respect to the original pixel space. This property is fundamentally different to the primate ventral visual stream, where neural populations have been found to be increasingly robust to changes in view or lighting conditions of the same object \citep{dicarlo2007untangling}.

Taking inspiration from this property of the visual cortex, we have proposed a label-free objective to encourage learning more robust features, using data augmentation as the framework to perform identity-preserving transformations on the input data. We constructed mini-batches with $M$ augmented versions of each image and modified the loss function to maximise the similarity between the activations of the same seed images, as compared to other images in the batch. Aiming to approximate the observations in the biological visual system, higher layers were set to achieve exponentially more invariance than the early layers. An interesting avenue for future work will be to investigate whether this increased robustness also allows for better modelling of neural data.

Data augmentation invariance effectively produced more robust representations, unlike standard models optimised only for object categorisation, at little or no cost in classification performance and with an affordable, slight increase (10~\%) in training time. Ideally, object recognition models should be reasonably invariant to all the transformations of the objects to which human perception is also invariant. Data augmentation is just an approximation to analyse and encourage invariance to a set of transformations that can be applied on still, 2D images. Future work should analyse the invariance of models trained with video \citep{taylor2010spatiotemporal} and even 3D data \citep{achlioptas20183d}.

These results provide additional empirical evidence that deep supervised models optimised only according to the standard categorisation objective---the categorical cross-entropy between the true object labels and the model predictions---learn discriminative but non-robust features. This is likely due to their large capacity to learn discriminative features in too unconstrained a setting \citep{geirhos2020shortcutlearning, malhotra2020bioconstraints}, which has been recently suggested to be at the source of adversarial vulnerability \citep{ilyas2019advfeatures}.

In order to explore whether getting rid of the categorisation objective can produce features more aligned with our intuitions from visual perception, less vulnerable to adversarial perturbations and potentially more similar to the brain representations, we proposed to replace the categorical cross-entropy with purely invariance objectives. In particular, we combined layer-wise data augmentation invariance with class-wise invariance, which encourages similarity of the features from images of the same class. Although our preliminary experiments did not succeed at achieving competitive categorisation performance, we made interesting observations in our analysis.

First, training with invariance objectives yields representations that are clustered hierarchically, while the dissimilarity matrix of the standard model is fairly homogeneous. Remarkably, the main clusters formed through invariance learning correspond to the animate and inanimate classes, a separation consistently observed in the primate visual cortex \cite{kriegeskorte2008manandmonkey, bao2020itmaps}. Furthermore, we trained a model on a binary classification task using invariance objectives only and found that the adversarial vulnerability is very low, in contrast to categorisation models, which exhibit very high sensitivity to adversarial perturbations. While these conclusions are still speculative, they set a promising path for future research on alternative objective functions that encode inductive biases from visual perception and biological vision.

\chapterbibliography
}

{
\chapter[Global visual salience of competing stimuli]{Global visual salience\\of competing stimuli}
\label{ch:globsal}
\renewcommand{\chapterpath}{includes/global-salience}
\begin{contributors}
    Ricardo Ramos Gameiro designed the first prototype of the experiments, collected the eye tracking data and contributed to the original draft of the manuscript. Alessandro Grillini contributed to the comparisons between global and local salience and to the corresponding part of the original draft. Peter K{\"o}nig contributed to the conceptualisation of the project and supervised the work. Ricardo, Alessandro and Peter reviewed and edited the manuscript submitted to the Journal of Vision.
\end{contributors}
\begin{outreach}
    \item \textit{Global visual salience of competing stimuli.} \textbf{Alex Hern{\'a}ndez-Garc{\'i}a}, Ricardo Ramos Gameiro, Alessandro Grillini, Peter K{\"o}nig. PsyArXiv preprint PsyArXiv:z7qp5 \& Journal of Vision (accepted), 2019.
\end{outreach}
Visual attention is a highly complex mechanism that facilitates our understanding and navigation of the world around us, by enabling the coherent processing of the large amount of information that enters our eyes. Therefore, a fundamental component of vision and hence cognition is the guidance of eye movements \citep{liversedge2000eyemovements, geisler2011eyemovements, konig2016eyemovements}. We constantly have to decide where to look next and which regions of interest to explore, in order to process and interpret relevant information of a scene. As a consequence, the investigation of eye movement behaviour has become a major field in many research areas \citep{kowler2011eyemovements, kaspar2013visualattention}. 

In this regard, a number of studies have shown that visual behaviour is controlled by three major mechanisms: bottom-up, top-down, and spatial biases \citep{desimone1995visualattention, egeth1997visualattention, kastner2000visualattention, corbetta2002topdownbottomup, connor2004buttomuptopdown, kollmorgen2010topdownbottomup}. Bottom-up factors describe features of the observed image, which attract eye fixations, involving primary contrasts, such as colour, luminance, brightness, and saturation \citep{itti1998salience, reinagel1999bottomup, baddeley2006bottomup}. Hence, bottom-up factors are typically based on the sensory input. In contrast, top-down factors comprise internal states of the observer \citep{connor2004buttomuptopdown, kaspar2013visualattention}. That is, eye movement behaviour is also guided by specific characteristics, such as personal motivation, specific search tasks, and emotions \citep{wadlinger2006topdown, einhauser2008topdown, rauthmann2012topdown, kaspar2012topdown}. Finally, spatial properties of the image, such as the image size, and motor constraints of the visual system in the brain may affect eye movement behaviour \citep{ramosgameiro2017explorationexploitation, ossandon2014spatialbiases}. As a result, spatial properties and motor constraints then lead to specific bias effects, such as the central bias in natural static images \citep{tatler2007centralbias}. Thus, investigating visual behaviour necessarily implies an examination of bottom-up and top-down factors as well as spatial biases. 

Based on these three mechanisms---bottom-up, top-down and spatial biases---guiding visual behaviour, \citet{koch1987salience} first revealed a method to highlight salient points in static image scenes. Whereas this model was purely conceptual, \citet{niebur1996salience} later developed an actual implementation of salience maps. This was the first prominent proposal of topographically organised features maps that guide visual attention. Salience maps describe these topographic representations of an image scene, revealing where people will most likely look at while observing the respective scene \citep{itti2001salience}. That is, salience maps can be interpreted as a prediction of the distribution of eye movements on images. Usually, salience maps include only bottom-up image features, predicting eye fixations on image regions with primary contrasts in colour changes, saturation, luminance or brightness among others \citep{itti1998salience}. However, in their first implementation, \citet{niebur1996salience} also tried to include top-down factors to build up salience maps and thus predict where people will look at most likely in image scenes. Current state-of-the-art computational salience models are artificial neural networks pre-trained on large data sets for visual object recognition and subsequently tuned to predict fixations, as is the case of Deep Gaze II \citep{kuemmerer2016deepgaze}. Such models do not rely only on bottom-up features any more, but also incorporate higher-level features learned on object recognition tasks. Still, despite the better performance on salience benchmarks, deep nets-based models seem to fail at predicting the salience driven by low-level features \citep{kuemmerer2017icfdeepgaze}.

Salience maps provide a highly accurate and robust method to predict human eye movement behaviour on static images, by relying on local features to determine which parts of an image are most salient \citep{niebur1996salience, itti2001salience, kowler2011eyemovements}. However, these methods do not provide any information about the salience of the image as whole, which may depend on both local properties and also the overall semantic and contextual information of the image. Such global salience is of great relevance when an observer is faced with two or more independent visual stimuli in one context. These combinations describe situations when several stimuli compete with each other with regard to their individual semantic content, despite being in the same overall context. Such cases appear frequently in real life, for instance when two billboards hang next to each other in a mall, when several windows are open on a computer screen, or a monitor on intensive care unit, to name a few examples. Thus, by placing two or more independent image contexts side by side, as described in the previous examples, classical salience maps may well predict eye movement behaviour within each of the individual images as a closed system, but they will most likely fail to predict visual behaviour across the whole scene involving all images. Specifically, they will fail at answering the question: which stimulus is most likely to attract the observers' visual attention?

\section{Hypotheses and contributions}
\label{sec:globsal-contributions}
In this chapter, we present the work of a study where we postulate several hypotheses. Our primary hypothesis (H1) is that it is possible to measure and calculate the global salience of natural images. That is, the likelihood of a visual stimulus to attract the first fixation of a human observer, when it is presented in competition alongside another stimulus, can be systematically modelled. In the experiment presented here, participants were confronted with stimuli containing two individual natural images---one on the left and one on the right side of the screen---at the same time. The set of images used to build our stimuli consisted of urban, indoor and nature scenes, close-ups of human faces and scenes with people in a social context. During the observation of the image pairs, we recorded the participants' eye movements. Specifically, to characterise the global salience we were interested in the direction---left or right---of the initial saccade the participant made after the stimulus onset. For further analysis, we also collected all binary saccade decisions on all the image pairs presented to the participants. We used the behavioural data collected from the participants to train a logistic regression model that successfully predicts the location of the first fixation for a given pair of images. This allowed us to use the coefficients of the model to characterise the likelihood of each image to attract the first fixation, relative to the other images in the set. In general, images that were fixated more often are ranked higher than other images. Hence, we computed a unique \textit{attraction score} for each image that we denote \textit{global salience}, which depends on the individual contextual information of the image as a whole. 

We also analysed the local salience properties of the individual images and compared it to the global salience. We hereby claimed that the global salience cannot be explained by the feature-driven salience maps. Formally, we hypothesise that (H2): Natural images have a specific global salience, independent of their local salience properties, that characterises their likelihood to attract the first fixation of human observers, when presented alongside another competing stimulus. A larger global salience leads to a higher attraction of initial eye movements.  

In order to properly calculate the global salience, we accounted for general effects of visual behaviour in stimuli with two paired images. Previous studies have shown that humans tend to exhibit a left bias in scanning visual stimuli. \citet{barton2006leftbias} showed that subjects looking at faces fixated longer the eye on their left side, even if the faces were inverted, and the effect was later confirmed and extended to dogs and monkeys \citep{guo2009leftbias}. For an extensive review about spatial biases see the work by \citet{ossandon2014spatialbiases}, where the authors presented evidence of a marked initial left bias in right-handers, but not in left-handers, regardless of their habitual reading direction. In sum, there is a large body of evidence of lateral asymmetry in viewing behaviour, although the specific sources are yet to be fully confirmed. With respect to our study, we hypothesise that (H3): Presenting images in horizontal pairs leads to a general spatial bias in favour of the image on the left side. 

In addition to the general left bias, in half of the trials of the experimental sessions, one of the images had been already seen by the participant in a previous trial, while the other was new. The participants also had to indicate which of the images was new or old. Thus, we also addressed the questions of whether the familiarity with one of the images or the task have any effect in the visual behaviour and thus in the global salience of the images. Do images that show the task-relevant scene attract more initial saccades? Likewise, are novel images more likely to attract the first fixation? This challenge sheds some light on central-peripheral interaction in visual processing. \citet{guo2007topdown}, for instance, showed that during face-processing, humans indeed rely on top-down information in scanning images. However, \citet{acik2010bottomuptopdown} proposed that young adults usually rely on bottom-up rather than top-down information during visual search. In this regard, we thus hypothesise that (H4): Task-relevance and familiarity of images will not lead to higher probability of being fixated first. In order to account for any spatial bias effects that could influence the global salience model, we added coefficients to the logistic regression algorithm that could potentially capture any lateral, familiarity and bias effects. This not only makes the model more accurate, but allows us to analyse the influence of these effects. Furthermore, the location of the images in the experiments was randomised across trials and participants. 

Finally, in order to better understand the properties of the global salience of competing stimuli, we also analysed the exploration time of each image. In this regard, we hypothesise the following (H5): Images with larger global salience will be explored longer than images with low global salience.

\section{Methods: experimental setup}
\label{sec:globsal-methods}
The present study was conducted in the Neurobiopsychology lab at the Institute of Cognitive Science of the University of Osnabr\"uck, Germany. The experimental methods were approved by the Ethical Committee of the University of Osnabr\"uck, Germany, and performed in accordance with the guidelines of the German Psychological Society. All participants gave written consent to participate in this study. 

\subsection{Participants}
\label{sec:globsal-methods_participants}
Forty-nine healthy participants (33 females, mean age = 22.39 years, \textit{SD} = 3.63) with normal or corrected-to-normal vision took part in this study. All participants were instructed to freely observe the stimuli on the screen. In part of the measurements, they had to indicate after the trial the old or new image of a pair as further described below.  

\subsection{Apparatuses}
\label{sec:globsal-methods_apparatuses}
We presented the stimuli on a 32'' widescreen Samsung monitor (Apple, California, USA) with a native resolution of 3840 $\times$ 2160 pixels. For eye movement recordings, we used a stationary Eye Link 1000 eye tracker (SR Research Ltd.) providing binocular recordings with one head camera and two eye cameras with sampling rate of 500 Hz.

Participants were seated in a darkened room at a distance of 80 cm from the monitor, resulting in 80.4 pixels per visual degree in the centre of the monitor. We did not fixate the participant's head with a headrest but verbally instructed the participants not to make head movements during the experiment. This facilitated comfortable conditions for the participants. However, eye tracker constantly recorded four edge markers on the screen with the head camera, in order to correct for small head movements. This guaranteed stable gaze recordings based on eye movements, independent of residual involuntary head movements. 

The eye tracker measured binocular eye movements. For calibration of the eye tracking camera, each participant had to fixate on 13 black circles (size 0.5\textdegree) that appeared consecutively at different screen locations. The calibration was validated afterwards by calculating the drift error for each point. The calibration was repeated until the system reached an average accuracy of \textless~0.5\textdegree\ for both eyes of the participant.

\subsection{Stimuli}
\label{sec:globsal-methods_stimuli}
The images set consisted of 200 images, of which 197 were natural photographs and 3 were randomly generated pink noise images. Altogether, the stimulus set was divided into 6 categories, according to the image content: human faces, urban scenes, natural landscapes, indoor scenes, social activities and pink noise. All the photographs were obtained from either the internal image database of the Neurobiopsychology laboratory at the University of Osnabr\"uck, Germany or the NimStim database. Each image was scaled to a resolution of 1800 $\times$ 1440 pixels. Some examples are shown in Figure~\ref{fig:globsal-stimuli_samples}.

Each trial consisted of one stimulus with a resolution of 3840 $\times$ 2160 pixels, matching the full-size screen resolution of the display monitor (32'' diagonal; 47.8\textdegree\ $\times$ 26.9\textdegree). Within each presented stimulus, two images were randomly paired that is, one image was shown on the left screen side and the other image on the right screen side. Between both images, each stimulus contained a central gap of 240 pixels, as illustrated by Figure~\ref{fig:globsal-stimuli_pair}. The background area of the stimuli was set to middle grey.

\begin{figure}[ht]
  \centering
  \begin{subfigure}{0.48 \linewidth}
      \includegraphics[width = \linewidth]{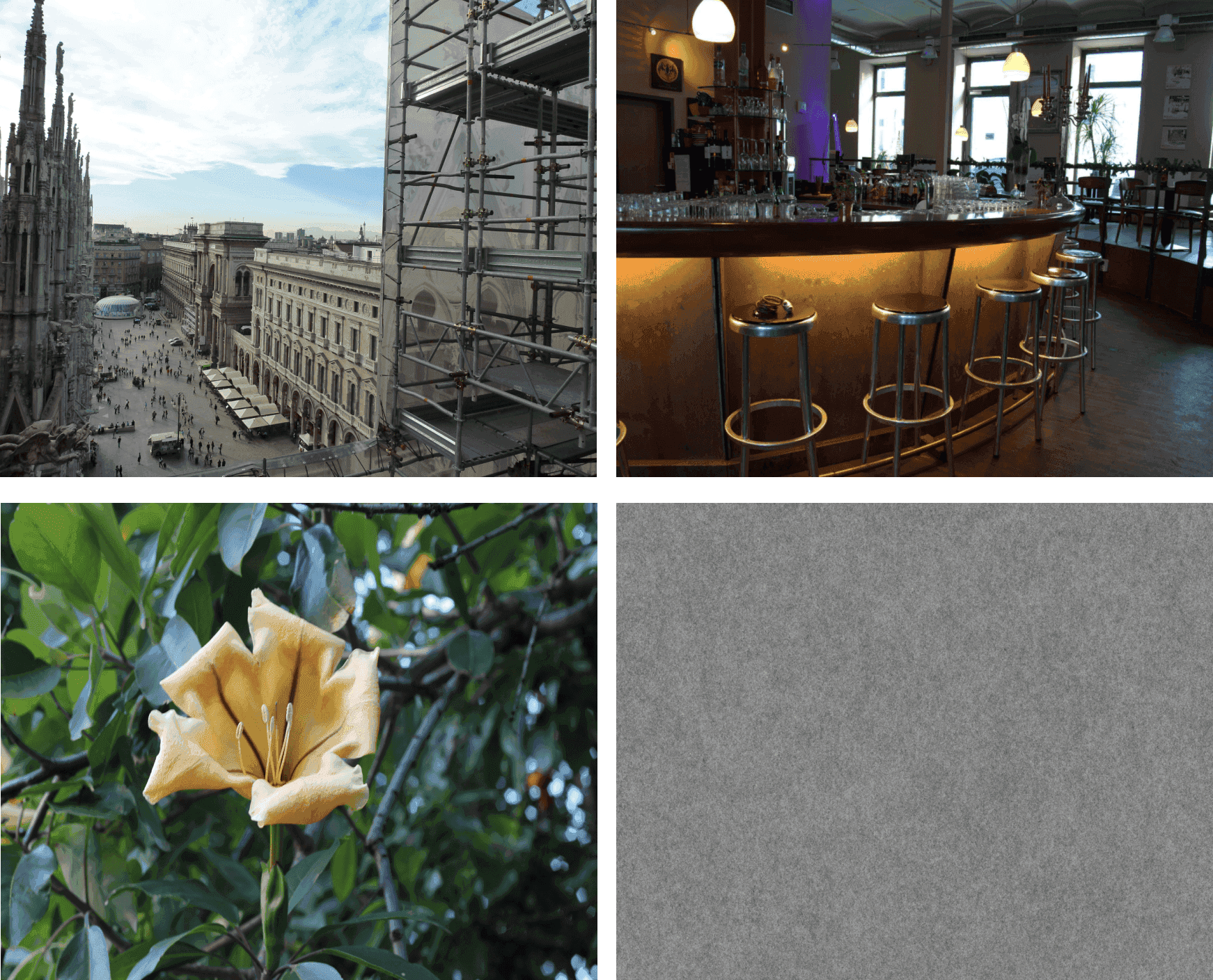}
      \caption{Four images from the set}
    \label{fig:globsal-stimuli_samples}
  \end{subfigure}
  \hspace{0.02 \linewidth}
  \begin{subfigure}{0.48 \linewidth}
      \centering
      \includegraphics[width = \linewidth]{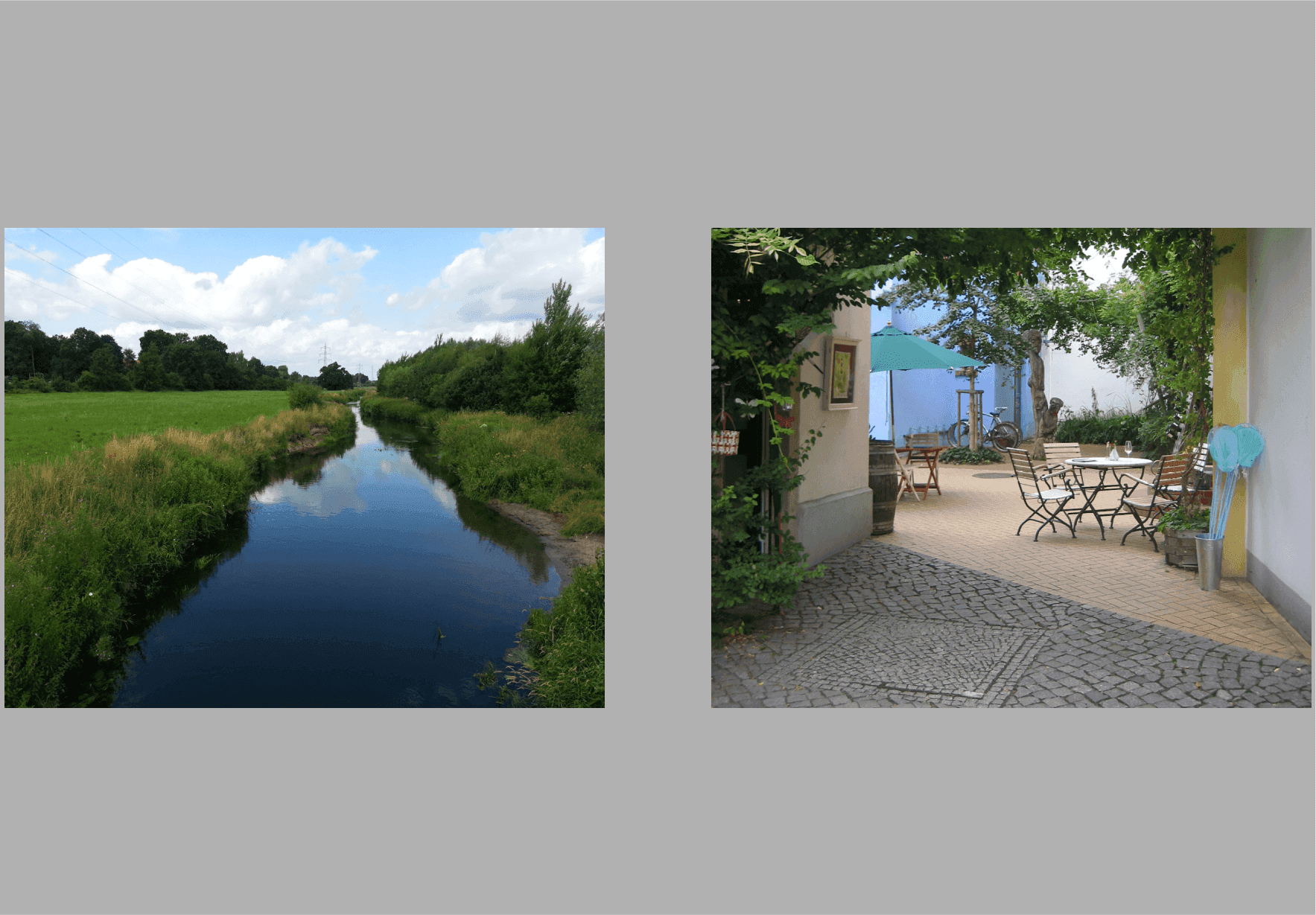}
      \caption{The stimulus at one trial}
    \label{fig:globsal-stimuli_pair}
  \end{subfigure}
  \\
  \begin{subfigure}{0.43 \linewidth}
      \centering
      \includegraphics[width = \linewidth]{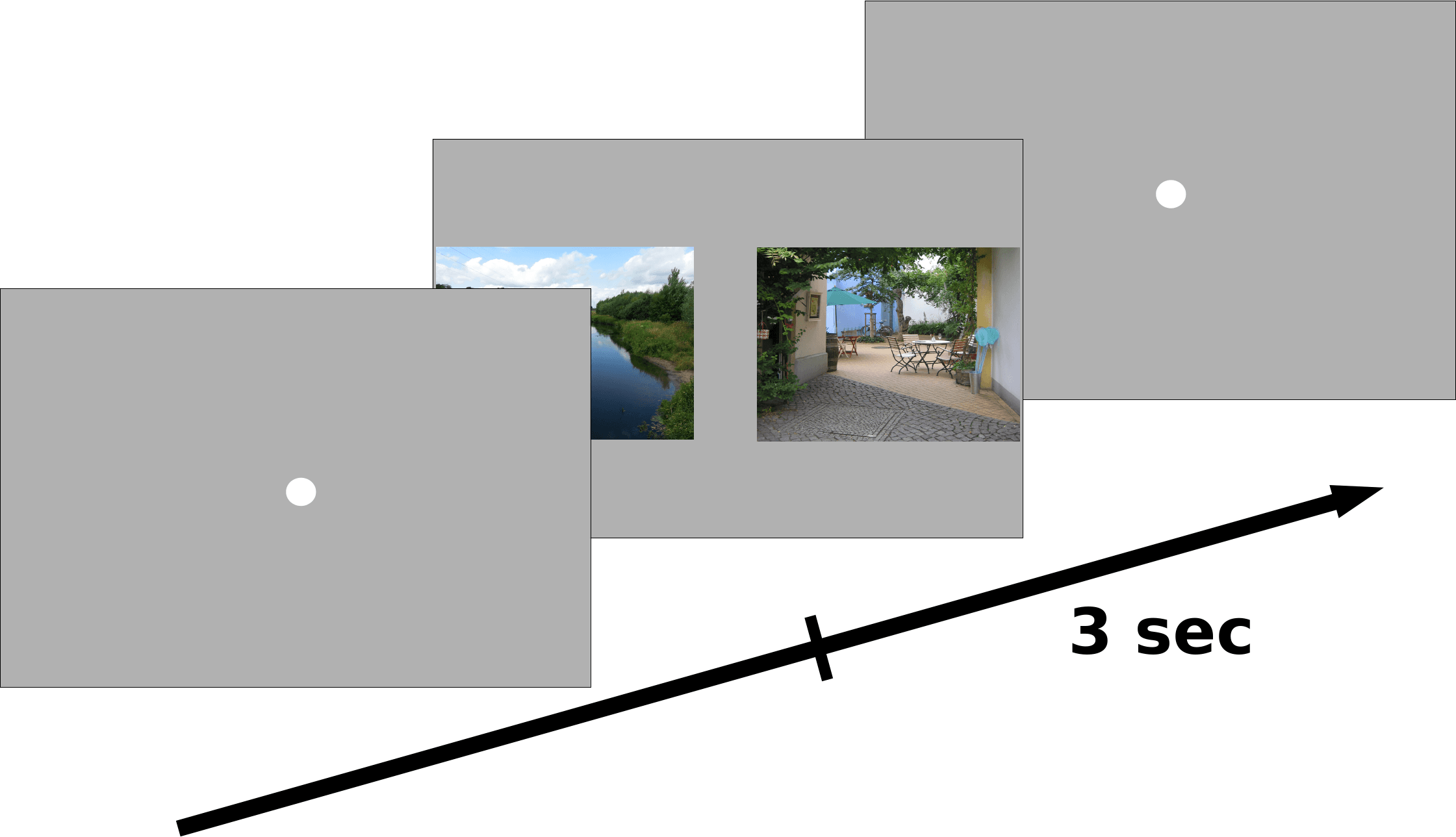}
      \caption{Procedure in blocks 1 and 4, where no task was given}
    \label{fig:globsal-trial_notask}
  \end{subfigure}
  \hspace{0.02 \linewidth}
  \begin{subfigure}{0.53 \linewidth}
      \centering
      \includegraphics[width = \linewidth]{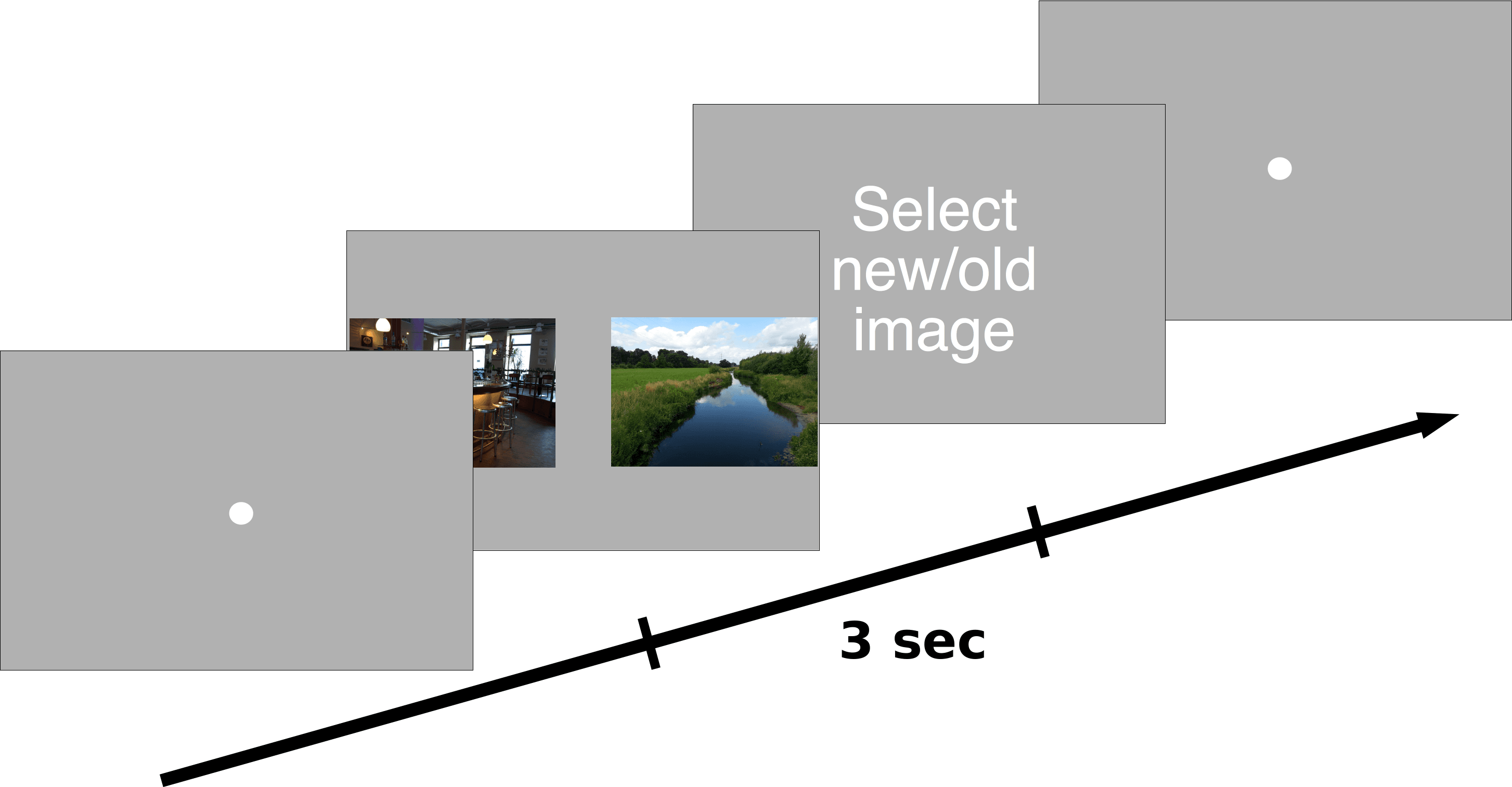}
      \caption{Procedure in blocks 2 and 3, where participants had to perform a simple task after the stimulus offset}
    \label{fig:globsal-trial_task}
  \end{subfigure}
  \caption{Experimental setup}
\label{fig:globsal-experiment}
\end{figure}

\subsection{Procedure}
\label{sec:globsal-methods_procedure}

The experiment consisted of 200 trials divided into four blocks, at the beginning of which the eye-tracking system was re-calibrated. The blocks were designed such that each had a different combination of task and image novelty:

\begin{itemize}
    \item \textbf{Block 1} consisted of 25 trials formed by 50 distinct, novel images (new/new). This block was task-free, that is participants were guided to freely observe the stimuli (Figure~\ref{fig:globsal-trial_notask}).
    \item \textbf{Block 2} consisted of 75 trials, each formed by one new image and one of the previously seen images. (new/old or old/new). In this block, the participants were guided to freely observe the stimuli and, additionally, they were asked to indicate the \textit{new} image of the pair after the stimulus offset (Figure~\ref{fig:globsal-trial_task}).
    \item \textbf{Block 3} consisted of 75 trials, each formed by one new image and one of the previously seen images. (new/old or old/new). In this block, the participants were asked to indicate the \textit{old} image of the pair.
    \item \textbf{Block 4} consisted of 25 trials formed by 50 previously seen images (old/old). Like block 1, this block was also task-free.
\end{itemize}

The decision in blocks 2 and 3 was indicated by either pressing the left---task-relevant image is on the left side---or right---task relevant-image is on the right side---arrow button on a computer keyboard. 

The image pairs were formed by randomly sampling from the set of 200 images, but some constraints were set in order to satisfy the characteristics of each block and keep a balance in the number of times each image was seen by the participant. The sampling process was as follows: In block 1, 50 images were randomly sampled to form the 25 pairs. In blocks 2 and 3, in order to construct the new/old and old/new pairs, the new image was randomly sampled from the set of remaining unseen images and the old image was randomly sampled of previously seen images, with two additional constraints: it must have been chosen only one time before and not in the previous 5 trials. Finally, in block 4, a set of exactly 50 images which had been shown only once remained. These were used to randomly sample the remaining 25 trials. In all blocks, after sampling the two images, the left/right configuration was also randomly chosen with probability 0.5.
    
The sampling process was different for each participant, that is they saw different sets of pairs from the 40,000 different pairs and in different order. This aimed at reducing the predictability of the process while satisfying the experimental constraints. Overall, we collected data from 9,800 pairs, some of which might have been repeated across participants. However, note that each participant saw each image exactly twice, therefore the frequency of presentation of the images was balanced across the whole experiment. As we will see in the following section, the amount of data was enough to fit the computational model. 

In all cases, the presentation time for each stimulus was 3 seconds and it was always preceded by a blank, grey screen with a white, central fixation dot. The stimulus was displayed only after the participant fixated the central dot. 

The majority of our analyses focused on the first fixation. As a pre-processing stage, we discarded the fixations 1) due to anticipatory saccades, 2) shorter than 50 ms or longer than $\mu_{dur} + 2 \sigma_{dur}$ ms, where $\mu_{dur} = 198$ ms and $\sigma_{dur} = 90$ ms are the mean and standard deviation of all fixation durations, respectively, and 3) located outside any of the two images. The discarded fixations were less than 4~\% of the total.

\section{Methods: computation of global salience}
\label{sec:globsal}
In order to characterise the global salience of competing stimuli, we trained a logistic regression model with the behavioural data from the eye-tracking experiments. Provided that the model can accurately predict the location of the first fixation---left or right---the coefficients for each image will represent the likelihood of the image to attract the first fixation and this, in turn, can then be interpreted as the global image salience. The intuition is that images that are more often the target of the first fixation after the stimulus onset have a higher global salience, and vice versa.

\subsection{Logistic regression for pairwise estimates}
\label{sec:globsal-logreg_plain}
Typically, logistic regression is used in binary classification problems, as is this case where the initial fixation after stimulus onset can land either on the left ($y = -1$) or on the right ($y = 1$) image. The classifier simply estimates a probability $h_{w}(\mathbf{x})$ for the binary event on the linear hypothesis $\mathbf{w}^{T}\mathbf{x}$ by applying a logistic function:

\begin{equation}
\label{eq:logreg}
 h_{w}(\mathbf{x}) = \frac{1}{1 + e^{-\mathbf{w}^{T}\mathbf{x}}} = \frac{e^{\mathbf{w}^{T}\mathbf{x}}}{1 + e^{\mathbf{w}^{T}\mathbf{x}}}
\end{equation}
where $\mathbf{x}$ is a vector that represents the independent or explanatory variables (features) and $\mathbf{w}$ the coefficients to be learned. Thus, the likelihood of the binary outcome given the data is the following:

\[   
P(y|\mathbf{x}) = 
     \begin{cases}
       h_{w}(\mathbf{x})  &\quad\text{if } y = 1\\
       1 - h_{w}(\mathbf{x})  &\quad\text{if } y = -1\\
     \end{cases}
= \frac{e^{y\mathbf{w}^{T}\mathbf{x}}}{1 + e^{y\mathbf{w}^{T}\mathbf{x}}}
\]

The coefficients are then optimised by minimising the negative log-likelihood $-log(P(y|\mathbf{x}))$ through gradient descent. Typically, a regularisation penalty is added on the coefficients, controlled by the parameter $C$---inverse of the regularisation strength. In our case, we applied $L_2$ regularisation and therefore the algorithm solves the following optimisation problem, given a set of $N$ training data points (trials):

\begin{equation}
\label{eq:objective}
\min_{\mathbf{w}} \frac{1}{2}\mathbf{w}^{T}\mathbf{w} + C\sum_{i=1}^{N}log(1 + e^{-y_{i}\mathbf{w}^{T}\mathbf{x}_{i}})
\end{equation}

The optimisation problem was solved through the \textit{LIBLINEAR} algorithm \citep{fan2008liblinear}, available in the \texttt{scikit-learn} Python toolbox.

In our particular case, for every trial $i$---stimulus pair seen by a participant---each feature $x_{ij}$ corresponded to one image $j$ and only two images were shown at each trial. Therefore, we were interested in modelling the probability that one image $u$ receives the first fixation when presented next to another image $v$; hence $p(u > v)$. This simplifies the standard logistic regression model to a special case for pairwise probability estimates, known as the Bradley-Terry-Luce model \citep{bradley1952pairwisecomp, luce2005pairwisecomp}, where the probability $h_{w}$ is the following:

\begin{equation}
\label{eq:btl}
 h_{w}(u,v) = p(u > v) = \frac{e^{w_{u}}}{e^{w_{u}} + e^{w_{v}}} = \frac{e^{w_{u}-w_{v}}}{1 + e^{w_{u}-w_{v}}}
\end{equation}
where $w_{u}$ and $w_{v}$ are the coefficients of image $u$ and $v$. This is a special case of the function in Equation \ref{eq:logreg}, where all the elements in the feature vector $\mathbf{x}$ are zero except for the two paired features $x_{u}$ and $x_{v}$, which are set to 1 and -1 respectively. Note that in the Bradley-Terry-Luce model the coefficients still refer to the whole set of features and therefore are described by an $M$-dimensional vector $\mathbf{w} = \{w_{1}, w_{2}, ..., w_{M}\}$, where in our case $M=200$, the total number of images in the set. After training the model, each learned coefficient $w_{j}$ will be related to the average likelihood of image $j$ of receiving the first fixation when presented next to other images from the set. As stated above, we interpret these coefficients $\mathbf{w}$ as a measure of the global image salience.

In order to estimate the coefficients $\mathbf{w}$, the logistic regression model was trained on the data set arranged into a design matrix $X$ of the following form:

\begin{equation}
\label{eq:matrix_x}
X = 
\begin{bmatrix}
    x_{1}^{(1)} & x_{2}^{(1)} & \dots  & x_{M}^{(1)} \\
    x_{1}^{(2)} & x_{2}^{(2)} & \dots  & x_{M}^{(2)} \\
    \vdots      & \vdots      & \ddots & \vdots      \\
    x_{1}^{(N)} & x_{2}^{(N)} & \dots  & x_{M}^{(N)} \\
\end{bmatrix}
\end{equation}
where each row represents one measured data point: one trial where one participant was presented a pair of images $u$ and $v$---the total number of trials was in our case $N = 49~\text{participants} \times 200~\text{trials per participant} = 9800$---and where the columns represent the values of the different features (images) that were tested ($M = 200$). According to Equation~\ref{eq:btl}, if image $u$ is presented on the right and image $v$ is presented on the left at trial $i$, then $x_{u}^{(i)}=1$, $x_{v}^{(i)}=-1$ and $x_{j}^{(i)}=0, \forall~j \neq u,v$. Finally, the outcome of each trial is given as a vector $\mathbf{y}$:

\[
\mathbf{y} =
\begin{bmatrix}
    y^{(1)} \\
    y^{(2)} \\
    \vdots  \\
    y^{(N)} \\
\end{bmatrix}
\]
such that $y^{(i)} = 1$ if the right image was fixated first, and $y^{(i)} = -1$ if the left image was fixated first at trial $i$. 

\subsection{Task, familiarity and lateral bias} 
\label{sec:globsal-biases}
Not only were we interested in modelling the likelihood of every image of receiving the first fixation, but also the contribution of other aspects of the experiment, namely the effect of having to perform a small task when observing the pair of images and the familiarity with one of the two images. More specifically, we were interested in answering the following questions: Do light task demands, such as having to determine  which image is new or old, influence the direction of the first saccade? And, are unseen stimuli more likely to receive the initial saccade than previously observed stimuli when presented together, or vice versa?

We addressed these questions by adding new features to the model that capture these characteristics of the experimental setup. These features were assigned coefficients that, after training, will indicate the magnitude of the contributions of the effects. In particular, we added the following features columns to every row $i$ of the design matrix:

\begin{itemize}
 \item $t^{(i)}$: 1 if the target of the task (select new/old image) was on the right at trial $i$, -1  image if on the left, 0 if no task.
 \item $f^{(i)}$: 1 if at trial $i$, the image on the right had been already shown at a previous trial (familiar), while the image on the left was still unseen; -1 if the familiar image was on the left; 0 if both images were new or familiar.
\end{itemize}

Not only did these new features enable new elements for the analysis, but also added more representational power to the model, which could potentially learn better coefficients to describe the global salience of each image. In this line, we added one more feature to the model to capture one important aspect of visual exploration: the lateral bias. Although a single intercept term in the argument of the logistic function ($\mathbf{w}^{T}\mathbf{x} + b)$ would capture most of the lateral bias, since the outcome $\mathbf{y}$ describes exactly the lateral direction, left or right, of the first saccade, we instead added subject-specific features to model the fact that the trials were generated by different subjects with an individual lateral bias. This was done by adding $K = 49$ (number of participants) features $s_{k}^{(i)}$, with value 1 if the trial $i$ was performed by subject $k$ and 0 otherwise. Altogether, the final design matrix $X^{\prime}$ extends the design matrix $X$ defined in Equation~\ref{eq:matrix_x} as follows:

\begin{equation}
\label{eq:matrix_x_prime}
X^{\prime} = 
\begin{bmatrix}[ccc|c|c|ccc]
    x_{1}^{(1)} & \dots  & x_{M}^{(1)} & t^{(1)} & f^{(1)} & s_{1}^{(1)} & \dots  & s_{K}^{(1)} \\
    x_{1}^{(2)} & \dots  & x_{M}^{(2)} & t^{(2)} & f^{(2)} & s_{1}^{(2)} & \dots  & s_{K}^{(2)} \\
    \vdots      & \ddots & \vdots      & \vdots  & \vdots  & \vdots      & \ddots & \vdots      \\
    x_{1}^{(N)} & \dots  & x_{M}^{(N)} & t^{(N)} & f^{(N)} & s_{1}^{(N)} & \dots  & s_{K}^{(N)} \\
\end{bmatrix}
\end{equation}

Note that the leftmost block of $X^{\prime}$ is identical to $X$ (defined in Equation~\ref{eq:matrix_x}). While the shape of $X$ is $9800 \times 200$, $X^{\prime}$ is a $9800 \times 251$ matrix, since $200 + 1 + 1 + 49 = 251$.

\subsection{Validation and evaluation of the model} 
\label{sec:globsal-model_eval}
In order to ensure the successful training of the model, we carried out a 5-fold cross-validation of the regularisation parameter $C$ of the model, described in Equation~\ref{eq:objective}. That is, we split our data set into 5 different folds of 39 subjects for training and 10 for validation---7,800 and 2,000 trials, respectively---and evaluated the performance with 10 different values of $C$, according to the following search space:

\[C = 10^{p} \quad\text{with}~p = -3 + \frac{2}{3}(n-1) \quad\text{and}~n=1, ..., 10 \]

The value that provided the best average performance across the folds was selected.

In order to reliably assess the model performance while taking the most out of the data set, we embedded the cross-validated model into a \textit{leave-2-participants-out} cross-evaluation. That is, we constructed 25 different folds of data, each with the trials of 23 participants for training and of 2 participants for evaluation. We report here the average performance across the 25 test and train partitions together with the standard deviation (within brackets). In particular, in Table \ref{tab:performance} we include the area under the curve (AUC), the Tjur\footnote{While there is no consensus about the best metric for the evaluation of logistic regression, the coefficient of discrimination $R^{2}$ proposed by \citet{tjur2009r2} has been widely adopted recently, as it is more intuitive than other definitions of coefficients of determination and still asymptotically related to them.} coefficient of discrimination $R^{2}$ and the accuracy. For the sake of an easier interpretation, we include the theoretical baseline values of the AUC and $R^{2}$, and the empirical baseline accuracy on our test partitions.

\begin{table}[ht]
\centering
  \begin{tabular}{rccc}
    \multicolumn{1}{l}{} & AUC             & Tjur $R^{2}$         & Accuracy        \\
    \hline \\
    Test                 & 0.8884 (0.0180) & 0.4287 (0.0460) & 81.36 \% (0.32) \\
    Train                & 0.8865 (0.0040) & 0.4240 (0.0214) & 81.99 \% (1.52) \\
    Random baseline      & 0.5             & 0.0             & 60.70 \% (2.32)
  \end{tabular}
\caption{Test, train and baseline performance of the logistic regression model. Values within brackets indicate the standard deviation across the folds.}
\label{tab:performance}
\end{table}

The results in Table~\ref{tab:performance} show that the logistic regression model successfully learned the behavioural patterns from the experimental data and hence accurately predicted the direction of the first saccade, with very low overfitting, since train and test performance were very similar and have low variance. As a conclusion, this implies that the learned coefficients can be meaningfully used for further analysis, as will be presented in Section~\ref{sec:results}.

\section{Methods: salience maps of competing stimuli}
\label{sec:locsal}

In order to test whether the global salience is independent from the lower-level, salience properties of the stimuli (H2), we also computed salience maps both of each individual image and of each full stimulus shown at each trial, that is the pair of images with grey background, as shown in Figure~\ref{fig:globsal-stimuli_pair}.  For the computation of the salience maps we used the Graph-Based Visual Salience algorithm (GBVS) \citep{harel2007gbvs}, which is a computational salience model that makes use of well-defined low-level features.

Moreover, we also analysed the connection between global salience and a less restricted salience model, Deep Gaze II \citep{kuemmerer2016deepgaze}, whose features include higher level cues, since it is a deep neural network model pre-trained for large scale, image object recognition tasks, with additional features optimised for salience prediction.

In order to compare the salience maps with the behavioural data from the observation of competing stimuli, as well as with our derived global salience, we performed the following tests:

\subsection{Predictivity of salience maps for the first fixation}
\label{sec:locsal-firstfix}

In this case, our aim was to evaluate the performance of salience maps in predicting the landing location of the first fixation when two competing stimuli are presented. To do so, we computed the Kullback-Leibler Divergence between the first fixation distribution $F_{j}(b)$ and the salience distribution $S_{j}(b)$ for every image $j$ in the set of 200 images:

\begin{equation}
 D_{KL}(F_{j}||S_{j}) = \sum_{b=1}^{B} F_{j}(b) \log(\frac{F_{j}(b)}{S_{j}(b)+\epsilon}+\epsilon)
 \label{eq:kld}
\end{equation}
where is $\epsilon$ is a small constant to ensure numerical stability and $b$ refers to $B$ bins of one $1 \times 1$ degrees of visual field angle.

The first fixation distribution, $F_{j}(b)$, is the probability density distribution of all the first fixations made by all observers on each image $j$. To compute $F_{j}(b)$, we divided every image into sections of one squared degree of visual field angle and counted the number of first fixations made by all participants on each bin to obtain a histogram. Then, the histogram was smoothed using a Gaussian kernel with a size of one degree of visual field angle and normalised such that it became a probability distribution. The salience distribution, $S_{j}(b)$, is the smoothed and normalised (likewise) salience map---computed with GBVS or Deep Gaze II---of each individual image $j$.

Hence, according to the definition in Equation~\ref{eq:kld}, a low $D_{KL}(F_{j}||S_{j})$ would imply a good match between the location of the first fixations and the salience map of image $j$.

\subsection{Comparison between global and local salience}
\label{sec:locsal-glob_vs_loc}

In order to compare the local salience maps and the global salience scores learned by the computational model presented in Section~\ref{sec:globsal}, we analysed the GBVS and Deep Gaze salience maps of both the individual images and the whole stimuli, in relation to the global salience scores.

\subsubsection{Individual images}

First, we compared the Kullback-Leibler Divergence between the first fixations distribution and the salience maps of the individual images, as computed in Equation~\ref{eq:kld}, and the global salience scores, that is the coefficients learned by the optimisation defined in Equation~\ref{eq:objective}. This aimed at analysing whether, for instance, images whose local salience properties indeed drove the location of the first fixation have a higher global salience score, and vice versa.

\subsubsection{Trials}

Second, we looked at the properties of the salience map of the final stimulus seen by the participants at each trial, that is the paired competing images with a grey background (see Figure~\ref{fig:globsal-stimuli_pair}). As a metric of the contribution of each image to the salience map, for each trial $i$ we computed the relative salience mass $M$ of each image, left and right:

\begin{table}[ht]
\centering
\label{tab:salience_mass}
  \begin{tabular}{cc}
    $M_{i}^{L} = \int_{x \in X_{L}} S_{i}(x)$ & $M_{i}^{R} = \int_{x \in X_{R}} S_{i}(x)$
  \end{tabular}
\end{table}
where $S_{i}(x)$ is the normalised salience map of the whole stimulus presented at trial $i$ and $X_{L}$ and $X_{R}$ are the stimulus locations corresponding to the left and right images, respectively. A significant positive correlation between $\Delta_{M}^{(i)} = M_{i}^{L} - M_{i}^{R}$ and the difference between the global salience scores of the images on the left and right, $\Delta_{GS}^{(i)}= w_{L}^{(i)} - w_{R}^{(i)}$, would indicate that the local salience properties can partly explain the direction of the first fixation.

\section{Results}
\label{sec:results}

In this section, we present the main results of our analyses and discuss the validity of the hypotheses presented in Section~\ref{sec:globsal-contributions}. Each of the sub-sections focuses on one of the five hypotheses, in the natural order. All the scatter plots that show the relationship between two variables include the value of the Pearson correlation, as well as the line fit by a linear regression model, with 95 \% confidence intervals estimated using bootstrap with 1,000 resamples.

\subsection{Global visual salience}
\label{sec:results-global_salience}

In our first hypothesis (H1), we stated that images can be ranked according to a specific global salience that leads to the attraction of initial eye fixations. In order to quantify the global salience of individual images, we have presented in Section~\ref{sec:globsal} a computational model that successfully predicts the direction of the first fixation from the behavioural data, as validated by the results in Section~\ref{sec:globsal-model_eval}, and thus we can analyse the coefficients of the model as indicators of the global salience of each image in the data set.

Importantly, the fact that the first fixation direction of the participants when exploring such competitive stimuli can be predicted by a computational model means that their behaviour was not random, but followed certain patterns. In order to shed some light on the nature of these patterns, in Figure~\ref{fig:globsal-stimuli_sorted} we show the complete set of stimuli ranked according the global salience score learned by our model and in Figure~\ref{fig:globsal-global_salience_categories} the value of the global salience scores of each image, highlighting the differences between the image categories.

\begin{figure}[ht]
  \centering
  \begin{subfigure}{\linewidth}
      \includegraphics[width = \linewidth]{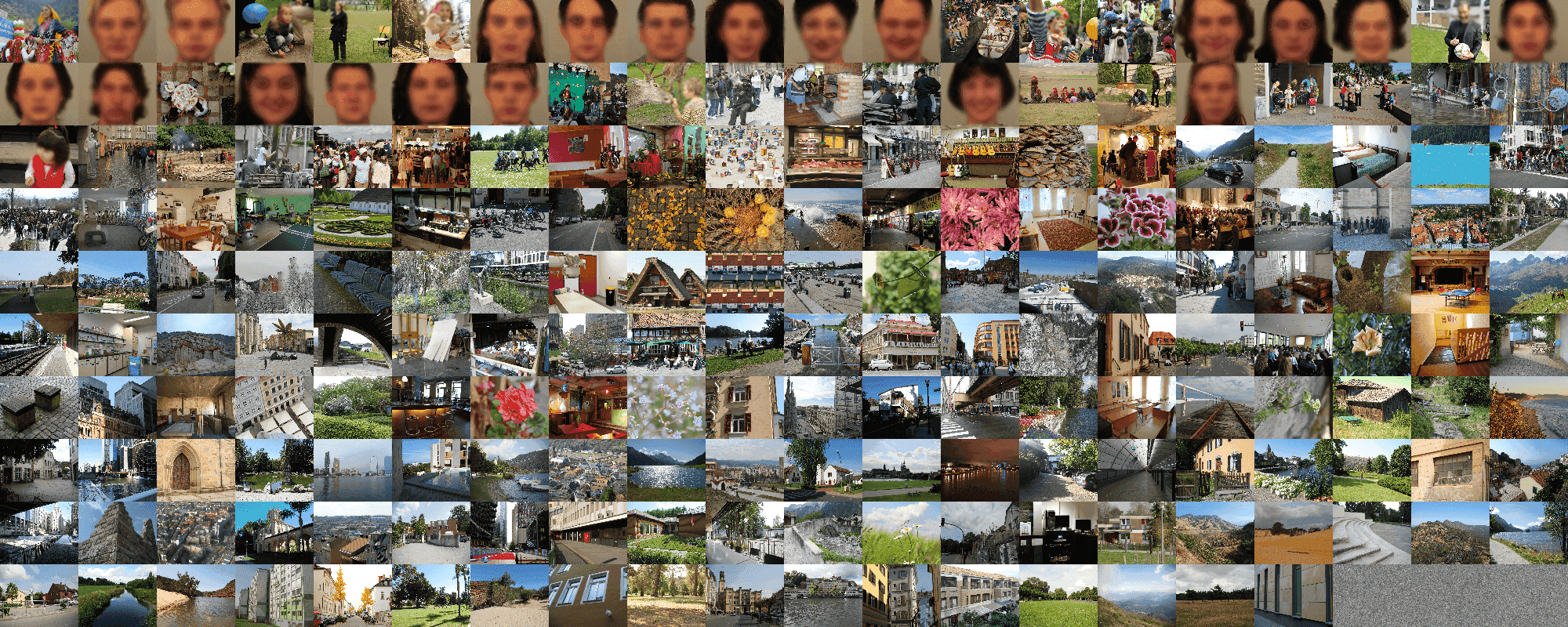}
      \caption{Experimental stimuli, ranked according to the learned global salience. The stimulus with the highest global salience score is on the top-left corner and the rest are sorted with the x-axis changing fastest (row-major order). Faces have been blurred to preserve the identity.}
    \label{fig:globsal-stimuli_sorted}
  \end{subfigure}
  \\
  \begin{subfigure}{\linewidth}
      \centering
      \includegraphics[width = \linewidth]{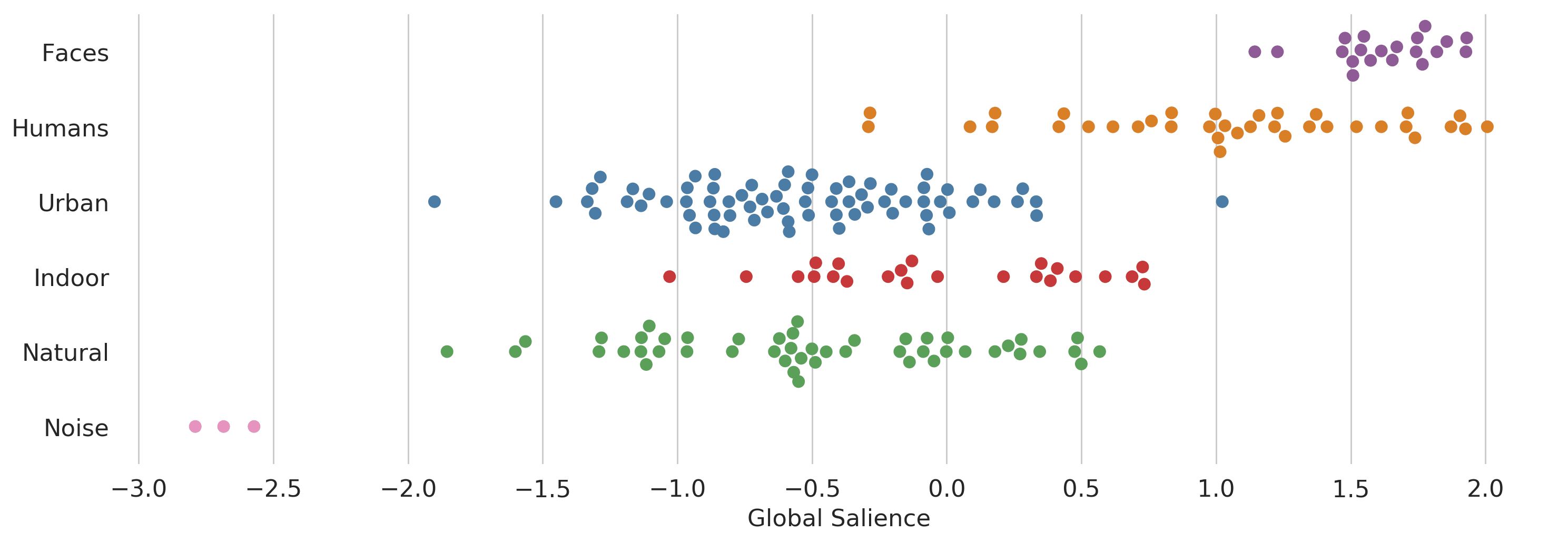}
      \caption{Global salience score of each stimulus and image categories.}
    \label{fig:globsal-global_salience_categories}
  \end{subfigure}
  \caption{Global salience scores of the experimental stimuli}
\label{fig:globsal-global_salience}
\end{figure}

Figure~\ref{fig:globsal-global_salience} shows that there exists a clear, general tendency to first fixate on the images that contain either close-up faces or scenes with humans, even though the first fixations may occur, on average, as early as after 242 ms ($\sigma_{SD} = 66$ ms) from the stimulus onset. These two categories, faces and humans, were assigned the highest global salience scores. Then, urban, indoor and natural landscapes obtained significantly lower scores, with no big differences among the three categories. Finally, the three pink-noise images were assigned very low scores, which serves as sanity-check of our proposed method.

\subsection{Global vs. local salience}
\label{sec:results-local_salience}

A reasonable question in view of the results presented in Figure~\ref{fig:globsal-global_salience} is whether the global salience scores---and the ranking of the stimuli that arises from the scores---is a unique measure that assesses the initial visual behaviour when facing competing stimuli, or whether this behaviour and thus our proposed global salience can be explained by the low-level properties of the respective stimuli.

In our second hypothesis (H2) we stated, instead, that the global salience is independent from the low-level local salience properties. So as to test this, we performed several tests, described in Section~\ref{sec:locsal}.

In Figure~\ref{fig:globsal-kld_distr_gbvs} we plot the distribution of the Kullback-Leibler Divergence between the first fixations maps and the GBVS local salience maps of the individual images (see Equation~\ref{eq:kld}). The mean of the distribution is significantly non-zero (two-tail t-test p-value $< .001$, $\mu_{KLD} = 1.44$, $\sigma_{KLD} = 0.33$), which means that there is a significant loss of information when using a local salience map to predict the landing locations of the first fixations on a given image \citep{riche2013saliency}. In order to illustrate the mismatch, in Figure~\ref{fig:globsal-kld_distr_gbvs} we display three example images with the overlaid salience maps and the location of all the first fixations that landed on them. When the KLD value is minimum (a), the salience maps can approximate the fixations, although this happened rarely. Already with KLD values around the mean, the performance of a salience map in predicting the landing location of fixations is rather mediocre (b) and deteriorates further as the KLD increases (c).

\begin{figure}[ht]
  \centering
  \begin{subfigure}{0.75 \linewidth}
      \includegraphics[width = \linewidth]{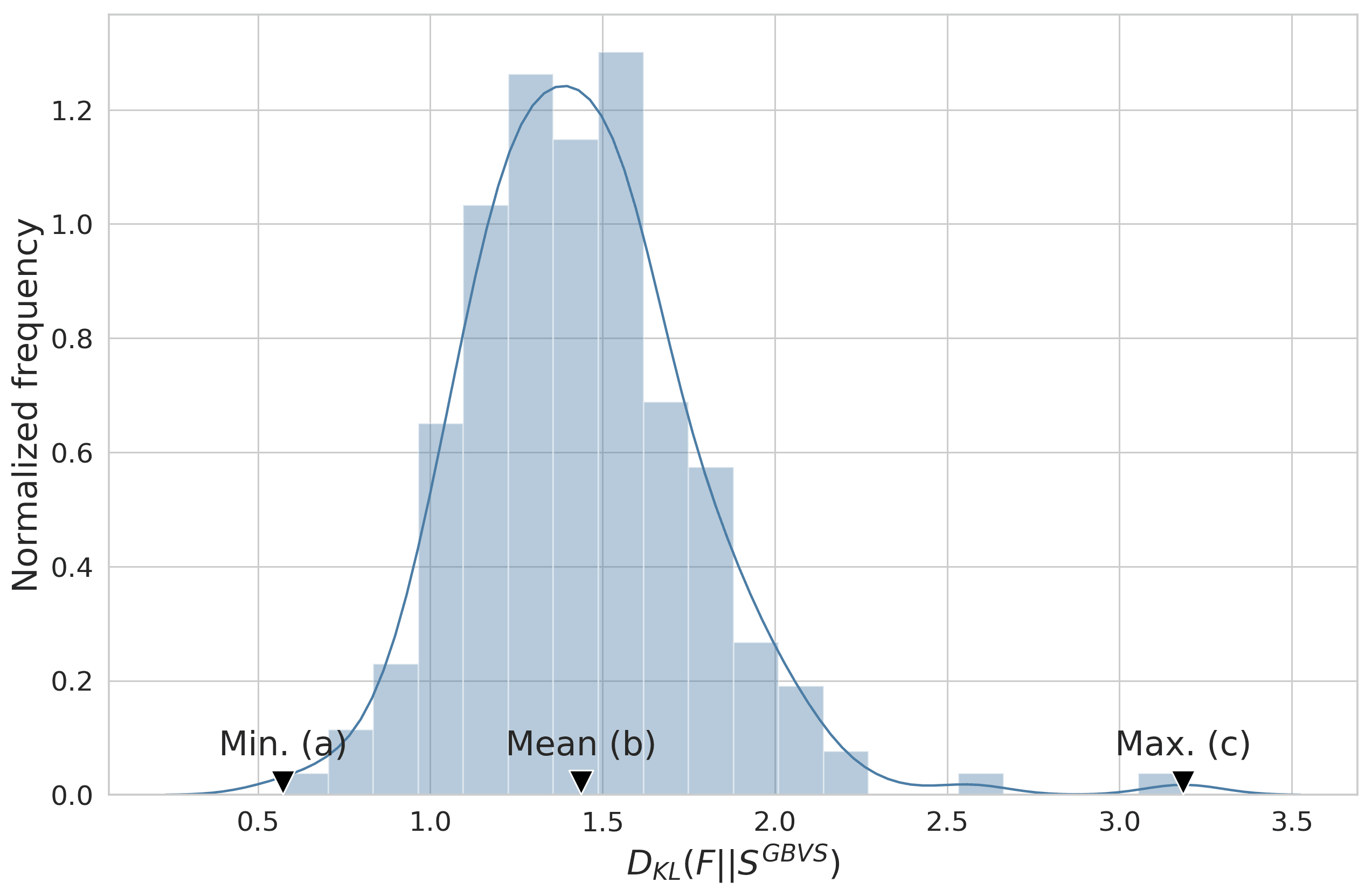}
      \label{fig:globsal-kld_distr_gbvs_sub}
  \end{subfigure}
  \\
  \begin{subfigure}{0.3 \linewidth}
      \centering
      \includegraphics[width = \linewidth]{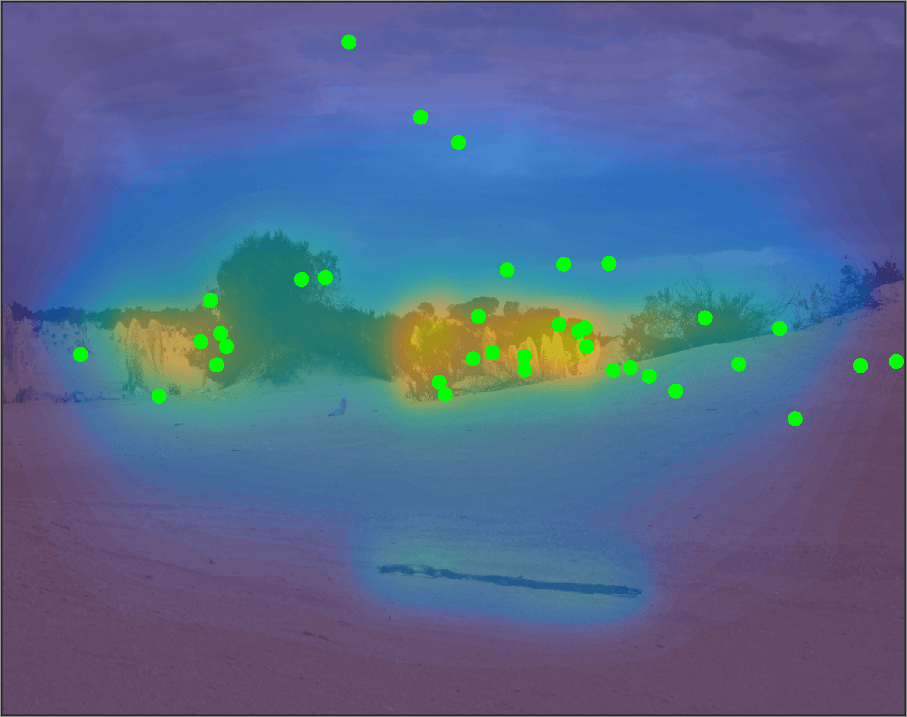}
      \caption{$D_{KL}(F||S) = 0.57$}
    \label{fig:globsal-kld_gbvs_min_image}
  \end{subfigure}
  \hspace{0.02 \linewidth}
  \begin{subfigure}{0.3 \linewidth}
      \centering
      \includegraphics[width = \linewidth]{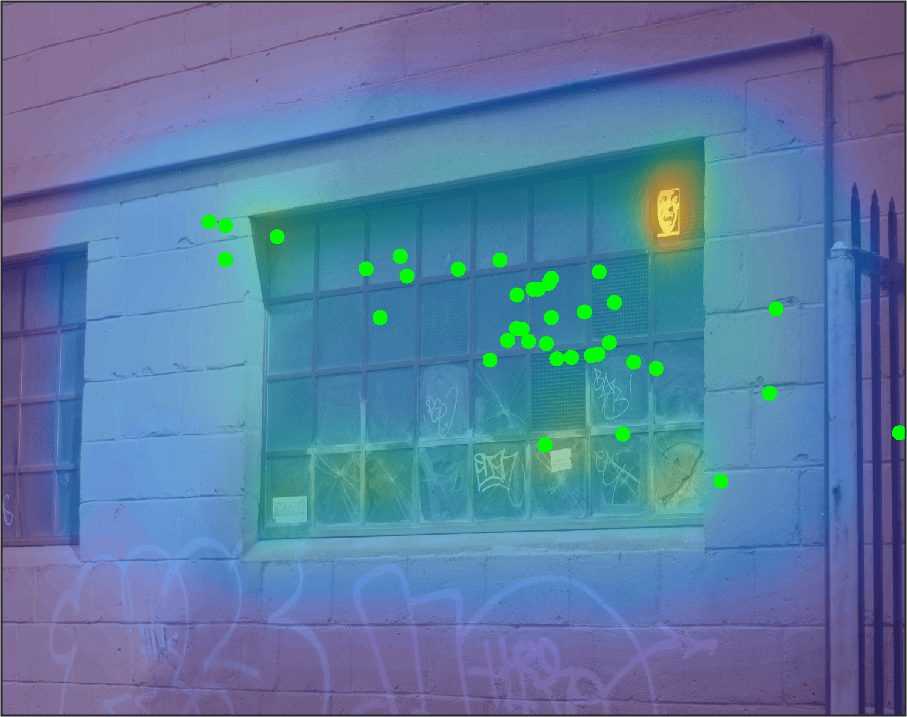}
      \caption{$D_{KL}(F||S) = 1.44$}
    \label{fig:globsal-kld_gbvs_mean_image}
  \end{subfigure}
  \hspace{0.02 \linewidth}
  \begin{subfigure}{0.3 \linewidth}
      \centering
      \includegraphics[width = \linewidth]{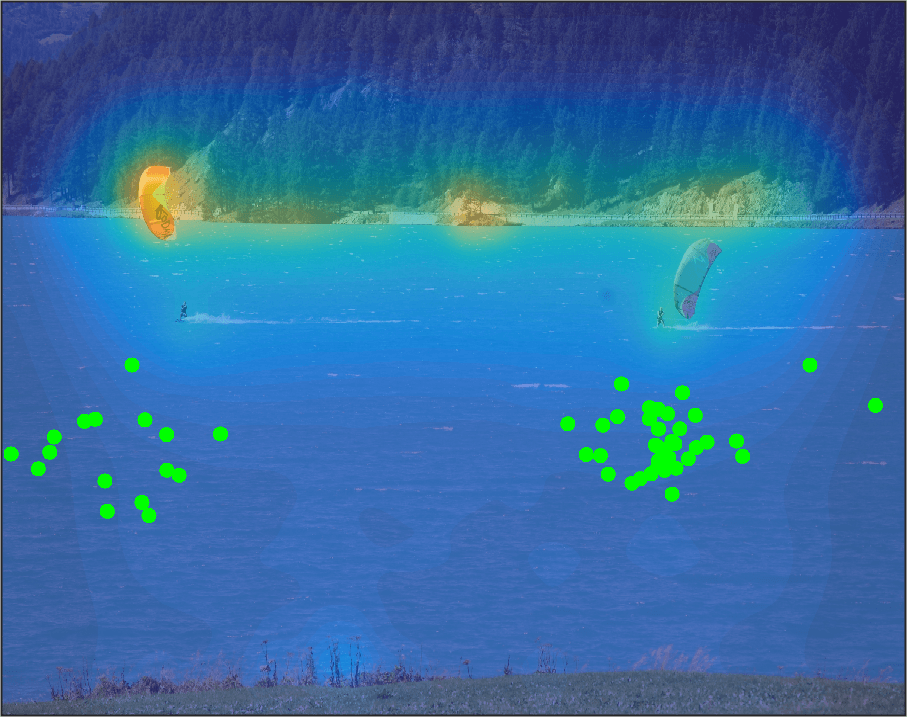}
      \caption{$D_{KL}(F||S) = 3.19$}
    \label{fig:globsal-kld_gbvs_max_image}
  \end{subfigure}
  \caption{Top row: distribution of the Kullback-Leibler Divergence between the first fixations map and the GBVS local salience maps. Bottom row: images with the minimum, closest to the mean and maximum KLD, with their overlaid salience map and the location of the first fixations.}
\label{fig:globsal-kld_distr_gbvs}
\end{figure}

Perhaps not surprisingly, in view of the poor match between the salience maps and the first fixation maps, Figure~\ref{fig:globsal-kld_vs_globsal_gbvs} shows that the Kullback-Leibler Divergence between them does not correlate with the global salience scores. This means that the images which attract the first fixations towards salient regions (low KLD) do not tend to have high global salience scores neither vice versa.

\begin{figure}[ht]
  \centering
  \begin{subfigure}{0.6 \linewidth}
      \includegraphics[width = \linewidth]{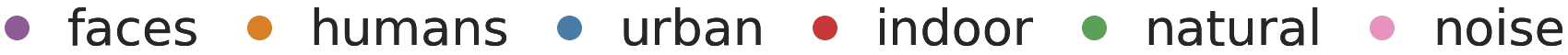}
  \end{subfigure}
  \\
  \begin{subfigure}{0.48 \linewidth}
      \includegraphics[width = \linewidth]{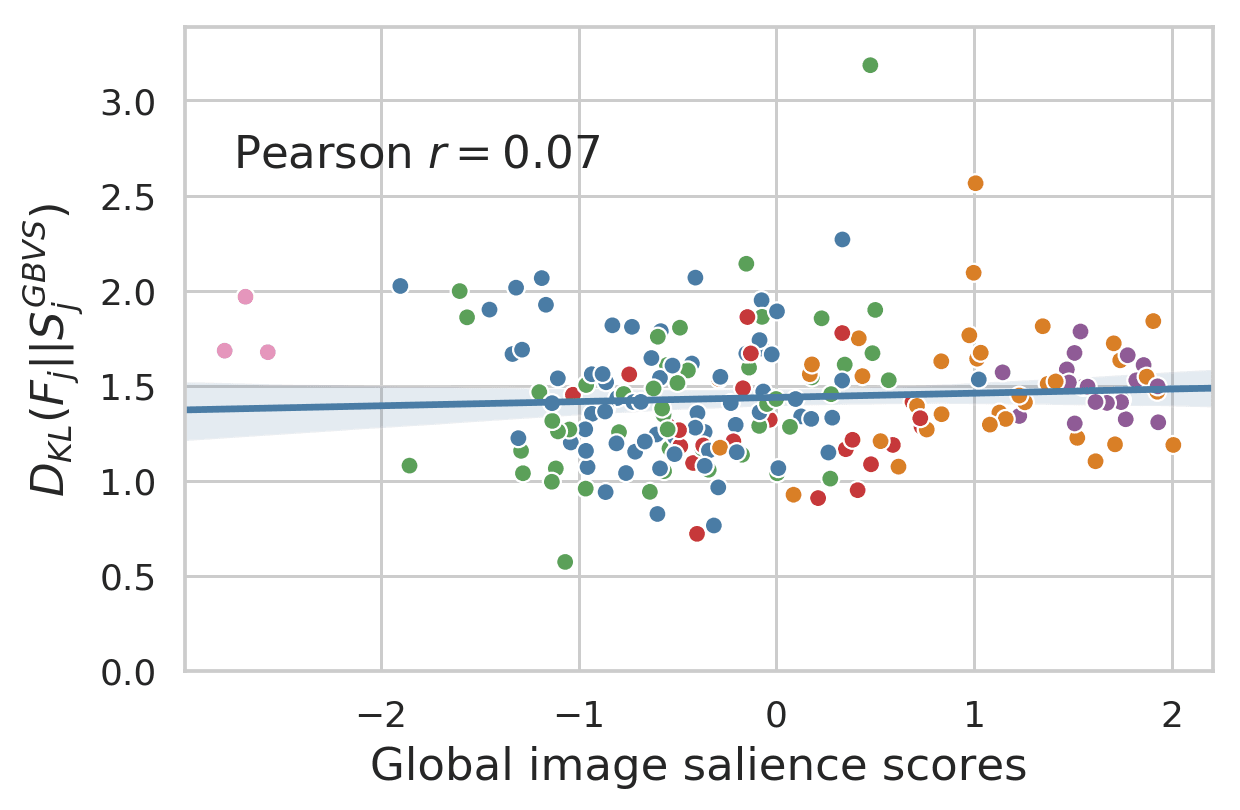}
      \caption{GBVS}
    \label{fig:globsal-kld_vs_globsal_gbvs}
  \end{subfigure}
  \hspace{0.01 \linewidth}
  \begin{subfigure}{0.48 \linewidth}
      \includegraphics[width = \linewidth]{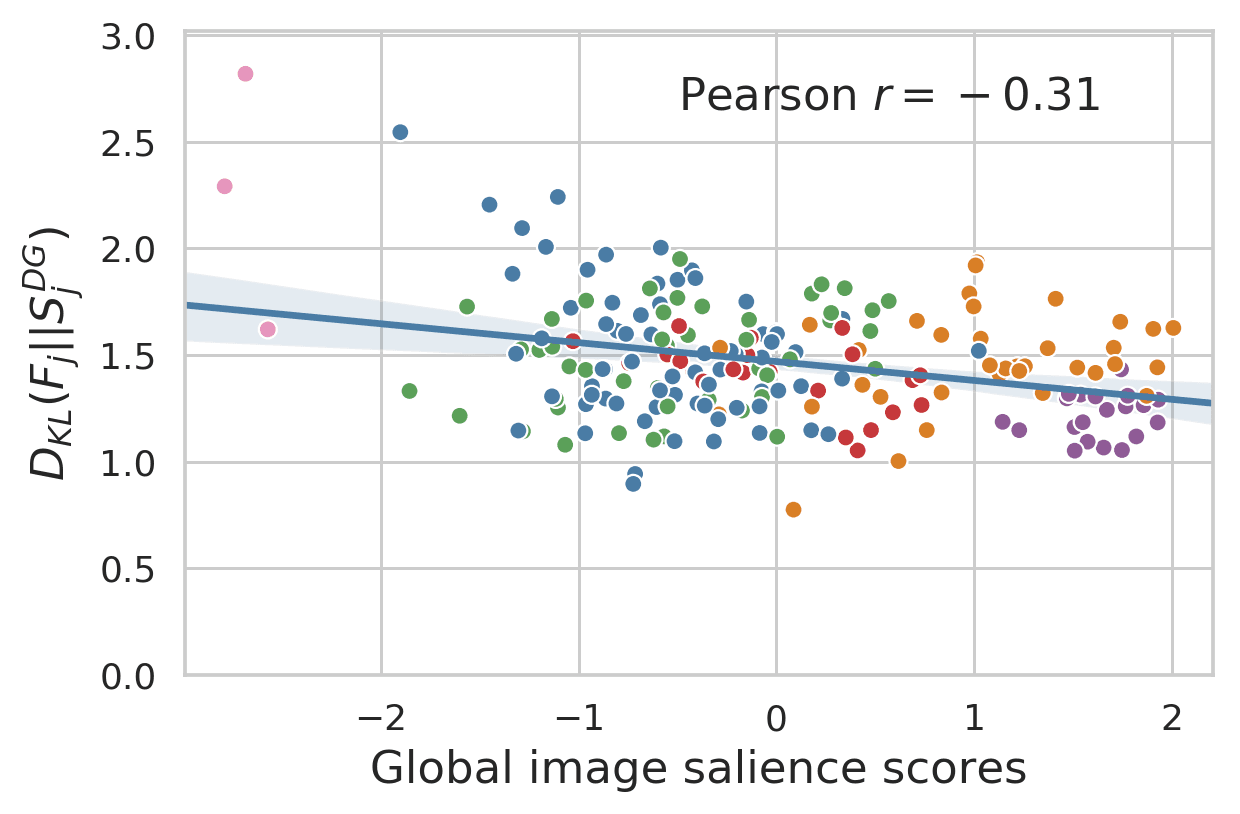}
      \caption{Deep Gaze}
    \label{fig:globsal-kld_vs_globsal_deepgaze}
  \end{subfigure}
  \caption{Comparison between the global salience scores the KLD between the first fixation distribution and the salience maps from the computational models}
  \label{fig:globsal-kld_vs_globsal}
\end{figure}

Finally, we analyse in Figure~\ref{fig:globsal-mass_gbvs_vs_globsal} whether the direction of the first fixation when looking at competing stimuli, as modelled by our proposed global salience scores, can be explained by the difference in the low-level salience properties of the competing stimuli, as measured by the GBVS salience mass of each image (see Section~\ref{sec:locsal-glob_vs_loc}). Also in this case we found no significant correlation.

The noisy images included in the stimulus set serve once more as a validation of the expected results. When one of the images (left or right) was pink noise the difference in GBVS salience mass was either very high or very low, as is the difference in global salience scores. In this case, both metrics do correlate, but as shown by the central scatter plot of Figure~\ref{fig:globsal-mass_gbvs_vs_globsal}, the feature-driven (GBVS) salience mass cannot explain the global salience scores learned by the model.

\begin{figure}[ht]
  \centering
  \includegraphics[width = \linewidth]{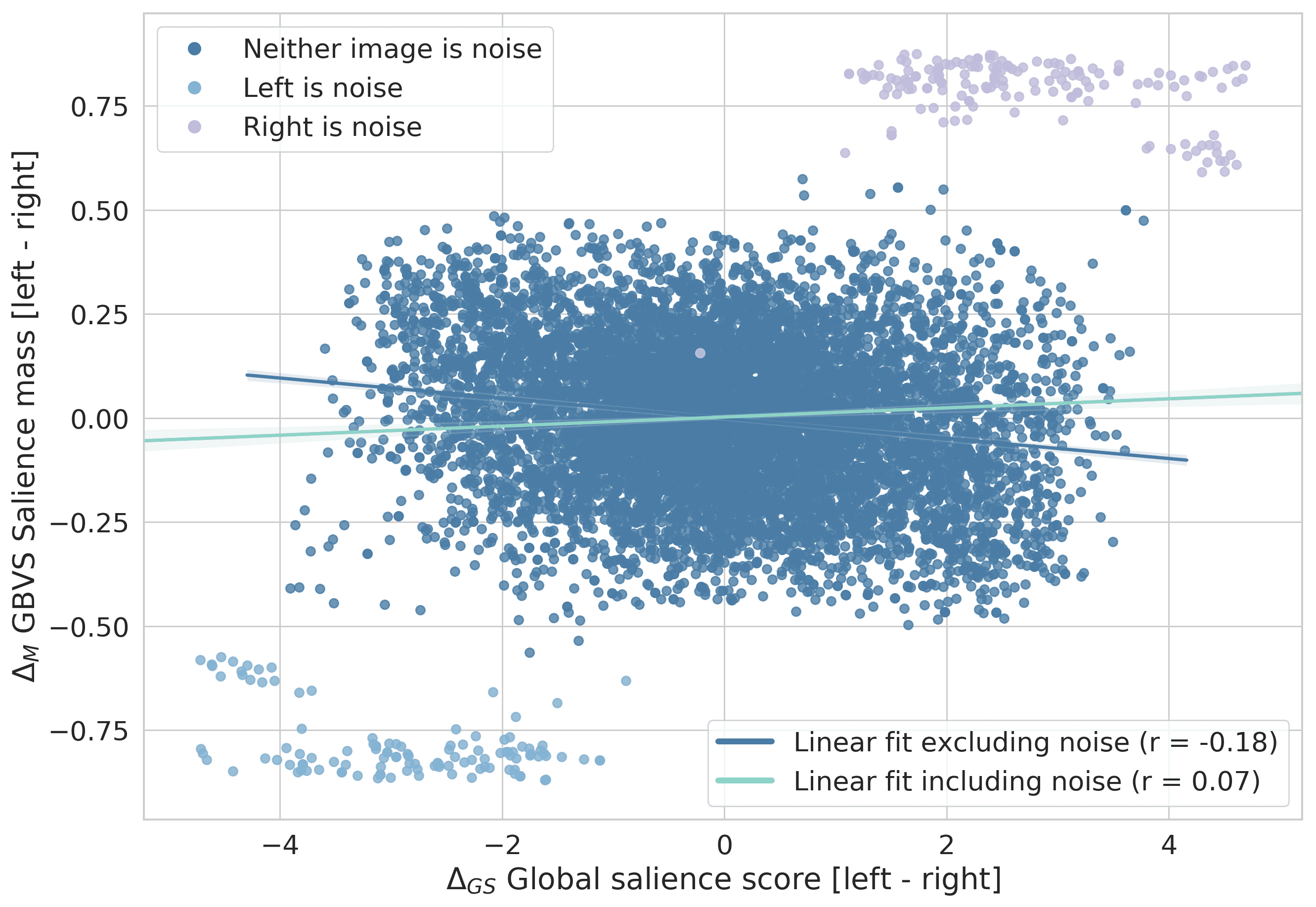}
  \caption{Correlation between the GBVS image salience mass and the global salience scores.}
\label{fig:globsal-mass_gbvs_vs_globsal}
\end{figure}

In order to better understand what drives the direction of the first fixation when faced with competing stimuli, we also compared our proposed global salience with properties of Deep Gaze II salience maps. As presented in Section~\ref{sec:locsal}, unlike GBVS, Deep Gaze does make use of higher-level information of the images to predict the salience maps, since it is a neural network pre-trained on image object recognition tasks. This allows it to model salience driven by faces or objects \citep{kuemmerer2017icfdeepgaze} and it becomes an interesting model to which compare our global salience model, since we have seen in Section~\ref{sec:results-global_salience} that images containing faces and humans tend to get a higher global salience score.

In general, we observe that unlike GBVS, measures derived from Deep Gaze salience maps exhibit a non-zero, yet moderate correlation with our proposed global salience. For instance, Figure~\ref{fig:globsal-kld_vs_globsal_deepgaze} shows a slight negative correlation between global salience scores and the KLD between first fixation distributions and Deep Gaze salience maps. However, looking at the distribution of the Kullback-Leibler divergence in Figure~\ref{fig:globsal-kld_distr_deepgaze}, we see that the salience maps are also far from matching the location of the first fixations on the images. Finally, we also observed (see Figure~\ref{fig:globsal-mass_deepgaze_vs_globsal} a non-zero correlation between the difference of global salience scores between the left and the right image, and the difference in salience mass computed with Deep Gaze.

\begin{figure}[ht]
  \centering
  \begin{subfigure}{0.75 \linewidth}
      \includegraphics[width = \linewidth]{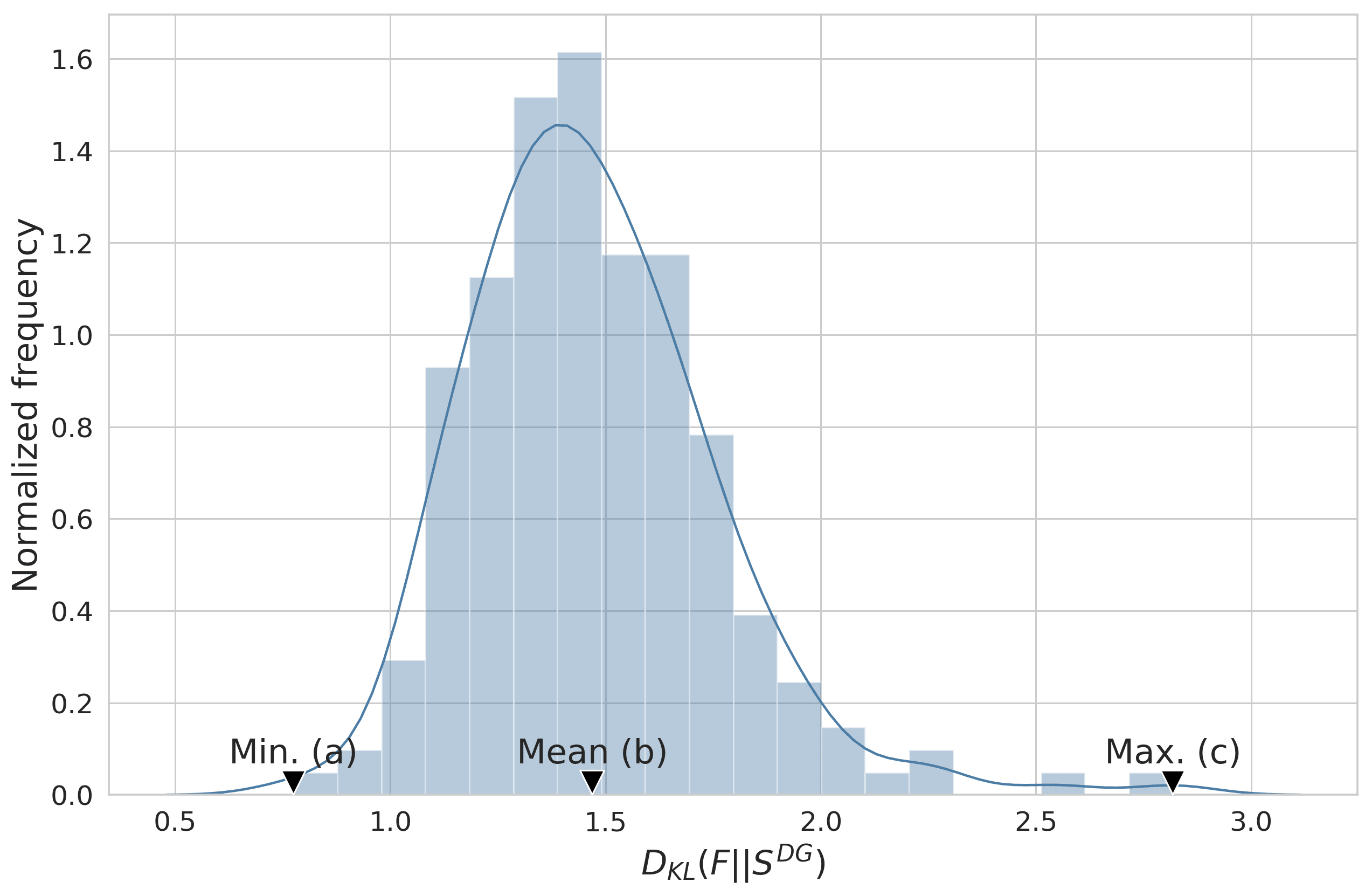}
      \label{fig:globsal-kld_distr_deepgaze_sub}
  \end{subfigure}
  \\
  \begin{subfigure}{0.3 \linewidth}
      \centering
      \includegraphics[width = \linewidth]{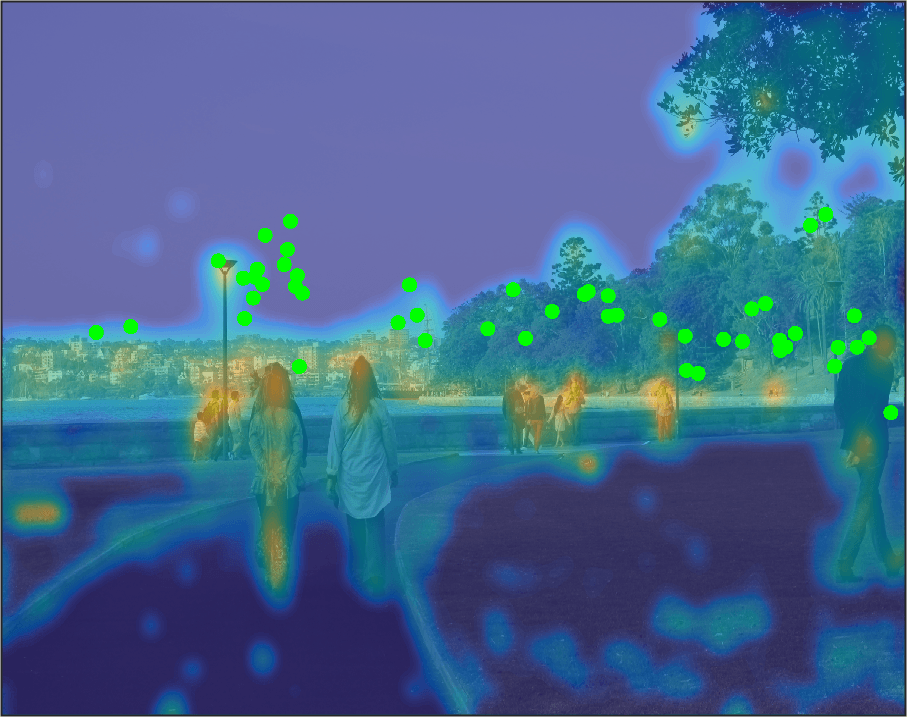}
      \caption{$D_{KL}(F||S) = 0.57$}
    \label{fig:globsal-kld_deepgaze_min_image}
  \end{subfigure}
  \hspace{0.02 \linewidth}
  \begin{subfigure}{0.3 \linewidth}
      \centering
      \includegraphics[width = \linewidth]{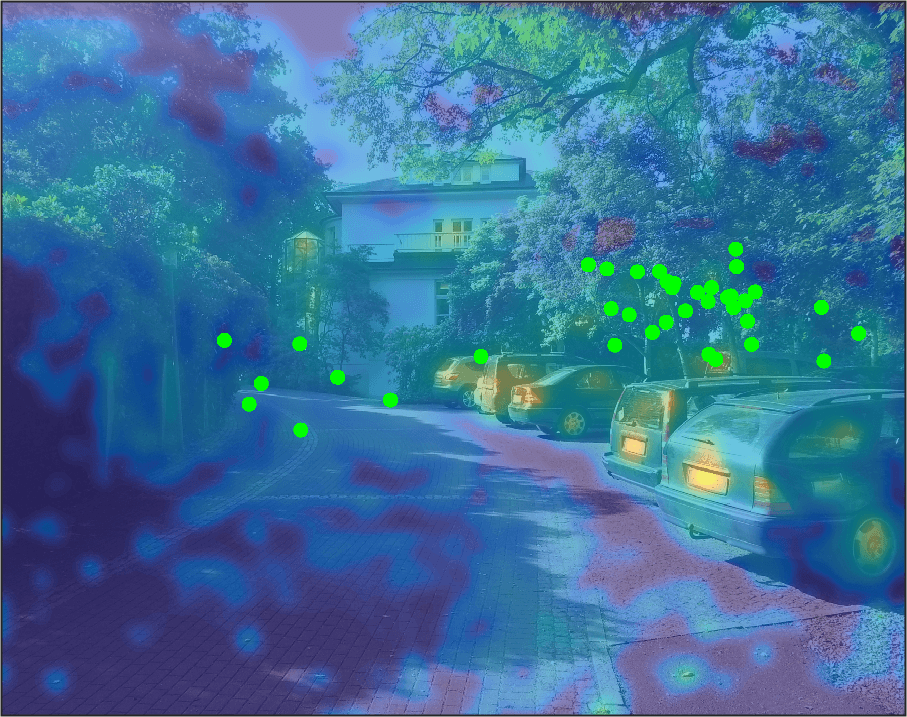}
      \caption{$D_{KL}(F||S) = 1.44$}
    \label{fig:globsal-kld_deepgaze_mean_image}
  \end{subfigure}
  \hspace{0.02 \linewidth}
  \begin{subfigure}{0.3 \linewidth}
      \centering
      \includegraphics[width = \linewidth]{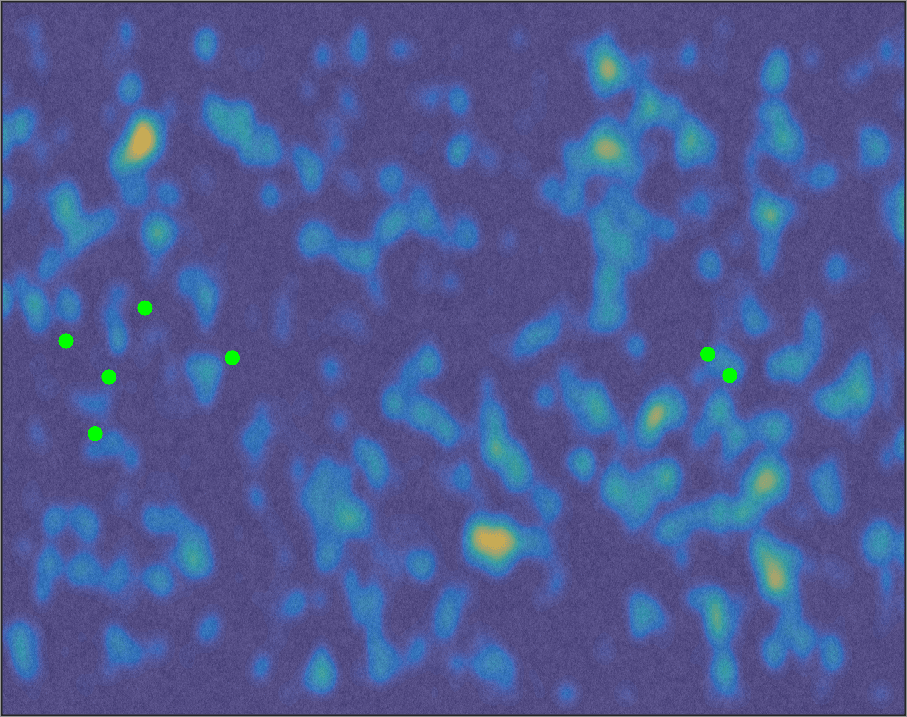}
      \caption{$D_{KL}(F||S) = 3.19$}
    \label{fig:globsal-kld_deepgaze_max_image}
  \end{subfigure}
  \caption{Top row: distribution of the Kullback-Leibler Divergence between the first fixations map and the Deep Gaze local salience maps. Bottom row: images with the minimum, closest to the mean and maximum KLD, with their overlaid salience map and the location of the first fixations.}
  \label{fig:globsal-kld_distr_deepgaze}
\end{figure}

\begin{figure}[ht]
  \centering
  \includegraphics[width = \linewidth]{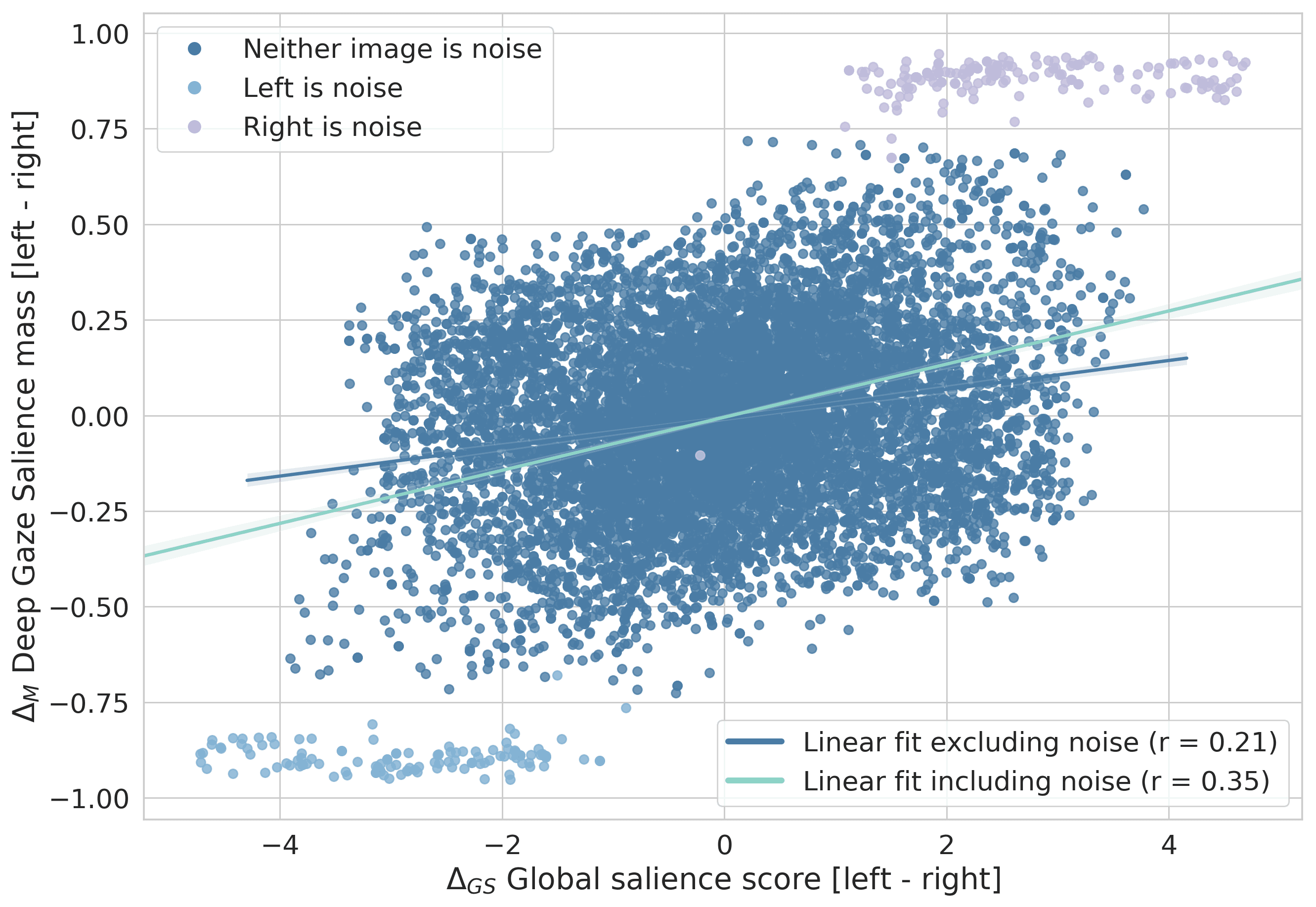}
  \caption{Correlation between the Deep Gaze image salience mass and the global salience scores.}
\label{fig:globsal-mass_deepgaze_vs_globsal}
\end{figure}

Taken together, we can conclude that our proposed computational model provided a robust method to rank images according to a unique global image salience that is independent of the low-level local salience properties of the stimuli, and we observed a non-zero, yet moderate correlation with a computational salience model that incorporates higher-level cues.

\subsection{Lateral bias}
\label{sec:results-spatialbias}
Our third hypothesis (H3) stated that a general spatial bias leads to a higher likelihood to first fixate on the left rather than the right image. We thus calculated the number of first saccades that landed onto the left and the right image for each block separately (Figure~\ref{fig:globsal-first_saccade_blocks}). A 4 $\times$ 2 (block: 1, 2, 3, 4 $\times$ image side: left, right) repeated measures ANOVA (Greenhouse-Geisser corrected) revealed a general spatial bias of the initial saccade towards the left image as indicated by a significant main effect according to the image side [\textit{F}(1, 48) = 30.833; \textit{p} \textless\ .001; \textit{$\eta_p^2$} = .391]. No further effects were found [all \textit{F} $\leq$ 2.594; all \textit{p} $\geq$.074, all \textit{$\eta_p^2$} $\leq$ .051], showing that the left bias was present in all blocks with similar extent. Thus, we can conclude that the participants generally targeted their initial saccades more on left than right sided images.

\begin{figure}[ht]
  \centering
  \includegraphics[width = \linewidth]{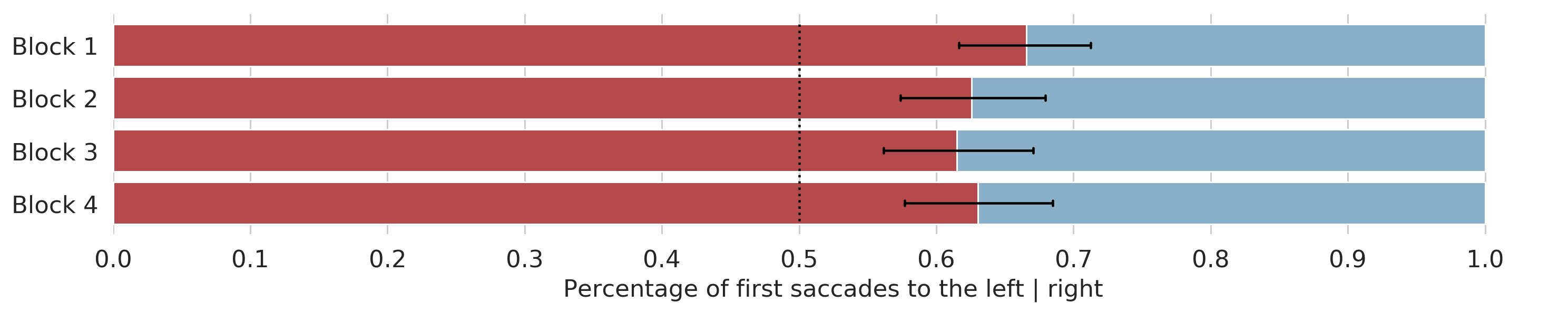}
  \caption{Percentage of first saccades that targeted on the left (red) and right (blue) images, at each block of the experimental session. Error bars depict the standard deviation of the mean. Note that considerably more first fixations landed on the left image, highlighting the lateral bias.}
\label{fig:globsal-first_saccade_blocks}
\end{figure}

Nonetheless, the error bars in Figure~\ref{fig:globsal-first_saccade_blocks} suggest a high variability of the lateral bias across subjects. In order to investigate this, we calculated the number of first saccades on the right image for each participant separately. Moreover, since our model included an individual bias term for each participant, as described in Section~\ref{sec:globsal}, we can also look at the magnitude of the coefficients learned by the model. In Figure~\ref{fig:globsal-subject_bias} we plot, for each participant, the percentage of first saccades towards the right image and their corresponding lateral bias term learned by the computational model. Both metrics are highly correlated---further highlighting the validity of the model---and reveal a high variability in the lateral bias across participants. Overall, 63 \% of all first fixations landed on the left image.

\begin{figure}[ht]
  \centering
  \includegraphics[width = 0.6 \linewidth]{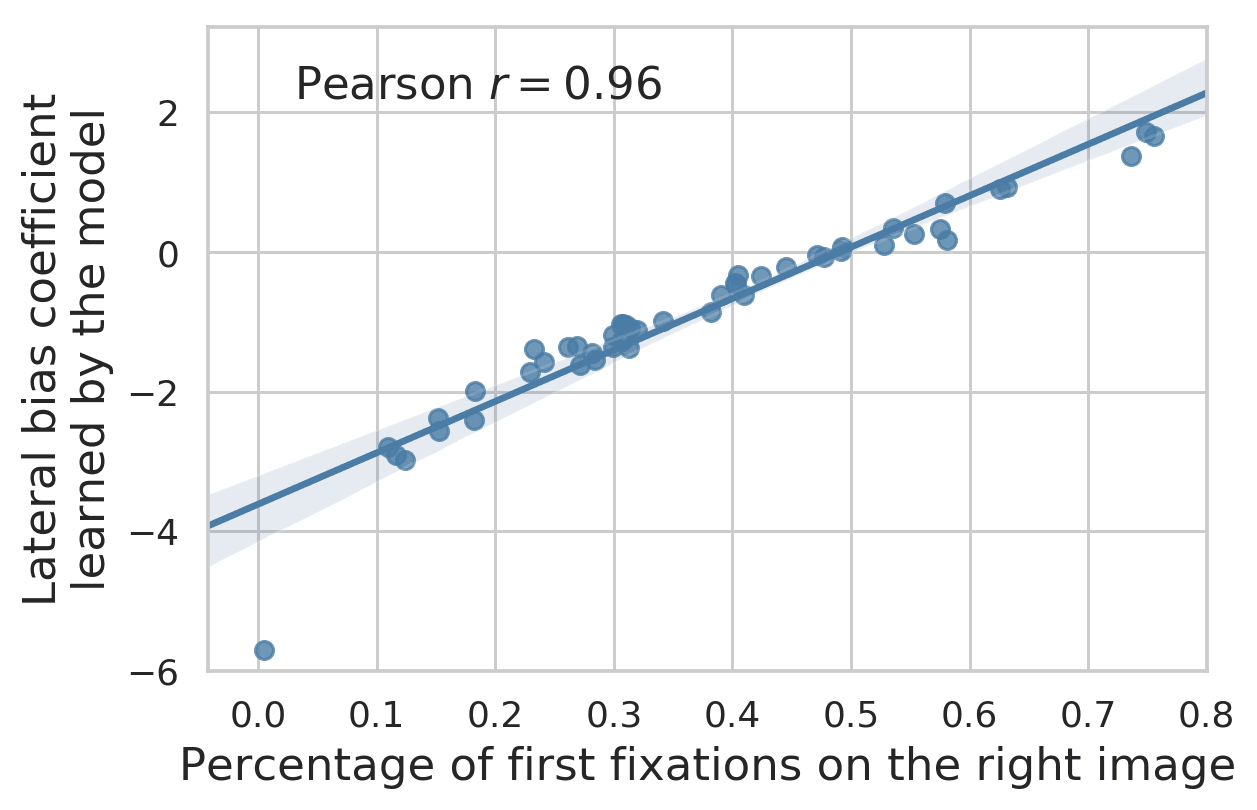}
  \caption{Lateral bias of each participant, as measured by the percentage of first fixations onto the right image and the lateral bias terms learned by our computational model. Both metrics are highly correlated and reveal the average left bias, but with high variability across participants.}
\label{fig:globsal-subject_bias}
\end{figure}

\subsection{Task and familiarity}
\label{sec:results-taskandfam}
Next, we investigated the effect of the familiarity with one of the images and of the task of selecting the already seen or unseen image, which the participants had to perform in blocks 2 and 3 of the experiment, respectively. In particular, we were interested in finding out whether there is a tendency to direct the initial saccade towards the task-relevant images or towards the new images, for instance. In our fourth hypothesis (H4) we stated that our task and familiarity should have little or no influence in the initial saccade. For that purpose, we first performed a 2 $\times$ 2 (task: select new, select old $\times$ fixated image: new image, old image) repeated measures ANOVA analysis (Greenhouse-Geisser corrected). The results revealed no significant effects [all \textit{F} $\leq$ 1.936; all \textit{p} $\geq$ .170, all \textit{$\eta_p^2$} $\leq$ .039] (Figure~\ref{fig:globsal-first_saccade_tasks}). Thus, the provided tasks did not bias the initial saccade decision to target one of the two presented images. Nevertheless, we found that participants correctly identified 91.43\% of the new images in block 2 and 91.16\% of the old images in block 3. Hence, the task performance was highly above chance (50\%) and the participants were accurate in identifying the new and old images respectively.

Also in this case, the same conclusion can be extracted from the coefficients learned by the model to capture the task and familiarity effects, which are -0.04 and -0.10, respectively, that is, very small and only slightly higher for the familiarity.

\begin{figure}[ht]
  \centering
  \includegraphics[width = \linewidth]{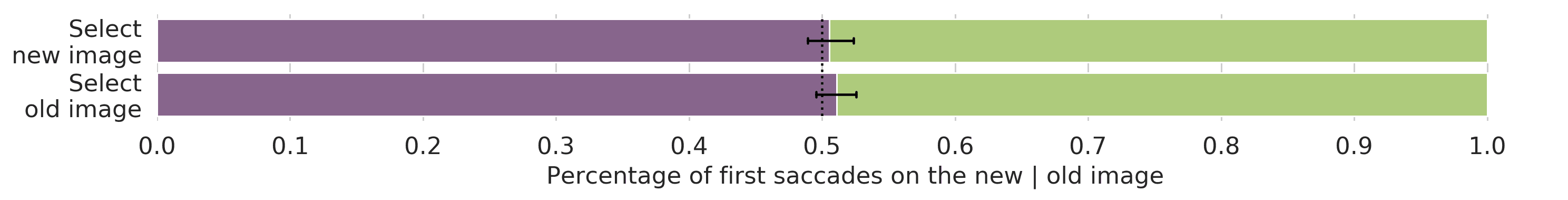}
  \caption{Percentage of first saccades that targeted on the new (purple) and old (green) images, at blocks 2 and 3, where participants had the task of indicating the new and old image, respectively. Error bars depict the standard deviation of the mean. No significant bias can be appreciated in this case.}
\label{fig:globsal-first_saccade_tasks}
\end{figure}

Taken together, spatial properties influenced the initial saccade in favour to fixate left sided images first. Although task performance was very high, neither the task nor the familiarity with one of the images had an influence in the direction of the first fixation after stimulus onset. These results fully support our third and fourth hypotheses.

\subsection{Total exploration of images}
\label{sec:results-totalexplorationofimages}
In our fifth hypothesis (H5), we stated that images with higher global image salience lead to a longer exploration time than images with lower global salience. We thus calculated the relative dwell time on each image, left and right, for each trial. As an initial step, similarly to the analysis of the initial saccade, we analysed the potential effect of the spatial image location as well as the task and familiarity relevance on the exploration time. 

With respect to the spatial image location, a 4 $\times$ 2 (block: 1, 2, 3, 4 $\times$ image side: left, right) repeated measures ANOVA (Greenhouse-Geisser corrected) revealed a significant main effect according to the block [\textit{F}(2.368, 113.668) = 12.066; \textit{p} \textless\ .001; \textit{$\eta_p^2$} = .201] but no further effects [all \textit{F} $\leq$ 2.232; all \textit{p} $\geq$.109, all \textit{$\eta_p^2$} $\leq$ .044]. Thus, the total time of exploration did not depend on the spatial location of the images, as also shown in Figure \ref{fig:globsal-dwell_blocks}. 

With respect to the task relevance---recall: block 2 - select new image; block 3 - select old image---we calculated a 2 $\times$ 2---task: select new, select old $\times$ fixated image: new image, old image---repeated measures ANOVA (Greenhouse-Geisser corrected). The results revealed a significant main effect according to the task [\textit{F}(1, 48) = 4.298; \textit{p} \textless\ .050; \textit{$\eta_p^2$} = .082] and fixated image [\textit{F}(1, 48) = 64.524; \textit{p} \textless\ .001; \textit{$\eta_p^2$} = .573], as well as an interaction between task and fixated image [\textit{F}(1, 48) = 36.728; \textit{p} \textless\ .001; \textit{$\eta_p^2$} = .433]. As shown by Figure \ref{fig:globsal-dwell_fam_task}, our results showed that, in general, participants tended to spend more time exploring new instead of previously seen images. Furthermore, this effect was noticeably larger in block 2, where the task was to select the new images, than in block 3 (select old image). 

Consequently, we found that the spatial location of images did not affect the total time of exploration. Instead, the task and familiarity had a considerable impact on the exploration time, revealing that new images were explored during a longer time than the counterpart.

\begin{figure}[ht]
  \centering
  \includegraphics[width = \linewidth]{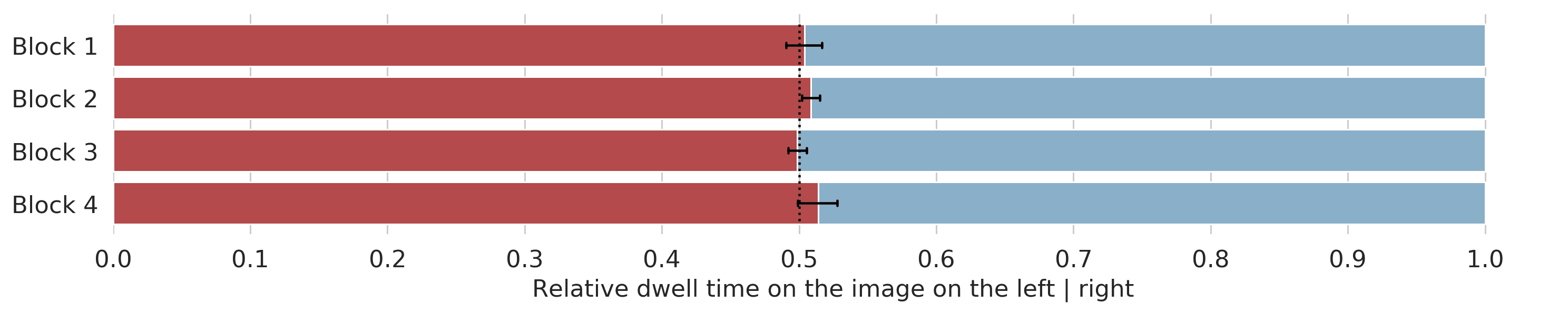}
  \caption{Exploration as measured by the relative dwell time on the left (red) and right (blue) images, at each block of the experimental session. Error bars depict the standard deviation of the mean.}
\label{fig:globsal-dwell_blocks}
\end{figure}

\begin{figure}[ht]
  \centering
  \includegraphics[width = \linewidth]{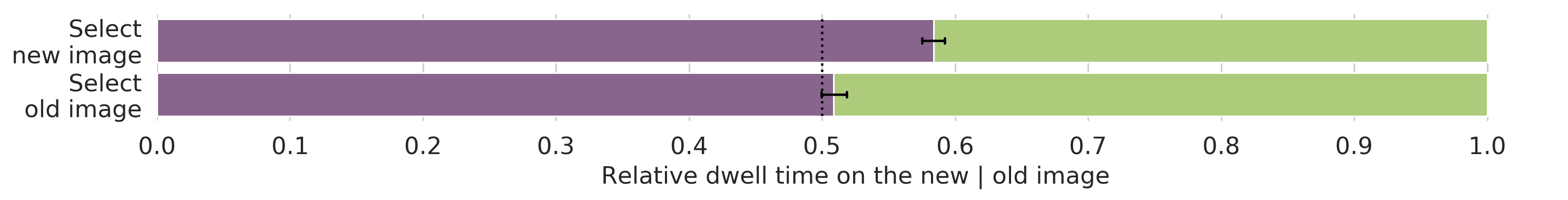}
  \caption{Exploration as measured by the relative dwell time on the new (purple) and old (green) images, at blocks 2 and 3, where participants had the task of indicating the new and old image, respectively. Error bars depict the standard deviation of the mean.}
\label{fig:globsal-dwell_fam_task}
\end{figure}

For our main analysis regarding the interaction between exploration time and global image salience, we then contrasted the global salience score learned for each image with its respective dwell time averaged over all trials and subjects. The results revealed a significant positive correlation, indicating that images with larger global image salience led to a more intense exploration (Figure \ref{fig:globsal-dwell_scores}). Thus, global image salience describes not only a measure of which image attracts initial eye movements, but is also connected to longer exploration time, suggesting that global salience may describe the relative engagement of images. 

\begin{figure}[ht]
  \centering
  \begin{subfigure}{0.6 \linewidth}
      \includegraphics[width = \linewidth]{\imgpath/legend_categories.png}
  \end{subfigure}
  \\
  \begin{subfigure}{0.6 \linewidth}
      \includegraphics[width = \linewidth]{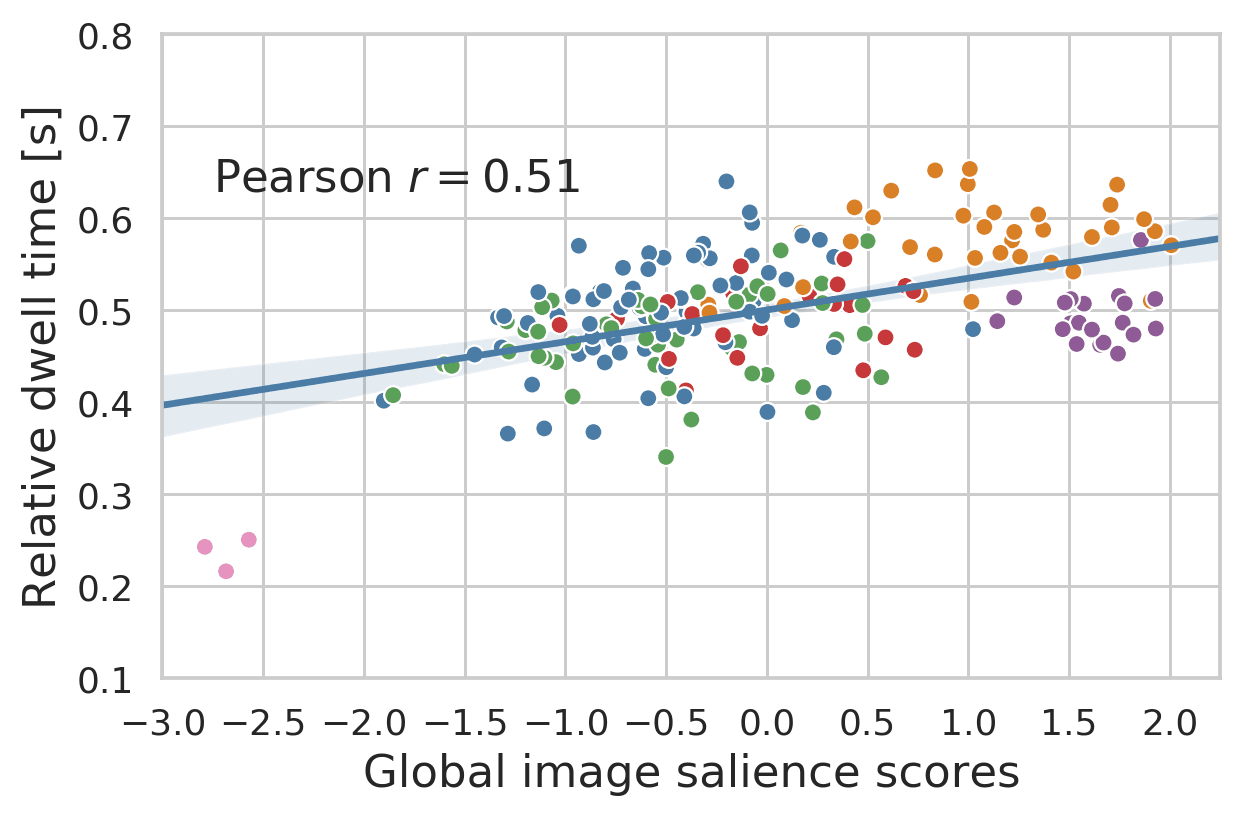}
  \end{subfigure}
  \caption{Dwell time vs. global salience scores}
\label{fig:globsal-dwell_scores}
\end{figure}

Taken together, our results suggest that the task and familiarity---but not the spatial location of images---influenced the exploration time with respect to higher dwell times on unseen images in combination with the task to select the new image. Note, however, that regarding the effects of task our findings are restricted to the specific task assigned in our experiments, that is selecting which image is new or old. The effects of task in visual attention is an active field in visual perception and the results of multiple contributions should be taken together into consideration to draw robust conclusions. Finally, we also found that images with higher global salience correspondingly led to a larger time of exploration. These results fully support our fifth hypothesis.

\section{Discussion}
\label{sec:discussion}

We have presented a computational model trained on the saccadic behaviour of participants freely looking at pairs of competing stimuli, which is able to learn a robust score for each image, related to its likelihood of attracting the first fixation. This fully supports our first hypothesis and we refer to this property of natural images as the global visual salience.

The computational model consists of a logistic regression classifier, trained with the behavioural data of 49 participants who were presented 200 pairs of images. In order to reliably assess the performance of the model, we carried out a careful 25-fold cross-evaluation, with disjoint sets of participants for training, validating and testing. Given a pair of images from the set of 200, the model predicted the direction of the first saccade with 82 \% accuracy and 0.88 area under the ROC curve.

Throughout this chapter, we have analysed the general lateral bias towards the left image (H2), as well as other possible influences such as the familiarity with one of the images and the effect of a simple task (H3). Moreover, we have analysed the relationship of our proposed global salience with the local salience properties of the individual images (H4). Finally, we have also studied the total exploration time of each image in the eye-tracking experiment and compared it to the global salience, which is based upon the first fixation (H5).

Regarding the lateral bias, we found that participants tended to look more frequently towards the image on the left. Such left bias is typical in visual behaviour and has been found in many previous studies \citep{barton2006leftbias, guo2009leftbias, calenwalshe2014assymetricfixation, ossandon2014spatialbiases}. However, most of these studies presented only single images per stimulus. In this regard, it has been argued that cultural factors of the Western population who mostly take part in the research experiments may lead to a semantic processing of natural visual stimuli similar to the reading direction, that is from left to right \citep{spalek2005leftbias, zaeinab2016leftbias}.

In our study, about 63 \% of the first fixations landed on the left image. However, we also observed a high variability across participants, successfully captured by our computational model. In contrast, we showed that the given task in certain trials did not influence initial saccade behaviour. Participants equally distributed the target location of saccades on the presented images, regardless of familiarity and task relevance. Consequently, the spatial location of an image affected saccade behaviour, whereas the task as well as familiarity had no influence.

Importantly, we found that that global salience, that is the likelihood of an image attracting the first fixation when presented next another competing image, is independent of the low-level local salience properties of the respective images. The location of the first fixations made by the participants in the study did not correlate with the GBVS salience maps of the images and the saccadic choice---left or right---was neither explained by the GBVS salience mass difference. Hence, our results provide some new insights in the understanding of visual perception of natural images, showing that the global salience of an image is rather affected by the semantics of the content. For instance, images involving socially relevant content such as humans or faces led to higher global salience than images containing purely indoor, urban or natural scenes.

To gain further insight regarding this aspect, we computed the salience maps using Deep Gaze II \citep{kuemmerer2016deepgaze}, a computational salience model that is not limited to low-level features, but also makes use of high-level cues, obtained by pre-training the model with image object recognition tasks. We repeated the same analyses as with the GBVS model and we found that metrics derived from Deep Gaze salience maps did have a non-zero, yet moderate correlation with our proposed global salience. This, together with previous evidence about the importance of low- and high-level features in detecting fixations \citep{kuemmerer2017icfdeepgaze}, matches our finding that global salience cannot be explained by low-level properties of the images. However, the relatively low correlation further suggests that the initial preference for one of the images does not depend only on properties of the individual salience maps.

According to previous research, initial eye movements in young adults are based on bottom-up image features, whereas socially relevant content is fixated later in time \citep{acik2010bottomuptopdown}. Interestingly, as described above, we found that this was not the case when two images have been shown at the same time. Considering the very short reaction time between stimulus onset and the observers reaction to fixate one of the two images, it seems surprising that participants had to pre-scan both images in their peripheral visual field before initialising the first saccade. Thus, in contrast to classical salience maps, we might argue that the global salience of an image highly relates to the semantic and socially relevant content.

In order to further investigate the effects of the global image salience, we also evaluated the total time of image exploration, that is the dwell time. We hereby found that, different to the initial saccade, the spatial location of images did not affect the time participants explored the individual images of each image pair. However, the task and familiarity had an effect. We saw that in the task where participants had to select the new image, new images were explored longer than previously seen images. In contrast, the task asking to select the old image led to an almost equal exploration time on new and familiar images. Therefore, we conclude that participants in general tended to explore new images for a slightly longer time. Nevertheless and most importantly, we saw generally---and independent of the spatial location, task and familiarity---that images with higher global salience were explored longer in time. Thus, images with larger global salience did not only attract initial eye movements after stimulus onset, but also led to longer exploration times. These results support our assumption, that the global salience score of an image can also be interpreted as a measure of the general attraction of an image, in comparison to other images. 

In this regard, note that although we considered the location of the first fixation as the target variable to model the global salience scores and carry out the subsequent analyses, the same computational model and procedures can be used to model alternative aspects of the behavioural responses. For instance, the model could be trained to fit the dwell time---which we have found to be positively correlated with the global salience based on the first fixations---, the engagement---time until fixating away---or the number of saccades.

In spite of the high performance of our computational model and its potential to assign reliable global salience scores to natural images, an important limitation is that the model and thus the scores are dependent on the image set that we used. Whereas local salience maps rely on image features, our proposed global salience model relies on the differences between the stimuli and the behavioural differences that they elicit on the participants. We observed significant differences between image categories, for example humans versus indoor scenes, but this is only one initial step and future work should investigate what other factors influence the global image salience. For example, it would be interesting to train a deep neural network with a possibly larger  set of images and the global salience scores learned by our model as labels, similarly to how Deep Gaze was trained to predict fixation locations. This could shed more light on what features make an image more globally salient.

Another related, interesting avenue for future work is investigating the global salience in homogeneous data sets, that is with images of similar content. Our work has shown that large differences exist between images with somehow different content, for instance containing humans or not. However, we did not observe significantly difference global salience between natural and urban scenes (see Figure~\ref{fig:globsal-global_salience_categories}), although significant difference do exist between specific images. An interesting question is: \textit{what} makes one image more likely to attract the first fixation, when presented alongside a semantically similar image? We think an answer to this question can be sought by combining a similar experimental setup as the one presented in this work, with additional data, and making use of advanced feature analysis, such as deep artificial neural networks, as mentioned above.

For instance, small changes in the context information of single images, might already have a dramatic influence on reaction times in decision tasks \citep{kietzmann2015topdown}. In addition, the global salience was based on eye movement behaviour of human data. Depending on the choice of participants, e.g. different culture, age, personal interests and emotions, our model could have revealed different results \citep{balcetis2006topdown, dowiasch2015aging}. Again, further studies might use the model on a wider range of participants, in order to validate the specific global salience and thus attraction of images. 

In contrast, differences in the global salience between participant groups could be a great advantage in certain research fields. In medical applications for instance, researchers could identify specific diseases, such autistic spectrum disorder (ASD). In such example, our method could generate a model of the global visual salience of both control people and individuals with certain condition, and then be used for diagnosis. Another use case of our model would be marketing research, where the attraction of different images could be compared adequately based on intuitive visual behaviour. Thus, depending on the research question, the global image salience might provide a new insight in prediction and analysis of visual behaviour.

\section{Conclusion}
\label{sec:conclusion}

Previous research has investigated the local salience properties of single images, which has helped understand visual behaviour. However, assigning a single and unique global salience score to an image as a whole has been neglected. Here, we thus trained a logistic regression model to learn unique, global salience scores for each tested image. We hereby showed that images can indeed be ranked according to their global salience, providing a new method to predict eye movement behaviour across images with distinct semantic content. These results could be used in a variety of research, such as medicine or marketing.

\chapterbibliography
}

{
\chapter[Image identification from brain activity]{Image identification\\from brain activity}
\label{ch:imageid}
\renewcommand{\chapterpath}{includes/image-id}
\begin{contributors}
    Wietske Zuiderbaan, Ben M. Harvey and Serge O. Dumoulin are the authors of the article upon which this work builds. Wietske and Serge supervised the work during my internship at their lab. Akhil Edadan and Peter K{\"onig} participated in the discussions.
\end{contributors}
\begin{outreach}
    \item \textit{Saliency and the population receptive field model to identify images from brain activity.} \textbf{Alex Hern{\'a}ndez-Garc{\'i}a}, Wietske Zuiderbaan, Akhil Edadan, Serge O. Dumoulin, Peter K{\"o}nig. Annual Meeting of the Visual Sciences Society (VSS, poster presentation), 2019.
\end{outreach}
The goal of computational neuroscience is to understand the mechanisms and principles by which the brain yields behaviour and cognition. One of the approaches to deepen our understanding of the brain is to develop models that can make predictions about brain activity. These methods, sometimes referred to as \textit{brain}/\textit{mind reading}, not only have some practical applications, but they also provide insights about the underlying neural processes that enable the prediction \citep{tong2012mindreading}. 

One particular example for the study of the visual system is the identification of presented images from brain imaging recordings, such as fMRI, in the visual cortex. While a successful approach to this challenge is to use statistical models that can learn activations patterns from a set of recordings \citep{kay2008imageid}, an alternative approach is to rely on biologically inspired encoding models. A recent proposal of this type showed that it is possible to identify the presented stimulus from a set of natural images by comparing the measured activity on areas V1, V2 and V3 with a predicted response profile encoded by the low-parametric population receptive field (pRF) model and contrast information from the images \citep{zuiderbaan2017imageidentification}. In the work presented in this chapter, we follow up this methodology by studying the predictive power of salience information from the images---instead of contrast---combined with the pRF model, and extend the analysis to higher visual areas: V1, V2, V3, hV4, LO12 and V3AB\footnote{This work was carried out during a 2-months internship at the Spinoza Centre for Neuroimaging in Amsterdam, in early 2018, supervised by Dr. Wietske Zuiderbaan and Professor Serge Dumoulin. Besides the potential scientific interest of the work, which was presented as a poster at the 19th Annual Meeting of the Vision Sciences Society (VSS), the internship was part of the training programme of my PhD fellowship, a Marie Skłodowska-Curie Innovative Training Network. A requirement of the programme is to carry out interdisciplinary internships at laboratories that are part of the partnership. As an additional contribution of my internship, I developed interactive visualisation tools in Python to better interpret the kind of data used in this project which became part of the laboratory's repository.}.

\section{Methods}
\label{sec:imageid-methods}
The goal of this study was to assess whether the salience information of images is predictive of brain activations in the visual cortex elicited by the visualisation of natural images. As a follow-up study of the work by \citet{zuiderbaan2017imageidentification}, a significant part of the methodology is borrowed from the first publication and, in particular, the neuroimaging data is the same. In this section, we will outline the most relevant aspects of the methodology that is not original of this work and refer to the original publication for further details, so as to focus on the aspects that are novel here.

\subsection{Visual stimuli}
As a data set of images we used a subset of 45 images from the Berkeley Segmentation Dataset and Benchmark \citep{martin2001berkeley}, which was already used in some of the experiments by \citep{zuiderbaan2017imageidentification}. The data set was processed such that the resulting images had a resolution of $538\times538$ pixels and were applied a circular fading mask. The set of 45 images as they were used in the experiments are shown in Figure~\ref{fig:imageid-berkely}.

\begin{figure}[htb]
  \begin{center}
    \includegraphics[width = \linewidth]{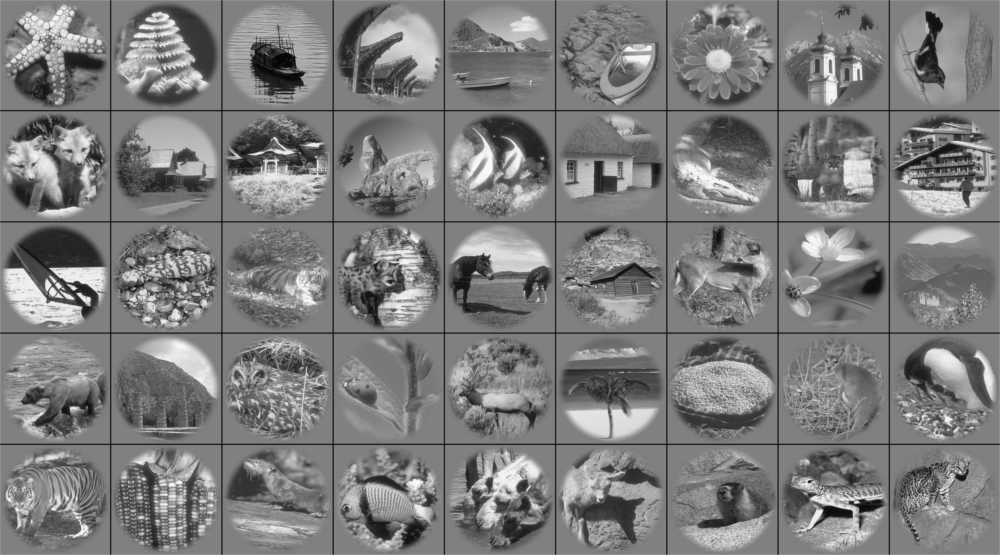}
  \end{center}
  \caption{The set of 45 natural images used in the experiments}
\label{fig:imageid-berkely}
\end{figure}

\subsection{Functional imaging}
\label{sec:imageid-functional_imaging}
The brain activations in the visual cortex were acquired in a neuroimaging study with functional magnetic resonance imaging (fMRI), carried out originally for the work presented by \citet{zuiderbaan2017imageidentification}. Therefore, the details can be found in that first article and here we will summarise the most relevant information: Two participants (one female; ages 28-38 years) took part in the fMRI study, were their brain responses to the natural images were collected. Each image was shown 18 times for a duration of 300 ms, with a radius of $5.5^{\circ}$ from the participant point of view. The neural response at each cortical location (voxel) was obtained through standard GLM-analysis \citep{friston1995fmrianalysis}: each voxel was summarised by the t-value of the fit between the predicted time series---the stimulus presentation convolved with the hemodynamic response function---and the measured BOLD signal.

In addition to the voxel responses to each stimulus, the optimal parameters of the population receptive field (pRF) model were estimated for each participant using the standard approach described by \citet{dumoulin2008prf}: For each voxel, the model estimates the position of the receptive field $(x, y)$ and the size $\sigma$ (standard deviation) using a Gaussian kernel. The parameters of the pRF were estimated using conventional contrast-defined moving-bar apertures with natural image content from images of the same data set but distinct from the set of 45 images used in the image identification experiments.

\citet{zuiderbaan2017imageidentification} analysed areas V1, V2 and V3 of the visual cortex as regions of interest. Here, we extended the analysis to higher visual areas: on the lateral occipital complex (LO-1 and LO-2: LO12), on the ventral occipital (hV4) and on the dorsal occipital (V3A and V3B: V3AB) \citep{wandell2007visualfield}. See Figure~\ref{fig:imageid-visual_field_maps} for an illustration of the regions of interest in the human visual cortex.

\begin{figure}[htb]
  \begin{center}
    \includegraphics[width = \linewidth]{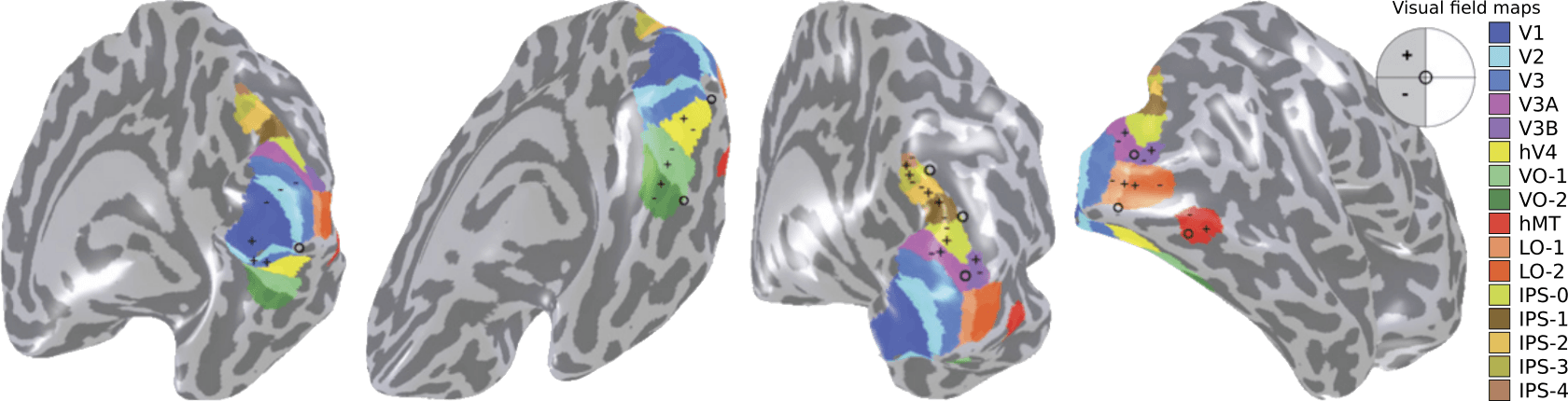}
  \end{center}
  \caption{Adapted from \citep{wandell2007visualfield} with permission from the authors: Visual field maps in the human visual cortex. In our analyses, V3A and V3B were considered a single region of interest (V3AB), and so were LO-1 and LO-2 (LO12).}
\label{fig:imageid-visual_field_maps}
\end{figure}

\subsection{Salience and contrast maps}
One of the goals of this work was to contrast the correlation of salience maps with the activations in the visual cortex, and the predictive ability of contrast maps---calculated as the root mean squared contrast---which was assessed in the original article by \citet{zuiderbaan2017imageidentification}. The choice of contrast is motivated by the evidence that the early visual cortex responds strongly to differences in contrast \citep{boynton1999contrast, olman2004contrast}. However, the visual cortex certainly responds to other properties of the stimuli and hence the interest in studying salience.

In order to analyse the correlation of salience with the activations in the visual cortex, we computed the salience maps of each image using two distinct salience models. One of the models is  DeepGaze II\footnote{Note that in Chapter~\ref{ch:globsal} we also used DeepGaze to analyse whether our proposed global salience was related or independent from the local salience properties of images, proposed by \citet{kuemmerer2016deepgaze}. DeepGaze II---for better readability, in what follows we will simply refer to it as DeepGaze---is a model that uses the features extracted by a deep neural network, VGG \citep{simonyan2014vgg}, trained on image object categorisation tasks as inputs to a 4-layer readout neural network optimised for salience prediction}, the then state-of-the-art salience model for various metrics. The second model is ICF, which stands for Intensity Contrast Features, proposed by \citet{kuemmerer2017icfdeepgaze}. ICF is trained on the same readout neural network as DeepGaze, but instead of using the VGG features, the input to the network are 5 intensity and 5 contrast feature maps.

Our choice of salience models, out of the many models proposed in the vast literature on image salience, was motivated by various reasons. First of all, we chose DeepGaze as it was the state-of-the-art model in various metrics of image salience evaluation \citep{kuemmerer2016deepgaze} and considering a model that accurately predicts the salient regions of images is also desirable for studying how salience information is encoded in the visual cortex. Nonetheless, as we have discussed in Chapter~\ref{ch:globsal}, visual attention is a complex brain mechanism and salience maps capture different aspects of it. For instance, we have discussed how visual attention can be guided by both bottom-up factors, as well as top-down factors and higher-level features. This motivated the inclusion of a model that is limited to lower-level---intensity and contrast---features, in this case ICF. The fact that DeepGaze and ICF share part of the architecture facilitates the comparisons. 

Interestingly, \citet{kuemmerer2017icfdeepgaze} analysed and compared the properties of DeepGaze and ICF, and found that while DeepGaze generally outperformed ICF in terms of the evaluation metrics used to assess salience models, ICF was more accurate in a large number of images. In particular, DeepGaze succeeds at predicting salient regions associated with higher-level factors such as objects and faces, while ICF was superior in the cases were fixations are driven by, for instance, high contrast. By way of illustration, in Figure~\ref{fig:imageid-sample_maps} we show the DeepGaze and ICF salience maps of three images from the data set, as well as their contrast maps, which were used in the original study.

\begin{figure}[htb]
  \begin{center}
    \includegraphics[width = 0.8 \linewidth]{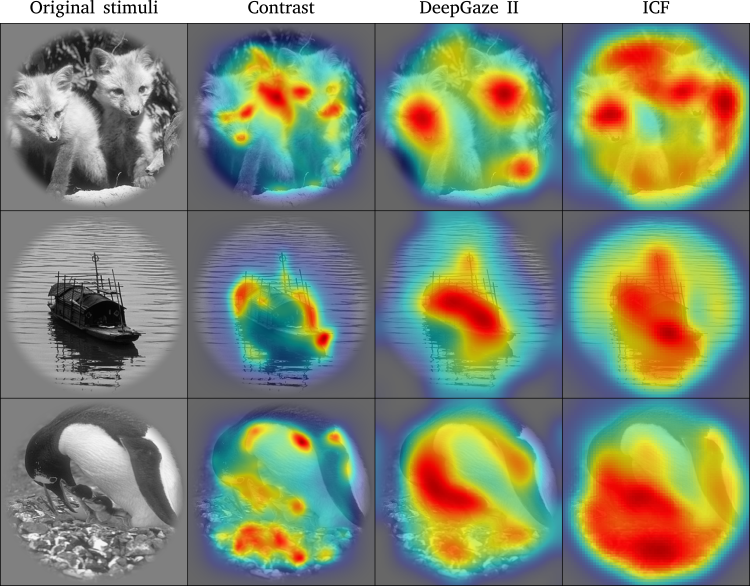}
  \end{center}
  \caption{Contrast, DeepGaze II and ICF maps of three images of the data set.}
\label{fig:imageid-sample_maps}
\end{figure}

\subsection{Brain response predictions}
To assess the correspondence between the salience and contrast maps and the brain activations, we first calculated a prediction response profile of every image as the summed overlap of the map with the pRF weighting function of each cortical location, normalised by the total volume of the pRF: 

\begin{equation}
\label{eq:imageid-pred}
    p = \frac{\sum_{i=1}^{N}w_i S_i}{\sum_{j=1}^{M}w_j}
\end{equation}
where $S_i$ denotes the value of the feature map $S$---either DeepGaze, ICF or contrast---normalised as a probability distribution, at pixel $i$; $N$ is the number of pixels within the window of the pRF and $M$ is the total number of pixels in the stimulus area. $w_i$ is the pRF weighting function, whose parameters were obtained as described in Section~\ref{sec:imageid-functional_imaging} and in \citep{zuiderbaan2017imageidentification} in more detail:

\begin{equation}
\label{eq:imageid-prf}
    w_i = \exp\left(-\frac{(x_i-x_c)^2+(y_i-y_c)^2}{2\sigma^2}\right)
\end{equation}
where $x_i$ and $y_i$ are the locations of pixel $i$; $x_c$ and $y_c$ are the centre of the pRF in the visual field; and $\sigma$ is the size of the Gaussian kernel of the pRF. Note that in \citep{zuiderbaan2017imageidentification} this is the procedure used to compute the predictions of synthetic images, while for natural images a different method was used, which was specific for the way the contrast information was obtained. Here, we use this method, since we can generalise the prediction $p$ in Equation~\ref{eq:imageid-pred} to any feature map $S$---DeepGaze, ICF or contrast.

\subsection{Evaluation metrics}
Let us formally express the measurements and predictions taking all variables into account---feature maps, areas, images, voxels---in order to describe the evaluation metrics we used in our analysis. First note that, as in the original study by \citet{zuiderbaan2017imageidentification}, we only considered for the analysis the voxels with positive t-value, pRF eccentricity values within $0.5\text{---}4.5^{\circ}$ and pRF variance explained larger than 55 \%. For every image $k$ and every visual area $A$ studied---V1, V2, V3, hV4, LO12 and V3AB---we have a \textit{measured} response profile 
\[
    \mathbf{m}_{k}^{A} = m_{k, 1}^{A}, \ldots, m_{k, D_A}^{A}
\]
where $D_A$ is the number of (\textit{valid}) voxels in area $A$. Correspondingly, we have the \textit{predicted} response profiles for every image $k$, every visual area $A$ and by every feature map $S$, that is DeepGaze, ICF and contrast:

\[
    \mathbf{p}_{k}^{A, S} = p_{k, 1}^{A, S}, \ldots, p_{k, D_A}^{A, S}
\]
where every $p_{k, v}^{A, S}$ is computed as in Equation~\ref{eq:imageid-pred}. As a similarity metric we compute the Pearson correlation between measured responses and predicted profiles. We will denote by $r_{k, l}^{A, S}$, or simply $r_{k, l}$ abusing notation, to the correlation between the measured response of image $k$, $\mathbf{m}_{k}^{A}$, and the predicted profile of image $l$, $\mathbf{p}_{l}^{A, S}$: 

\begin{equation}
\label{eq:imageid-correlation}
	r_{k, l}^{A, S} = corr(\mathbf{m}_{k}^{A}, \mathbf{p}_{l}^{A, S})
\end{equation}

In order to assess the image identification accuracy we simply considered a correct identification if $r_{k, k} > r_{k, l}$, $\forall~k \neq l$. Additionally, as a more informative metric, we combined the correlation values into a confidence score that represents how hard it is to distinguish the actual presented image $k$ from the other candidate images in the data set, based on the correlation values:

\begin{equation}
\label{eq:imageid-confidence}
c_k^{A, S} = r_{k, k}^{A, S} - \frac{1}{K}\sum_{l=1}^{K}r_{k, l}^{A, S}
\end{equation}

Finally, in order to have a compact measure to compare the predictivity of the different feature maps on every visual area, we also performed a representational similarity analysis\footnote{In Chapter~\ref{ch:daugit}, we also used representational similarity analysis to compare the features learnt by artificial neural networks and the representations measured in the inferior temporal cortex.} (RSA) \citep{kriegeskorte2008rsa}. In order to perform RSA, we constructed representational dissimilarity matrices (RDM) for both the measured responses and the predicted profiles, $M_{k, l}^{A}$ and $P_{k, l}^{A, S}$ respectively, where each entry $(k, l)$ of the matrices is 1 minus the Pearson correlation between the profile for image $k$ and image $j$. In this case, the correlation is not computed between the measured and the predicted profiles, but between two measured responses for $M_{k, l}$ and two predicted profiles for $P_{k, l}$. Thus, the RDMs are symmetric and the values in the diagonals are zero. As a summary metric to compare $M$ and $P$ we computed the Kendall correlation.

\section{Results and discussion}
\label{sec:imageid-results}
We first analyse the distribution of the confidence of the predictions for each model---feature map---and visual area, shown in Figure~\ref{fig:imageid-boxplot_confidence}. Although the results are complex and not highly consistent, several interesting conclusions can be drawn. First, we observe that the identification ability of all models is best on V1 and decreases in higher visual areas. This was observed by \citep{zuiderbaan2017imageidentification} too, who analysed V1, V2 and V3; and we here confirmed it, although we had hypothesised that salience maps might be more discriminative in higher visual areas. Unfortunately, it is not possible to conclude from our results that contrast and salience are less predictive of the activations in higher visual areas because this result could be also explained by the increased pRF sizes \citep{smith2001rfsizes}. In what follows, our analysis will focus in the earlier areas---V1, V2, V3 and, to a lesser extent, hV4---where the identification performance is better and the differences between models larger.

\begin{figure}[htb]
  \begin{center}
    \includegraphics[width = \linewidth]{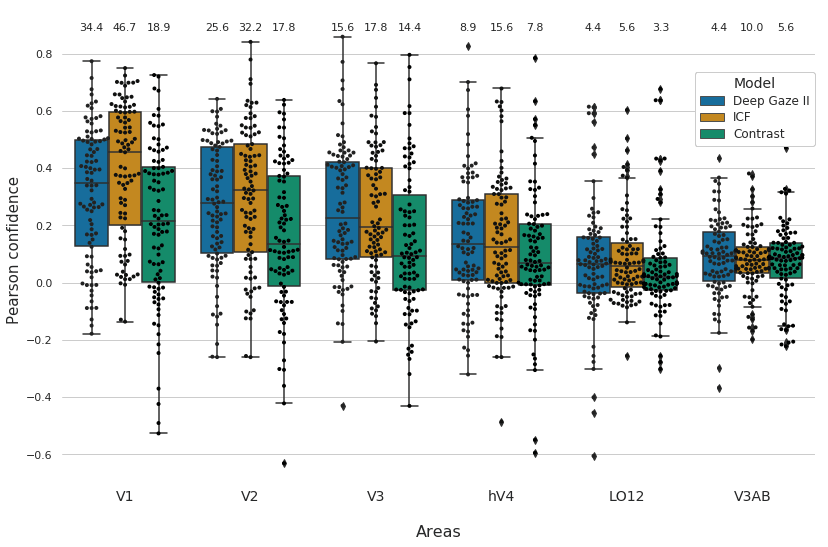}
  \end{center}
  \caption{Distribution of the confidence values (see Equation~\ref{eq:imageid-confidence}) for every feature map and visual area. Each black dot corresponds to the confidence of the prediction for one image $k$, from the data of one of the experimental participants. The numbers over each box represent the identification accuracy (as a percentage).}
\label{fig:imageid-boxplot_confidence}
\end{figure}

Second, the most relevant observation is that both salience models, DeepGaze and ICF, seem more predictive of the measured activations in the visual cortex than the contrast maps of the images. A significant difference can be observed in both the median of the distribution of the confidence values and in the identification accuracy, shown over each box. For instance, the identification accuracy on V1 using DeepGaze and ICF maps is 34.4 \% and 46.7 \% respectively, while it is below 20 \% using contrast\footnote{In the first publication by \citet{zuiderbaan2017imageidentification} the prediction model for natural images using contrast is computed differently and the evaluation procedures are not the same, hence the results differ slightly to the ones presented here. Nonetheless, in our study we analysed both methods and the results are qualitatively very similar. We here present the results using contrast maps, since the method is identical for salience maps.}.

Although image identification from brain activity has been shown before \citep{kay2008imageid}, the identification performance of the methods presented here is remarkable due to the simplicity of the model. The main differences between this pRF-based method and the model presented by \citet{kay2008imageid} were reviewed by \citet{zuiderbaan2017imageidentification}, but most importantly, note that here we estimated only 3 pRF parameters for each voxel---location and size of the receptive field---which were obtained from the fMRI recordings of a standard pRF session: responses to conventional moving-bar apertures during about half an hour. Then, the 3 pRF parameters are combined with the contrast or salience maps of the images to predict the activity of each voxel. In contrast, \citet{kay2008imageid} recorded the activations to 1,750 natural images (about 5 hours of scanning time) to fit a Gabor-Wavelet-Pyramid model with 2,730 parameters per voxel. Summarised, image identification with the pRF model requires a short scanning time and does not require training a model with brain activations towards natural image. Thus, it can be easily applied to any other image.

The fact that identifying natural images from brain activity using salience maps is more accurate than using contrast information from the images tells us that the image salience \textit{under} the receptive field of each voxel is more discriminative than the contrast. This may come as a surprise, since the early visual cortex is well known to respond strongly to differences in contrast \citep{boynton1999contrast, olman2004contrast}. Salience maps also certainly contain contrast information, in our analysis especially the maps obtained with the ICF model, but not only. Therefore, our results can be interpreted as additional evidence of the activations in the early visual cortex being shaped by multiple factors, likely top-down influences \citep{treue2003salience}.

\begin{figure}[htb]
  \begin{center}
    \includegraphics[width = \linewidth]{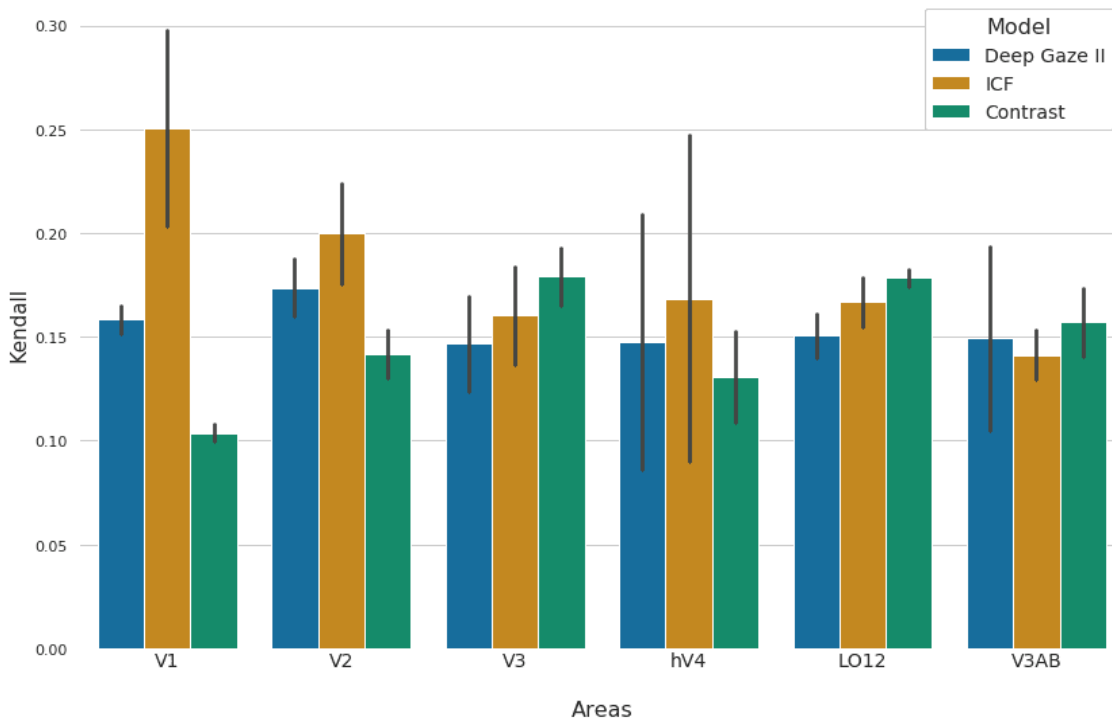}
  \end{center}
  \caption{Results of the representational similarity analysis (RSA): Kendall correlation between the representational dissimilarity matrices of the measured responses and the predicted profiles. The error bars correspond to the variation across experimental participants.}
\label{fig:imageid-kendall}
\end{figure}

Another interesting conclusion derives from the comparison between the two salience models analysed. Recall that DeepGaze computes the salience map by first extracting features through a deep neural network trained on image object recognition tasks. Therefore, these features likely contain high-level information such as face and object detectors and the model is particularly accurate at spotting salience driven by such factors, as reflected in Figure~\ref{fig:imageid-sample_maps}. On the contrary, ICF does not extract high-level features but is restricted to intensity and contrast features. Here, we found that image identification is more accurate by using ICF than DeepGaze salience maps. While this is observable in Figure~\ref{fig:imageid-boxplot_confidence}, we additionally perform a representational similarity analysis (RSA) to derive a compact metric of the ability of each model to discriminate the brain activations of each image. We present these results in Figure~\ref{fig:imageid-kendall}, which confirms the conclusion that ICF is more discriminative than DeepGaze in the earlier visual areas.

One more way to visualise the superiority of ICF at predicting brain activations in the early visual cortex is by directly analysing the correlation matrices. In Figure~\ref{fig:imageid-corr}, we plot the correlation matrices of each model on V1, where each element in the matrix encodes the correlation between the measured responses of one image (rows) and the predicted profile of another image (columns), as in Equation~\ref{eq:imageid-correlation}. Note that a correct identification occurs when the element in the diagonal---correlation between measured and predicted profile of the same image---has the highest value in the row. Visually, it becomes apparent that the diagonal in the ICF model has higher values and is better discriminated from the rest of the matrix, than in the DeepGaze model, and yet more in the contrast model. In view of these results, we conclude that image salience is more predictive of the brain activity in the early visual cortex than contrast information, and hypothesise that ICF may be more accurate than DeepGaze because the salience of the latter is driven by high-level information that may not correlate with the activity in the early areas of the visual cortex.

\begin{figure}[htb]
  \begin{center}
    \includegraphics[width = \linewidth]{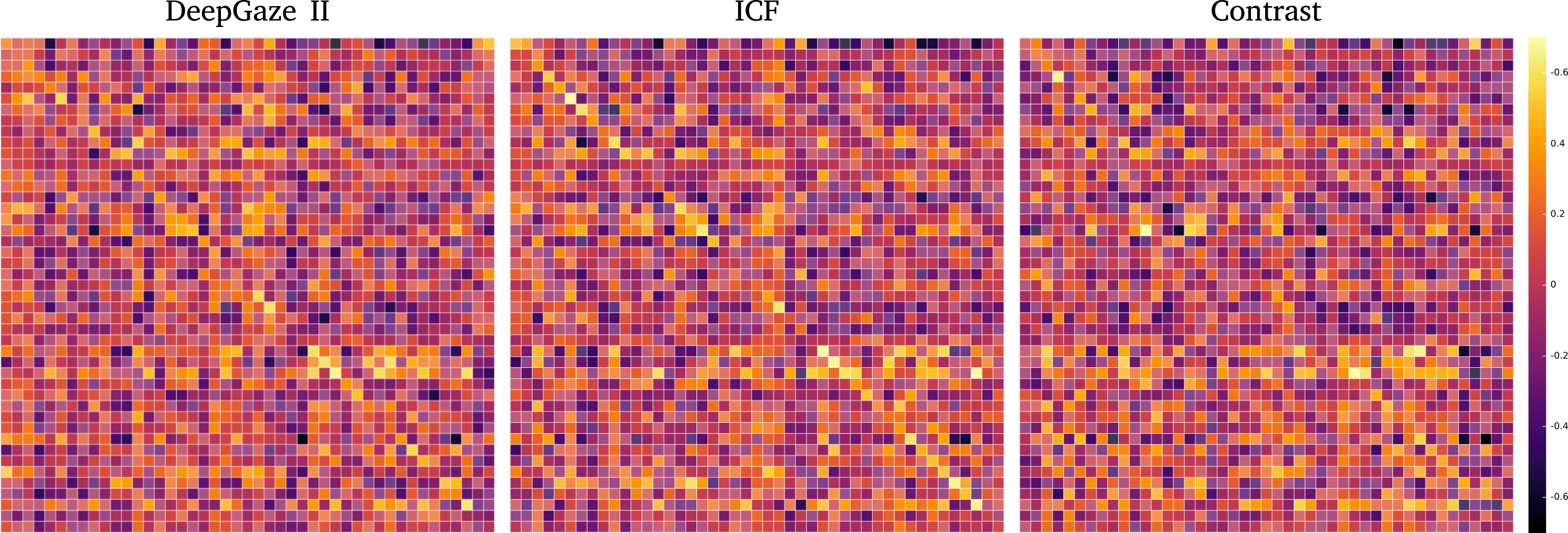}
  \end{center}
  \caption{Correlation matrices of the three models for the predictions in visual area V1. Each entry $(i, j)$ of the matrices represent $r_{i, j}^{V1, S} = corr(\mathbf{m}_{i}^{V1}, \mathbf{p}_{j}^{V1, S})$.}
\label{fig:imageid-corr}
\end{figure}

The data we obtained from the predictions allows for multiple levels of analysis, since there are several factors at play: three feature maps, six visual areas, two subjects, etc. During the two-months internship in which this project was carried out, I developed multiple interactive visualisation using Bokeh\footnote{Bokeh is an interactive visualisation library for Python: \href{https://bokeh.org/}{www.bokeh.org}} and the interactive functionality of Jupyter notebooks. Interactive visualisation provides insights that are hardly accessed otherwise and is useful to guide the more systematic, conventional, statistical analyses. In particular, it can be used to analysed the results at different levels of analysis and to look into specific details or data points. Some examples are shown in Figure~\ref{fig:imageid-interactive_visualisation}.

\begin{figure}[htb]
  \centering
  \begin{subfigure}{0.32 \linewidth}
      \includegraphics[width = \linewidth]{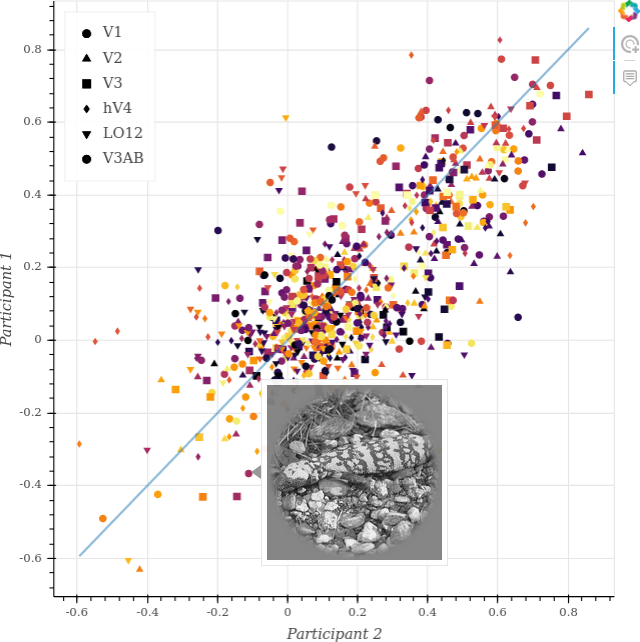}
      \label{fig:imageid-participants_all}
  \end{subfigure}
  \begin{subfigure}{0.32 \linewidth}
      \includegraphics[width = \linewidth]{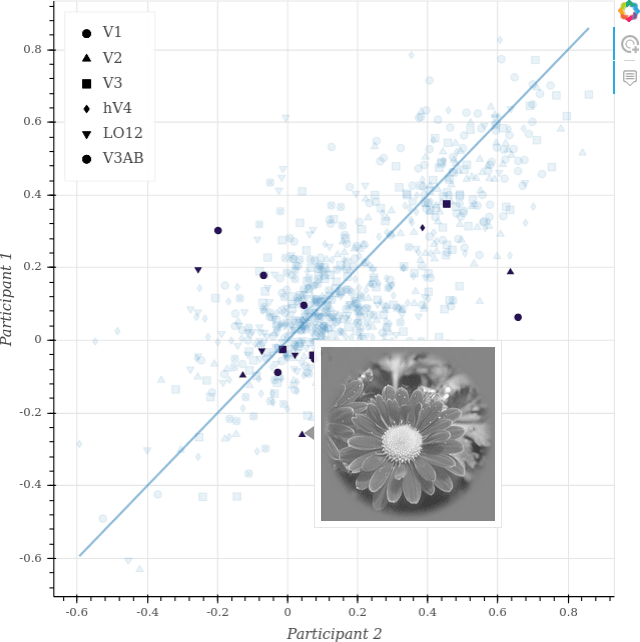}
      \label{fig:imageid-participants_specific}
  \end{subfigure}
  \begin{subfigure}{0.32 \linewidth}
      \includegraphics[width = \linewidth]{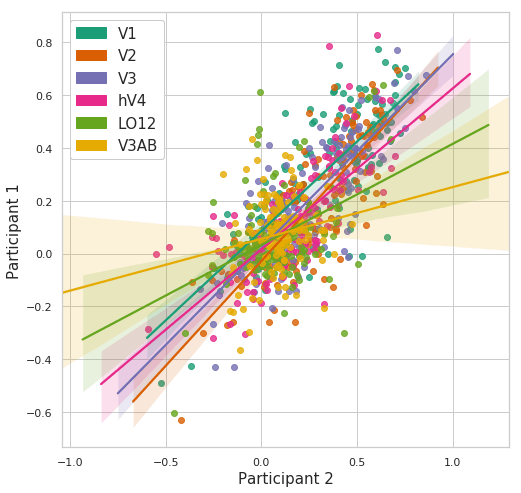}
      \label{fig:imageid-participants_fit}
  \end{subfigure}
  \\
  \begin{subfigure}{\linewidth}
      \includegraphics[width = \linewidth]{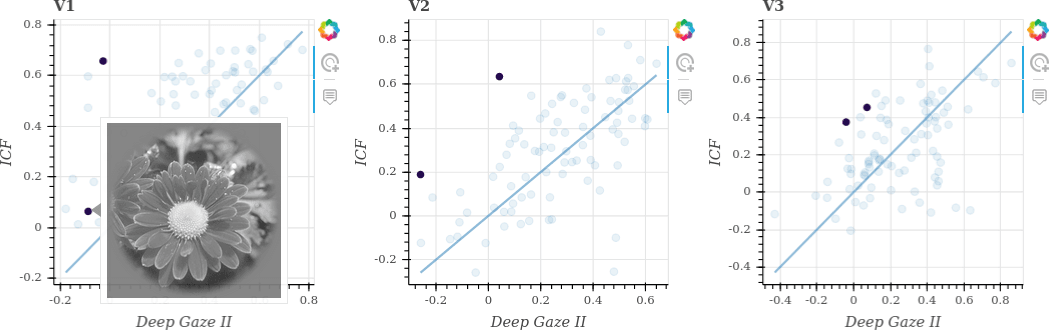}
      \label{fig:imageid-icf_vs_dg}
  \end{subfigure}
  \caption{Examples of interactive scatter plots of the Pearson confidence (Equation~\ref{eq:imageid-confidence}) to contrast different experimental variables. The interactive plot allows to hover over the data points to visualise the image that they represent and highlight all data points of one image by clicking on them. Bottom row: DeepGaze vs. ICF on areas V1, V2 and V3. Top row: comparison of the confidence values of the predictions from the data of the two experimental participants. From left to right: data from the three models, represented with different shapes for each visual area and different colours for each image; highlight of one image, after clicking on one point; linear fit for each visual area separately.}
  \label{fig:imageid-interactive_visualisation}
\end{figure}

Besides the main conclusions discussed above, some other observations are the following: In general, we found a positive linear correlation in the prediction confidence between the predictions on the data from both participants in areas V1, V2, V3 and hV4 ($r = 0.76, 0.80, 0.78, 0.71$ respectively) and less so in LO12 and V3AB ($r = 0.42, 0.17$), where the predictions were also less confident. We also found a correlation between the predictions across visual areas, that is images that were confidently predicted in V1 were also in V2 ($r = 73$) and in V3 ($r = 0.56$), and the correlation decreases further in higher visual areas, as expected. 

The prediction confidence was also positively correlated between DeepGaze and ICF ($r = 0.58, 0.62, 0.50, 0.58$ in areas V1, V2, V3 and hV4), although with exceptions, that is several images were confidently predicted by ICF but not by DeepGaze, and vice versa. The interactive visualisation was useful to identify and study these cases. Not surprisingly, the correlation was much lower between the salience  and the contrast models. For instance, in V1, the correlation between DeepGaze and contrast and ICF and contrast was $r = 0.25, 0.35$, respectively.

\section{Conclusion}
\label{sec:imageid-conclusion}
In this chapter we have presented the results of a project carried out during a two-months internship at the Spinoza Centre for Neuroimaging in Amsterdam, in which we extended the work presented by \citet{zuiderbaan2017imageidentification}. In particular we analysed the discriminability of salience maps to identify images from brain activity in the visual cortex, using a low-parametric model of the receptive field of the measured cortical locations, the population receptive field (pRF) model \citep{dumoulin2008prf}.

We contrasted the identification performance of two distinct salience models---DeepGaze and ICF--- and the contrast-based model originally presented by \citet{zuiderbaan2017imageidentification} and found that the salience information within the receptive fields of the voxels is more predictive of the brain activity than the contrast information. Furthermore, we observed that ICF, whose salience maps are computed using low-level features, is more predictive than DeepGaze, which encodes high-level information such as salience driven by faces and objects.

Overall, this analysis demonstrates that this simple method of prediction of brain activity using the pRF model can be extended to other types of image information beyond contrast, enabling further analysis. Future work may use this kind of analysis to further understand the computations in the early visual cortex and extend this methodology to other cortical areas.

\chapterbibliography
}

{
\chapter{General discussion}
\label{ch:discussion}
\renewcommand{\chapterpath}{includes/discussion}
In this dissertation I have presented a series of experimental studies and discussions revolving around machine learning for image understanding, visual perception and visual neuroscience. An overarching objective of this work was to explore and exploit the connections between these fields, combining the tools and techniques common to each discipline. 

A central subject of the dissertation has been data augmentation. Data augmentation has been ubiquitously used to train machine learning models on image tasks since the early 1990s, but it has received little scientific attention. In the first part of the thesis, we tried to bring data augmentation to the fore and study its role as implicit regularisation of machine learning algorithms and its potential to incorporate inductive biases from visual perception and biological vision. While on the surface data augmentation is just a method to synthetically increase the number of examples in a data set, we have here analysed it as a technique that encodes effective priors from perception: The image transformations typically included in data augmentation techniques---rotations, translations, scaling, changes in illumination, etc.---coincide with those that are plausible in the real world as we perceive it. Likely not by coincidence, the visual cortex of our brains represents objects under these transformations in a largely robust way.

From a machine learning point of view, data augmentation can be seen as a form of regularisation, in that it helps improve generalisation. Nonetheless, we discussed an important distinction between the type of regularisation provided by data augmentation---implicit regularisation---and explicit regularisation techniques (Chapter~\ref{ch:reg}). The terms explicit and implicit regularisation have appeared frequently in the deep learning literature, but no formal definition had been provided, to the best of our knowledge. Hence, the terms have been in used in an inconsistent and subjective manner. We here provided formal definitions of the two concepts based on their effect on the representational capacity of the model they are applied on, alongside several examples of each category for illustration. Importantly, we argued that data augmentation does not reduce the representational capacity and therefore is not explicit but implicit regularisation. We hope our definitions find consensus in the machine learning community and foster more rigorous discussions about regularisation.

In Chapter~\ref{ch:daugreg}, we delved into the distinction between data augmentation and explicit regularisation. We departed from the hypothesis that data augmentation improves generalisation by increasing the number of training examples through transformations that resemble those that can be find in the real world, while explicit regularisation \textit{simply} relies on the inductive bias that simpler models should generalise better. Although this inductive bias is at the root of the feasibility of learning from data and has proven effective in uncountable applications, the prior knowledge encoded by data augmentation seems intuitively more effective. Accordingly, we challenged the need for explicit regularisation techniques such as weight decay and dropout to train deep neural networks, provided data augmentation is also employed. If large networks with orders of magnitude more learnable parameters than training examples are able to generalise well, is it necessary to constrain their representational capacity? We first derived some theoretical insights from the literature that suggest that weight and dropout can be seen as \textit{naive} data augmentation, that is without domain knowledge. We then confirmed through an empirical evaluation that models trained with data augmentation alone outperform the combination of explicit regularisation and data augmentation typically used in practice.

Although the experimental setup of our empirical study included several network architectures and data sets, with results of over 300 trained models, extended experimentation would be of course desirable. All the experimental results from training neural networks presented in this thesis have been conducted with one---occasionally two---graphical processing unit (GPU) available. It would be highly beneficial if researchers without such computational limitations extended this analysis to confirm or reject our conclusions, and therefore we made the code available alongside the publications. Another desirable extension of this part of the dissertation would be to compare data augmentation and explicit regularisation in other data modalities beyond natural images, such as speech, text or medical images.

Since one of the motivations for analysing image data augmentation was its connection with visual perception and biological vision, we hypothesised that larger variation in the image transformations seen by a neural network may induce better representational similarity with the inferior temporal cortex. This is the region in the visual cortex where it is possible to decode object classes from measured activations and invariance to transformations has been repeatedly observed. In Chapter~\ref{ch:daugit}, we used representational similarity analysis to compare the features learnt by artificial neural networks and the activations measured in the inferior temporal cortex through fMRI. As hypothesised, we found that models trained with heavier transformations exhibit higher similarity with the visual cortex. This study was the result of a short collaboration in which we tested the idea with a limited experimental setup. Therefore, it would also be desirable to find more evidence of our conclusion in future work, as well as delving into what specific transformations drive invariance in the higher visual cortex.

The last chapter of the block on data augmentation made the connection with visual perception and biological vision more explicit. We departed from the idea that simply applying transformations to the input images and optimising a neural network for classification may not be enough to learn robust features as in the higher visual cortex. We first observed that useful information is lost in the way data augmentation is commonly applied: every time an image is transformed according to a data augmentation scheme, it is fed into the network to compute the classification loss just as any other new image. The transformed image is not just one more image, but a perceptually plausible transformation of another image in the set. With the standard classification objectives this potentially valuable information is simply lost. Could it not be used as an inductive bias?

In order to further exploit the potential inductive bias of data augmentation, in Chapter~\ref{ch:invariance} we proposed \textit{data augmentation invariance}, a simple learning objective inspired by the increasing invariance to identity-preserving transformations observed in the ventral visual stream. Data augmentation invariance combines several novel aspects: First, we perform data augmentation within the training batches, that is we construct the mini-batches by including $M$ transformations of each image. In this way, the model has access to the multiple transformations of an example at once---instead of separated by many iterations---and it potentially reduces the variance of the gradients. Second, we proposed a contrastive loss term that encourages similar representations of images that are transformations of each other. This has been suggested to be a key property of the inferior temporal cortex. Third, we define the data augmentation invariance objective in a layer-wise fashion, that is the representational invariance is optimised at multiple layers of the network. However, we distribute the weights of the loss terms of each layer exponentially along the hierarchy. This aimed to loosely mimic the increasing invariance along the visual cortex. We trained several architectures with data augmentation invariance and the models effectively and efficiently learnt robust representations, without detriment of the classification performance. In contrast, the representations of models trained with the standard categorical cross-entropy loss did not become more invariant to transformations than at the pixel space, in spite of being exposed to data augmentation during training.

Although our results were remarkably consistent across architectures and data sets, future work should find more evidence for the benefits of data augmentation invariance. Furthermore, we are interested in exploring other potential benefits of training with this objective. In particular, we would like to test the representational similarity of the learnt features with the inferior temporal cortex, which inspired this approach. Furthermore, it would be interesting to study whether encouraging invariance to some transformations---rotations, translations, illumination changes---induces invariance to other transformations, such as occlusions as in cutout augmentation.

In the second part of the dissertation, we moved the focus from data augmentation and artificial neural networks to visual attention and salience, using tools of cognitive science and neuroscience, such as eye-tracking and neuroimaging. In Chapter~\ref{ch:globsal}, we proposed and analysed the concept of \textit{global visual salience}. While a large body of scientific literature has studied visual attention and the salience properties of images, it has mostly focused in analysing what parts and features of an image drive eye movements and are more likely to attract fixations. Here, we studied the likelihood of natural images as a whole to attract the initial fixation of a human observer, when presented in competition with other images. For this purpose, we carried out an eye tracking experiment in which we showed participants pairs of images side by side. We trained a simple machine learning algorithm with the behavioural data from the experiment and found that it is possible to predict the direction of the first saccade---left or right---given a pair of images from the data set. This implies that some images have a higher \textit{global visual salience} than others. Specifically, faces and images with social content are most likely to be fixated first. Importantly, we also found that global salience is largely independent from the local salience properties of the images. 

We believe our experimental data can be further used to study aspects of human visual attention of competing stimuli, since we mostly focused on the direction of the first fixation upon stimulus presentation. Therefore, we open sourced the data and the code of our analyses. In particular, it would be interesting to study the reaction times and engagement with the stimuli during the duration of the trials so as to find if there exist differences depending on the nature of the two images, for instance. Another interesting direction would be to more deeply study the visual properties of the images and find out whether it is possible to predict the global visual attention of novel images. Further, we hypothesised that global salience could be used as a tool or metric to better understand the visual attention behaviour of humans with conditions such as the autism spectrum disorder.

Finally, in Chapter~\ref{ch:imageid}, we analysed the relationship between the local salience maps of natural images and the brain activations in the early visual cortex. In particular, we followed up a previous study that demonstrated the possibility to identify natural images from brain activity using the low-parametric population receptive field (pRF) model and contrast information from the images. In our work, we extended that study by analysing the discriminability of salience maps. We compared contrast and salience maps computed with two distinct image salience models, one based on low-level features and the other based on high-level features learnt by a deep neural network. We found that salience, especially based on low-level features, is significantly more predictive of brain activity than contrast. This suggests that the activations in the early visual cortex contain information about various properties of the images, likely driven by feedback connectivity from higher areas. Moreover, the results in this chapter provided additional evidence for the possibility of studying properties of the visual cortex through predictive models based on simple tools such as the pRF model.

Before concluding this dissertation, I would like to briefly discuss some ethical considerations and the societal and environmental impact of the work presented here. First, although compared to much of the deep learning literature the computational resources used for this work were small, some of the results of this thesis required training multiple neural network models, especially for Chapter~\ref{ch:daugreg}. Training these models certainly contributed negatively on the environment with emmision of carbon dioxide, as reported in the chapter. In order to disseminate my work and engage with other scientists, I travelled by plane to attend several conferences, which also had a negative impact on climate change. I strongly advocate minimising the impact of scientific activity on the environment. One way of positively contributing to reduce this impact is through data sharing. Hence, we have made available much of the data collected for this work, which will also hopefully contribute to more open science. Currently, deep neural networks are remarkably energy-inneficient compared to brains. Incorporating better inductive biases, as we have discussed in this thesis, may contribute to more efficient machine learning algorithms. Second, while I do not envision a direct negative use of the work presented here, I believe that as work that aims to advance our technology, it has the potential of being misused or negatively impact our society. As Professor Ruha Benjamin puts it, ``technology can exclude without being explicitly designed for it''. I hope this is not the case of my work and I explicitly disapprove the use of the results, conclusions, data and code related to this work for applications that incite racism, sexism or unequal treatment of marginalised groups.

In sum, in this dissertation we have presented the results of various projects connecting different fields, such as machine learning, cognitive science and computational neuroscience. While science clearly needs the depth of very narrow studies, we have here tried to show the supplementary value of an interdisciplinary approach to science. In particular, I believe that understanding the nature of learning systems---both algorithms and brains---requires the collaboration of scientists of multiple disciplines, as many other researchers have argued before me. Learning algorithms will become more effective and efficient by incorporating insights from the brain; and we will deepen our understanding of the brain by using the tools of improved machine learning. 
}

\includepdf[pages=-]{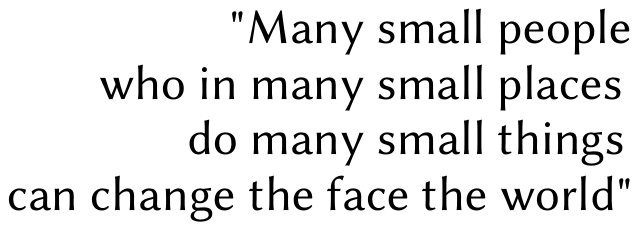}


\begin{thebibliography}{38}
\providecommand{\natexlab}[1]{#1}
\providecommand{\url}[1]{\texttt{#1}}
\expandafter\ifx\csname urlstyle\endcsname\relax
  \providecommand{\doi}[1]{doi: #1}\else
  \providecommand{\doi}{doi: \begingroup \urlstyle{rm}\Url}\fi

\bibitem[Abu-Mostafa et~al.(2012)Abu-Mostafa, Magdon-Ismail, and
  Lin]{abu2012learningfromdata}
Abu-Mostafa, Y.~S., Magdon-Ismail, M., and Lin, H.-T.
\newblock \emph{Learning from data}.
\newblock AMLBooks, 2012.

\bibitem[Allen(1974)]{allen1974crossval}
Allen, D.~M.
\newblock The relationship between variable selection and data agumentation and
  a method for prediction.
\newblock \emph{Technometrics}, 1974.

\bibitem[Alpaydin(2009)]{alpaydin2009machinelearning}
Alpaydin, E.
\newblock \emph{Introduction to machine learning}.
\newblock MIT Press, 2009.

\bibitem[Babenko(2018)]{babenko2018wdvsl2}
Babenko, B.
\newblock weight decay vs {L2} regularization.
\newblock Accessed: 2020-05-20, 2018.
\newblock URL \url{https://bbabenko.github.io/weight-decay/}.

\bibitem[Baldi \& Sadowski(2013)Baldi and Sadowski]{baldi2013dropout}
Baldi, P. and Sadowski, P.~J.
\newblock Understanding dropout.
\newblock In \emph{Advances in Neural Information Processing Systems
  (NeurIPS)}, 2013.

\bibitem[Bartlett et~al.(2002)Bartlett, Boucheron, and
  Lugosi]{bartlett2002complexity}
Bartlett, P.~L., Boucheron, S., and Lugosi, G.
\newblock Model selection and error estimation.
\newblock \emph{Machine Learning}, 2002.

\bibitem[Bishop(1995)]{bishop1995tikhonov}
Bishop, C.~M.
\newblock Training with noise is equivalent to {Tikhonov} regularization.
\newblock \emph{Neural Computation}, 1995.

\bibitem[Bousquet et~al.(2003)Bousquet, Boucheron, and
  Lugosi]{bousquet2003learningtheory}
Bousquet, O., Boucheron, S., and Lugosi, G.
\newblock Introduction to statistical learning theory.
\newblock In \emph{Advanced Lectures on Machine Learning}. 2003.

\bibitem[Bouthillier et~al.(2015)Bouthillier, Konda, Vincent, and
  Memisevic]{bouthillier2015dropoutasdaug}
Bouthillier, X., Konda, K., Vincent, P., and Memisevic, R.
\newblock Dropout as data augmentation.
\newblock \emph{arXiv preprint arXiv:1506.08700}, 2015.

\bibitem[Breiman(1996)]{breiman1994bagging}
Breiman, L.
\newblock Bagging predictors.
\newblock \emph{Machine Learning}, 1996.

\bibitem[Gal \& Ghahramani(2016)Gal and Ghahramani]{gal2016dropout}
Gal, Y. and Ghahramani, Z.
\newblock Dropout as a {Bayesian} approximation: Representing model uncertainty
  in deep learning.
\newblock In \emph{International Conference on Machine Learning (ICML)}, 2016.

\bibitem[Girosi et~al.(1995)Girosi, Jones, and
  Poggio]{girosi1995regularization}
Girosi, F., Jones, M., and Poggio, T.
\newblock Regularization theory and neural networks architectures.
\newblock \emph{Neural Computation}, 1995.

\bibitem[Hadamard(1902)]{hadamard1902illposed}
Hadamard, J.
\newblock Sur les probl{\`e}mes aux d{\'e}riv{\'e}es partielles et leur
  signification physique.
\newblock \emph{Princeton University Bulletin}, 1902.

\bibitem[Helmbold \& Long(2017)Helmbold and Long]{helmbold2017dropout}
Helmbold, D.~P. and Long, P.~M.
\newblock Surprising properties of dropout in deep networks.
\newblock \emph{Journal of Machine Learning Research (JMLR)}, 2017.

\bibitem[Hertz et~al.(1991)Hertz, Krogh, and Palmer]{hertz1991dilution}
Hertz, J., Krogh, A., and Palmer, R.~G.
\newblock \emph{Introduction to the Theory of Neural Computation}.
\newblock Addison-Wesley Longman Publishing Co., Inc., USA, 1991.

\bibitem[Hinton(1987)]{hinton1987wd}
Hinton, G.~E.
\newblock Learning translation invariant recognition in a massively parallel
  networks.
\newblock In \emph{International Conference on Parallel Architectures and
  Languages Europe}. 1987.

\bibitem[Hinton et~al.(2012)Hinton, Srivastava, Krizhevsky, Sutskever, and
  Salakhutdinov]{hinton2012dropout}
Hinton, G.~E., Srivastava, N., Krizhevsky, A., Sutskever, I., and
  Salakhutdinov, R.~R.
\newblock Improving neural networks by preventing co-adaptation of feature
  detectors.
\newblock \emph{arXiv preprint arXiv:1207.0580}, 2012.

\bibitem[Hoeffding(1963)]{hoeffding1963hoeffding}
Hoeffding, W.
\newblock Probability inequalities for sums of bounded random variables.
\newblock \emph{Journal of the American Statistical Association}, 1963.

\bibitem[Ivanov(1976)]{ivanov1976regularisation}
Ivanov, V.~V.
\newblock \emph{The theory of approximate methods and their applications to the
  numerical solution of singular integral equations}.
\newblock Nordhof International, 1976.

\bibitem[Krogh \& Hertz(1992)Krogh and Hertz]{krogh1992wd}
Krogh, A. and Hertz, J.~A.
\newblock A simple weight decay can improve generalization.
\newblock In \emph{Advances in Neural Information Processing Systems
  (NeurIPS)}, 1992.

\bibitem[Livnat et~al.(2010)Livnat, Papadimitriou, Pippenger, and
  Feldman]{livnat2010sex}
Livnat, A., Papadimitriou, C., Pippenger, N., and Feldman, M.~W.
\newblock Sex, mixability, and modularity.
\newblock \emph{Proceedings of the National Academy of Sciences (PNAS)}, 2010.

\bibitem[Mou et~al.(2018)Mou, Zhou, Gao, and Wang]{mou2018dropout}
Mou, W., Zhou, Y., Gao, J., and Wang, L.
\newblock Dropout training, data-dependent regularization, and generalization
  bounds.
\newblock In \emph{International Conference on Machine Learning (ICML)}, 2018.

\bibitem[Murphy(2012)]{murphy2012machinelearning}
Murphy, K.~P.
\newblock \emph{Machine learning: a probabilistic perspective}.
\newblock MIT Press, 2012.

\bibitem[Neyshabur et~al.(2015)Neyshabur, Tomioka, and
  Srebro]{neyshabur2015regularization}
Neyshabur, B., Tomioka, R., and Srebro, N.
\newblock Norm-based capacity control in neural networks.
\newblock In \emph{Conference on Learning Theory}, 2015.

\bibitem[Phillips(1962)]{phillips1962regularisation}
Phillips, D.~L.
\newblock A technique for the numerical solution of certain integral equations
  of the first kind.
\newblock \emph{Association for Computing Machinery}, 1962.

\bibitem[Poggio \& Girosi(1990)Poggio and Girosi]{poggio1990regularisation}
Poggio, T. and Girosi, F.
\newblock Networks for approximation and learning.
\newblock \emph{Proceedings of the IEEE}, 1990.

\bibitem[Srivastava et~al.(2014)Srivastava, Hinton, Krizhevsky, Sutskever, and
  Salakhutdinov]{srivastava2014dropout}
Srivastava, N., Hinton, G.~E., Krizhevsky, A., Sutskever, I., and
  Salakhutdinov, R.~R.
\newblock Dropout: a simple way to prevent neural networks from overfitting.
\newblock \emph{Journal of Machine Learning Research (JMLR)}, 2014.

\bibitem[Stone(1974)]{stone1974crossval}
Stone, M.
\newblock Cross-validatory choice and assessment of statistical predictions.
\newblock \emph{Journal of the Royal Statistical Society: Series B
  (Methodological)}, 1974.

\bibitem[Tikhonov(1963)]{tikhonov1963regularisation}
Tikhonov, A.~N.
\newblock On solving ill-posed problem and method of regularization.
\newblock \emph{Doklady Akademii Nauk USSR}, 1963.

\bibitem[Vapnik(1982)]{vapnik1982erm}
Vapnik, V.~N.
\newblock \emph{Estimation of dependences based on empirical data}.
\newblock Springer, 1982.

\bibitem[Vapnik(1992)]{vapnik1992erm}
Vapnik, V.~N.
\newblock Principles of risk minimization for learning theory.
\newblock In \emph{Advances in Neural Information Processing Systems
  (NeurIPS)}, 1992.

\bibitem[Vapnik(1995)]{vapnik1995learningtheory}
Vapnik, V.~N.
\newblock \emph{The nature of statistical learning theory}.
\newblock Springer Verlag, 1995.

\bibitem[Vapnik \& Chervonenkis(1971)Vapnik and Chervonenkis]{vapnik1971vc}
Vapnik, V.~N. and Chervonenkis, A.~Y.
\newblock On the uniform convergence of relative frequencies of events to their
  probabilities.
\newblock \emph{Theory of Probability and its Applications}, 1971.

\bibitem[Vapnik \& Chervonenkis(1974)Vapnik and Chervonenkis]{vapnik1974srm}
Vapnik, V.~N. and Chervonenkis, A.~Y.
\newblock Theory of pattern recognition.
\newblock 1974.

\bibitem[Vapnik \& Chervonenkis(1991)Vapnik and
  Chervonenkis]{vapnik1991necessary}
Vapnik, V.~N. and Chervonenkis, A.~Y.
\newblock The necessary and sufficient conditions for consistency of the method
  of empirical risk minimization.
\newblock \emph{Pattern Recognition and Image Analysis}, 1991.

\bibitem[Von~Luxburg \& Sch{\"o}lkopf(2011)Von~Luxburg and
  Sch{\"o}lkopf]{vonluxburg2011learningtheory}
Von~Luxburg, U. and Sch{\"o}lkopf, B.
\newblock Statistical learning theory: Models, concepts, and results.
\newblock In \emph{Handbook of the History of Logic}, volume~10, pp.\
  651--706. Elsevier, 2011.

\bibitem[Wager et~al.(2013)Wager, Wang, and Liang]{wager2013dropout}
Wager, S., Wang, S., and Liang, P.~S.
\newblock Dropout training as adaptive regularization.
\newblock In \emph{Advances in Neural Information Processing Systems
  (NeurIPS)}, 2013.

\bibitem[Zhang et~al.(2018)Zhang, Wang, Xu, and Grosse]{zhang2018wd}
Zhang, G., Wang, C., Xu, B., and Grosse, R.
\newblock Three mechanisms of weight decay regularization.
\newblock \emph{arXiv preprint arXiv:1810.12281}, 2018.

\end{thebibliography}


\begin{thebibliography}{15}
\providecommand{\natexlab}[1]{#1}
\providecommand{\url}[1]{\texttt{#1}}
\expandafter\ifx\csname urlstyle\endcsname\relax
  \providecommand{\doi}[1]{doi: #1}\else
  \providecommand{\doi}{doi: \begingroup \urlstyle{rm}\Url}\fi

\bibitem[Girshick et~al.(2014)Girshick, Donahue, Darrell, and
  Malik]{girshick2014dlhierarchy}
Girshick, R., Donahue, J., Darrell, T., and Malik, J.
\newblock Rich feature hierarchies for accurate object detection and semantic
  segmentation.
\newblock In \emph{IEEE Conference on Computer Vision and Pattern Recognition
  (CVPR)}, 2014.

\bibitem[G{\"u}{\c{c}}l{\"u} \& van Gerven(2015)G{\"u}{\c{c}}l{\"u} and van
  Gerven]{gucclu2015annbrains}
G{\"u}{\c{c}}l{\"u}, U. and van Gerven, M.~A.
\newblock Deep neural networks reveal a gradient in the complexity of neural
  representations across the ventral stream.
\newblock \emph{Journal of Neuroscience}, 2015.

\bibitem[He et~al.(2016)He, Zhang, Ren, and Sun]{he2016resnet}
He, K., Zhang, X., Ren, S., and Sun, J.
\newblock Deep residual learning for image recognition.
\newblock In \emph{IEEE Conference on Computer Vision and Pattern Recognition
  (CVPR)}, 2016.

\bibitem[Khaligh-Razavi \& Kriegeskorte(2014)Khaligh-Razavi and
  Kriegeskorte]{khaligh2014annbrains}
Khaligh-Razavi, S.-M. and Kriegeskorte, N.
\newblock Deep supervised, but not unsupervised, models may explain it cortical
  representation.
\newblock \emph{PLOS Computational Biology}, 2014.

\bibitem[Kietzmann et~al.(2019)Kietzmann, McClure, and
  Kriegeskorte]{kietzmann2019dnncompneuro}
Kietzmann, T.~C., McClure, P., and Kriegeskorte, N.
\newblock Deep neural networks in computational neuroscience.
\newblock \emph{Oxford Research Encyclopedia of Neuroscience}, 2019.

\bibitem[Kriegeskorte et~al.(2008{\natexlab{a}})Kriegeskorte, Mur, and
  Bandettini]{kriegeskorte2008rsa}
Kriegeskorte, N., Mur, M., and Bandettini, P.~A.
\newblock Representational similarity analysis-connecting the branches of
  systems neuroscience.
\newblock \emph{Frontiers in Systems Neuroscience}, 2008{\natexlab{a}}.

\bibitem[Kriegeskorte et~al.(2008{\natexlab{b}})Kriegeskorte, Mur, Ruff, Kiani,
  Bodurka, Esteky, Tanaka, and Bandettini]{kriegeskorte2008manandmonkey}
Kriegeskorte, N., Mur, M., Ruff, D.~A., Kiani, R., Bodurka, J., Esteky, H.,
  Tanaka, K., and Bandettini, P.~A.
\newblock Matching categorical object representations in inferior temporal
  cortex of man and monkey.
\newblock \emph{Neuron}, 2008{\natexlab{b}}.

\bibitem[Mehrer et~al.(2017)Mehrer, Kietzmann, and Kriegeskorte]{mehrer2017ccn}
Mehrer, J., Kietzmann, T.~C., and Kriegeskorte, N.
\newblock Deep neural networks trained on ecologically relevant categories
  better explain human {IT}.
\newblock In \emph{Conference on Cognitive Computational Neuroscience (CCN)},
  2017.

\bibitem[Nili et~al.(2014)Nili, Wingfield, Walther, Su, Marslen-Wilson, and
  Kriegeskorte]{nili2014rsatoolbox}
Nili, H., Wingfield, C., Walther, A., Su, L., Marslen-Wilson, W., and
  Kriegeskorte, N.
\newblock A toolbox for representational similarity analysis.
\newblock \emph{PLOS Computational Biology}, 2014.

\bibitem[Russakovsky et~al.(2015)Russakovsky, Deng, Su, Krause, Satheesh, Ma,
  Huang, Karpathy, Khosla, Bernstein, Berg, and
  Fei-Fei]{russakovsky2015imagenet}
Russakovsky, O. et~al.
\newblock {ImageNet} large scale visual recognition challenge.
\newblock \emph{International Journal of Computer Vision (IJCV)}, 2015.

\bibitem[Springenberg et~al.(2014)Springenberg, Dosovitskiy, Brox, and
  Riedmiller]{springenberg2014allcnn}
Springenberg, J.~T., Dosovitskiy, A., Brox, T., and Riedmiller, M.
\newblock Striving for simplicity: The all convolutional net.
\newblock In \emph{International Conference on Learning Representations (ICLR),
  arXiv:1412.6806}, 2014.

\bibitem[Storrs et~al.(2017)Storrs, Mehrer, Walther, and
  Kriegeskorte]{storrs2017ccn}
Storrs, K., Mehrer, J., Walther, A., and Kriegeskorte, N.
\newblock Architecture matters: How well neural networks explain {IT}
  representation does not depend on depth and performance alone.
\newblock In \emph{Conference on Cognitive Computational Neuroscience (CCN)},
  2017.

\bibitem[Yamins \& DiCarlo(2016)Yamins and DiCarlo]{yamins2016computneuro}
Yamins, D.~L. and DiCarlo, J.~J.
\newblock Using goal-driven deep learning models to understand sensory cortex.
\newblock \emph{Nature Neuroscience}, 2016.

\bibitem[Yamins et~al.(2014)Yamins, Hong, Cadieu, Solomon, Seibert, and
  DiCarlo]{yamins2014annsbrains}
Yamins, D.~L., Hong, H., Cadieu, C.~F., Solomon, E.~A., Seibert, D., and
  DiCarlo, J.~J.
\newblock Performance-optimized hierarchical models predict neural responses in
  higher visual cortex.
\newblock \emph{Proceedings of the National Academy of Sciences (PNAS)}, 2014.

\bibitem[Zagoruyko \& Komodakis(2016)Zagoruyko and Komodakis]{zagoruyko2016wrn}
Zagoruyko, S. and Komodakis, N.
\newblock Wide residual networks.
\newblock In \emph{British Machine Vision Conference (BMVC)}, 2016.

\end{thebibliography}


\begin{thebibliography}{148}
\providecommand{\natexlab}[1]{#1}
\providecommand{\url}[1]{\texttt{#1}}
\expandafter\ifx\csname urlstyle\endcsname\relax
  \providecommand{\doi}[1]{doi: #1}\else
  \providecommand{\doi}{doi: \begingroup \urlstyle{rm}\Url}\fi

\bibitem[Abadi et~al.(2015)Abadi, Agarwal, Barham, Brevdo, Chen, Citro,
  Corrado, Davis, Dean, Devin, Ghemawat, Goodfellow, Harp, Irving, Isard, Jia,
  Jozefowicz, Kaiser, Kudlur, Levenberg, Man\'{e}, Monga, Moore, Murray, Olah,
  Schuster, Shlens, Steiner, Sutskever, Talwar, Tucker, Vanhoucke, Vasudevan,
  Vi\'{e}gas, Vinyals, Warden, Wattenberg, Wicke, Yu, and
  Zheng]{tensorflow2015}
Abadi, M. et~al.
\newblock {TensorFlow}: Large-scale machine learning on heterogeneous systems,
  2015.
\newblock Software available from tensorflow.org.

\bibitem[Abu-Mostafa(1990)]{abumostafa1990hints}
Abu-Mostafa, Y.~S.
\newblock Learning from hints in neural networks.
\newblock \emph{Journal of Complexity}, 1990.

\bibitem[Abu-Mostafa et~al.(2012)Abu-Mostafa, Magdon-Ismail, and
  Lin]{abu2012learningfromdata}
Abu-Mostafa, Y.~S., Magdon-Ismail, M., and Lin, H.-T.
\newblock \emph{Learning from data}.
\newblock AMLBooks, 2012.

\bibitem[Achille \& Soatto(2018)Achille and Soatto]{achille2018emergence}
Achille, A. and Soatto, S.
\newblock Emergence of invariance and disentanglement in deep representations.
\newblock \emph{Journal of Machine Learning Research (JMLR)}, 2018.

\bibitem[Aljundi(2019)]{aljundi2019continuallearning}
Aljundi, R.
\newblock \emph{Continual Learning in Neural Networks}.
\newblock PhD thesis, KU Leuven, Faculty of Engineering Science, 2019.

\bibitem[Atkinson(2002)]{atkinson2002developmental}
Atkinson, J.
\newblock \emph{The developing visual brain}.
\newblock Oxford Scholarship Online, 2002.

\bibitem[Audebert et~al.(2019)Audebert, Le~Saux, and
  Lef{\`e}vre]{audebert2019multispectral}
Audebert, N., Le~Saux, B., and Lef{\`e}vre, S.
\newblock Deep learning for classification of hyperspectral data: A comparative
  review.
\newblock \emph{IEEE Geoscience and Remote Sensing Magazine}, 2019.

\bibitem[Azzopardi \& Cowey(1993)Azzopardi and Cowey]{azzopardi1993fovea}
Azzopardi, P. and Cowey, A.
\newblock Preferential representation of the fovea in the primary visual
  cortex.
\newblock \emph{Nature}, 1993.

\bibitem[Baydin et~al.(2017)Baydin, Pearlmutter, Radul, and
  Siskind]{baydin2017automaticdifferentiation}
Baydin, A.~G., Pearlmutter, B.~A., Radul, A.~A., and Siskind, J.~M.
\newblock Automatic differentiation in machine learning: a survey.
\newblock \emph{Journal of Machine Learning Research (JMLR)}, 2017.

\bibitem[Becker(1999)]{becker1999temporalstability}
Becker, S.
\newblock Implicit learning in {3D} object recognition: The importance of
  temporal context.
\newblock \emph{Neural Computation}, 1999.

\bibitem[Belkin et~al.(2019)Belkin, Hsu, Ma, and
  Mandal]{belkin2019biasvariance}
Belkin, M., Hsu, D., Ma, S., and Mandal, S.
\newblock Reconciling modern machine-learning practice and the classical
  bias--variance trade-off.
\newblock \emph{Proceedings of the National Academy of Sciences (PNAS)}, 2019.

\bibitem[Bengio et~al.(2015)Bengio, Lee, Bornschein, Mesnard, and
  Lin]{bengio2015dlandneuroscience}
Bengio, Y., Lee, D.-H., Bornschein, J., Mesnard, T., and Lin, Z.
\newblock Towards biologically plausible deep learning.
\newblock \emph{arXiv preprint arXiv:1502.04156}, 2015.

\bibitem[Betz et~al.(2010)Betz, Kietzmann, Wilming, and
  K{\"o}nig]{betz2010topdown}
Betz, T., Kietzmann, T.~C., Wilming, N., and K{\"o}nig, P.
\newblock Investigating task-dependent top-down effects on overt visual
  attention.
\newblock \emph{Journal of Vision}, 2010.

\bibitem[Booth \& Rolls(1998)Booth and Rolls]{booth1998invariantitmacaque}
Booth, M. and Rolls, E.~T.
\newblock View-invariant representations of familiar objects by neurons in the
  inferior temporal visual cortex.
\newblock \emph{Cerebral Cortex}, 1998.

\bibitem[Bornstein \& Arterberry(2010)Bornstein and
  Arterberry]{bornstein2010hierarchychildren}
Bornstein, M.~H. and Arterberry, M.~E.
\newblock The development of object categorization in young children:
  Hierarchical inclusiveness, age, perceptual attribute, and group versus
  individual analyses.
\newblock \emph{Developmental Psychology}, 2010.

\bibitem[Bowers(2017)]{bowers2017pdp}
Bowers, J.~S.
\newblock Parallel distributed processing theory in the age of deep networks.
\newblock \emph{Trends in Cognitive Sciences}, 2017.

\bibitem[B{\"u}lthoff \& Newell(2006)B{\"u}lthoff and
  Newell]{bulthoff2006viewpointdependence}
B{\"u}lthoff, I. and Newell, F.~N.
\newblock The role of familiarity in the recognition of static and dynamic
  objects.
\newblock \emph{Progress in Brain Research}, 2006.

\bibitem[Cao et~al.(2018)Cao, Shen, Xie, Parkhi, and
  Zisserman]{cao2018vggface2}
Cao, Q., Shen, L., Xie, W., Parkhi, O.~M., and Zisserman, A.
\newblock {VGGFace2}: A dataset for recognising faces across pose and age.
\newblock In \emph{IEEE International Conference on Automatic Face \& Gesture
  Recognition}. 2018.

\bibitem[Cohen et~al.(2020)Cohen, Botch, and Robertson]{cohen2020colour}
Cohen, M.~A., Botch, T.~L., and Robertson, C.~E.
\newblock The limits of color awareness during active, real-world vision.
\newblock \emph{Proceedings of the National Academy of Sciences (PNAS)}, 2020.

\bibitem[Cohen \& Welling(2016)Cohen and Welling]{cohen2016groupequivcnns}
Cohen, T. and Welling, M.
\newblock Group equivariant convolutional networks.
\newblock In \emph{International Conference on Machine Learning (ICML)}, 2016.

\bibitem[Connor et~al.(2004)Connor, Egeth, and
  Yantis]{connor2004buttomuptopdown}
Connor, C.~E., Egeth, H.~E., and Yantis, S.
\newblock Visual attention: bottom-up versus top-down.
\newblock \emph{Current Biology}, 2004.

\bibitem[Cortes \& Vapnik(1995)Cortes and Vapnik]{cortes1995svm}
Cortes, C. and Vapnik, V.
\newblock Support-vector networks.
\newblock \emph{Machine Learning}, 1995.

\bibitem[Dalal \& Triggs(2005)Dalal and Triggs]{dalal2005hog}
Dalal, N. and Triggs, B.
\newblock Histograms of oriented gradients for human detection.
\newblock In \emph{IEEE Conference on Computer Vision and Pattern Recognition
  (CVPR)}. 2005.

\bibitem[Desimone \& Duncan(1995)Desimone and
  Duncan]{desimone1995visualattention}
Desimone, R. and Duncan, J.
\newblock Neural mechanisms of selective visual attention.
\newblock \emph{Annual Review of Neuroscience}, 1995.

\bibitem[Desimone et~al.(1984)Desimone, Albright, Gross, and
  Bruce]{desimone1984invariantitmacaque}
Desimone, R., Albright, T.~D., Gross, C.~G., and Bruce, C.
\newblock Stimulus-selective properties of inferior temporal neurons in the
  macaque.
\newblock \emph{Journal of Neuroscience}, 1984.

\bibitem[DeVries \& Taylor(2017)DeVries and Taylor]{devries2017daugfeatspace}
DeVries, T. and Taylor, G.~W.
\newblock Dataset augmentation in feature space.
\newblock In \emph{International Conference on Learning Representations (ICLR),
  arXiv:1702.05538}, 2017.

\bibitem[DiCarlo \& Cox(2007)DiCarlo and Cox]{dicarlo2007untangling}
DiCarlo, J.~J. and Cox, D.~D.
\newblock Untangling invariant object recognition.
\newblock \emph{Trends in Cognitive Sciences}, 2007.

\bibitem[Douglas et~al.(1989)Douglas, Martin, and
  Whitteridge]{douglas1989neocortex}
Douglas, R.~J., Martin, K.~A., and Whitteridge, D.
\newblock A canonical microcircuit for neocortex.
\newblock \emph{Neural Computation}, 1989.

\bibitem[Duchi et~al.(2011)Duchi, Hazan, and Singer]{duchi2011adagrad}
Duchi, J., Hazan, E., and Singer, Y.
\newblock Adaptive subgradient methods for online learning and stochastic
  optimization.
\newblock \emph{Journal of Machine Learning Research (JMLR)}, 2011.

\bibitem[Duda \& Hart(1972)Duda and Hart]{duda1972hough}
Duda, R.~O. and Hart, P.~E.
\newblock Use of the {Hough} transformation to detect lines and curves in
  pictures.
\newblock \emph{Communications of the ACM}, 1972.

\bibitem[Dujmovi{\'c} et~al.(2020)Dujmovi{\'c}, Malhotra, and
  Bowers]{dujmovic2020adversarial}
Dujmovi{\'c}, M., Malhotra, G., and Bowers, J.
\newblock What do adversarial images tell us about human vision?
\newblock \emph{bioRxiv preprint 2020.02.25.964361}, 2020.

\bibitem[Dumoulin \& Wandell(2008)Dumoulin and Wandell]{dumoulin2008prf}
Dumoulin, S.~O. and Wandell, B.~A.
\newblock Population receptive field estimates in human visual cortex.
\newblock \emph{Neuroimage}, 2008.

\bibitem[Edelman \& B{\"u}lthoff(1992)Edelman and
  B{\"u}lthoff]{edelman1992viewpointdependence}
Edelman, S. and B{\"u}lthoff, H.~H.
\newblock Orientation dependence in the recognition of familiar and novel views
  of three-dimensional objects.
\newblock \emph{Vision Research}, 1992.

\bibitem[Einh{\"a}user et~al.(2005)Einh{\"a}user, Hipp, Eggert, K{\"o}rner, and
  K{\"o}nig]{einhauser2005viewpointinvariance}
Einh{\"a}user, W., Hipp, J., Eggert, J., K{\"o}rner, E., and K{\"o}nig, P.
\newblock Learning viewpoint invariant object representations using a temporal
  coherence principle.
\newblock \emph{Biological Cybernetics}, 2005.

\bibitem[Fukushima \& Miyake(1982)Fukushima and
  Miyake]{fukushima1982neocognitron}
Fukushima, K. and Miyake, S.
\newblock Neocognitron: A new algorithm for pattern recognition tolerant of
  deformations and shifts in position.
\newblock \emph{Pattern Recognition}, 1982.

\bibitem[Gauthier et~al.(1999)Gauthier, Tarr, Anderson, Skudlarski, and
  Gore]{gauthier1999greebles}
Gauthier, I., Tarr, M.~J., Anderson, A.~W., Skudlarski, P., and Gore, J.~C.
\newblock Activation of the middle fusiform'face area'increases with expertise
  in recognizing novel objects.
\newblock \emph{Nature Neuroscience}, 1999.

\bibitem[Geirhos et~al.(2020)Geirhos, Jacobsen, Michaelis, Zemel, Brendel,
  Bethge, and Wichmann]{geirhos2020shortcutlearning}
Geirhos, R., Jacobsen, J.-H., Michaelis, C., Zemel, R., Brendel, W., Bethge,
  M., and Wichmann, F.~A.
\newblock Shortcut learning in deep neural networks.
\newblock \emph{arXiv preprint arXiv:2004.07780}, 2020.

\bibitem[Gelman \& Meyer(2011)Gelman and Meyer]{gelman2011childcategorization}
Gelman, S.~A. and Meyer, M.
\newblock Child categorization.
\newblock \emph{Wiley Interdisciplinary Reviews: Cognitive Science}, 2011.

\bibitem[Gencoglu et~al.(2019)Gencoglu, van Gils, Guldogan, Morikawa,
  S{\"u}zen, Gruber, Leinonen, and Huttunen]{gencoglu2019hark}
Gencoglu, O., van Gils, M., Guldogan, E., Morikawa, C., S{\"u}zen, M., Gruber,
  M., Leinonen, J., and Huttunen, H.
\newblock {HARK} side of deep learning--from grad student descent to automated
  machine learning.
\newblock \emph{arXiv preprint arXiv:1904.07633}, 2019.

\bibitem[Glorot \& Bengio(2010)Glorot and Bengio]{glorot2010glorot}
Glorot, X. and Bengio, Y.
\newblock Understanding the difficulty of training deep feedforward neural
  networks.
\newblock In \emph{International Conference on Artificial Intelligence and
  Statistics (AISTATS)}, 2010.

\bibitem[Glorot et~al.(2011)Glorot, Bordes, and Bengio]{glorot2011relu}
Glorot, X., Bordes, A., and Bengio, Y.
\newblock Deep sparse rectifier neural networks.
\newblock In \emph{International Conference on Artificial Intelligence and
  Statistics (AISTATS)}, 2011.

\bibitem[Gonzalez \& Woods(2018)Gonzalez and
  Woods]{gonzalez2018imageprocessing}
Gonzalez, R. and Woods, R.
\newblock \emph{Digital image processing}.
\newblock Pearson, 4th edition, 2018.

\bibitem[Goodfellow et~al.(2016)Goodfellow, Bengio, and
  Courville]{goodfellow2016dlbook}
Goodfellow, I., Bengio, Y., and Courville, A.
\newblock \emph{Deep learning}.
\newblock MIT press, 2016.

\bibitem[Goodfellow et~al.(2013)Goodfellow, Warde{-}Farley, Mirza, Courville,
  and Bengio]{goodfellow2013maxout}
Goodfellow, I.~J., Warde{-}Farley, D., Mirza, M., Courville, A.~C., and Bengio,
  Y.
\newblock Maxout networks.
\newblock In \emph{International Conference on Machine Learning (ICML)}, 2013.

\bibitem[Graham(2014)]{graham2014fracmaxpool}
Graham, B.
\newblock Fractional max-pooling.
\newblock \emph{arXiv preprint arXiv:1412.6071}, 2014.

\bibitem[Gross et~al.(1972)Gross, Rocha-Miranda, and Bender]{gross1972it}
Gross, C.~G., Rocha-Miranda, C.~d., and Bender, D.
\newblock Visual properties of neurons in inferotemporal cortex of the macaque.
\newblock \emph{Journal of Neurophysiology}, 1972.

\bibitem[G{\"u}{\c{c}}l{\"u} \& van Gerven(2015)G{\"u}{\c{c}}l{\"u} and van
  Gerven]{gucclu2015annbrains}
G{\"u}{\c{c}}l{\"u}, U. and van Gerven, M.~A.
\newblock Deep neural networks reveal a gradient in the complexity of neural
  representations across the ventral stream.
\newblock \emph{Journal of Neuroscience}, 2015.

\bibitem[Harris \& Shepherd(2015)Harris and Shepherd]{harris2015neocortex}
Harris, K.~D. and Shepherd, G.~M.
\newblock The neocortical circuit: themes and variations.
\newblock \emph{Nature Neuroscience}, 2015.

\bibitem[Harwerth et~al.(1986)Harwerth, Smith, Duncan, Crawford, and
  Von~Noorden]{harwerth1986criticalperiods}
Harwerth, R.~S., Smith, E.~L., Duncan, G.~C., Crawford, M., and Von~Noorden,
  G.~K.
\newblock Multiple sensitive periods in the development of the primate visual
  system.
\newblock \emph{Science}, 1986.

\bibitem[Hassabis et~al.(2017)Hassabis, Kumaran, Summerfield, and
  Botvinick]{hassabis2017aiandneuroscience}
Hassabis, D., Kumaran, D., Summerfield, C., and Botvinick, M.
\newblock Neuroscience-inspired artificial intelligence.
\newblock \emph{Neuron}, 2017.

\bibitem[Hasson et~al.(2020)Hasson, Nastase, and
  Goldstein]{hasson2020directfit}
Hasson, U., Nastase, S.~A., and Goldstein, A.
\newblock Direct fit to nature: An evolutionary perspective on biological and
  artificial neural networks.
\newblock \emph{Neuron}, 2020.

\bibitem[Haugeland(1989)]{haugeland1989ai}
Haugeland, J.
\newblock \emph{Artificial intelligence: The very idea}.
\newblock MIT Press, 1989.

\bibitem[He et~al.(2015)He, Zhang, Ren, and Sun]{he2015he}
He, K., Zhang, X., Ren, S., and Sun, J.
\newblock Delving deep into rectifiers: Surpassing human-level performance on
  {ImageNet} classification.
\newblock In \emph{IEEE International Conference on Computer Vision (ICCV)},
  2015.

\bibitem[He et~al.(2016)He, Zhang, Ren, and Sun]{he2016resnet}
He, K., Zhang, X., Ren, S., and Sun, J.
\newblock Deep residual learning for image recognition.
\newblock In \emph{IEEE Conference on Computer Vision and Pattern Recognition
  (CVPR)}, 2016.

\bibitem[Hebb(1949)]{hebb1949biologicallearning}
Hebb, D.~O.
\newblock \emph{The organization of behavior: a neuropsychological theory}.
\newblock Wiley, 1949.

\bibitem[Hern{\'a}ndez-Garc{\'i}a(2014)]{hernandez2014bscthesis}
Hern{\'a}ndez-Garc{\'i}a, A.
\newblock Aesthetics assessment of videos through visual descriptors and
  automatic polarity annotation.
\newblock {B.S.} thesis, Universidad Carlos III de Madrid, 2014.

\bibitem[Hern{\'a}ndez-Garc{\'i}a \&
  K{\"o}nig(2018{\natexlab{a}})Hern{\'a}ndez-Garc{\'i}a and
  K{\"o}nig]{hergar2018daugadvantages}
Hern{\'a}ndez-Garc{\'i}a, A. and K{\"o}nig, P.
\newblock Further advantages of data augmentation on convolutional neural
  networks.
\newblock In \emph{International Conference on Artificial Neural Networks
  (ICANN)}. 2018{\natexlab{a}}.

\bibitem[Hern{\'a}ndez-Garc{\'i}a \&
  K{\"o}nig(2018{\natexlab{b}})Hern{\'a}ndez-Garc{\'i}a and
  K{\"o}nig]{hergar2018daugreg}
Hern{\'a}ndez-Garc{\'i}a, A. and K{\"o}nig, P.
\newblock Data augmentation instead of explicit regularization.
\newblock \emph{arXiv preprint arXiv:1806.03852}, 2018{\natexlab{b}}.

\bibitem[Hern{\'a}ndez-Garc{\'i}a \&
  K{\"o}nig(2018{\natexlab{c}})Hern{\'a}ndez-Garc{\'i}a and
  K{\"o}nig]{hergar2018wddropout}
Hern{\'a}ndez-Garc{\'i}a, A. and K{\"o}nig, P.
\newblock Do deep nets really need weight decay and dropout?
\newblock \emph{arXiv preprint arXiv:1802.07042}, 2018{\natexlab{c}}.

\bibitem[Hern{\'a}ndez-Garc{\'i}a et~al.(2017)Hern{\'a}ndez-Garc{\'i}a,
  Fern\'{a}ndez-Mart\'{\i}nez, and D\'{\i}az-de
  Mar\'{\i}a]{hernandez2017mscthesis}
Hern{\'a}ndez-Garc{\'i}a, A., Fern\'{a}ndez-Mart\'{\i}nez, F., and D\'{\i}az-de
  Mar\'{\i}a, F.
\newblock Emotion and attention: Predicting electrodermal activity through
  video visual descriptors.
\newblock In \emph{International Conference on Web Intelligence}, 2017.

\bibitem[Hern{\'a}ndez-Garc{\'i}a et~al.(2018)Hern{\'a}ndez-Garc{\'i}a, Mehrer,
  Kriegeskorte, K{\"o}nig, and Kietzmann]{hergar2018daugit}
Hern{\'a}ndez-Garc{\'i}a, A., Mehrer, J., Kriegeskorte, N., K{\"o}nig, P., and
  Kietzmann, T.~C.
\newblock Deep neural networks trained with heavier data augmentation learn
  features closer to representations in {hIT}.
\newblock In \emph{Conference on Cognitive Computational Neuroscience (CCN)},
  2018.

\bibitem[Hern{\'a}ndez-Garc{\'i}a
  et~al.(2019{\natexlab{a}})Hern{\'a}ndez-Garc{\'i}a, Gameiro-Ramos, Grillini,
  and K{\"o}nig]{hergar2019globsal}
Hern{\'a}ndez-Garc{\'i}a, A., Gameiro-Ramos, R., Grillini, A., and K{\"o}nig,
  P.
\newblock Global visual salience of competing stimuli.
\newblock \emph{PsyArXiv preprint PsyArXiv:z7qp5}, 2019{\natexlab{a}}.

\bibitem[Hern{\'a}ndez-Garc{\'i}a
  et~al.(2019{\natexlab{b}})Hern{\'a}ndez-Garc{\'i}a, K{\"o}nig, and
  Kietzmann]{hergar2019dauginv}
Hern{\'a}ndez-Garc{\'i}a, A., K{\"o}nig, P., and Kietzmann, T.
\newblock Learning robust visual representations using data augmentation
  invariance.
\newblock \emph{arXiv preprint arXiv:1906.04547}, 2019{\natexlab{b}}.

\bibitem[Hern{\'a}ndez-Garc{\'i}a
  et~al.(2019{\natexlab{c}})Hern{\'a}ndez-Garc{\'i}a, Zuiderbaan, Edadan,
  Dumoulin, and K{\"o}nig]{hergar2019imageid}
Hern{\'a}ndez-Garc{\'i}a, A., Zuiderbaan, W., Edadan, A., Dumoulin, S.~O., and
  K{\"o}nig, P.
\newblock Saliency and the population receptive field model to identify images
  from brain activity.
\newblock In \emph{Annual Meeting of the Vision Sciences Society (VSS)}.
  2019{\natexlab{c}}.

\bibitem[Hochreiter \& Schmidhuber(1997)Hochreiter and
  Schmidhuber]{hochreiter1997lstm}
Hochreiter, S. and Schmidhuber, J.
\newblock Long short-term memory.
\newblock \emph{Neural Computation}, 1997.

\bibitem[Holm et~al.(2008)Holm, Eriksson, and
  Andersson]{holm2008attentionprecedes}
Holm, L., Eriksson, J., and Andersson, L.
\newblock Looking as if you know: Systematic object inspection precedes object
  recognition.
\newblock \emph{Journal of Vision}, 2008.

\bibitem[Hornik(1991)]{hornik1991functionapproximation}
Hornik, K.
\newblock Approximation capabilities of multilayer feedforward networks.
\newblock \emph{Neural Networks}, 1991.

\bibitem[Huang et~al.(2017)Huang, Liu, Van Der~Maaten, and
  Weinberger]{huang2017densenet}
Huang, G., Liu, Z., Van Der~Maaten, L., and Weinberger, K.~Q.
\newblock Densely connected convolutional networks.
\newblock In \emph{IEEE Conference on Computer Vision and Pattern Recognition
  (CVPR)}. 2017.

\bibitem[Hubel \& Wiesel(1959)Hubel and Wiesel]{hubelwiesel1959}
Hubel, D.~H. and Wiesel, T.~N.
\newblock Receptive fields of single neurones in the cat's striate cortex.
\newblock \emph{The Journal of Physiology}, 1959.

\bibitem[Hubel \& Wiesel(1962)Hubel and Wiesel]{hubelwiesel1962}
Hubel, D.~H. and Wiesel, T.~N.
\newblock Receptive fields, binocular interaction and functional architecture
  in the cat's visual cortex.
\newblock \emph{The Journal of Physiology}, 1962.

\bibitem[Ioffe \& Szegedy(2015)Ioffe and Szegedy]{ioffe2015batchnorm}
Ioffe, S. and Szegedy, C.
\newblock Batch normalization: Accelerating deep network training by reducing
  internal covariate shift.
\newblock In \emph{International Conference on Machine Learning (ICML)}, 2015.

\bibitem[Itti et~al.(1998)Itti, Koch, and Niebur]{itti1998salience}
Itti, L., Koch, C., and Niebur, E.
\newblock A model of saliency-based visual attention for rapid scene analysis.
\newblock \emph{IEEE Transactions on Pattern Analysis and Machine Intelligence
  (TPAMI)}, 1998.

\bibitem[Ivanov(1976)]{ivanov1976regularisation}
Ivanov, V.~V.
\newblock \emph{The theory of approximate methods and their applications to the
  numerical solution of singular integral equations}.
\newblock Nordhof International, 1976.

\bibitem[Jing \& Tian(2020)Jing and Tian]{jing2020selfsupervised}
Jing, L. and Tian, Y.
\newblock Self-supervised visual feature learning with deep neural networks: A
  survey.
\newblock \emph{IEEE Transactions on Pattern Analysis and Machine Intelligence
  (TPAMI)}, 2020.

\bibitem[Khaligh-Razavi \& Kriegeskorte(2014)Khaligh-Razavi and
  Kriegeskorte]{khaligh2014annbrains}
Khaligh-Razavi, S.-M. and Kriegeskorte, N.
\newblock Deep supervised, but not unsupervised, models may explain it cortical
  representation.
\newblock \emph{PLOS Computational Biology}, 2014.

\bibitem[Kietzmann et~al.(2011)Kietzmann, Geuter, and
  K{\"o}nig]{kietzmann2011attentionprecedes}
Kietzmann, T.~C., Geuter, S., and K{\"o}nig, P.
\newblock Overt visual attention as a causal factor of perceptual awareness.
\newblock \emph{PLOS ONE}, 2011.

\bibitem[Kietzmann et~al.(2019)Kietzmann, McClure, and
  Kriegeskorte]{kietzmann2019dnncompneuro}
Kietzmann, T.~C., McClure, P., and Kriegeskorte, N.
\newblock Deep neural networks in computational neuroscience.
\newblock \emph{Oxford Research Encyclopedia of Neuroscience}, 2019.

\bibitem[Kingma \& Ba(2014)Kingma and Ba]{kingma2014adam}
Kingma, D.~P. and Ba, J.
\newblock Adam: A method for stochastic optimization.
\newblock \emph{arXiv preprint arXiv:1412.6980}, 2014.

\bibitem[Kollmorgen et~al.(2010)Kollmorgen, Nortmann, Schr{\"{o}}der, and
  K{\"{o}}nig]{kollmorgen2010topdownbottomup}
Kollmorgen, S., Nortmann, N., Schr{\"{o}}der, S., and K{\"{o}}nig, P.
\newblock {Influence of low-level stimulus features, task dependent factors,
  and spatial biases on overt visual attention}.
\newblock \emph{PLOS Computational Biology}, may 2010.

\bibitem[Kording et~al.(2004)Kording, Kayser, Einhauser, and
  Konig]{kording2004complexcells}
Kording, K.~P., Kayser, C., Einhauser, W., and Konig, P.
\newblock How are complex cell properties adapted to the statistics of natural
  stimuli?
\newblock \emph{Journal of Neurophysiology}, 2004.

\bibitem[Krizhevsky \& Hinton(2009)Krizhevsky and Hinton]{krizhevsky2009cifar}
Krizhevsky, A. and Hinton, G.
\newblock Learning multiple layers of features from tiny images.
\newblock \emph{Technical report, University of Toronto}, 2009.

\bibitem[Krizhevsky et~al.(2012)Krizhevsky, Sutskever, and
  Hinton]{krizhevsky2012alexnet}
Krizhevsky, A., Sutskever, I., and Hinton, G.~E.
\newblock {ImageNet} classification with deep convolutional neural networks.
\newblock In \emph{Advances in Neural Information Processing Systems
  (NeurIPS)}, 2012.

\bibitem[K{\"u}mmerer et~al.(2017)K{\"u}mmerer, Wallis, Gatys, and
  Bethge]{kuemmerer2017icfdeepgaze}
K{\"u}mmerer, M., Wallis, T.~S., Gatys, L.~A., and Bethge, M.
\newblock Understanding low-and high-level contributions to fixation
  prediction.
\newblock In \emph{IEEE International Conference on Computer Vision (ICCV)},
  2017.

\bibitem[K{\"u}mmerer et~al.(2018)K{\"u}mmerer, Wallis, and
  Bethge]{kuemmerer2018salience}
K{\"u}mmerer, M., Wallis, T.~S., and Bethge, M.
\newblock Saliency benchmarking made easy: Separating models, maps and metrics.
\newblock In \emph{European Conference on Computer Vision (ECCV)}, 2018.

\bibitem[Larsson et~al.(2016)Larsson, Maire, and
  Shakhnarovich]{larsson2016fractalnet}
Larsson, G., Maire, M., and Shakhnarovich, G.
\newblock {FractalNet}: Ultra-deep neural networks without residuals.
\newblock \emph{arXiv preprint arXiv:1605.07648}, 2016.

\bibitem[Latif et~al.(2019)Latif, Rasheed, Sajid, Ahmed, Ali, Ratyal, Zafar,
  Dar, Sajid, and Khalil]{latif2019imageretrieval}
Latif, A., Rasheed, A., Sajid, U., Ahmed, J., Ali, N., Ratyal, N.~I., Zafar,
  B., Dar, S.~H., Sajid, M., and Khalil, T.
\newblock Content-based image retrieval and feature extraction: a comprehensive
  review.
\newblock \emph{Mathematical Problems in Engineering}, 2019.

\bibitem[LeCun et~al.(1998)LeCun, Bottou, Bengio, and Haffner]{lecun1998mnist}
LeCun, Y., Bottou, L., Bengio, Y., and Haffner, P.
\newblock Gradient-based learning applied to document recognition.
\newblock \emph{Proceedings of the IEEE}, 1998.

\bibitem[LeCun et~al.(2015)LeCun, Bengio, and
  Hinton]{lecun2015deeplearningnature}
LeCun, Y., Bengio, Y., and Hinton, G.
\newblock Deep learning.
\newblock \emph{Nature}, 2015.

\bibitem[Lienhart \& Maydt(2002)Lienhart and Maydt]{lienhart2002haar}
Lienhart, R. and Maydt, J.
\newblock An extended set of {Haar}-like features for rapid object detection.
\newblock In \emph{International Conference on Image Processing (ICIP)}. 2002.

\bibitem[Lindsay(2020)]{lindsay2020dlandneuroscience}
Lindsay, G.
\newblock Convolutional neural networks as a model of the visual system: past,
  present, and future.
\newblock \emph{Journal of Cognitive Neuroscience}, 2020.

\bibitem[Lindsay \& Miller(2018)Lindsay and Miller]{lindsay2018bioconstraints}
Lindsay, G.~W. and Miller, K.~D.
\newblock How biological attention mechanisms improve task performance in a
  large-scale visual system model.
\newblock \emph{{eLife}}, 2018.

\bibitem[Lindsey et~al.(2019)Lindsey, Ocko, Ganguli, and
  Deny]{lindsey2019bioconstraints}
Lindsey, J., Ocko, S.~A., Ganguli, S., and Deny, S.
\newblock A unified theory of early visual representations from retina to
  cortex through anatomically constrained deep {CNNs}.
\newblock In \emph{International Conference on Learning Representations (ICLR),
  arXiv:1901.00945}, 2019.

\bibitem[Locatello et~al.(2018)Locatello, Bauer, Lucic, R{\"a}tsch, Gelly,
  Sch{\"o}lkopf, and Bachem]{locatello2018disentanglement}
Locatello, F., Bauer, S., Lucic, M., R{\"a}tsch, G., Gelly, S., Sch{\"o}lkopf,
  B., and Bachem, O.
\newblock Challenging common assumptions in the unsupervised learning of
  disentangled representations.
\newblock \emph{arXiv preprint arXiv:1811.12359}, 2018.

\bibitem[Logothetis \& Sheinberg(1996)Logothetis and
  Sheinberg]{logothetis1996objectrecognition}
Logothetis, N.~K. and Sheinberg, D.~L.
\newblock Visual object recognition.
\newblock \emph{Annual Review of Neuroscience}, 1996.

\bibitem[Lowe(2004)]{lowe2004sift}
Lowe, D.~G.
\newblock Distinctive image features from scale-invariant keypoints.
\newblock \emph{International Journal of Computer Vision (IJCV)}, 2004.

\bibitem[Malhotra et~al.(2020)Malhotra, Evans, and
  Bowers]{malhotra2020bioconstraints}
Malhotra, G., Evans, B.~D., and Bowers, J.~S.
\newblock Hiding a plane with a pixel: examining shape-bias in {CNNs} and the
  benefit of building in biological constraints.
\newblock \emph{Vision Research}, 2020.

\bibitem[Marblestone et~al.(2016)Marblestone, Wayne, and
  Kording]{marblestone2016dlandneuroscience}
Marblestone, A.~H., Wayne, G., and Kording, K.~P.
\newblock Toward an integration of deep learning and neuroscience.
\newblock \emph{Frontiers in Computational Neuroscience}, 2016.

\bibitem[Marcus(2018)]{marcus2018critiquedl}
Marcus, G.
\newblock Deep learning: A critical appraisal.
\newblock \emph{arXiv preprint arXiv:1801.00631}, 2018.

\bibitem[McCulloch \& Pitts(1943)McCulloch and
  Pitts]{mcculloch1943biologicallearning}
McCulloch, W.~S. and Pitts, W.
\newblock A logical calculus of the ideas immanent in nervous activity.
\newblock \emph{The Bulletin of Mathematical Biophysics}, 1943.

\bibitem[Milivojevic(2012)]{milivojevic2012viewpointdependence}
Milivojevic, B.
\newblock Object recognition can be viewpoint dependent or invariant--it's just
  a matter of time and task.
\newblock \emph{Frontiers in Computational Neuroscience}, 2012.

\bibitem[Morgenstern et~al.(2019)Morgenstern, Schmidt, and
  Fleming]{morgenstern2019oneshot}
Morgenstern, Y., Schmidt, F., and Fleming, R.~W.
\newblock One-shot categorization of novel object classes in humans.
\newblock \emph{Vision Research}, 2019.

\bibitem[Mundt et~al.(2019)Mundt, Majumder, Pliushch, and
  Ramesh]{mundt2019continuallearning}
Mundt, M., Majumder, S., Pliushch, I., and Ramesh, V.
\newblock Unified probabilistic deep continual learning through generative
  replay and open set recognition.
\newblock \emph{arXiv preprint arXiv:1905.12019}, 2019.

\bibitem[Munoz \& Everling(2004)Munoz and Everling]{munoz2004bottomup}
Munoz, D.~P. and Everling, S.
\newblock Look away: the anti-saccade task and the voluntary control of eye
  movement.
\newblock \emph{Nature Reviews Neuroscience}, 2004.

\bibitem[Murphy(2012)]{murphy2012machinelearning}
Murphy, K.~P.
\newblock \emph{Machine learning: a probabilistic perspective}.
\newblock MIT Press, 2012.

\bibitem[Nayebi \& Ganguli(2017)Nayebi and Ganguli]{nayebi2017bioconstraints}
Nayebi, A. and Ganguli, S.
\newblock Biologically inspired protection of deep networks from adversarial
  attacks.
\newblock \emph{arXiv preprint arXiv:1703.09202}, 2017.

\bibitem[Paszke et~al.(2019)Paszke, Gross, Massa, Lerer, Bradbury, Chanan,
  Killeen, Lin, Gimelshein, Antiga, Desmaison, Kopf, Yang, DeVito, Raison,
  Tejani, Chilamkurthy, Steiner, Fang, Bai, and Chintala]{pytorch2019}
Paszke, A. et~al.
\newblock {PyTorch}: An imperative style, high-performance deep learning
  library.
\newblock In \emph{Advances in Neural Information Processing Systems
  (NeurIPS)}. 2019.

\bibitem[Perronnin \& Dance(2007)Perronnin and Dance]{perronnin2007fisher}
Perronnin, F. and Dance, C.
\newblock Fisher kernels on visual vocabularies for image categorization.
\newblock In \emph{IEEE Conference on Computer Vision and Pattern Recognition
  (CVPR)}. 2007.

\bibitem[Phillips(1962)]{phillips1962regularisation}
Phillips, D.~L.
\newblock A technique for the numerical solution of certain integral equations
  of the first kind.
\newblock \emph{Association for Computing Machinery}, 1962.

\bibitem[Pozzi et~al.(2018)Pozzi, Boht{\'e}, and
  Roelfsema]{pozzi2018biologicallyplausible}
Pozzi, I., Boht{\'e}, S., and Roelfsema, P.
\newblock A biologically plausible learning rule for deep learning in the
  brain.
\newblock \emph{arXiv preprint arXiv:1811.01768}, 2018.

\bibitem[Quiroga et~al.(2005)Quiroga, Reddy, Kreiman, Koch, and
  Fried]{quiroga2005invariantithuman}
Quiroga, R.~Q., Reddy, L., Kreiman, G., Koch, C., and Fried, I.
\newblock Invariant visual representation by single neurons in the human brain.
\newblock \emph{Nature}, 2005.

\bibitem[Richards et~al.(2019)Richards, Lillicrap, Beaudoin, Bengio, Bogacz,
  Christensen, Clopath, Costa, de~Berker, Ganguli,
  et~al.]{richards2019dlandneuroscience}
Richards, B.~A. et~al.
\newblock A deep learning framework for neuroscience.
\newblock \emph{Nature Neuroscience}, 2019.

\bibitem[Rosenblatt(1958)]{rosenblatt1958perceptron}
Rosenblatt, F.
\newblock The perceptron: a probabilistic model for information storage and
  organization in the brain.
\newblock \emph{Psychological Review}, 1958.

\bibitem[Rumelhart et~al.(1986)Rumelhart, Hinton, and
  Williams]{rumelhart1986backprop}
Rumelhart, D.~E., Hinton, G.~E., and Williams, R.~J.
\newblock Learning representations by back-propagating errors.
\newblock \emph{Nature}, 1986.

\bibitem[Russakovsky et~al.(2015)Russakovsky, Deng, Su, Krause, Satheesh, Ma,
  Huang, Karpathy, Khosla, Bernstein, Berg, and
  Fei-Fei]{russakovsky2015imagenet}
Russakovsky, O. et~al.
\newblock {ImageNet} large scale visual recognition challenge.
\newblock \emph{International Journal of Computer Vision (IJCV)}, 2015.

\bibitem[Saxe et~al.(2020)Saxe, Nelli, and
  Summerfield]{saxe2020dlandneuroscience}
Saxe, A., Nelli, S., and Summerfield, C.
\newblock If deep learning is the answer, then what is the question?
\newblock \emph{arXiv preprint arXiv:2004.07580}, 2020.

\bibitem[Sch{\"u}tt et~al.(2019)Sch{\"u}tt, Rothkegel, Trukenbrod, Engbert, and
  Wichmann]{schutt2019bottomuptopdown}
Sch{\"u}tt, H.~H., Rothkegel, L.~O., Trukenbrod, H.~A., Engbert, R., and
  Wichmann, F.~A.
\newblock Disentangling bottom-up versus top-down and low-level versus
  high-level influences on eye movements over time.
\newblock \emph{Journal of Vision}, 2019.

\bibitem[Shapiro(2012)]{shapiro2019embodied}
Shapiro, L.
\newblock \emph{Embodied cognition}.
\newblock Oxford Handbooks Online, 2012.

\bibitem[Simard et~al.(1992)Simard, Victorri, LeCun, and
  Denker]{simard1992daug}
Simard, P., Victorri, B., LeCun, Y., and Denker, J.
\newblock Tangent prop-a formalism for specifying selected invariances in an
  adaptive network.
\newblock In \emph{Advances in Neural Information Processing Systems
  (NeurIPS)}, 1992.

\bibitem[Simonyan \& Zisserman(2014)Simonyan and Zisserman]{simonyan2014vgg}
Simonyan, K. and Zisserman, A.
\newblock Very deep convolutional networks for large-scale image recognition.
\newblock \emph{arXiv preprint arXiv:1409.1556}, 2014.

\bibitem[Sinz et~al.(2019)Sinz, Pitkow, Reimer, Bethge, and
  Tolias]{sinz2019dlvsbrain}
Sinz, F.~H., Pitkow, X., Reimer, J., Bethge, M., and Tolias, A.~S.
\newblock Engineering a less artificial intelligence.
\newblock \emph{Neuron}, 2019.

\bibitem[Sivic \& Zisserman(2003)Sivic and Zisserman]{sivic2003bog}
Sivic, J. and Zisserman, A.
\newblock Video google: A text retrieval approach to object matching in videos.
\newblock In \emph{IEEE Conference on Computer Vision and Pattern Recognition
  (CVPR)}. 2003.

\bibitem[Springenberg et~al.(2014)Springenberg, Dosovitskiy, Brox, and
  Riedmiller]{springenberg2014allcnn}
Springenberg, J.~T., Dosovitskiy, A., Brox, T., and Riedmiller, M.
\newblock Striving for simplicity: The all convolutional net.
\newblock In \emph{International Conference on Learning Representations (ICLR),
  arXiv:1412.6806}, 2014.

\bibitem[Tacchetti et~al.(2018)Tacchetti, Isik, and
  Poggio]{tacchetti2018invariance}
Tacchetti, A., Isik, L., and Poggio, T.~A.
\newblock Invariant recognition shapes neural representations of visual input.
\newblock \emph{Annual Review of Vision Science}, 2018.

\bibitem[Tarr et~al.(1998)Tarr, Williams, Hayward, and
  Gauthier]{tarr1998viewpointdependence}
Tarr, M.~J., Williams, P., Hayward, W.~G., and Gauthier, I.
\newblock Three-dimensional object recognition is viewpoint dependent.
\newblock \emph{Nature Neuroscience}, 1998.

\bibitem[Taylor et~al.(2011)Taylor, Spiro, Bregler, and
  Fergus]{taylor2011temporalstability}
Taylor, G.~W., Spiro, I., Bregler, C., and Fergus, R.
\newblock Learning invariance through imitation.
\newblock In \emph{IEEE Conference on Computer Vision and Pattern Recognition
  (CVPR)}. 2011.

\bibitem[{Theano Development Team}(2016)]{theano2016}
{Theano Development Team}.
\newblock {Theano: A {Python} framework for fast computation of mathematical
  expressions}.
\newblock \emph{arXiv preprint arXiv:1605.02688}, 2016.

\bibitem[Tikhonov(1963)]{tikhonov1963regularisation}
Tikhonov, A.~N.
\newblock On solving ill-posed problem and method of regularization.
\newblock \emph{Doklady Akademii Nauk USSR}, 1963.

\bibitem[Turing(1968)]{turing1948intelligentmachinery}
Turing, A.~M.
\newblock \emph{Cybernetics; (Key papers)}.
\newblock University Park Press, 1968.

\bibitem[Valentine(1988)]{valentine1988invertedfaces}
Valentine, T.
\newblock Upside-down faces: A review of the effect of inversion upon face
  recognition.
\newblock \emph{British Journal of Psychology}, 1988.

\bibitem[Van~Essen(2003)]{vanessen2003visualcortex}
Van~Essen, D.~C.
\newblock Organization of visual areas in macaque and human cerebral cortex.
\newblock \emph{The Visual Neurosciences}, 2003.

\bibitem[Vapnik \& Chervonenkis(1971)Vapnik and Chervonenkis]{vapnik1971vc}
Vapnik, V.~N. and Chervonenkis, A.~Y.
\newblock On the uniform convergence of relative frequencies of events to their
  probabilities.
\newblock \emph{Theory of Probability and its Applications}, 1971.

\bibitem[Vinyals et~al.(2016)Vinyals, Blundell, Lillicrap, Wierstra,
  et~al.]{vinyals2016oneshot}
Vinyals, O., Blundell, C., Lillicrap, T., Wierstra, D., et~al.
\newblock Matching networks for one shot learning.
\newblock In \emph{Advances in Neural Information Processing Systems
  (NeurIPS)}, 2016.

\bibitem[Von~Stein et~al.(2000)Von~Stein, Chiang, and
  K{\"o}nig]{vonstein2000topdown}
Von~Stein, A., Chiang, C., and K{\"o}nig, P.
\newblock Top-down processing mediated by interareal synchronization.
\newblock \emph{Proceedings of the National Academy of Sciences (PNAS)}, 2000.

\bibitem[Wandell et~al.(2007)Wandell, Dumoulin, and
  Brewer]{wandell2007visualfield}
Wandell, B.~A., Dumoulin, S.~O., and Brewer, A.~A.
\newblock Visual field maps in human cortex.
\newblock \emph{Neuron}, 2007.

\bibitem[W{\"a}ssle et~al.(1990)W{\"a}ssle, Gr{\"u}nert, R{\"o}hrenbeck, and
  Boycott]{wassle1990fovea}
W{\"a}ssle, H., Gr{\"u}nert, U., R{\"o}hrenbeck, J., and Boycott, B.~B.
\newblock Retinal ganglion cell density and cortical magnification factor in
  the primate.
\newblock \emph{Vision Research}, 1990.

\bibitem[Wolpert(1996)]{wolpert1996nofreelunch}
Wolpert, D.~H.
\newblock The lack of a priori distinctions between learning algorithms.
\newblock \emph{Neural Computation}, 1996.

\bibitem[Wyss et~al.(2003)Wyss, K{\"o}nig, and
  Verschure]{wyss2003temporalstability}
Wyss, R., K{\"o}nig, P., and Verschure, P.~F.
\newblock Invariant representations of visual patterns in a temporal population
  code.
\newblock \emph{Proceedings of the National Academy of Sciences (PNAS)}, 2003.

\bibitem[Yamins et~al.(2014)Yamins, Hong, Cadieu, Solomon, Seibert, and
  DiCarlo]{yamins2014annsbrains}
Yamins, D.~L., Hong, H., Cadieu, C.~F., Solomon, E.~A., Seibert, D., and
  DiCarlo, J.~J.
\newblock Performance-optimized hierarchical models predict neural responses in
  higher visual cortex.
\newblock \emph{Proceedings of the National Academy of Sciences (PNAS)}, 2014.

\bibitem[Yin(1969)]{yin1969invertedfaces}
Yin, R.~K.
\newblock Looking at upside-down faces.
\newblock \emph{Journal of Experimental Psychology}, 1969.

\bibitem[Zador(2019)]{zador2019purelearning}
Zador, A.~M.
\newblock A critique of pure learning and what artificial neural networks can
  learn from animal brains.
\newblock \emph{Nature Communications}, 2019.

\bibitem[Zhang et~al.(2017{\natexlab{a}})Zhang, Bengio, Hardt, Recht, and
  Vinyals]{zhang2016understandingdl}
Zhang, C., Bengio, S., Hardt, M., Recht, B., and Vinyals, O.
\newblock Understanding deep learning requires rethinking generalization.
\newblock In \emph{International Conference on Learning Representations (ICLR),
  arXiv:1611.03530}, 2017{\natexlab{a}}.

\bibitem[Zhang et~al.(2017{\natexlab{b}})Zhang, Cisse, Dauphin, and
  Lopez-Paz]{zhang2017mixup}
Zhang, H., Cisse, M., Dauphin, Y.~N., and Lopez-Paz, D.
\newblock mixup: Beyond empirical risk minimization.
\newblock \emph{arXiv preprint arXiv:1710.09412}, 2017{\natexlab{b}}.

\bibitem[Zhao et~al.(2019)Zhao, Zheng, Xu, and Wu]{zhao2019objectdetection}
Zhao, Z.-Q., Zheng, P., Xu, S.-t., and Wu, X.
\newblock Object detection with deep learning: A review.
\newblock \emph{IEEE Transactions on Neural Networks and Learning Systems},
  2019.

\bibitem[Zhaoping \& Guyader(2007)Zhaoping and Guyader]{zhaoping2007topdown}
Zhaoping, L. and Guyader, N.
\newblock Interference with bottom-up feature detection by higher-level object
  recognition.
\newblock \emph{Current Biology}, 2007.

\bibitem[Zhou et~al.(2017)Zhou, Li, and Tian]{zhou2017imageretrieval}
Zhou, W., Li, H., and Tian, Q.
\newblock Recent advance in content-based image retrieval: A literature survey.
\newblock \emph{arXiv preprint arXiv:1706.06064}, 2017.

\bibitem[Zhuang et~al.(2019)Zhuang, Qi, Duan, Xi, Zhu, Zhu, Xiong, and
  He]{zhuang2019transferlearning}
Zhuang, F., Qi, Z., Duan, K., Xi, D., Zhu, Y., Zhu, H., Xiong, H., and He, Q.
\newblock A comprehensive survey on transfer learning.
\newblock \emph{arXiv preprint arXiv:1911.02685}, 2019.

\bibitem[Zoph \& Le(2016)Zoph and Le]{zoph2016nas}
Zoph, B. and Le, Q.~V.
\newblock Neural architecture search with reinforcement learning.
\newblock \emph{arXiv preprint arXiv:1611.01578}, 2016.

\bibitem[Zuiderbaan et~al.(2017)Zuiderbaan, Harvey, and
  Dumoulin]{zuiderbaan2017imageidentification}
Zuiderbaan, W., Harvey, B.~M., and Dumoulin, S.~O.
\newblock Image identification from brain activity using the population
  receptive field model.
\newblock \emph{PLOS ONE}, 2017.

\end{thebibliography}


\begin{thebibliography}{0}
\providecommand{\natexlab}[1]{#1}
\providecommand{\url}[1]{\texttt{#1}}
\expandafter\ifx\csname urlstyle\endcsname\relax
  \providecommand{\doi}[1]{doi: #1}\else
  \providecommand{\doi}{doi: \begingroup \urlstyle{rm}\Url}\fi

\end{thebibliography}


\begin{thebibliography}{42}
\providecommand{\natexlab}[1]{#1}
\providecommand{\url}[1]{\texttt{#1}}
\expandafter\ifx\csname urlstyle\endcsname\relax
  \providecommand{\doi}[1]{doi: #1}\else
  \providecommand{\doi}{doi: \begingroup \urlstyle{rm}\Url}\fi

\bibitem[A{\c{c}}ık et~al.(2010)A{\c{c}}ık, Sarwary, Schultze-Kraft, Onat,
  and K{\"{o}}nig]{acik2010bottomuptopdown}
A{\c{c}}ık, A., Sarwary, A., Schultze-Kraft, R., Onat, S., and K{\"{o}}nig, P.
\newblock Developmental changes in natural viewing behavior: Bottom-up and
  top-down differences between children, young adults and older adults.
\newblock \emph{Frontiers in Psychology}, 2010.

\bibitem[Baddeley \& Tatler(2006)Baddeley and Tatler]{baddeley2006bottomup}
Baddeley, R.~J. and Tatler, B.~W.
\newblock {High frequency edges (but not contrast) predict where we fixate: A
  Bayesian system identification analysis}.
\newblock \emph{Vision Research}, sep 2006.

\bibitem[Balcetis \& Dunning(2006)Balcetis and Dunning]{balcetis2006topdown}
Balcetis, E. and Dunning, D.
\newblock {See what you want to see: Motivational influences on visual
  perception.}
\newblock \emph{Journal of Personality and Social Psychology}, 2006.

\bibitem[Barton et~al.(2006)Barton, Radcliffe, Cherkasova, Edelman, and
  Intriligator]{barton2006leftbias}
Barton, J. J.~S., Radcliffe, N., Cherkasova, M.~V., Edelman, J., and
  Intriligator, J.~M.
\newblock Information processing during face recognition: The effects of
  familiarity, inversion, and morphing on scanning fixations.
\newblock \emph{Perception}, aug 2006.

\bibitem[Bradley \& Terry(1952)Bradley and Terry]{bradley1952pairwisecomp}
Bradley, R.~A. and Terry, M.~E.
\newblock Rank analysis of incomplete block designs: I. the method of paired
  comparisons.
\newblock \emph{Biometrika}, 1952.

\bibitem[{Calen Walshe} \& Nuthmann(2014){Calen Walshe} and
  Nuthmann]{calenwalshe2014assymetricfixation}
{Calen Walshe}, R. and Nuthmann, A.
\newblock Asymmetrical control of fixation durations in scene viewing.
\newblock \emph{Vision Research}, jul 2014.

\bibitem[Connor et~al.(2004)Connor, Egeth, and
  Yantis]{connor2004buttomuptopdown}
Connor, C.~E., Egeth, H.~E., and Yantis, S.
\newblock Visual attention: bottom-up versus top-down.
\newblock \emph{Current Biology}, 2004.

\bibitem[Corbetta \& Shulman(2002)Corbetta and
  Shulman]{corbetta2002topdownbottomup}
Corbetta, M. and Shulman, G.~L.
\newblock {Control of goal-directed and stimulus-driven attention in the
  brain}.
\newblock \emph{Nature Reviews Neuroscience}, mar 2002.

\bibitem[Desimone \& Duncan(1995)Desimone and
  Duncan]{desimone1995visualattention}
Desimone, R. and Duncan, J.
\newblock Neural mechanisms of selective visual attention.
\newblock \emph{Annual Review of Neuroscience}, 1995.

\bibitem[Dowiasch et~al.(2015)Dowiasch, Marx, Einh{\"{a}}user, and
  Bremmer]{dowiasch2015aging}
Dowiasch, S., Marx, S., Einh{\"{a}}user, W., and Bremmer, F.
\newblock {Effects of aging on eye movements in the real world}.
\newblock \emph{Frontiers in Human Neuroscience}, feb 2015.

\bibitem[Egeth \& Yantis(1997)Egeth and Yantis]{egeth1997visualattention}
Egeth, H.~E. and Yantis, S.
\newblock {Visual attention: Control, representation, and time course}.
\newblock \emph{Annual Review of Psychology}, feb 1997.

\bibitem[Einh{\"{a}}user et~al.(2008)Einh{\"{a}}user, Spain, and
  Perona]{einhauser2008topdown}
Einh{\"{a}}user, W., Spain, M., and Perona, P.
\newblock {Objects predict fixations better than early saliency}.
\newblock \emph{Journal of Vision}, nov 2008.

\bibitem[Fan et~al.(2008)Fan, Chang, Hsieh, Wang, and Lin]{fan2008liblinear}
Fan, R.-E., Chang, K.-W., Hsieh, C.-J., Wang, X.-R., and Lin, C.-J.
\newblock {LIBLINEAR: A library for large linear classification}.
\newblock \emph{Journal of Machine Learning Research (JMLR)}, 2008.

\bibitem[Geisler \& Cormack(2011)Geisler and Cormack]{geisler2011eyemovements}
Geisler, W.~S. and Cormack, L.~K.
\newblock {Models of overt attention}.
\newblock In \emph{The Oxford Handbook of Eye Movements}, pp.\  439--454.
  Oxford University Press, 2011.

\bibitem[Guo(2007)]{guo2007topdown}
Guo, K.
\newblock {Initial fixation placement in face images is driven by top–down
  guidance}.
\newblock \emph{Experimental Brain Research}, jul 2007.

\bibitem[Guo et~al.(2009)Guo, Meints, Hall, Hall, and Mills]{guo2009leftbias}
Guo, K., Meints, K., Hall, C., Hall, S., and Mills, D.
\newblock {Left gaze bias in humans, rhesus monkeys and domestic dogs}.
\newblock \emph{Animal Cognition}, may 2009.

\bibitem[Harel et~al.(2007)Harel, Koch, and Perona]{harel2007gbvs}
Harel, J., Koch, C., and Perona, P.
\newblock Graph-based visual saliency.
\newblock In \emph{Advances in Neural Information Processing Systems
  (NeurIPS)}, 2007.

\bibitem[Itti \& Koch(2001)Itti and Koch]{itti2001salience}
Itti, L. and Koch, C.
\newblock {Computational modelling of visual attention}.
\newblock \emph{Nature Reviews Neuroscience}, mar 2001.

\bibitem[Itti et~al.(1998)Itti, Koch, and Niebur]{itti1998salience}
Itti, L., Koch, C., and Niebur, E.
\newblock A model of saliency-based visual attention for rapid scene analysis.
\newblock \emph{IEEE Transactions on Pattern Analysis and Machine Intelligence
  (TPAMI)}, 1998.

\bibitem[Kaspar(2013)]{kaspar2013visualattention}
Kaspar, K.
\newblock {What guides visual overt attention under natural conditions? Past
  and future research}.
\newblock \emph{ISRN Neuroscience}, 2013.

\bibitem[Kaspar \& K{\"{o}}nig(2012)Kaspar and K{\"{o}}nig]{kaspar2012topdown}
Kaspar, K. and K{\"{o}}nig, P.
\newblock {Emotions and personality traits as high-level factors in visual
  attention: a review}.
\newblock \emph{Frontiers in Human Neuroscience}, 2012.

\bibitem[Kastner \& Ungerleider(2000)Kastner and
  Ungerleider]{kastner2000visualattention}
Kastner, S. and Ungerleider, L.~G.
\newblock {Mechanisms of visual attention in the human cortex}.
\newblock \emph{Annual Review of Neuroscience}, mar 2000.

\bibitem[Kietzmann \& K{\"{o}}nig(2015)Kietzmann and
  K{\"{o}}nig]{kietzmann2015topdown}
Kietzmann, T.~C. and K{\"{o}}nig, P.
\newblock {Effects of contextual information and stimulus ambiguity on overt
  visual sampling behavior}.
\newblock \emph{Vision Research}, may 2015.

\bibitem[Koch \& Ullman(1987)Koch and Ullman]{koch1987salience}
Koch, C. and Ullman, S.
\newblock Shifts in selective visual attention: Towards the underlying neural
  circuitry.
\newblock In Vaina, L. (ed.), \emph{Matters of Intelligence}, chapter~4, pp.\
  115--141. Springer, Dordrecht, 1987.
\newblock \doi{10.1007/978-94-009-3833-5_5}.

\bibitem[Kollmorgen et~al.(2010)Kollmorgen, Nortmann, Schr{\"{o}}der, and
  K{\"{o}}nig]{kollmorgen2010topdownbottomup}
Kollmorgen, S., Nortmann, N., Schr{\"{o}}der, S., and K{\"{o}}nig, P.
\newblock {Influence of low-level stimulus features, task dependent factors,
  and spatial biases on overt visual attention}.
\newblock \emph{PLOS Computational Biology}, may 2010.

\bibitem[K{\"{o}}nig et~al.(2016)K{\"{o}}nig, Wilming, Kietzmann,
  Ossand{\'{o}}n, Onat, Ehinger, {Ramos Gameiro}, and
  Kaspar]{konig2016eyemovements}
K{\"{o}}nig, P., Wilming, N., Kietzmann, T.~C., Ossand{\'{o}}n, J.~P., Onat,
  S., Ehinger, B.~V., {Ramos Gameiro}, R., and Kaspar, K.
\newblock {Eye movements as a window to cognitive processes}.
\newblock \emph{Journal of Eye Movement Research}, 2016.

\bibitem[Kowler(2011)]{kowler2011eyemovements}
Kowler, E.
\newblock {Eye movements: The past 25 years}.
\newblock \emph{Vision Research}, 2011.

\bibitem[K{\"u}mmerer et~al.(2016)K{\"u}mmerer, Wallis, and
  Bethge]{kuemmerer2016deepgaze}
K{\"u}mmerer, M., Wallis, T.~S., and Bethge, M.
\newblock {DeepGaze II: Reading fixations from deep features trained on object
  recognition}.
\newblock \emph{arXiv preprint arXiv:1610.01563}, 2016.

\bibitem[K{\"u}mmerer et~al.(2017)K{\"u}mmerer, Wallis, Gatys, and
  Bethge]{kuemmerer2017icfdeepgaze}
K{\"u}mmerer, M., Wallis, T.~S., Gatys, L.~A., and Bethge, M.
\newblock Understanding low-and high-level contributions to fixation
  prediction.
\newblock In \emph{IEEE International Conference on Computer Vision (ICCV)},
  2017.

\bibitem[Liversedge \& Findlay(2000)Liversedge and
  Findlay]{liversedge2000eyemovements}
Liversedge, S.~P. and Findlay, J.~M.
\newblock {Saccadic eye movements and cognition}.
\newblock \emph{Trends in Cognitive Sciences}, jan 2000.

\bibitem[Luce(2005)]{luce2005pairwisecomp}
Luce, R.~D.
\newblock \emph{{Individual Choice Behavior: A Theoretical Analysis}}.
\newblock Dover Publications, Inc., Mineola, New York, dover edit edition,
  2005.

\bibitem[Niebur \& Koch(1996)Niebur and Koch]{niebur1996salience}
Niebur, E. and Koch, C.
\newblock {Control of selective visual attention: Modeling the "where"
  pathway}.
\newblock \emph{Advances in Neural Information Processing Systems (NeurIPS)},
  1996.

\bibitem[Ossand{\'{o}}n et~al.(2014)Ossand{\'{o}}n, Onat, and
  K{\"{o}}nig]{ossandon2014spatialbiases}
Ossand{\'{o}}n, J.~P., Onat, S., and K{\"{o}}nig, P.
\newblock {Spatial biases in viewing behavior}.
\newblock \emph{Journal of Vision}, 2014.

\bibitem[{Ramos Gameiro} et~al.(2017){Ramos Gameiro}, Kaspar, K{\"{o}}nig,
  Nordholt, and K{\"{o}}nig]{ramosgameiro2017explorationexploitation}
{Ramos Gameiro}, R., Kaspar, K., K{\"{o}}nig, S.~U., Nordholt, S., and
  K{\"{o}}nig, P.
\newblock {Exploration and Exploitation in Natural Viewing Behavior}.
\newblock \emph{Scientific Reports}, dec 2017.

\bibitem[Rauthmann et~al.(2012)Rauthmann, Seubert, Sachse, and
  Furtner]{rauthmann2012topdown}
Rauthmann, J.~F., Seubert, C.~T., Sachse, P., and Furtner, M.~R.
\newblock {Eyes as windows to the soul: Gazing behavior is related to
  personality}.
\newblock \emph{Journal of Research in Personality}, apr 2012.

\bibitem[Reinagel \& Zador(1999)Reinagel and Zador]{reinagel1999bottomup}
Reinagel, P. and Zador, A.~M.
\newblock {Natural scene statistics at the center of gaze}.
\newblock \emph{Computation in Neural Systems}, 1999.

\bibitem[Riche et~al.(2013)Riche, Duvinage, Mancas, Gosselin, and
  Dutoit]{riche2013saliency}
Riche, N., Duvinage, M., Mancas, M., Gosselin, B., and Dutoit, T.
\newblock Saliency and human fixations: State-of-the-art and study of
  comparison metrics.
\newblock In \emph{IEEE Conference on Computer Vision and Pattern Recognition
  (CVPR)}, 2013.

\bibitem[Spalek \& Hammad(2005)Spalek and Hammad]{spalek2005leftbias}
Spalek, T.~M. and Hammad, S.
\newblock {The left-to-right bias in inhibition of return is due to the
  direction of reading}.
\newblock \emph{Psychological Science}, jan 2005.

\bibitem[Tatler(2007)]{tatler2007centralbias}
Tatler, B.~W.
\newblock {The central fixation bias in scene viewing: Selecting an optimal
  viewing position independently of motor biases and image feature
  distributions}.
\newblock \emph{Journal of Vision}, nov 2007.

\bibitem[Tjur(2009)]{tjur2009r2}
Tjur, T.
\newblock Coefficients of determination in logistic regression models—a new
  proposal: The coefficient of discrimination.
\newblock \emph{The American Statistician}, 2009.

\bibitem[Wadlinger \& Isaacowitz(2006)Wadlinger and
  Isaacowitz]{wadlinger2006topdown}
Wadlinger, H.~A. and Isaacowitz, D.~M.
\newblock {Positive mood broadens visual attention to positive stimuli}.
\newblock \emph{Motivation and Emotion}, mar 2006.

\bibitem[Zaeinab et~al.(2016)Zaeinab, Ossand{\'{o}}n, and
  K{\"{o}}nig]{zaeinab2016leftbias}
Zaeinab, A., Ossand{\'{o}}n, J.~P., and K{\"{o}}nig, P.
\newblock {The dynamic effect of reading direction habit on spatial asymmetry
  of image perception}.
\newblock \emph{Journal of Vision}, 2016.

\end{thebibliography}


\begin{thebibliography}{15}
\providecommand{\natexlab}[1]{#1}
\providecommand{\url}[1]{\texttt{#1}}
\expandafter\ifx\csname urlstyle\endcsname\relax
  \providecommand{\doi}[1]{doi: #1}\else
  \providecommand{\doi}{doi: \begingroup \urlstyle{rm}\Url}\fi

\bibitem[Boynton et~al.(1999)Boynton, Demb, Glover, and
  Heeger]{boynton1999contrast}
Boynton, G.~M., Demb, J.~B., Glover, G.~H., and Heeger, D.~J.
\newblock Neuronal basis of contrast discrimination.
\newblock \emph{Vision Research}, 1999.

\bibitem[Dumoulin \& Wandell(2008)Dumoulin and Wandell]{dumoulin2008prf}
Dumoulin, S.~O. and Wandell, B.~A.
\newblock Population receptive field estimates in human visual cortex.
\newblock \emph{Neuroimage}, 2008.

\bibitem[Friston et~al.(1995)Friston, Holmes, Poline, Grasby, Williams,
  Frackowiak, Turner, et~al.]{friston1995fmrianalysis}
Friston, K.~J., Holmes, A.~P., Poline, J., Grasby, P., Williams, S.,
  Frackowiak, R.~S., Turner, R., et~al.
\newblock Analysis of fmri time-series revisited.
\newblock \emph{Neuroimage}, 1995.

\bibitem[Kay et~al.(2008)Kay, Naselaris, Prenger, and Gallant]{kay2008imageid}
Kay, K.~N., Naselaris, T., Prenger, R.~J., and Gallant, J.~L.
\newblock Identifying natural images from human brain activity.
\newblock \emph{Nature}, 2008.

\bibitem[Kriegeskorte et~al.(2008)Kriegeskorte, Mur, and
  Bandettini]{kriegeskorte2008rsa}
Kriegeskorte, N., Mur, M., and Bandettini, P.~A.
\newblock Representational similarity analysis-connecting the branches of
  systems neuroscience.
\newblock \emph{Frontiers in Systems Neuroscience}, 2008.

\bibitem[K{\"u}mmerer et~al.(2016)K{\"u}mmerer, Wallis, and
  Bethge]{kuemmerer2016deepgaze}
K{\"u}mmerer, M., Wallis, T.~S., and Bethge, M.
\newblock {DeepGaze II: Reading fixations from deep features trained on object
  recognition}.
\newblock \emph{arXiv preprint arXiv:1610.01563}, 2016.

\bibitem[K{\"u}mmerer et~al.(2017)K{\"u}mmerer, Wallis, Gatys, and
  Bethge]{kuemmerer2017icfdeepgaze}
K{\"u}mmerer, M., Wallis, T.~S., Gatys, L.~A., and Bethge, M.
\newblock Understanding low-and high-level contributions to fixation
  prediction.
\newblock In \emph{IEEE International Conference on Computer Vision (ICCV)},
  2017.

\bibitem[Martin et~al.(2001)Martin, Fowlkes, Tal, and
  Malik]{martin2001berkeley}
Martin, D., Fowlkes, C., Tal, D., and Malik, J.
\newblock A database of human segmented natural images and its application to
  evaluating segmentation algorithms and measuring ecological statistics.
\newblock In \emph{IEEE International Conference on Computer Vision (ICCV)}.
  2001.

\bibitem[Olman et~al.(2004)Olman, Ugurbil, Schrater, and
  Kersten]{olman2004contrast}
Olman, C.~A., Ugurbil, K., Schrater, P., and Kersten, D.
\newblock Bold fmri and psychophysical measurements of contrast response to
  broadband images.
\newblock \emph{Vision Research}, 2004.

\bibitem[Simonyan \& Zisserman(2014)Simonyan and Zisserman]{simonyan2014vgg}
Simonyan, K. and Zisserman, A.
\newblock Very deep convolutional networks for large-scale image recognition.
\newblock \emph{arXiv preprint arXiv:1409.1556}, 2014.

\bibitem[Smith et~al.(2001)Smith, Singh, Williams, and
  Greenlee]{smith2001rfsizes}
Smith, A.~T., Singh, K.~D., Williams, A., and Greenlee, M.~W.
\newblock Estimating receptive field size from fmri data in human striate and
  extrastriate visual cortex.
\newblock \emph{Cerebral Cortex}, 2001.

\bibitem[Tong \& Pratte(2012)Tong and Pratte]{tong2012mindreading}
Tong, F. and Pratte, M.~S.
\newblock Decoding patterns of human brain activity.
\newblock \emph{Annual Review of Psychology}, 2012.

\bibitem[Treue(2003)]{treue2003salience}
Treue, S.
\newblock Visual attention: the where, what, how and why of saliency.
\newblock \emph{Current Opinion in Neurobiology}, 2003.

\bibitem[Wandell et~al.(2007)Wandell, Dumoulin, and
  Brewer]{wandell2007visualfield}
Wandell, B.~A., Dumoulin, S.~O., and Brewer, A.~A.
\newblock Visual field maps in human cortex.
\newblock \emph{Neuron}, 2007.

\bibitem[Zuiderbaan et~al.(2017)Zuiderbaan, Harvey, and
  Dumoulin]{zuiderbaan2017imageidentification}
Zuiderbaan, W., Harvey, B.~M., and Dumoulin, S.~O.
\newblock Image identification from brain activity using the population
  receptive field model.
\newblock \emph{PLOS ONE}, 2017.

\end{thebibliography}


\begin{thebibliography}{41}
\providecommand{\natexlab}[1]{#1}
\providecommand{\url}[1]{\texttt{#1}}
\expandafter\ifx\csname urlstyle\endcsname\relax
  \providecommand{\doi}[1]{doi: #1}\else
  \providecommand{\doi}{doi: \begingroup \urlstyle{rm}\Url}\fi

\bibitem[Achille \& Soatto(2018)Achille and Soatto]{achille2018emergence}
Achille, A. and Soatto, S.
\newblock Emergence of invariance and disentanglement in deep representations.
\newblock \emph{Journal of Machine Learning Research (JMLR)}, 2018.

\bibitem[Achlioptas et~al.(2018)Achlioptas, Diamanti, Mitliagkas, and
  Guibas]{achlioptas20183d}
Achlioptas, P., Diamanti, O., Mitliagkas, I., and Guibas, L.
\newblock Learning representations and generative models for {3D} point clouds.
\newblock In \emph{International Conference on Machine Learning (ICML)}, 2018.

\bibitem[Bao et~al.(2020)Bao, She, McGill, and Tsao]{bao2020itmaps}
Bao, P., She, L., McGill, M., and Tsao, D.~Y.
\newblock A map of object space in primate inferotemporal cortex.
\newblock \emph{Nature}, 2020.

\bibitem[Becker(1999)]{becker1999temporalstability}
Becker, S.
\newblock Implicit learning in {3D} object recognition: The importance of
  temporal context.
\newblock \emph{Neural Computation}, 1999.

\bibitem[Belharbi et~al.(2017)Belharbi, Chatelain, H{\'e}rault, and
  Adam]{belharbi2017classinvariance}
Belharbi, S., Chatelain, C., H{\'e}rault, R., and Adam, S.
\newblock Neural networks regularization through class-wise invariant
  representation learning.
\newblock \emph{arXiv preprint arXiv:1709.01867}, 2017.

\bibitem[Bengio et~al.(2015)Bengio, Lee, Bornschein, Mesnard, and
  Lin]{bengio2015dlandneuroscience}
Bengio, Y., Lee, D.-H., Bornschein, J., Mesnard, T., and Lin, Z.
\newblock Towards biologically plausible deep learning.
\newblock \emph{arXiv preprint arXiv:1502.04156}, 2015.

\bibitem[Booth \& Rolls(1998)Booth and Rolls]{booth1998invariantitmacaque}
Booth, M. and Rolls, E.~T.
\newblock View-invariant representations of familiar objects by neurons in the
  inferior temporal visual cortex.
\newblock \emph{Cerebral Cortex}, 1998.

\bibitem[Cohen \& Welling(2016)Cohen and Welling]{cohen2016groupequivcnns}
Cohen, T. and Welling, M.
\newblock Group equivariant convolutional networks.
\newblock In \emph{International Conference on Machine Learning (ICML)}, 2016.

\bibitem[DiCarlo \& Cox(2007)DiCarlo and Cox]{dicarlo2007untangling}
DiCarlo, J.~J. and Cox, D.~D.
\newblock Untangling invariant object recognition.
\newblock \emph{Trends in Cognitive Sciences}, 2007.

\bibitem[Dujmovi{\'c} et~al.(2020)Dujmovi{\'c}, Malhotra, and
  Bowers]{dujmovic2020adversarial}
Dujmovi{\'c}, M., Malhotra, G., and Bowers, J.
\newblock What do adversarial images tell us about human vision?
\newblock \emph{bioRxiv preprint 2020.02.25.964361}, 2020.

\bibitem[Geirhos et~al.(2020)Geirhos, Jacobsen, Michaelis, Zemel, Brendel,
  Bethge, and Wichmann]{geirhos2020shortcutlearning}
Geirhos, R., Jacobsen, J.-H., Michaelis, C., Zemel, R., Brendel, W., Bethge,
  M., and Wichmann, F.~A.
\newblock Shortcut learning in deep neural networks.
\newblock \emph{arXiv preprint arXiv:2004.07780}, 2020.

\bibitem[Hadsell et~al.(2006)Hadsell, Chopra, and
  LeCun]{hadsell2006contrastive}
Hadsell, R., Chopra, S., and LeCun, Y.
\newblock Dimensionality reduction by learning an invariant mapping.
\newblock In \emph{IEEE Conference on Computer Vision and Pattern Recognition
  (CVPR)}. 2006.

\bibitem[Hoffer et~al.(2019)Hoffer, Ben{-}Nun, Hubara, Giladi, Hoefler, and
  Soudry]{hoffer2019batchaugmentation}
Hoffer, E., Ben{-}Nun, T., Hubara, I., Giladi, N., Hoefler, T., and Soudry, D.
\newblock Augment your batch: better training with larger batches.
\newblock \emph{arXiv preprint arXiv:1901.09335}, 2019.

\bibitem[Huang et~al.(2017)Huang, Liu, Van Der~Maaten, and
  Weinberger]{huang2017densenet}
Huang, G., Liu, Z., Van Der~Maaten, L., and Weinberger, K.~Q.
\newblock Densely connected convolutional networks.
\newblock In \emph{IEEE Conference on Computer Vision and Pattern Recognition
  (CVPR)}. 2017.

\bibitem[Ilyas et~al.(2019)Ilyas, Santurkar, Tsipras, Engstrom, Tran, and
  Madry]{ilyas2019advfeatures}
Ilyas, A., Santurkar, S., Tsipras, D., Engstrom, L., Tran, B., and Madry, A.
\newblock Adversarial examples are not bugs, they are features.
\newblock \emph{arXiv preprint arXiv:1905.02175}, 2019.

\bibitem[Isik et~al.(2013)Isik, Meyers, Leibo, and Poggio]{isik2013dynamics}
Isik, L., Meyers, E.~M., Leibo, J.~Z., and Poggio, T.
\newblock The dynamics of invariant object recognition in the human visual
  system.
\newblock \emph{Journal of Neurophysiology}, 2013.

\bibitem[Jo \& Bengio(2017)Jo and Bengio]{jo2017surfaceregularities}
Jo, J. and Bengio, Y.
\newblock Measuring the tendency of {CNNs} to learn surface statistical
  regularities.
\newblock \emph{arXiv preprint arXiv:1711.11561}, 2017.

\bibitem[Kietzmann et~al.(2019)Kietzmann, McClure, and
  Kriegeskorte]{kietzmann2019dnncompneuro}
Kietzmann, T.~C., McClure, P., and Kriegeskorte, N.
\newblock Deep neural networks in computational neuroscience.
\newblock \emph{Oxford Research Encyclopedia of Neuroscience}, 2019.

\bibitem[Kording et~al.(2004)Kording, Kayser, Einhauser, and
  Konig]{kording2004complexcells}
Kording, K.~P., Kayser, C., Einhauser, W., and Konig, P.
\newblock How are complex cell properties adapted to the statistics of natural
  stimuli?
\newblock \emph{Journal of Neurophysiology}, 2004.

\bibitem[Kriegeskorte et~al.(2008)Kriegeskorte, Mur, Ruff, Kiani, Bodurka,
  Esteky, Tanaka, and Bandettini]{kriegeskorte2008manandmonkey}
Kriegeskorte, N., Mur, M., Ruff, D.~A., Kiani, R., Bodurka, J., Esteky, H.,
  Tanaka, K., and Bandettini, P.~A.
\newblock Matching categorical object representations in inferior temporal
  cortex of man and monkey.
\newblock \emph{Neuron}, 2008.

\bibitem[Krizhevsky \& Hinton(2009)Krizhevsky and Hinton]{krizhevsky2009cifar}
Krizhevsky, A. and Hinton, G.
\newblock Learning multiple layers of features from tiny images.
\newblock \emph{Technical report, University of Toronto}, 2009.

\bibitem[Kubilius et~al.(2018)Kubilius, Schrimpf, Nayebi, Bear, Yamins, and
  DiCarlo]{kubilius2018cornet}
Kubilius, J., Schrimpf, M., Nayebi, A., Bear, D., Yamins, D.~L., and DiCarlo,
  J.~J.
\newblock Cornet: Modeling the neural mechanisms of core object recognition.
\newblock \emph{bioRxiv:408385}, 2018.

\bibitem[Laine \& Aila(2016)Laine and Aila]{laine2016ssl}
Laine, S. and Aila, T.
\newblock Temporal ensembling for semi-supervised learning.
\newblock \emph{arXiv preprint arXiv:1610.02242}, 2016.

\bibitem[Malhotra et~al.(2020)Malhotra, Evans, and
  Bowers]{malhotra2020bioconstraints}
Malhotra, G., Evans, B.~D., and Bowers, J.~S.
\newblock Hiding a plane with a pixel: examining shape-bias in {CNNs} and the
  benefit of building in biological constraints.
\newblock \emph{Vision Research}, 2020.

\bibitem[Marblestone et~al.(2016)Marblestone, Wayne, and
  Kording]{marblestone2016dlandneuroscience}
Marblestone, A.~H., Wayne, G., and Kording, K.~P.
\newblock Toward an integration of deep learning and neuroscience.
\newblock \emph{Frontiers in Computational Neuroscience}, 2016.

\bibitem[Quiroga et~al.(2005)Quiroga, Reddy, Kreiman, Koch, and
  Fried]{quiroga2005invariantithuman}
Quiroga, R.~Q., Reddy, L., Kreiman, G., Koch, C., and Fried, I.
\newblock Invariant visual representation by single neurons in the human brain.
\newblock \emph{Nature}, 2005.

\bibitem[Richards et~al.(2019)Richards, Lillicrap, Beaudoin, Bengio, Bogacz,
  Christensen, Clopath, Costa, de~Berker, Ganguli,
  et~al.]{richards2019dlandneuroscience}
Richards, B.~A. et~al.
\newblock A deep learning framework for neuroscience.
\newblock \emph{Nature Neuroscience}, 2019.

\bibitem[Russakovsky et~al.(2015)Russakovsky, Deng, Su, Krause, Satheesh, Ma,
  Huang, Karpathy, Khosla, Bernstein, Berg, and
  Fei-Fei]{russakovsky2015imagenet}
Russakovsky, O. et~al.
\newblock {ImageNet} large scale visual recognition challenge.
\newblock \emph{International Journal of Computer Vision (IJCV)}, 2015.

\bibitem[Saxe et~al.(2019)Saxe, Bansal, Dapello, Advani, Kolchinsky, Tracey,
  and Cox]{saxe2019informationbottleneck}
Saxe, A.~M., Bansal, Y., Dapello, J., Advani, M., Kolchinsky, A., Tracey,
  B.~D., and Cox, D.~D.
\newblock On the information bottleneck theory of deep learning.
\newblock \emph{Journal of Statistical Mechanics: Theory and Experiment}, 2019.

\bibitem[Shwartz-Ziv \& Tishby(2017)Shwartz-Ziv and
  Tishby]{shwartz2017bottleneck}
Shwartz-Ziv, R. and Tishby, N.
\newblock Opening the black box of deep neural networks via information.
\newblock \emph{arXiv preprint arXiv:1703.00810}, 2017.

\bibitem[Simard et~al.(1992)Simard, Victorri, LeCun, and
  Denker]{simard1992daug}
Simard, P., Victorri, B., LeCun, Y., and Denker, J.
\newblock Tangent prop-a formalism for specifying selected invariances in an
  adaptive network.
\newblock In \emph{Advances in Neural Information Processing Systems
  (NeurIPS)}, 1992.

\bibitem[Sinz et~al.(2019)Sinz, Pitkow, Reimer, Bethge, and
  Tolias]{sinz2019dlvsbrain}
Sinz, F.~H., Pitkow, X., Reimer, J., Bethge, M., and Tolias, A.~S.
\newblock Engineering a less artificial intelligence.
\newblock \emph{Neuron}, 2019.

\bibitem[Springenberg et~al.(2014)Springenberg, Dosovitskiy, Brox, and
  Riedmiller]{springenberg2014allcnn}
Springenberg, J.~T., Dosovitskiy, A., Brox, T., and Riedmiller, M.
\newblock Striving for simplicity: The all convolutional net.
\newblock In \emph{International Conference on Learning Representations (ICLR),
  arXiv:1412.6806}, 2014.

\bibitem[Szegedy et~al.(2013)Szegedy, Zaremba, Sutskever, Bruna, Erhan,
  Goodfellow, and Fergus]{szegedy2013adversarial}
Szegedy, C., Zaremba, W., Sutskever, I., Bruna, J., Erhan, D., Goodfellow, I.,
  and Fergus, R.
\newblock Intriguing properties of neural networks.
\newblock \emph{arXiv preprint arXiv:1312.6199}, 2013.

\bibitem[Tacchetti et~al.(2018)Tacchetti, Isik, and
  Poggio]{tacchetti2018invariance}
Tacchetti, A., Isik, L., and Poggio, T.~A.
\newblock Invariant recognition shapes neural representations of visual input.
\newblock \emph{Annual Review of Vision Science}, 2018.

\bibitem[Taylor et~al.(2010)Taylor, Fergus, LeCun, and
  Bregler]{taylor2010spatiotemporal}
Taylor, G.~W., Fergus, R., LeCun, Y., and Bregler, C.
\newblock Convolutional learning of spatio-temporal features.
\newblock In \emph{European Conference on Computer Vision (ECCV)}. 2010.

\bibitem[Tishby \& Zaslavsky(2015)Tishby and
  Zaslavsky]{tishby2015infobottleneck}
Tishby, N. and Zaslavsky, N.
\newblock Deep learning and the information bottleneck principle.
\newblock In \emph{{IEEE Information Theory Workshop (ITW)}}. 2015.

\bibitem[Wang et~al.(2019)Wang, Wu, Yin, and Xing]{wang2019highfreq}
Wang, H., Wu, X., Yin, P., and Xing, E.~P.
\newblock High frequency component helps explain the generalization of
  convolutional neural networks.
\newblock \emph{arXiv preprint arXiv:1905.13545}, 2019.

\bibitem[Wyss et~al.(2006)Wyss, K{\"o}nig, and
  Verschure]{wyss2006temporalstability}
Wyss, R., K{\"o}nig, P., and Verschure, P. F.~J.
\newblock A model of the ventral visual system based on temporal stability and
  local memory.
\newblock \emph{PLOS Biology}, 2006.

\bibitem[Zagoruyko \& Komodakis(2016)Zagoruyko and Komodakis]{zagoruyko2016wrn}
Zagoruyko, S. and Komodakis, N.
\newblock Wide residual networks.
\newblock In \emph{British Machine Vision Conference (BMVC)}, 2016.

\bibitem[Zhang(2019)]{zhang2019convolutions}
Zhang, R.
\newblock Making convolutional networks shift-invariant again.
\newblock In \emph{International Conference on Machine Learning (ICML)}, 2019.

\end{thebibliography}


\begin{thebibliography}{25}
\providecommand{\natexlab}[1]{#1}
\providecommand{\url}[1]{\texttt{#1}}
\expandafter\ifx\csname urlstyle\endcsname\relax
  \providecommand{\doi}[1]{doi: #1}\else
  \providecommand{\doi}{doi: \begingroup \urlstyle{rm}\Url}\fi

\bibitem[Achille \& Soatto(2018)Achille and Soatto]{achille2018emergence}
Achille, A. and Soatto, S.
\newblock Emergence of invariance and disentanglement in deep representations.
\newblock \emph{Journal of Machine Learning Research (JMLR)}, 2018.

\bibitem[Belkin et~al.(2019)Belkin, Hsu, Ma, and
  Mandal]{belkin2019biasvariance}
Belkin, M., Hsu, D., Ma, S., and Mandal, S.
\newblock Reconciling modern machine-learning practice and the classical
  bias--variance trade-off.
\newblock \emph{Proceedings of the National Academy of Sciences (PNAS)}, 2019.

\bibitem[Goodfellow et~al.(2016)Goodfellow, Bengio, and
  Courville]{goodfellow2016dlbook}
Goodfellow, I., Bengio, Y., and Courville, A.
\newblock \emph{Deep learning}.
\newblock MIT press, 2016.

\bibitem[Guo et~al.(2019)Guo, Mao, and Zhang]{guo2018mixup}
Guo, H., Mao, Y., and Zhang, R.
\newblock Mixup as locally linear out-of-manifold regularization.
\newblock In \emph{Association for the Advancement of Artificial Intelligence
  (AAAI)}, 2019.

\bibitem[Hanson \& Pratt(1989)Hanson and Pratt]{hanson1989wd}
Hanson, S.~J. and Pratt, L.~Y.
\newblock Comparing biases for minimal network construction with
  back-propagation.
\newblock In \emph{Advances in Neural Information Processing Systems
  (NeurIPS)}, 1989.

\bibitem[Huang et~al.(2016)Huang, Sun, Liu, Sedra, and
  Weinberger]{huang2016stochasticdepth}
Huang, G., Sun, Y., Liu, Z., Sedra, D., and Weinberger, K.~Q.
\newblock Deep networks with stochastic depth.
\newblock In \emph{European Conference on Computer Vision (ECCV)}. 2016.

\bibitem[Huang et~al.(2017)Huang, Liu, Van Der~Maaten, and
  Weinberger]{huang2017densenet}
Huang, G., Liu, Z., Van Der~Maaten, L., and Weinberger, K.~Q.
\newblock Densely connected convolutional networks.
\newblock In \emph{IEEE Conference on Computer Vision and Pattern Recognition
  (CVPR)}. 2017.

\bibitem[Ioffe \& Szegedy(2015)Ioffe and Szegedy]{ioffe2015batchnorm}
Ioffe, S. and Szegedy, C.
\newblock Batch normalization: Accelerating deep network training by reducing
  internal covariate shift.
\newblock In \emph{International Conference on Machine Learning (ICML)}, 2015.

\bibitem[Kuka{\v{c}}ka et~al.(2017)Kuka{\v{c}}ka, Golkov, and
  Cremers]{kukavcka2017regularization}
Kuka{\v{c}}ka, J., Golkov, V., and Cremers, D.
\newblock Regularization for deep learning: A taxonomy.
\newblock \emph{arXiv preprint arXiv:1710.10686}, 2017.

\bibitem[LeCun et~al.(1990)LeCun, Boser, Denker, Henderson, Howard, Hubbard,
  and Jackel]{lecun1990conv}
LeCun, Y., Boser, B.~E., Denker, J.~S., Henderson, D., Howard, R.~E., Hubbard,
  W.~E., and Jackel, L.~D.
\newblock Handwritten digit recognition with a back-propagation network.
\newblock In \emph{Advances in Neural Information Processing Systems
  (NeurIPS)}, 1990.

\bibitem[Martin \& Mahoney(2018)Martin and
  Mahoney]{martin2018selfregularisation}
Martin, C.~H. and Mahoney, M.~W.
\newblock Implicit self-regularization in deep neural networks: Evidence from
  random matrix theory and implications for learning.
\newblock \emph{arXiv preprint arXiv:1810.01075}, 2018.

\bibitem[Mesnil et~al.(2011)Mesnil, Dauphin, Glorot, Rifai, Bengio, Goodfellow,
  Lavoie, Muller, Desjardins, Warde-Farley, et~al.]{mesnil2011transferlearning}
Mesnil, G. et~al.
\newblock Unsupervised and transfer learning challenge: a deep learning
  approach.
\newblock In \emph{Workshop on Unsupervised and Transfer Learning (ICML)},
  2011.

\bibitem[Neyshabur(2017)]{neyshabur2017thesis}
Neyshabur, B.
\newblock \emph{Implicit regularization in deep learning}.
\newblock PhD thesis, Toyota Technological Institute at Chicago,
  arXiv:1709.01953, 2017.

\bibitem[Neyshabur et~al.(2014)Neyshabur, Tomioka, and
  Srebro]{neyshabur2014implicitreg}
Neyshabur, B., Tomioka, R., and Srebro, N.
\newblock In search of the real inductive bias: On the role of implicit
  regularization in deep learning.
\newblock In \emph{International Conference on Learning Representations (ICLR),
  arXiv:1412.6614}, 2014.

\bibitem[Poggio et~al.(2017)Poggio, Kawaguchi, Liao, Miranda, Rosasco, Boix,
  Hidary, and Mhaskar]{poggio2017theory3}
Poggio, T., Kawaguchi, K., Liao, Q., Miranda, B., Rosasco, L., Boix, X.,
  Hidary, J., and Mhaskar, H.
\newblock Theory of deep learning {III}: explaining the non-overfitting puzzle.
\newblock \emph{arXiv preprint arXiv:1801.00173}, 2017.

\bibitem[Santurkar et~al.(2018)Santurkar, Tsipras, Ilyas, and
  Madry]{santurkar2018bn}
Santurkar, S., Tsipras, D., Ilyas, A., and Madry, A.
\newblock How does batch normalization help optimization?
\newblock In \emph{Advances in Neural Information Processing Systems
  (NeurIPS)}, 2018.

\bibitem[Springenberg et~al.(2014)Springenberg, Dosovitskiy, Brox, and
  Riedmiller]{springenberg2014allcnn}
Springenberg, J.~T., Dosovitskiy, A., Brox, T., and Riedmiller, M.
\newblock Striving for simplicity: The all convolutional net.
\newblock In \emph{International Conference on Learning Representations (ICLR),
  arXiv:1412.6806}, 2014.

\bibitem[Srivastava et~al.(2014)Srivastava, Hinton, Krizhevsky, Sutskever, and
  Salakhutdinov]{srivastava2014dropout}
Srivastava, N., Hinton, G.~E., Krizhevsky, A., Sutskever, I., and
  Salakhutdinov, R.~R.
\newblock Dropout: a simple way to prevent neural networks from overfitting.
\newblock \emph{Journal of Machine Learning Research (JMLR)}, 2014.

\bibitem[Tan \& Le(2019)Tan and Le]{tan2019efficientnet}
Tan, M. and Le, Q.~V.
\newblock {EfficientNet}: Rethinking model scaling for convolutional neural
  networks.
\newblock In \emph{International Conference on Machine Learning (ICML)}, 2019.

\bibitem[Tikhonov(1963)]{tikhonov1963regularisation}
Tikhonov, A.~N.
\newblock On solving ill-posed problem and method of regularization.
\newblock \emph{Doklady Akademii Nauk USSR}, 1963.

\bibitem[Wilson et~al.(2017)Wilson, Roelofs, Stern, Srebro, and
  Recht]{wilson2017neurips}
Wilson, A.~C., Roelofs, R., Stern, M., Srebro, N., and Recht, B.
\newblock The marginal value of adaptive gradient methods in machine learning.
\newblock In \emph{Advances in Neural Information Processing Systems
  (NeurIPS)}, 2017.

\bibitem[Yao et~al.(2007)Yao, Rosasco, and Caponnetto]{yao2007earlystopping}
Yao, Y., Rosasco, L., and Caponnetto, A.
\newblock On early stopping in gradient descent learning.
\newblock \emph{Constructive Approximation}, 2007.

\bibitem[Zagoruyko \& Komodakis(2016)Zagoruyko and Komodakis]{zagoruyko2016wrn}
Zagoruyko, S. and Komodakis, N.
\newblock Wide residual networks.
\newblock In \emph{British Machine Vision Conference (BMVC)}, 2016.

\bibitem[Zhang et~al.(2017{\natexlab{a}})Zhang, Bengio, Hardt, Recht, and
  Vinyals]{zhang2016understandingdl}
Zhang, C., Bengio, S., Hardt, M., Recht, B., and Vinyals, O.
\newblock Understanding deep learning requires rethinking generalization.
\newblock In \emph{International Conference on Learning Representations (ICLR),
  arXiv:1611.03530}, 2017{\natexlab{a}}.

\bibitem[Zhang et~al.(2017{\natexlab{b}})Zhang, Liao, Rakhlin, Sridharan,
  Miranda, Golowich, and Poggio]{zhang2017sgd}
Zhang, C., Liao, Q., Rakhlin, A., Sridharan, K., Miranda, B., Golowich, N., and
  Poggio, T.
\newblock Theory of deep learning {III}: Generalization properties of {SGD}.
\newblock Technical report, Center for Brains, Minds and Machines (CBMM),
  2017{\natexlab{b}}.

\end{thebibliography}
\end{document}